%% file: main.tex
\documentclass{report}

\usepackage{blindtext}
\usepackage{etoolbox}
\usepackage{titlesec}
\usepackage{titletoc}

\usepackage[subpreambles=true]{standalone}

\usepackage[english]{babel}
\usepackage[sectionbib]{natbib}
\usepackage{chapterbib}
\usepackage{import}
\usepackage{braket}
\usepackage{subcaption}
\usepackage[resetlabels,labeled]{multibib}

\usepackage{graphicx}

\usepackage[a4paper]{geometry}
\usepackage{pdfpages}
\usepackage{setspace}
\usepackage{nomencl}
\usepackage{amsthm}
\usepackage{delimset} 
\usepackage{amssymb}
\usepackage{afterpage}

\newcommand\blankpage{%
    \null
    \thispagestyle{empty}%
    \addtocounter{page}{-1}%
    \newpage}
\input{math_commands.tex}

\usepackage{braket}
\usepackage{amssymb}

\usepackage{soul}

\usepackage{amsmath}
\usepackage{amsthm}

\newtheorem{theorem}{Theorem}[section]
\newtheorem{corollary}{Corollary}[theorem]
\newtheorem{lemma}[theorem]{Lemma}

\theoremstyle{definition}
\newtheorem{definition}{Definition}[section]
\usepackage{amssymb,amsmath,amsthm}

\newcommand{\X}{\mathcal{X}}
\newcommand{\Y}{\mathcal{Y}}
\newcommand{\K}{\mathcal{K}}
\newcommand{\N}{\mathcal{N}}
\newcommand{\ReLU}{\textsc{ReLU}}
\newcommand{\Erf}{\textsc{Erf}}
\newcommand{\kl}[2]{\KL\left[ #1 \Vert #2 \right]}

\providecommand{\printnomenclature}{\printglossary}
\providecommand{\makenomenclature}{\makeglossary}
\makenomenclature
\makeatletter

\providecommand{\LyX}{L\kern-.1667em\lower.25em\hbox{Y}\kern-.125emX\@}

\usepackage{tauthesis}
\usepackage[font={small,bf}, labelfont={small,bf}, margin=1cm]{caption}
\usepackage{titlesec}

\newcommand{\hsp}{\hspace{5pt}}

\titleformat{\chapter}[hang]{\Huge\bfseries}{\thechapter\hsp}{0pt}{\Huge\bfseries}

\Title{\textbf{Information Flow in Deep Neural Networks}}
\Authorpref{A thesis submitted for the degree of}
\Author{Ravid  Shwartz-Ziv}
\Year{June 2021}
\Supervisor{Prof. Naftali Tishby}
\Department{School of Computer Science and Engineering}
\Degree{Doctor of Philosophy}

\makeatother

\begin{document}

\prelimpages

\titlepage


\acknowledgments{First and foremost, I am extremely grateful to my supervisor and mentor, Professor Naftali Tishby,  who passed away recently. His invaluable advice, continuous support, and patience during my Ph.D. study were so important. His immense knowledge and plentiful experience have encouraged me in all the time of my academic research and daily life. He was a remarkable man who gave me so much, and I learned a lot from him. An incredible scholar and a lovely person. 

I would also like to thank Prof. Haim Sompolinsky and Dr. Alex Alemi for their support and guidance during my study. I wish to thank also my
collaborators, Zoe Piran and Dr. Amichai Painsky. It was a great privilege to work with them. 

Finally, I cannot begin to express my gratitude to my wife and children. Without their tremendous understanding and encouragement in the past few years, it would be impossible for me to complete my study. There were always there to support me and encourage me to succeed.}
\include{Chapters/Abstract}
\include{Chapters/letter}

\titleformat{\chapter}[hang]{\Huge\bfseries}{\thechapter\hsp}{0pt}{\Huge\bfseries}

\tableofcontents{}

\textpages

\printnomenclature{}
\setcounter{chapter}{1}

\include{Chapters/Intro}

\include{Chapters/opening}

\include{Chapters/rep_compression}

\include{Chapters/abbi}

\include{Chapters/dualIB}

\include{Chapters/general_discussion}




\end{document}

%% file: math_commands.tex
\usepackage{amsmath,amsfonts,bm}

\def\eqref#1{equation~\ref{#1}}

\def\1{\bm{1}}

\def\rvx{{\mathbf{x}}}
\def\rvy{{\mathbf{y}}}

\DeclareMathAlphabet{\mathsfit}{\encodingdefault}{\sfdefault}{m}{sl}
\SetMathAlphabet{\mathsfit}{bold}{\encodingdefault}{\sfdefault}{bx}{n}

\newcommand{\KL}{D_{\mathrm{KL}}}

\DeclareMathOperator*{\argmin}{arg\,min}

\DeclareMathOperator{\Tr}{Tr}

%% file: Chapters/Abstract.tex
\chapter*{Abstract}
\addcontentsline{toc}{chapter}{Abstract}
While deep neural networks have been immensely successful, a comprehensive theoretical understanding of how they work or how they are structured does not exist. Deep networks are often viewed as black boxes, where the interpretation of predictions and their reliability are still unclear.  Today, understanding the groundbreaking performance of deep neural networks is one of the greatest challenges facing the scientific community. To use these algorithms more effectively and improve them, we need to understand their dynamic behavior and their ability to learn new representations. 

This thesis addresses these issues by applying principles and techniques from information theory to deep learning models to increase our theoretical understanding and use it to design better algorithms. The main results and contributions of this thesis are structured in three parts, as detailed below.

Chapters 2 and 3 present our information-theoretic approach to deep learning models.  As an explanation for deep learning systems, we propose using the Information Bottleneck (IB) theory. The novel paradigm for analyzing networks sheds light on their layered structure, generalization capabilities, and learning dynamics. Based on our analysis, we find that deep networks optimize each layer's mutual information on input and output variables, resulting in a trade-off between compression and prediction for each layer.  Our analytical and numerical study of these networks demonstrated that the stochastic gradient descent (SGD) algorithm follows the IB trade-off principle by working in two phases: a fast empirical error minimization phase followed by a slow representation compression phase. These phases are distinguished by different signal-to-noise ratios (SNRs) for each layer. Moreover, we demonstrated that the SGD achieved this optimal bound due to the compression phase, derived a new Gaussian bound on the representation compression, and related it to the compression time. Furthermore, our results indicate that the network's layers converge to the IB theoretical bound, leading to a self-consistent relationship between the encoder and decoder distributions. 
 
Chapter 4 deals with one of the most difficult problems of applying the IB to deep neural networks --- estimating mutual information in  high dimnesional space. Despite being an important quantity in data science, mutual information has historically posed a computational challenge. Computing mutual information is only tractable for discrete variables or for a limited number of problems where probability distributions are known. To better estimate information-theoretic quantities and to investigate generalization signals, we research several frameworks and utilize recent theoretical developments, such as the neural tangent kernel (NTK) framework. In our study, we found that for infinite ensembles of infinitely wide neural networks, we could obtain tractable computations of many information-theoretic quantities and their bounds. Many quantities can be described in a closed-form solution by the network's kernels. By analyzing these derivations, we can learn the important information-theoretic quantities of the network and how compression, generalization, and the sample size are related.

Chapter 5 presents the dual Information Bottleneck (dualIB), a new information-theoretic framework. Despite the IB framework's advantages, it also has several drawbacks: The IB is completely non-parametric and operates only on the probability space. In addition, the IB formulation does not relate to the task of prediction over unseen patterns and assumes full access to the joint probability. Therefore, we developed the dualIB, which resolves some of the IB's drawbacks through a mere switch between terms in the distortion function. The dualIB can account for known features of the data and use them to make better predictions over unseen examples. We provide dualIB self-consistent equations, allowing us to obtain analytical solutions. A local stability analysis revealed the underlying structure of the critical points of the solutions, resulting in a full bifurcation diagram of the optimal pattern representations. We discovered several interesting properties of dualIB's objective. First, the dualIB retains its structure when expressed in a parametric form. It also optimizes the mean prediction error exponent, thereby improving prediction accuracy with respect to sample size. In addition to dualIB's analytic solutions, we provided a variational dualIB framework that optimizes the functional using deep neural networks. The framework enables a practical implementation of dualIB for real-world datasets. With it, we empirically evaluated its dynamics and validated the theoretical predictions in  modern deep neural networks.

In conclusion, this thesis proposes a new information-theoretic perspective for studying deep neural networks that draws upon the correspondence between deep learning and the IB framework. Our unique perspective can provide a number of benefits, such as attaining a deeper understanding of deep neural networks, explaining their behavior, and improving their performance. At the same time, our study opens up new theoretical and practical  research questions.

%% file: Chapters/letter.tex
\chapter*{Letter of Contribution}
\addcontentsline{toc}{chapter}{Letter of Contribution}
This dissertation includes four manuscripts that summarize Ravid Shwartz-Ziv's research under the supervision of Naftali Tishby. The work on this dissertation was also done in collaboration with Dr. Alex Alemi (Google), Dr. Amichai Painsky (Tel Aviv University) and Zoe Piran (The Hebrew University). They provided valuable guidance and contributed to the writing of the papers as co-authors. The lead author and primary contributor for three of these manuscripts is Ravid Shwartz-Ziv. Ravid Shwartz-Ziv and Zoe Piran each contributed equally to the manuscript ``The Dual Information Bottleneck.'' For this manuscript, Ravid Shwartz-Ziv did the variational dual IB section and the numerical experiments. The manuscripts are listed below. One was published in a peer-reviewed venue, and three are yet to be published. 
\begin{enumerate}
        \item \textbf{``Opening the Black Box of Deep Neural Networks via Information''}. \\Co-author: Naftali Tishby. 2018.
        \item \textbf{``Representation Compression and Generalization in Deep Neural Networks''}. \\Co-authors: Amichai Painsky and Naftali Tishby. 2019.
        \item \textbf{``Information in Infinite Ensembles of Infinitely-Wide Neural Network''}. \\Co-author: Alex Alemi. Published in the Proceedings of the Symposium on Advances in Approximate Bayesian Inference, PMLR, 2020.
        \item \textbf{``The Dual Information Bottleneck''}. \\Co-authors: Zoe Piran and Naftali Tishby. 2020.
\end{enumerate}

%% file: Chapters/Intro.tex
\chapter*{Introduction}
\addcontentsline{toc}{chapter}{1: Introduction}

Learning representations is at the core of many problems in computer vision, natural language processing, cognitive science, and machine learning \citep{bengio2013representation}. A complex data representation is required for classification and prediction since physical parameters, such as location, size, orientation, and intensity, are considered \citep{salakhutdinov2013learning}.  However, it is unclear what constitutes a good representation and how it is related to learning and to the specific problem type.

By combining multiple transformations of simple neurons, deep neural networks (DNNs) can produce a more useful (and, in most cases, more abstract) representation.  Due to their versatility and success across various domains, these systems have gained popularity over the past few years.  The performance of DNNs demonstrates great improvements over conventional machine learning methods in various domains, including images, audio, and text \citep{devlin2018bert, he2016deep, oord2016wavenet}. The latest deep learning models are more complex, and their architectures are becoming more complex with more parameters that need to be optimized. The ResNet-$52$ network, for example, contains about $23$ million parameters optimized over millions of images.

However, the reasons for these performances are only partially understood from a theoretical perspective, and we only have a heuristic understanding of them. It is unclear why deep models perform so well on real-world data and what their key components are.  In addition, current metrics do not provide insight into the internal structure of a network or the quality of its layers. As a result, even if the model is extremely accurate, it is difficult to use it as a basis for further scientific research.
To make these algorithms more effective and improve them, we must understand their underlying dynamic behavior and how they learn representations.

In this thesis, we propose studying DNNs from the perspective of information theory. As an explanation for modern deep learning systems, we propose the information bottleneck (IB) theory. We hope to shed light on their layered structures, generalization capabilities, and learning dynamics through this innovative approach for analyzing DNNs.

To better understand DNNs, the first question is as follows: How can information theory in general and the IB framework, in particular, be used to better understand DNNs? 

Shannon invented the information theory to determine the number of bits needed to transmit a message over a noisy channel. This theory has since been shown to be an invaluable measure of the influence between variables \citep{shannon1948mathematical}. Given two random variables $X$ and $Y$, the mutual information between them measures the divergence of their joint probability distribution $P(x, y)$ from the product of their marginals $P(x)P(y)$ to determine how dependent or independent they are. The notion of mutual information, unlike correlation, can capture nonlinear statistical relationships between variables, strengthening our ability to analyze complex system dynamics \citep{kinney2014equitability}. Although mutual information is an essential quantity in data science, it has historically been challenging to estimate  \citep{paninski2003estimation}. Exact computations are tractable only for a limited number of problems with well-defined probability distributions (e.g., the exponential family). The calculation of mutual information is not possible for finite samples of data or general problems.

This leads us to the following research question: How can we calculate the mutual information for large-scale DNNs?  To derive an exact calculation of information-theoretic quantities and to search for generalization signals, we examined several frameworks and utilized current theoretical developments, including the Neural Tangent Kernel (NTK) framework \citep{lee2019wide}. We obtained tractable calculations of information-theoretic quantities and their bounds for infinite ensembles of infinitely-wide neural networks. Our analysis revealed that the kernels described many quantities in a closed form. Furthermore, we found that the input's compression contributed to the generalization in this model family.

Although the IB framework has its advantages, it also has a few disadvantages, including an inability to preserve the structure of the data and suboptimal performance with finite data. The final research question concerns whether we can derive a new framework that can solve these problems and apply it to DNNs.

Therefore, we developed the dual IB (dualIB), which switches between the terms in the distortion function to solve some of the problems with the IB. A local stability analysis revealed the underlying structure and optimal pattern representations. We discovered that the dualIB retains its structure when expressed in a parametric form. Furthermore, it optimizes the mean prediction error exponent, improving the predictions' accuracy with respect to sample size. dualIB can be applied to real-world datasets using neural networks with the help of a variational framework. Using this framework, we evaluated the dynamics of the dualIB and validated the theoretical predictions.

First, we examine some of the information-theoretic principles upon which this thesis is based. 
The chapter concludes by highlighting the main contributions of the thesis.

\section{Information theory}
In 1948, Shannon's paper established the area of information theory \citep{shannon1948mathematical}. Communication, which refers to sending information with the receiver's intent, is one of the main topics in information theory. Shannon's work served as the basis for quantifying the questions regarding this information. Let us review the results of information theory relevant to this thesis. In the present work, we use the following notation: Upper-case letters denote random variables (e.g., $X$), calligraphic letters denote their support (e.g., $\mathcal X$), and lower-case letters denote specific realizations (e.g., $x$). 

\subsection{Theoretical information quantities}
As a formal concept, information is a function of several basic measures based solely on the probability distribution of random variables without any particular assumptions.

Consider $X \in \mathcal{X}$ and $Y \in \mathcal{Y}$ as random variables
with the joint distribution function of $p(x, y)$.  Our first definition of information is entropy, which takes into account the amount of uncertainty involved with a given distribution.
\begin{definition}{\textbf{Entropy}}

The entropy of $X\sim p(x)$ is defined as 
$$H(X)=-\sum_{x \in \mathcal{X}} p(x)\log(p(x))$$
\end{definition}

Shannon axiomatically derived this definition by defining three intuitive properties of uncertainty measures (continuity, additivity, and monotonicity). Entropy is the only function that satisfies these requirements. Intuitively, entropy reflects the minimum description length of $X$ because it is (roughly) the minimal number of bits or binary questions needed to determine the exact value of $X$.

If the distribution is uniform, then the entropy reaches its maximum. If it is deterministic, then it is zero. As $H(X)$ is concave, the more $X$ strays away from being completely unpredictable, the more it is punished by $H(X)$.

Our next informational measure is the Kullback-Leibler (KL) divergence, also known as relative entropy, which measures the divergence between two distributions.
\begin{definition}{\textbf{Kullback Leibler Divergence}}

The KL divergence between two distributions $p(x)$ and $q(x)$ is defined as
$$D[p||q]=\sum_{x\in \mathcal{X}} p(x)\log\left(\frac{p(x)}{q(x)}\right)$$
\end{definition}
In the case where $p(x) = q(x)$ for some $x$, then it will not contribute to $D[p||q]$. If for all $x$, $p(x) = q(x)$, then $D(p||q)=0$. When $p$ and $q$ have a low Kullback-Leibler distance, they are similar. Conversely, a high distance indicates they are dissimilar.

It can be shown that $D[p|q] \geq 0$ and that equality holds if and only if $p = q$ almost everywhere.  Note that the KL divergence is not a metric because it is not symmetric and does not obey the triangle inequality.
However, it is still useful to think of it as the natural "distance" between distributions for several reasons. First, notice that $D[p||q]$ is the expected log-likelihood ratio between $p(x)$ and $q(x)$. Thus, it controls the discriminability between these two distributions when $p$ is the true underlying distribution of $X$. Secondly, the KL divergence reflects the difference between the minimal description length of $X \sim p(x)$ and the description length if $q(x)$ is used instead of $p(x)$. To see this, notice that $D[p|q] =\sum_x p(x)log\frac{1}{q(x)}-H(X)$, and recall
that $H(X)$ is the minimal description length. Third, Pinsker's inequality implies that the KL divergence upper bounds the $L1$ distance between $p$ and $q$. Therefore, it also bounds any $L_\rho$ distance for $\rho\geq1$.

The last piece of information we will present is mutual information.
\begin{definition}{\textbf{Mutual Information}}

Let $(X,Y) \sim p(x,y)$, and let $p(x)$ and $p(y)$ be their marginal distributions, respectively. The mutual information between $X$ and $Y$ is defined as $$I(X;Y)=\sum_{x\in \mathcal{X}, y \in \mathcal{Y}}p(x,y)\log\frac{p(x,y)}{p(x)p(y)}$$
\end{definition}

Mutual information is a useful method for measuring statistical dependence between variables and has many beneficial properties. For example, it remains invariant under bijective transformations. Its predecessor, called the transmission rate, was first introduced by Shannon in 1948 \citep{shannon1948mathematical} for a communication system. Mutual information can be axiomatically derived as a function satisfying several natural `informativeness' conditions \citep{cover1999elements}.

Mutual information could be understood by thinking of the $KL$ divergence between $p(x,y)$ and the hypothetical joint distribution if $X$ and $Y$ were independent.
$$I(X;Y) = D[p(x,y)|| p(x)p(y)]$$
It thus follows that $I(X; Y ) \geq 0$ and $I(X; Y)=0$ if and only if $X$ and $Y$ are independent. Alternatively, mutual information can be interpreted as reducing uncertainty regarding one variable due to the knowledge of the other variable.
$$I(X;Y)=H(X)-H(X|Y)=H(Y)-H(Y|X)$$

Since $H(X|Y),H(Y |X) \geq 0$, it holds that
$I(X;Y) \leq \min\{H(X),H(Y)\}$. If $H(X|Y)=0$, that is, $X$ is completely known given $Y$, then $I(X;Y)=H(X)$. As a special case, it holds that $I(X;X)=H(X)$. In this sense, mutual information is more general than entropy, whereas entropy is sometimes referred to as self-information.

One of the important properties of mutual information is the data processing inequality (DPI), which implies that the information about $X$ contained in $Y$ cannot be increased by processing $Y$. Formally, we say that $(X,Y,Z)$ form a Markov chain if their joint distribution can be decomposed as $p(x,y,z) = p(x,y)p(z|y)$, $Z$ is a function of $Y$ and given $Y$, it is independent of $X$.  Then, the DPI implies 
\begin{equation}
I\left(X;Y\right)\ge I(X;Z) \label{eq:DPI-1}
\end{equation} for any 3 variables that form a Markov chain $X\rightarrow Y\rightarrow Z$.  

Mutual information is also invariant with respect to invertible transformations: 
\begin{equation}
I\left(X;Y\right)=I\left(\psi(X);\phi(Y))\right)\label{eqn:invariance-1}
\end{equation}
 for any invertible functions $\phi$ and $\psi$. 
\section{Representation learning}
Machine learning is a field of Artificial Intelligence (AI) that automatically learns and improves from experience.  These models use an internal representation of the input based on prior observations or data records to make better decisions. Learning representations lie at the core of many computer vision problems, natural language processing, cognitive science, and machine learning \citep{bengio2013representation}. Even so, the question of what constitutes a good representation or the relationship between representation, learning process, and specific features of a problem remains unanswered.

David Marr \citep{marr2010vision} defined representation as a formal system for making explicit particular entities and types of information that an algorithm can use to process information. In this thesis, we focus on the second part -- the idea that algorithms operate on representations. Representation learning refers to learning structures in the data. This learning makes it easier to extract useful information to classify and make predictions based on the data \citep{Goodfellow-et-al-2016-Book}. The question of what a good representation is has many answers. For probabilistic models, a good representation often captures the posterior distribution of the underlying explanatory factors for the observed input \citep{natureDeepLeraning}.
In their AI-tasks, Bengio and LeCun \citep{BengioAndLecunMIT} introduced the concept of complex but highly structured dependencies.
In most cases, the tasks require transforming a high-dimensional input structure into a low-dimensional output or learning low-level representations. Accordingly, most of the input's entropy is not relevant for the output, and it is difficult to extract the input's relevant features \citep{TishbyZ15}. 

\subsection{Large deviation theory and the Sanov theorem}
Statisticians, information theorists, and machine learning scientists share concepts for defining and extracting the optimal representation from observations. How would we characterize the optimal representation of the random input variable $X$?
In general, we would like to have a representation that achieves high values for some reward function. In addition, we aim to learn efficiently these representations by using an empirical sample of the (unknown) joint distribution, which will enable us to generalize to unknown datasets.

As we will see in the following section, the large deviation theory can be used to illustrate the idea that constraining information with the representation will produce a representation that is more likely to generalize. This approach is derived from the type analysis and the Sanov theorem \citep{cover1999elements}. It is presented here for an independent and identically distributed (i.i.d.) process, although a Markov type analysis is also available \citep{csiszar1987conditional}. 

With the large deviation theory, we can understand the probability of rare events, i.e., events with an exponentially small probability.  Type analysis is applied to analyze the most likely samples from a given probability distribution following a constraint. Sanov's theorem identifies the rate function for large deviations of the empirical measure of a sequence of i.i.d. random variables.  The type definition and the Sanov Theorem for i.i.d. processes \citep{cover1999elements} are presented here with an adaptation of the annotation to our needs.

Let $T_1,T_2,\ldots$ be i.i.d. random variables with values in a finite set $A = \{\alpha_1,\ldots ,\alpha_r\}$ and with a distribution $P$. Denote by $\mathcal{P}$ the set of probabilities on $\{\alpha_1,\ldots,\alpha_r \}$.  Let $\hat P_n$ be the empirical distribution of $T$:

\[\hat P_n = \frac{1}{n}\sum_{k=1}^n \delta_{T_k}.\]

The law of large numbers states that $\hat P_n \rightarrow P$ almost surely. To define a rare event, we fix $E \subseteq	 \mathcal{P}$ that does not contain $P$. We are interested in the behavior of probabilities of the form $\mathcal{P}[\hat P_n \in E]$, as $n \rightarrow \infty$.

 It is a trivial observation that each possible value $\{\alpha_1,\ldots,\alpha_r \}$ must appear an integer number of times among the samples $\{T_1,\ldots,T_n\}$. This implies, however, that the empirical measure $\hat P_n$ cannot take arbitrary values.
\begin{definition}
Type $P_t$ of the sequence $t = t_1, t_2,\dots ,t_n$ is the relative proportion of occurrences of each value. That is,  $P_t(\alpha) =\frac {N(\alpha | t)} {n}$ where $N(\alpha|x)$ is the number of times $\alpha$  occurs in the sequence t.
\end{definition}

By definition, each type contains only information about how often each value shows up in the sample, discarding the order in which they appear.
The importance of it is that large numbers of sequences are associated with each type, and each type is associated with a single reward value $\sum_{\alpha\in A} P(\alpha)R(\alpha)=\bar{R}$.
It is always the case that $\hat P_n \in \mathcal{P}_n$, where we define $\mathcal{P}_n$ as the set of types associated with sequences of samples with length $n$.

\begin{theorem}{\textbf{Sanov}}
Let $t_1, t_2, \dots t_{n-1}$ be i.i.d. sequence from  $P(t)$. Let $E\subseteq	 \mathcal{P}_n$ be a set of types that is equal to the closure of its interior.
\[\lim_{n \to \infty} \frac{1}{n} \log P^n[\hat P_n \in E] = -D_{KL}(P^\star| P^n)\]
where $P^\star=arg \min_{Q \in E\cap \mathcal{P}_n} D_{KL}(Q | P^n)$.
\end{theorem}

For our needs, we define the set $E$ to be all the types (and the
sequences that are associated with them) that achieve a reward above the desired threshold $\theta$; i.e., $E=\sum\{P_{\alpha \in A} P(\alpha)R(\alpha) \geq \bar{R}\}$.

Sanov's theorem gives us an important insight: the most likely sequence to be selected as a type that is the closest from the $D_{KL}$ perspective to the distribution $P(t)$ (denoted as $P^\star$). This insight can be formalized as a trade-off, where on one side, we have the reward $\bar{R}$, and on the other, we have $D_{KL}$. The selected solution is governed by a parameter $\beta$, which is the Lagrange multiplier coefficient. Given the nature of the types, they can be visualized and represented as points of a simplex whose axes are the probabilities of each symbol to appear. In this space, the two crucial points are the maximal reward point (a deterministic point at the edge of the simplex) and $P$. The trade-off parameter $\beta$ defines a set of distributions that form a line between these points. These are known as geodesic lines. No two points along this line have the same $\beta$, and therefore they do not have the same $D_{KL}$ value.
\subsection{Minimal sufficient statistic}
An alternative definition of what constitutes a good representation is based on minimal sufficient statistics.
\begin{definition}
Let $(X,Y)\sim P(X,Y)$ . Let $T:=t(X)$ , where $t$ is a deterministic function. We call T a sufficient statistic of $X$ for $Y$ if $Y-T-X$ forms a Markov chain.
\end{definition}

Therefore, a sufficient statistic captures all the information about $Y$ that is available in $X$. The following theorem states this property:
\begin{theorem}
\cite{cover1999elements}
Let $T$ be a probabilistic function of
$X$. Then, $T$ is a sufficient statistic for $Y$ if and
only if (iff ) $I(T(X);Y)=I(X;Y)$
\end{theorem}
As we can see, the sufficiency definition includes the trivial identity
statistic $T = X$. Such statistics accomplish nothing since all they do is "copy" rather than "extract" important information. Therefore, it is necessary to prevent statistics from using observations in an inefficient manner.

To address this issue, the concept of minimal sufficient statistics was introduced:
\begin{definition}(Minimal sufficient statistic (MSS))
A sufficient statistic $T$ is minimal if for any other sufficient statistic $S$ , there exists a function $f$ such that $T=f(S)$ almost surely (a.s.).
\end{definition}

MMS are the simplest sufficient statistics and induce the coarsest sufficient partition on $X$. MSS try to group the values of $X$ into as few partitions as possible without sacrificing any information.
In addition, MSS can be shown to be the statistic with all the available information about $Y$ while retaining as little information about $X$ as possible. Generally, sufficient statistics are restricted, in the sense that their dimension always depends on the sample size, unless the data comes from an exponential family distribution \citep{koopman1936distributions}.  
\subsection{The Information Bottleneck}
Since exact minimal sufficient statistics only exist for special
distributions, \cite{tishby2000information} relaxed this optimization problem in two ways: (i) allowing the map to be stochastic, defined as an encoder $p(T|X)$, and (ii) allowing
the map to capture only as much as possible of $I(X; Y)$, but not
necessarily all of it. They introduced the Information Bottleneck (IB) as a principled approach to extract relevant information from observed signals related to a target. For a random variable $x\in X$, this framework finds the best trade-off between the accuracy and the complexity related to another random variable $y\in Y$ with a joint distribution. The IB  has been used in several fields, including neuroscience \citep{buesing2010spiking,palmer2015predictive}, slow feature analysis \citep{turner2007maximum}, speech recognition \citep{hecht2009speaker} and deep learning \citep{TishbyZ15, shwartz2017opening, alemi2016deep}.

Let $X$ be an input random variable, $Y$ a target variable, and $p(x,y)$ their joint distribution. A representation $T$ is a stochastic function of $X$ defined by a mapping $p(t|x)$. This mapping can be viewed as transforming $X \sim P(X)$ into a representation of $T\sim P(T):=\int P_{T \mid X}\brk*{\cdot \mid x}dP_X(x)$ in the $\mathcal{T}$ space.

The triple $Y-X-T$ forms a Markov chain in that order w.r.t. the joint probability measure $P_{X,Y,T}=P_{X,Y}P_{T \mid X}$ and the mutual information terms $I(X; T)$ and $I(Y; T)$.

As part of the IB framework, we aim to find a representation $P(T  \mid X)$ that extracts as much information as possible about $Y$ (high performance), while compressing $X$ maximally (keeping $I(X; T)$ small).  We could also interpret it as extracting only the relevant information that $X$ contains about $Y$.

The DPI implies $I(Y; T)\leq I(X; Y)$, so the compressed representation $T$ cannot convey more information than the original signal. As a result, there is a tradeoff between compressed representation and the preservation of relevant information about $Y$.  The construction of an efficient representation variable is characterized by its encoder and decoder distributions, $P(T|X)$ and $P(Y|T)$, respectively. The efficient representation of $X$ means minimizing the complexity of the representation $I\left(T; X\right)$
while maximizing $I\left(T; Y\right)$.  Formally, the IB optimization involves minimizing the following objective function:
\begin{align}
\label{eq:IB}
    \mathcal{F}\brk[s]*{p_{\beta}\brk*{t \mid x}; p_{\beta}\brk*{y \mid t }}  =I(X;T) - \beta I(Y;T)~,
\end{align}
where $\beta$ is a parameter that controls the trade-off between the complexity of $T$ and the amount of relevant information it preserves. Intuitively,  we pass the information that $X$ contains about $Y$ through a ``bottleneck'' via the representation $T$.
It can be shown that
\begin{align}
    I(T:Y)=I(X:Y)-\mathbb{E}_{x\sim p(x), t\sim p(t|x)}\left[D\left[p(y|x)||p(y|t)\right]\right]
\end{align}

IB representations can be found using the IB method, a variant of the Blatu Arimoto algorithm \citep{blaArimo}. 
For any $\beta$, the conditions for a stationary point of equation \ref{eq:IB}, can be expressed via the following self-consistent equations \citep{tishby2000information}:
\begin{align} \label{eq:IB_opt}
    \begin{cases}
    \brk*{i}\ &p_{\beta}\brk*{t \mid x} = \frac{p_{\beta}\brk*{t} }{Z_{t \mid \rvx}\brk*{x;\beta} } e^{-\beta D\brk[s]*{p\brk*{y\mid x} \|  p_{\beta} \brk*{y \mid t}  }} \\
    \brk*{ii}\ &p_{\beta}\brk*{t} = \sum_{x} p_{\beta}\brk*{t \mid x} p\brk*{x} \\
    \brk*{iii}\ &p_{\beta}\brk*{y \mid t } = \sum_{x}   p\brk*{y \mid x}{p_{\beta}\brk*{x \mid t}} 
    \end{cases}
,\end{align}
where $Z_{t \mid \rvx}\brk*{x;\beta}$ is a normalization term.
If $X$, $Y$, and $T$ take values in finite sets, and $P(X,Y)$ is known, then alternating iterations of \ref{eq:IB_opt} locally converge to a solution, for any initial $P(T \mid X)$.

If we denote $I_{X}^{\beta}=I_{\beta}\left(T;X\right)$ and $I_{Y}^{\beta}=I_{\beta}\left(T;Y\right)$, the optimal information curve is defined as the optimal values of the trade-off $\left(I_{X}^{\beta},I_{Y}^{\beta}\right)$ for some $\beta$.
The information plane is the two-dimensional plane in which the IB curve resides. The equations in \ref{eq:IB_opt} are satisfied along the information curve, which separates the feasible and unfeasible regions of the information plane by a monotonic concave line.
\section{Deep neural networks}

In 1958, Frank Rosenblatt developed the perceptron algorithm, the first artificial neural network component \citep{rosenblatt1958perceptron}. The system was designed to mimic the way the human brain processes visual data and identifies recognizable objects. It was extended to pattern recognition in the late 1980s.

Deep learning can perform hierarchical learning of the data by applying nonlinear transformations, which distinguishes it from traditional neural networks. The data are cumulatively passed across multiple layers, which may be fully connected or partially connected. DNNs are multilayer structures constructed by processing units that are called neurons. Each neuron's activation involves the weighted summation of neuron inputs from the previous layer, followed by the transfer function's operation. Interconnected layers of these basic computing blocks are used to build complex deep learning architectures \citep{DBLP:journals/corr/Schmidhuber14}. Using this structure, DNNs can learn hierarchically sophisticated features directly from raw data without manually constructing them \citep{salakhutdinov2013learning}. Deep architectures are often more challenging to effectively train. They bring, however, two significant advantages: (1) they promote the reuse of features, and (2) they can potentially lead to progressively more abstract features at higher layers of representation, which hopefully make it easier to separate the explanatory factors in the data \citep{erhan2009difficulty}.

Although deep architectures have long existed, the term ``deep learning'' was first used in $2006$ by \cite{hinton2006fast}. This work showed that a multilayer feedforward neural network could be more efficient by applying pretraining of one layer at a time and considering each layer as an unsupervised Restricted Boltzmann Machine (RBM) by using supervised backpropagation for finetuning. In 2007, \cite{bengio2007greedy} developed the Stack AutoEncoder (SAE), which comprises a deep architecture of many AutoeEcoders (AEs). 

In 2012, \cite{krizhevsky2012imagenet} proposed the AlexNet architecture, which won the ImageNet challenge. This was a significant breakthrough in artificial neural networks. They proposed a deep convolutional neural network with nine layers and implemented it over GPUs for the first time.  Several extensions of the vanilla AlexNet network have been developed since then, including deeper convolutional nets as proposed by \cite{simonyan2014very} and residual connections as demonstrated by \cite{ren2015faster}. DNNs have demonstrated their ability to improve state-of-the-art results in a wide range of machine learning tasks over the past few years. In many areas, from visual object recognition to speech recognition and genomics to drug discovery, they work well and enhance state-of-the-art results dramatically \citep{DBLP:journals/corr/abs-1303-5778,DBLP:journals/corr/ZhangL15,DBLP:journals/corr/abs-1207-0580,DBLP:journals/corr/HeZRS15,natureDeepLeraning}.
\section{Information Bottleneck and DNNs}

Even though DNN has been highly successful, little is known about its reasons for success, and no underlying principles have driven its development. DNNs are often considered black boxes, where the interpretation of predictions and reliability are still open questions. Additionally, their internal structure and the optimization process are still not fully understood. To use these algorithms more efficiently and improve them, we need to understand their dynamic behavior and their ability to learn representations.

In the literature, two areas of work involve DNNs and the IB. One uses the IB concept to analyze DNNs, while the other uses the IB  to improve the learning algorithm for DNNs. The rest of this section is divided into these categories.

\subsection{Information Bottleneck as optimization objective}
Recently, the IB framework was explored as an objective for deep learning. This concept was achieved by optimizing the IB Lagrangian using a variational bound \citep{alemi2016deep,olchinsky2019nonlinear}.

The variational information bottleneck (VIB) approach presented in \citep{alemi2016deep} used DNNs to parameterize the IB model. A variational approximation of the objective is parametrized using DNN, and an efficient training algorithm is suggested for obtaining a stochastic network that maps inputs to randomized representations.

Given a DNN, denote its output representation as $T$, a randomized mapping operating on the input feature $X$, where the corresponding label is $Y$. The encoding of $X$ into $T$ is defined through a conditional probability distribution, which is parametrized as $p^\theta_{T\mid X}$. The VIB  optimization objective is
\begin{equation}
\mathcal {L}_{\beta }^{\mathsf {\left ({VIB}\right)}}\left ({\theta }\right) \mathrel {\mathrel {\mathop:}\hspace {-0.0672em}=}\max _{\theta \in \Theta }I\big (T^{\left ({\theta }\right)};Y\big)-\beta I\big (X;T^{\left ({\theta }\right)}\big). \label{eq:vib}
\end{equation}

However, since the data distributions $P_{X,Y}$ and $P_{T \mid X}$ are unknown, a direct optimization of Equation \ref{eq:vib} is intractable.
To overcome it, \cite{alemi2016deep} suggested to lower bound Equation \ref{eq:vib} in a form that we can optimize:
\begin{equation*} \mathbb {E}_{P^{\left ({\theta }\right)}_{Y,T}}\left [{\log Q^{\left ({\phi }\right)}_{Y|T}\left ({Y\Big |T^{\left ({\theta }\right)}}\right)}\right] -\beta \mathsf {D}_{\mathsf {KL}} \left ({P^{\left ({\theta }\right)}_{T|X}\Big \|P_{T}^{\left ({\theta }\right)}}\right), \tag{18}\end{equation*}

In this case, we parametrize the decoder  $Q^{phi}_{(Y \mid T}$  by a DNN. However, the main difficulty of this optimization is intractable marginal distribution $P^\theta_T$. To circumvent this, we take $p_{T\mid X}$ and $p_T$ distributions with a closed-form KL solution. In this case, we treat the network output's encoder as the parameters of $P_{T\mid X}$.

Using the reparametrization trick from \cite{kingma2013auto} and replacing $P(X,Y)$ with its empirical proxy, the loss function of IB can be approximated as
\begin{align*} \hat { \mathcal {L}}_{\beta }^{\left ({\mathsf {VIB}}\right)}\left ({\theta,\phi, \mathcal {D}_{n}}\right)\mathrel {\mathrel {\mathop:}\hspace {-0.0672em}=}&\frac {1}{n}\sum _{i=1}^{n} \mathbb {E}\left [{-\log Q^{\left ({\phi }\right)}_{Y|T}\left ({y_{n}\big | f\left ({x_{n},T}\right)}\right)}\right] \\&+ \beta \mathsf {D}_{\mathsf {KL}} \left ({P^{\left ({\theta }\right)}_{T|X}\left ({\cdot |x_{n}}\right)\Big \| R_{T}\left ({\cdot }\right)}\right). \tag{19}\end{align*}

$T$ is an auxiliary noise variable, and the expectation is w.r.t to its law. 
Note that if $t$ represents the output of a DNN, the first term is cross-entropy, which is derived from the loss function common in deep learning. The second term acts as a regularization term that penalizes the dependence of $X$ on $T$, thus encouraging compression. An unbiased estimate of the true variational lower bound can be obtained through standard stochastic gradient-based methods by calculating the estimator's gradient.

Since \cite{alemi2016deep}, various extensions have been developed, demonstrating promising attributes \citep{strouse2017deterministic, elad2019direct, kolchinsky2019nonlinear}. Recently, the conditional entropy bottleneck (CEB) \citep{fischer2020ceb} has been proposed. The CEB provides variational optimizing bounds on $I(Y ;T)$ and $I(X; T)$ using a variational decoder $q(y |x)$, variational conditional marginal, $q(t |  y)$, and a variational encoder, $p(t | x)$, all implemented by DNNs. \cite{alemi2016deep} also showed that the variational auto-encoder (VAE) \citep{kingma2013auto} can be considered as an estimation for a special case of IB when $Y=X$, $\beta=1$ and the prior distribution function is fixed.
\subsection{Information Bottleneck theory for deep learning}
\cite{TishbyZ15} proposed a theoretical framework to analyze DNNs based on the principle of the IB. They formulated the ultimate goal of the network as a trade-off between compression and prediction. An optimal point on the \textit{information curve} exists for each layer where this trade-off can be addressed effectively. The layer's network structure forms a Markov chain, where each layer processes inputs from the previous layer. Due to DPI, any loss of information about $Y$ in one layer is not recoverable in higher layers.

Formally, define $t^{i}\in T_{i}$ as the compressed
representation of $x$ in the i-th layer and $\hat{y}$ as the network's output. $T_{i}$ is uniquely mapped to a single point in the information-plane with coordinates $\left(I(X;T_{i}),I(T_{i};Y)\right)$. We map a network with $k$ layers to $K$ monotonic connected points in the plane. For any $k\ge j$, it holds that 

\begin{equation}
\begin{array}{c}
H\left(X\right)\ge I\left(X;T_{j}\right)\ge I\left(X;T_{k}\right)\ge I\left(X;\hat{Y}\right)\\
I\left(Y;X\right)\ge I\left(Y;T_{j}\right)\ge I\left(Y;T_{k}\right)\ge I\left(Y;\hat{Y}\right)\\
I\left(X;T_{j}\right)\geq I\left(Y;T_{j}\right),
\end{array}\label{eq:dpi}
\end{equation}

and the equality in the second line is achieved IFF each layer is a sufficient statistic of its input. Using this framework, each layer should extract a compact representation while preserving the relevant information.

By successively decreasing $\beta$, we shift the network
from low representations and construct higher and more abstract
ones. On the one hand, $I\left(Y; T_{i}\right)$ measures how much of the predictive
features in $ X $ for $ Y $ is captured by the layer and can view it as
an upper bound of the layer's quality. On the other hand, $I\left(X;T_{i}\right)$ can be interpreted as the complexity of the layer. As a result, we can assess the performance of DNNs not just in terms of output, as we do when we use other measures of error when evaluating DNNs.

Based on the theoretical IB limit and the limitations imposed by the DPI on the information flow between layers, we can get a good sense of each layer's optimality in the network. With each successive layer, the IB distortion level increases, but it also compresses the inputs, hopefully removing only non-relevant information \citep{TishbyZ15}.

\begin{figure}[ht]
.9\includegraphics[scale=0.5]{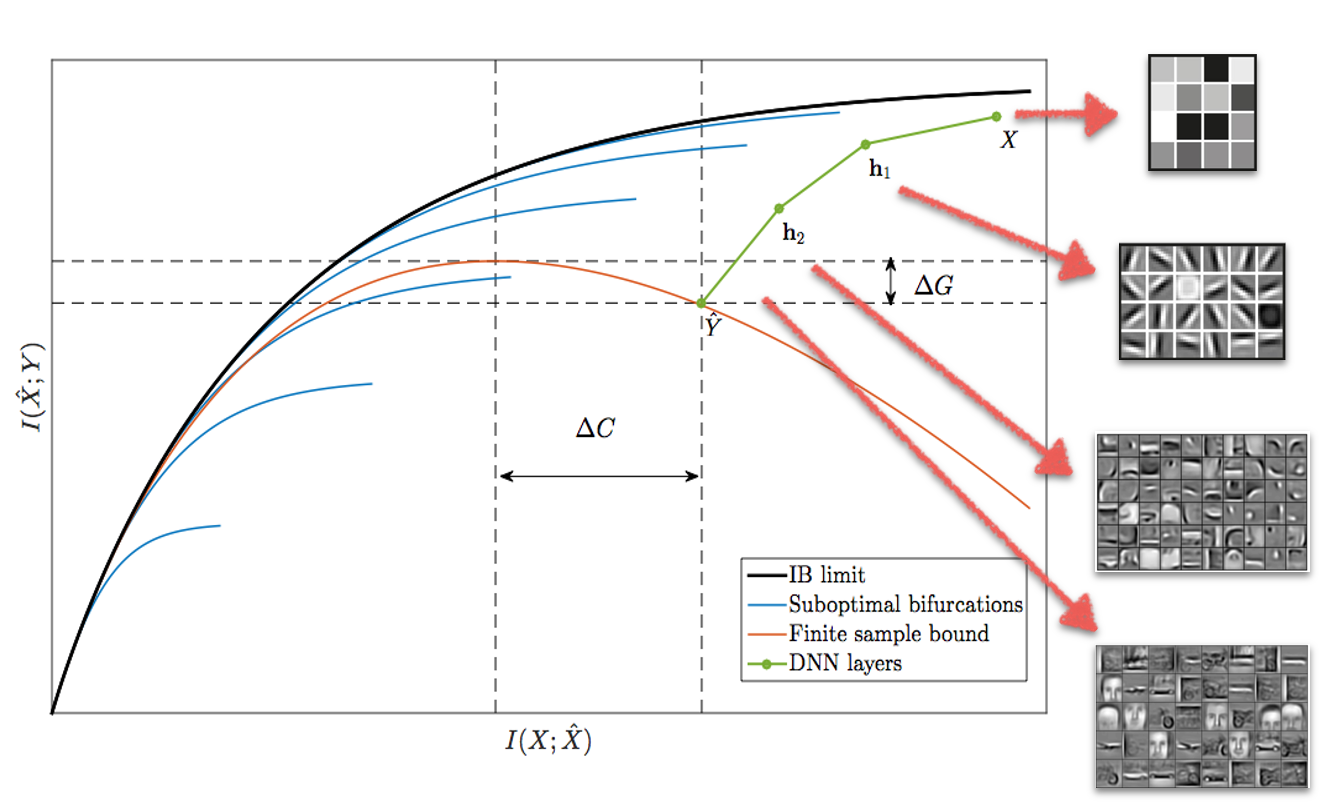}

\caption{A qualitative information plane from \cite{TishbyZ15},
a hypothesized path of the layers in a typical DNN (green line) on
the training data. The black line is the optimal achievable IB limit,
and the blue lines are sub-optimal IB bifurcations. The red line corresponds to the upper bound on the out-of-sample IB distortion when training from a finite sample. We want to shift the green DNN layers
closer to the optimal curve to obtain lower complexity and better
generalization by shifting the last layer to the maximum of the red curve.}
\end{figure}
\subsection{Information bounds on the generalization gap}
Based on the statistical learning theory, models with many parameters tend to overfit by modeling learned data too accurately, which reduces their ability to generalize to new data \citep{boucheron2005theory}. In practice, however, we see that DNNs have a minimal generalization gap between training and testing.  Recently, there has been much interest in understanding implicit regularization. Researchers' findings indicate that network size is not the most important factor involved in the process of learning multilayer feedforward networks and that some unknown factors are important  \citep{neyshabur2014search}. Moreover, conventional statistical learning theories, such as Rademacher complexity, VC-dimension, and uniform stability \citep{stavac, bartlett2002rademacher, bousquet2002stability}, do not explain all of the unexpected results of numerical experiments. According to \cite{zhang2016understanding}, regularization plays a unique role in deep learning that differs from empirical risk minimization. However, convincing experiments have indicated that the generalization gap can be reduced even without explicit regularization. In recent years, several works have demonstrated that the mutual information between the training inputs and the inferred parameters provides a concise bound on the generalization gap \citep{ xu2017information,pensia2018generalization,NIPS2019_9282,NIPS2018_7954, russo2016controlling, steinke2020reasoning, achille2019information}. \cite{achille2018emergence} investigated how using an IB objective on network parameters (rather than the representations) could avoid overfitting while enforcing invariant representations.
\section{Infinitely-wide neural networks}
Neural tangent kernel (NTK) is a powerful theoretical tool for modeling neural networks. \cite{lee2019wide} showed that if neural network training is modeled as gradient flow, the training trajectory can be modeled by ordinary differential equations (ODEs). Equations like these represent a finite-width tangent kernel encoded by the network architecture and its current time-dependent weights. Moreover, they showed that if one scales the learning rate per layer appropriately ("NTK parametrization") and lets the width tend to infinity, this kernel converges to the infinite-width NTK. By being independent of weights and staying constant throughout training, this model simplifies the ODE in this limit. They showed that for $l2$ loss, the predictor at convergence would be produced by a kernel regression using an infinite-width NTK. Importantly, the NTK depends only on the architecture of the network and is not learned.

Infinitely-wide neural networks behave as they are linear in their parameters \citep{lee2019wide}:
\begin{equation}
    z(x, \theta) = z_0(x) + \frac{\partial z_0}{\partial \theta} (\theta - \theta_0) \quad z_0(x) \equiv z(x, \theta_0).
\end{equation}
This makes them particularly analytically tractable.  
An infinitely-wide neural network, trained by gradient flow to minimize the squared loss, admits a closed form expression for the evolution of its predictions as a function of time:
\begin{equation}
    z(x, \tau) =  z_0(x) - \Theta(x, \X) \Theta^{-1} \left( I - e^{-\tau \Theta} \right)(z_0(\X) - \Y) .
\end{equation}
Here, $z$ denotes the output of our neural network acting on the input $x$.  $\tau$ is a dimensionless representation of the time of our training process.  $\X$ denotes the whole training set of examples, with their targets $\Y$; $z_0(x) \equiv z(x,\tau=0)$ denotes the neural networks
output at initialization.
The evolution is governed by $\Theta$ (the NTK). For a finite width network, the NTK corresponds to $JJ^T$, the neural network
gradients' gram matrix.  As the network's width increases to infinity, this kernel converges in probability to a fixed value. Tractable ways to calculate the exact infinite-width kernel for broad classes of neural networks are available \citep{lee2019wide}.
The shorthand $\Theta$ denotes the kernel function evaluated on the train data ($\Theta \equiv \Theta(\X, \X)$).

Observe that infinitely-wide networks trained with gradient flow and squared loss behave as affine transformations of their initial predictions.
As a result, for an infinite ensemble of such networks, if the initial weight configurations are drawn from a Gaussian distribution, the law of large numbers guarantees the distribution of the output conditioned on the input is Gaussian. As the evolution is an affine transformation of the initial predictions, the predictions remain Gaussian throughout.
\begin{align}
    p(z|x) &\sim \N(\mu(x,\tau), \Sigma(x,\tau)) 
    \label{eqn:rep} \\
    \mu(x,\tau) &= \Theta(x, \X) \Theta^{-1} \left( I - e^{-\tau \Theta} \right) \Y\\
    \Sigma(x,\tau) &=  \K(x, x) + \Theta(x, \X) \Theta^{-1} \left(I - e^{-\tau \Theta} \right) \left( \K \Theta^{-1} \left( I - e^{-\tau \Theta} \right) \Theta(\X, x) - 2 \K (\X, x) \right).
\end{align}
Here, $\K$ denotes another kernel, the \emph{neural network gaussian process} kernel (NNGP). For a finite width network, the NNGP corresponds to the expected gram matrix of the output- $\mathbb{E}\left[ z z^T \right]$. In the infinite width limit, this concentrates to a fixed value.
Just as for the NTK, the NNGP can be tractably computed \citep{lee2019wide} and should be considered only a function of the neural network architecture. 
These results, which give us a conditional posterior distribution for each time step, enable us to create a powerful model family to investigate DNNs. These tractable distributions can be used to derive many intractable information-theoretic quantities.

\section{The impact of the work}
\label{sec:work_impact}
In summary, the IB theory offers a new perspective on DNNs and their capability to learn meaningful representations. This theory suggests studying the system by grouping each layer in the network into two information pairs, one with input and another with output:  $(I(X;T_\ell),I(T_\ell; Y))$.

Following this study, which provided a theoretical foundation for understanding deep learning, several studies have offered further explorations of DNNs using information theory tools. We now discuss some of these works and whether they support or challenge our claims. A partial list includes \cite{achille2018emergence, saxe2019information, yu2020understanding, cheng2018evaluating, goldfeld2018estimating,wickstrom2019information, amjad2019learning, goldfeld2020convergence, cvitkovic2019minimal}.

Following this study, several researchers have analyzed the information plane using different estimation mechanisms in various DNN datasets, architectures, and activation functions. They found conflicting results:  The authors of \cite{saxe2019information} did not observe compression in DNNs with ReLU activation functions. However, according to \cite{chelombiev2019adaptive}, compression can occur earlier in training or later in training, depending on how the DNN parameters are initialized. 
\cite{goldfeld2018estimating}  developed a noisy DNNs framework with a rigorous estimator for $I(X;T)$. Using this estimator, they observed the input's compression in various models. The authors related the noisy DNN to an information-theoretic communication problem, demonstrating that compression is driven by the progressive clustering of inputs belonging to the same class. The study clarified the geometric effects of mutual information compression during training. Other methods that have been proposed for estimating mutual information in DNNs are generative decoder networks \citep{darlow2020information,nash2018inverting}, the mutual information neural estimator (MINE) \citep{elad2019direct}, ensemble dependency graph estimator (EDGE) \citep{noshad2018scalable}, adaptive approaches for density estimation \citep{chelombiev2019adaptive}, and noisy sounding entropy estimator \citep{goldfeld2018estimating}, and more.  
For a detailed review of these works, see \cite{geiger2020information}. 

In another line of research, our idea was used to analyze networks using information with the weights \citep{achille2018emergence, achille2018critical, achille2018information}. It has been demonstrated that flat minima, which have better generalization properties, bound the information with the weights, which bounds the information with the activations. In \cite{achille2018critical}, they used Fisher information on the weights to illustrate the two stages of learning that DNNs go through:  increasing the information followed by progressively decreasing the information. By compressing the information with the weights, the network also compresses the information with the activations.

As shown in \cite{achille2018information}, minimizing a stochastic network using an approximation of the compression term as a regularizer is equivalent to minimizing cross-entropy over deterministic DNNs with multiplicative noise (information dropout). In addition, they found that Bernoulli noise is a special case that leads to the dropout method.  \cite{elad2019direct} shows that the binned information can be interpreted as a weight decay penalty, which is typical of DNN training.

Furthermore, many competing information objectives have been proposed for training DNNs based on the IB principle, which can be difficult to compute for high dimensions. \cite{kirsch2020unpacking} examined these quantities and compared, unified, and related them to surrogate objectives that are more easily optimized. The unifying view provided insights into IB training limitations and demonstrated how to avoid the pathological behavior of IB objectives. In addition, they discussed how simple objectives, based on this intuition, could capture many desirable features of IB algorithms while also scaling to complex deep learning problems. Furthermore, they explored how applying practical constraints to a neural network's expressivity can provide new insights into measuring compression and possibly improving regularization.

\section{Overview and main contributions}
The purpose of our thesis is to provide an explanation of deep learning by using an information-theoretic framework, supported by empirical evidence, and overcoming some of the shortcomings of IB.  We have outlined the major results and contributions of the thesis in three sections below:
\begin{itemize}
    \item Part 1 -- \textbf{Opening the Black Box of Deep Neural Networks} -- The first contribution of this thesis is a presentation of an information-based theory for DNNs. Combining two papers shows that DNNs can learn to optimize the mutual information that each layer preserves on the input and output variables, resulting in a trade-off between compression and prediction. We present a theoretical and numerical analysis of DNNs in the information plane and explain how the SGD algorithm follows the information bottleneck trade-off principle. 
SGD achieves this optimal bound by compressing each layer to a maximum conditional entropy state subject to the constraints of the labels' information. Moreover, we find that the network's training is characterized by a rapid increase in the mutual information between the layers and the target label, followed by a slower decrease in the mutual information between the layers and the input variable. By introducing a new generalization error bound, which is exponential in the input representation compression, we propose a novel analytic bound on the mutual information between successive layers in the network. Combining these with the empirical case study enables a comprehensive understanding of optimization dynamics, SGD training properties, and deep architectures' computational benefits.
  
 \item Part 2 -- \textbf{Information in Infinite Ensembles of Infinitely-wide neural networks} -- As previously mentioned, it is often challenging to measure information-theoretic quantities. We developed tractable computations for a wide range of information-theoretic quantities using the NTK framework. These results are used to investigate the critical quantities that correlate with generalization, dynamics, and optimality during the learning process in high-dimensional networks.
    
  \item Part 3 -- \textbf{The Dual Information Bottleneck} -- As mentioned before, the IB framework has several known drawbacks, including that it is not optimal for training with finite samples and that it does not preserve the problem structure. We present the dualIB, a framework for addressing some of these shortcomings. By switching the order between the representation decoder and data in the IB's distortion function, the optimal decoding becomes the geometric mean of input points instead of the arithmetic mean. Whenever the data can be modeled in a parametric form, this conversion preserves its structure and, thus, the original properties of the data. Furthermore, the prediction accuracy is improved by optimizing the mean prediction error exponent for each data size. By analyzing the framework's structure, we show how to solve this new representation learning formulation. A variational formulation of the dualIB for DNNs, (VdualIB)  is presented to address large-scale problems. We examine its properties using real-world datasets and compare them to the original IB.
\end{itemize}

\bibliographystyle{dcu}

\bibliography{main}

%% file: Chapters/opening.tex
\chapter*{Opening the Black Box of Deep \\ Neural Networks via Information}
\addcontentsline{toc}{chapter}{2:   Opening the Black Box of Deep Neural Networks via Information}

\textbf{Unpublished} \\
Ravid Shwartz-Ziv and Naftali Tishby (2018)
\newpage

\begin{center}
        \vspace*{0.5cm}
        \LARGE
        \textbf{Opening the Black Box of Deep Neural Networks via Information} \\
        \vspace{0.8cm}
        \normalsize
    Ravid Shwartz-Ziv \textsuperscript{1}
    Naftali Tishby\textsuperscript{1,2} \\
            \vspace{2.cm}
    \textsuperscript{1} The Edmond and Lilly Safra Center for Brain Sciences, The Hebrew University, \\
  Jerusalem, Israel.\\
    \textsuperscript{2} School of Computer Science and Engineering, \\
    The Hebrew University, \\
  Jerusalem, Israel.\\
    \end{center}
\begin{center}
  \vspace*{0.5cm}
         \normalsize
        \textbf{Abstract} \\
\end{center}
Despite their great success, there is still no comprehensive theoretical understanding of learning with Deep Neural Networks (DNNs) or their inner organization.
Previous work suggested analyzing DNNs in the \textit{Information-Plane}; the plane of the mutual information values that each layer preserves on the input and output variables. 
They suggested that the network's goal is to optimize the Information Bottleneck (IB) trade-off between compression and prediction, successively, for each layer.
In this work, we demonstrate the effectiveness of the information-plane visualization of DNNs.  We first show that the stochastic gradient descent (SGD) epochs have two distinct phases: fast empirical error minimization followed by slow representation compression for each layer. We then argue that the DNN layers end up very close to the IB theoretical bound and present a new argument for the hidden layers' computational benefit.

\vspace*{0.8cm}

\section*{Introduction}
\addcontentsline{toc}{section}{Introduction}

\label{Introduction_1}
DNNs heralded a new era in predictive modeling and machine learning. Their ability to learn and generalize has set a new bar on performance, compared to state-of-the-art methods. This improvement is evident across almost every application domain
 \citep{DBLP:journals/corr/abs-1303-5778,DBLP:journals/corr/ZhangL15,DBLP:journals/corr/abs-1207-0580,DBLP:journals/corr/HeZRS15,natureDeepLeraning}.

However, despite their success, there is still little understanding of their internal organization or optimization process. They are often seen as mysterious "black boxes" \citep{probes2016}.

\begin{figure}[t]
\vskip 0.1in
\centerline{\includegraphics[width=0.7\columnwidth]{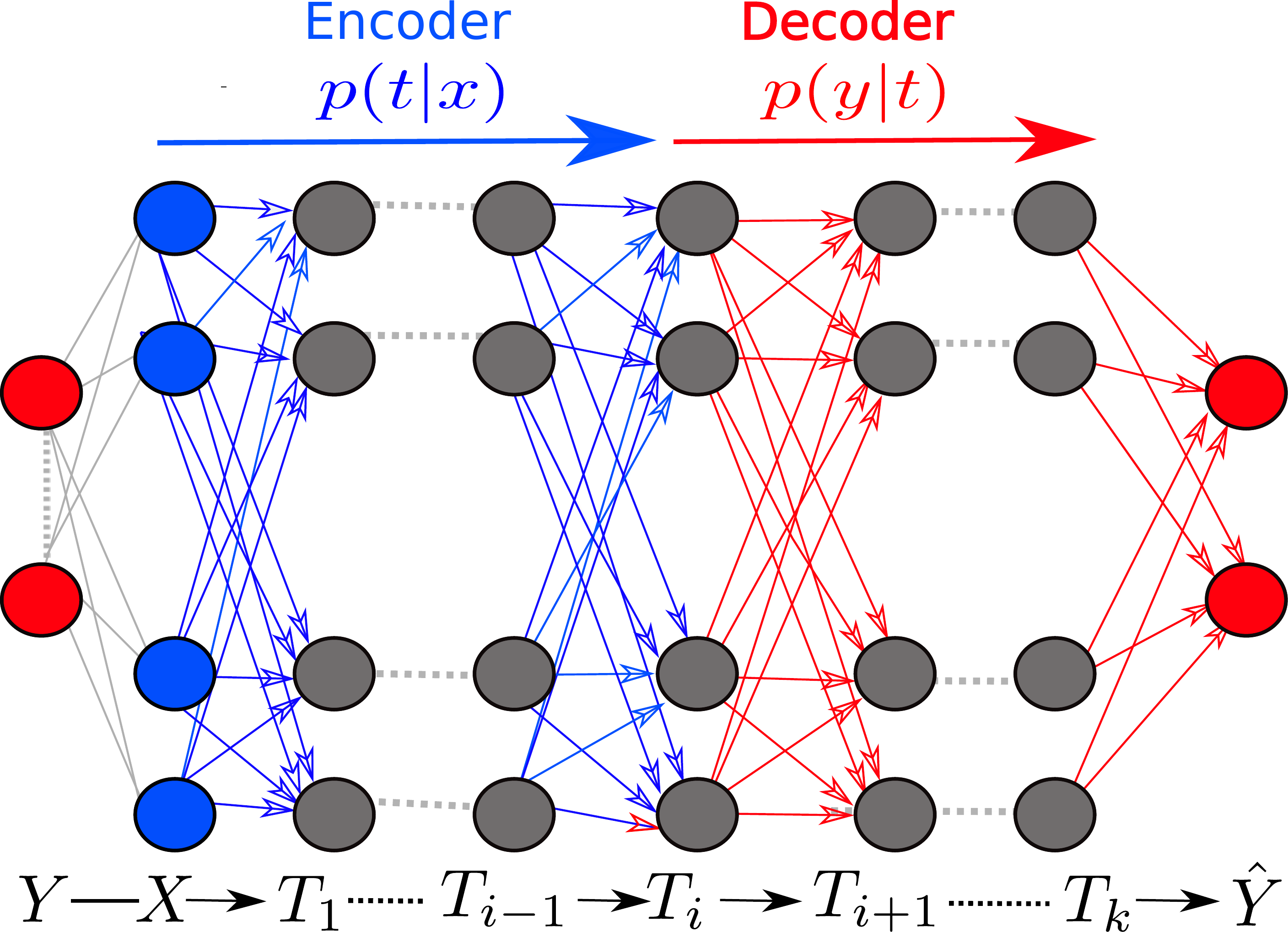}}
\caption{The DNN layers form a Markov chain of successive internal representations of the input. 
Any representation, $T$, is defined through an encoder, $P(T|X)$, and a decoder $P(\hat{Y}|T)$, and can be quantified by its \emph{information-plane} coordinates: $I(X;T)$ and $I(T;Y)$. The IB bound characterizes the optimal representations, which maximally compress the input $X$, for a given mutual information on the desired output $Y$.   }
\label{DNN-layers}
\vskip -0.1in
\end{figure}

The authors in \cite{TishbyZ15} pointed out that layered neural networks' representations of input layers form a Markov chain. They suggested studying these in the \textit{information-plane} - The 
plane of the mutual information values of any other variable with the input variable $X$ and the desired output variable $Y$ (Figure \ref{DNN-layers}).  The rationale for this analysis was based on the mutual information's invariance to invertible re-parameterization and the data processing inequalities (DPI) \citep{Cover:2006} along the Markov chain of the layers. Moreover, they suggested that optimized DNNs layers should approach the Information Bottleneck (IB) bound \citep{DBLP:journals/corr/Tishby1999} of the optimally achievable representations of the input $X$. 

In this paper, we extend their work and demonstrate the effectiveness of visualizing DNNs in the information-plane to understand better the training dynamics, learning processes, and internal representations of deep learning.

Our analysis reveals that the SGD optimization, commonly used in deep networks, has two different and distinct phases: empirical error minimization (ERM) and representation compression. 
Each of these phases displays a very different signal-to-noise ratio of the stochastic gradients. During the ERM phase, gradient norms are much larger than their stochastic fluctuations, leading to a sharp increase in the mutual information on the label variable $Y$. At the compression phase, the errors fluctuate much more than their means, causing the weights to change much like Weiner processes, or random diffusion, with a small influence of error gradients.  In this phase, the input variable $X$ undergoes slow representation compression or reduction of the mutual information.   

In our experiments, most of the optimization iterations are spent on compressing the internal representations under the training error constraint. This compression occurs by the SGD without any other explicit regularization or sparsity imposed. We suspect that this helps avoid overfitting in deep networks. This observation also suggests that many (exponential in the number of weights) different randomized networks have essentially optimal performance. Hence, the interpretation of a single neuron (or weight) in the layers is practically meaningless. 

Then we show that the optimized layers lay on or relatively close to the optimal IB bound for large enough training samples, resulting in a self-consistent relationship between each layer's encoder and decoder distributions (Figure \ref{DNN-layers}). The hidden layers in the optimized model converge along particular lines in the information-plane and move up as the training sample size increases.  
In addition to this, the diffusive nature of the SGD dynamics explains the computational benefit of the hidden layers. 
\section*{Information Theory of Deep Learning}
\addcontentsline{toc}{section}{Information Theory of Deep Learning}

In supervised learning, we are interested in good representations, $T(X)$, of the input patterns  $x\in X$, which enable good predictions of the label  $y \in Y$. Moreover, we want to efficiently learn such representations from an empirical sample of the (unknown) joint distribution $P(X,Y)$, in a manner that provides good generalization.

DNNs generate a Markov chain of such representations, the hidden layers, by minimizing the empirical error over the network's weights, layer by layer. This optimization occurs via SGD, using a noisy estimate of the empirical error gradient at each weight through the backpropagation algorithm \citep{rumelhart1986learning}.  

Our first important insight is to treat the whole layer $T$, as a single random variable, characterized by its encoder, $P(T|X)$, and decoder, $P(Y|T)$ distributions. As we are only interested in the information flowing through the network, invertible transformations of the representations that preserve information generate equivalent representations even if the individual neurons encode entirely different input features. For this reason, we quantify the representations by two numbers, or order parameters, that are invariant to any invertible re-parameterization of $T$, the mutual information of $T$ with the input $X$, $I(X;T)$, and the desired output $Y$, $I(T;Y)$.

Next, the layers' quality is quantified by comparing them to the information-theoretic optimal representation - the IB representations. Furthermore, we explore how the SGD can be used to obtain these optimal representations in DNNs. 

\subsection*{Mutual information}
Given any two random variables, $X$ and $Y$, with a joint distribution $p(x,y)$, their mutual information is defined as:
\small{
\begin{align}
I(X;Y) & = D_{KL}[p(x,y)||p(x)p(y)]\\ & = \sum_{x\in X, y\in Y} p(x,y) \log\left(\frac{p\left(x,y\right)}{p\left(x\right)p\left(y\right)}\right)\\
 &= H(X)-H(X|Y) ~,
 \label{MI}
\end{align}}
\normalsize
where $D_{KL}[p||q]$ is the Kullback-Liebler divergence of the distributions $p$ and $q$, and $H(X)$ and $H(X|Y)$ are the entropy and conditional entropy of $X$ and $Y$, respectively. 

The mutual information quantifies the number of relevant bits that the input variable $X$ contains about the label $Y$, on average. The optimal learning problem can be cast as the construction of an \textit{optimal encoder} of that relevant information via an efficient representation - a minimal sufficient statistic of $X$ with respect to $Y$. A minimal sufficient statistic can enable the \textit{decoding} of the relevant information with the smallest number of binary questions (on average);  i.e., an optimal code. The connection between mutual information and minimal sufficient statistics is discussed in the next section.

Among the characteristics of mutual information, two stand out particularly in relation to DNNs. 
The first is its invariance to invertible transformations:
\begin{equation}
\label{eqn:invariance}
I\left(X;Y\right)=I\left(\psi(X);\phi(Y))\right)
\end{equation}
for any invertible functions $\phi$ and $\psi$.
 
The second is the DPI -- \citep{Cover:2006}
For any three variables that form a Markov chain $X\rightarrow Y \rightarrow Z$,
 \[
 I\left(X;Y\right) \ge I(X;Z) ~.
 \]
\subsection*{The information-plane} 
\label{I-Plane}
\normalsize 
Any representation variable, $T$, defined as a (possibly stochastic) map of the input $X$, is characterized by its joint distributions with $X$ and $Y$, or by its encoder and decoder distributions, $P(T|X)$ and  $P(Y|T)$, respectively. 
Given $P(X;Y)$, $T$ is uniquely mapped to a point in the information-plane with coordinates 
$\left(I(X;T),I(T;Y)\right)$. When applied to the Markov chain of a K-layers DNN, with $T_i$ denoting the $i^{th}$ hidden layer as a single multivariate variable (Figure \ref{DNN-layers}), the layers are mapped to $K$ monotonic connected points in the plane. Therefore,  a unique \textit{information path}  which satisfies the following DPI chains:
\vskip -0.1in
\small
\begin{align*}
I(X;Y) & \ge I(T_1;Y) \ge I(T_2;Y) \ge ... I(T_k;Y) \ge I(\hat{Y};Y)\\
H(X) & \ge I(X;T_1) \ge I(X;T_2) \ge ... I(X;T_k) \ge I(X;\hat{Y}) .
 \label{DPI-1}
\end{align*}
\normalsize
Layers related by invertible re-parametrization appear at the same point, meaning that each information path in the plane corresponds to a multitude of different DNNs, possibly corresponding to very different architectures.
\subsection*{The Information Bottleneck optimal bound}
\label{sec:IB}

What characterizes the optimal representations of $X$ w.r.t. $Y$? 
The classical notion of minimal sufficient statistics provides good candidates for optimal representations. Sufficient statistics, in our context, are maps or partitions of $X$, $S(X)$, capturing all the information that $X$ has on $Y$. Namely, $I(S(X);Y)=I(X;Y)$ \citep{Cover:2006}. 

Minimal sufficient statistics, $T(X)$, are the simplest sufficient statistics and induce the coarsest sufficient partition on $X$. Thus, they are functions of any other sufficient statistic. A simple way of formulating this is through the Markov chain: $Y\rightarrow X \rightarrow S(X) \rightarrow T(X)$, which should hold for a minimal sufficient statistics $T(X)$ with any other sufficient statistics $S(X)$. Using  DPI, we can cast it into a constrained optimization problem:
\[
T(X) = \arg \min_{S(X): I(S(X);Y)=I(X;Y)} I(S(X);X) ~.
\]
Since exact minimal sufficient statistics only exist for special distributions (i.e., exponential families), \cite{DBLP:journals/corr/Tishby1999} relaxed this optimization problem by first allowing the map to be stochastic, defined as an encoder $P(T|X)$, and then, by enabling the map to capture \emph{as much as possible} of $I(X;Y)$, not necessarily all of it.

This leads to the IB trade off \citep{DBLP:journals/corr/Tishby1999}, which provides a computational framework for finding approximate minimal sufficient statistics, or the optimal trade-off between compression of $X$ and prediction of $Y$.  In that sense, efficient representations are approximate minimal sufficient statistics.

If we define $t\in T$ as the compressed representations of $x\in X$, the
representation of $x$ is now defined by the mapping $p\left(t|x\right)$.
This IB trade-off is formulated by the following optimization problem, carried independently for the distributions, $p(t|x), p(t), p(y|t)$, with the Markov chain: $Y\rightarrow X \rightarrow T$,
\begin{equation}
\min_{p\left(t|x\right),p\left(y|t\right),p\left(t\right)}\left\{ I\left(X;T\right)-\beta I\left(T;Y\right)\right\} ~.
\end{equation}

The Lagrange multiplier $\beta$ determines the level of relevant information captured by the representation $T$, $I(T;Y)$, which is bounding the error in the label prediction from this representation.  The local optimal solution to this problem is given by three self-consistent equations, the IB equations:
\begin{equation}
\label{eqn:IB}
\begin{cases}
p\left(t|x\right)=\frac{p\left(t\right)}{Z\left(x;\beta\right)}\exp\left(-\beta D_{KL}\left[p\left(y|x\right)\parallel p\left(y|t\right)\right]\right)\\
p\left(t\right)=\sum_{x}p\left(t|x\right)p\left(x\right)\\
p\left(y|t\right)=\sum_{x}p\left(y|x\right)p\left(x|t\right)~,
\end{cases}
\end{equation}
where $Z\left(x;\beta\right)$ is the normalization function. 
These equations must be satisfied along the \emph{information curve} which is a monotonic concave line that separates the achievable and unachievable regions in the information-plane. For smooth $P(X,Y)$ distributions; i.e., when $Y$ is not a completely deterministic function of $X$,  the information curve is strictly concave with a unique slope, $\beta^{-1}$, at every point, and a finite slope at the origin. 
In these cases, $\beta$ determines a single point on the information curve with specified \emph{encoder}, $P^{\beta}(T|X)$, and \emph{decoder},  $P^{\beta}(Y|T)$, distributions that are related through Eq.(\ref{eqn:IB}).  
\subsection*{Visualizing DNNs in the information-plane}
As proposed by \cite{TishbyZ15}, we study the \emph{information paths} of DNNs in the information-plane. It is possible to do so when the underlying distribution, $P(X,Y)$, is known and the encoder and decoder distributions, $P(T|X)$ and $P(Y|T)$, can be derived directly. In "real-world" problems, these distributions and mutual information values should be estimated from samples or other modeling assumptions. We use two order parameters, $I(T;X)$ and $I(T;Y)$, to visualize and compare different network architectures in terms of their efficiency in preserving the relevant information in $P(X;Y)$.

By visualizing the paths of different networks in the information-plane, we explore the following fundamental issues: 
\begin{enumerate}
\item The SGD layer's dynamics in the information-plane.
\item The effect of the training sample size on the layers.
\item The benefit of the hidden layers.
\item The final location in the information-plane of the hidden layers.
\item The optimality of the layers' representation.
\end{enumerate}
\section*{Experiments}
\addcontentsline{toc}{section}{Experiments}

This section is organized as follows; First, we describe the experimental settings. Next,
we discuss the results of our experiments, including the dynamics of the optimization process in the information-plane, the stochastic gradients, and the computational benefit of the hidden layers. In the last part, we present the layers' optimality and their evolution with the training size.

\subsection*{Experimental setup}
\label{sub_sec:setup}

The numerical studies in this paper explore fully connected feed-forward neural networks with no further architecture constraints.
The activation function of all hidden layers is the hyperbolic tangent function. For the final layer, we use a sigmoidal function.
The networks train using SGD and the cross-entropy loss function (with no explicit regularization). 
Unless otherwise noted, the DNNs use up to seven  fully connected hidden layers, with widths of $12-10-7-5-4-3-2$ neurons (see Figure \ref{fig:gradients}). 
In our results below, layer one is the hidden layer closest to the input and the highest is the output layer.

To simplify our analysis, the tasks were chosen as binary decision rules, which are invariant under $O(3)$ rotations of the sphere. We define the input $X$ to be $12$ binary inputs representing uniformly distributed points on a $2D$ sphere (for other, non-symmetric rules, see Supplementary material). 
The $4096$ different patterns of the input variable $X$ are divided into $64$ disjoint orbits of the rotation group. These orbits form a minimal sufficient statistics for spherically symmetric rules \citep{Kazhdan2003}.
  
Given a function $f$ defined on the sphere, we can obtain a rotation and reflection invariant representation  $\psi\left(f\right)$ by computing the spherical harmonic decomposition of the function. 
\[
\textbf{$f\left(\theta,\phi\right)=\sum_{l\geq{0}}\sum_{m=-l}^{l}{a_{l}^{m}Y_{l}^{m}\left(\theta, \phi\right)}$}
\]
and calculate the energy ($L_2$ norms) of the frequency components.
 \begin{equation}
 \psi\left(f\right)= \sqrt{\|a_{0}\|^2+\|a_{1}\|^2\dots}
 \end{equation}

We refer the reader to \cite{Kazhdan2003} for a good exposition on the above.
To generate the input-output distribution, $P(X,Y)$, we consider  spherically symmetric real valued function of the pattern
$f(x)$ minus a threshold, $\theta$, and apply a sigmoidal function, 
$\Omega(u)=1/(1+\exp(-\gamma u))$:
\begin{equation}
\label{eq:rule}
p(y=1|x)=\Omega\left(\psi\left(f_{x}\right)-\theta\right))~.
\end{equation}
The threshold $\theta$ was selected such that $p(y=1)=\sum_x p(y=1|x)p(x) \approx 0.5$, with uniform $p(x)$. The sigmoidal gain $\gamma$ was high enough to keep the mutual information $I(X;Y)\approx0.99$ bits. 

\subsubsection*{Estimating the mutual information of the layers}

As mentioned above, we model each layer in the network as a single variable $T_{i}, 1\le i\le K$ , and calculate its mutual information with the input and the labels. 
 
To calculate the mutual information of the networks' layers with the input and output variables, we binned the neuron's $arctan$ output activation into $30$ equal intervals between $-1$ and $1$. These discretized values for each neuron in the layer, $t\in T_i$, are used to directly calculate the 
joint distributions, over the $4096$ equally likely input patterns $x\in X$, $P(T_i,X)$ and  $P(T_i,Y)=\sum_x P(x,Y)P(T_i|x)$, using the Markov chain $Y\rightarrow X \rightarrow T_i$ for every hidden layer.
Using the above discrete joint distributions, we calculate the decoder mutual information, $I(Y;T_i)$, and the encoder mutual information, $I(T_i;X)$, for each hidden layer. Note that $I(T_i;Y)$ is calculate with the full data distribution; thus, it corresponds to the generalization error. 

We repeat these calculations with $50$ different randomized initialization of the network's weights and different random selections of the training samples, randomly distributed according to the rule $P(X,Y)$ in Eq.(\ref{eq:rule}).

\subsection*{The dynamics of the optimization process}
\label{sub_sec:results}

\label{sub_sub:dynamics_info_plane}
\begin{figure}[t]
\vskip 0.05in
\centerline{\includegraphics[width=0.9\columnwidth]{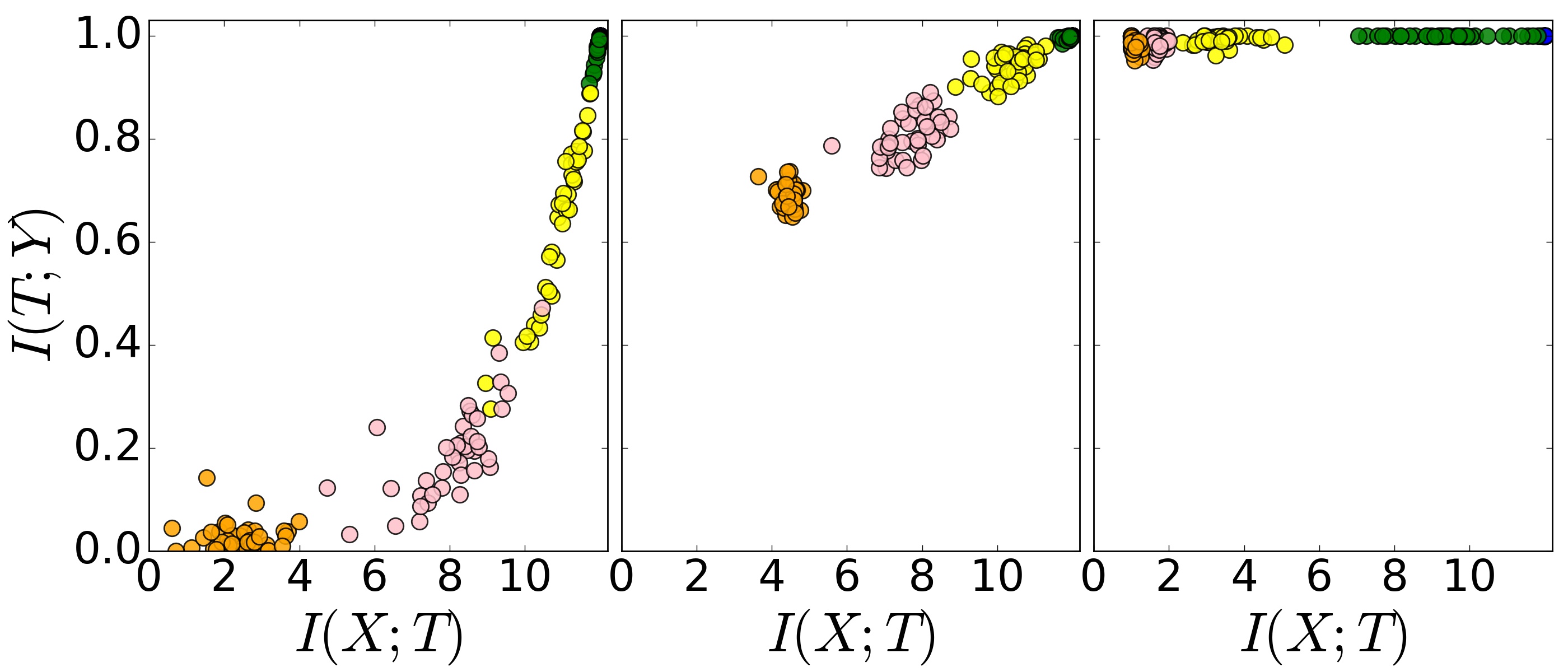}}
\caption{Snapshots of layers (different colors) of 50 randomized networks during the training process in the information-plane:  \textbf{left} - At the initialization time;  \textbf{center} -  After 400 epochs; \textbf{right} -  After 9000 epochs. }
\label{opt_process}
\vskip -0.1in
\end{figure}

To understand the network's SGD optimization dynamics, we evaluate and visualize pairs of $I(X;T_i)$ and $I(T_i;Y)$ for each layer along the learning process. We repeat this process for $50$ different randomized initializations with different randomized training samples. 
Figure \ref{opt_process} depicts the layers (each one in a different color) of all the $50$ networks, trained with a randomized $85$ percents of the input patterns, in the information-plane, at various times during the training. 

\begin{figure}[t]
\centerline{     
\includegraphics[width=0.9\columnwidth]{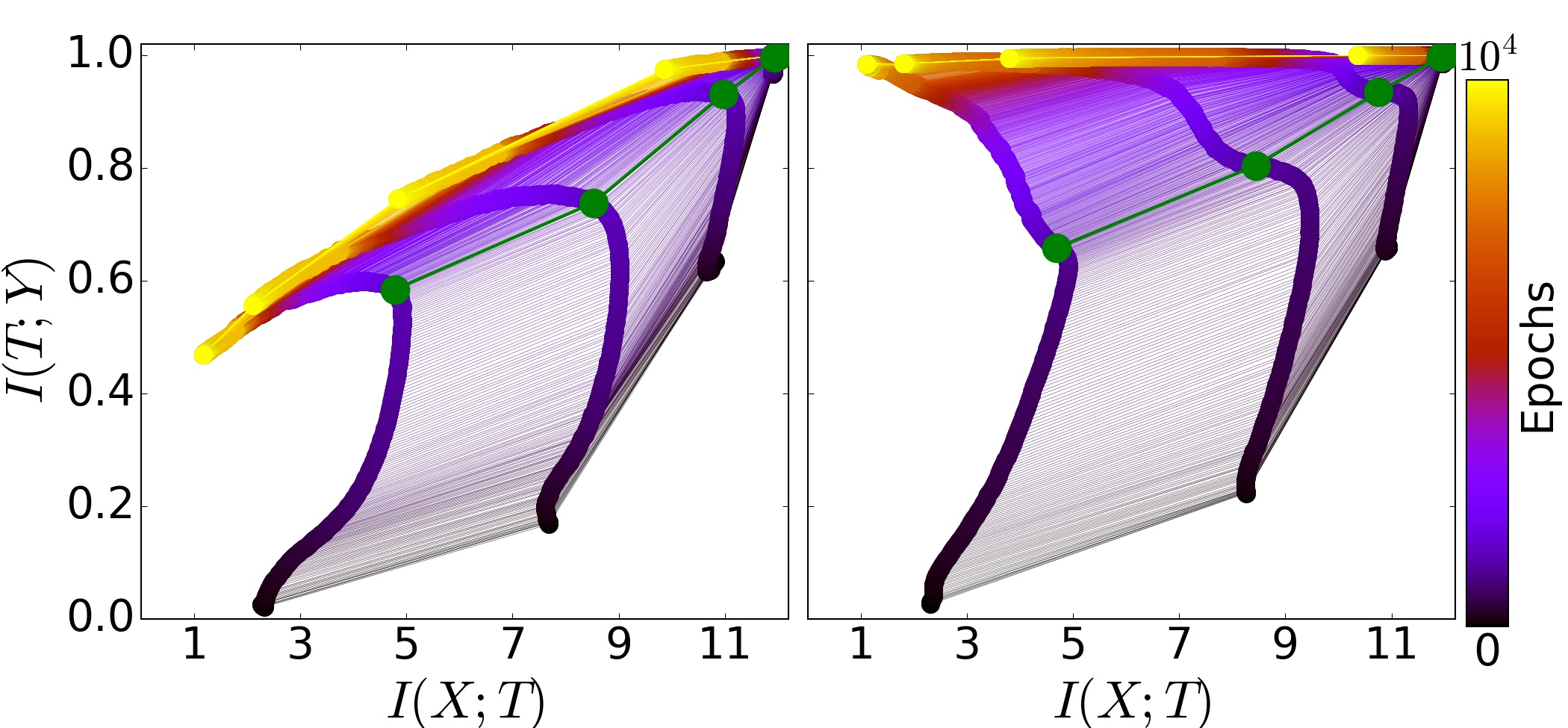}}

\caption{\textbf{The evolution of layers during the learning process -  the information-plane.}
\textbf{Left} - Paths of layers trained with only 5\% of the patterns.  \textbf{Right} - Paths of the layers, trained with 85\% of the input patterns.  The line color indicates the number of epochs. There are 6 points for each path's different hidden layers, averaged over all 50 randomized networks. The layer with the smallest $I(X;T)$ is the output layer, and the one with the highest $I(X;T)$ is the first hidden layer. The green paths correspond to the SGD phase transition grey line on Figure \ref{fig:gradients}} 
\label{network_epochs}
\vskip -0.1in 
\end{figure}

As can be seen, the deeper layers of the randomly-initialized network do not preserve the relevant information at the beginning of the optimization, and $I(T;Y)$ decreases sharply along the path. During the SGD optimization, the layers first increase $I(T;Y)$, then significantly decrease $I(X;T)$ -- compressing the representation. One striking observation is that the different randomized networks' layers follow similar paths through the optimization process and eventually converge to nearby points in the information-plane. The randomized networks are thus averaged over, and the average layer trajectory plot is shown in Figure \ref{network_epochs}.

The graph on the left shows those same trajectories when trained with $85$ percent of all patterns, whereas on the right, the same trajectories when trained with only $5$ percent. Note that the mutual information is calculated with the full rule distribution. Thus, $I(T;Y)$ corresponds to the test or generalization error.  
In each case, the two optimization phases are visible. During the fast - ERM phase, which lasts a few hundred epochs, the layers increase the information on the labels (increase $I(T;Y)$). However, the layers' information on the input, $I(X;T)$, decreases in the second phase of training, resulting in loss of irrelevant information until converging (yellow points). 
We call this phase the \emph{representation compression} phase. 
Because of the cross-entropy loss approximate $I(T;Y)$ up to a constant (see \cite{shwartz2018representation}), the increase of $I(T;Y)$ in the ERM phase is expected. However, the compression phase is surprising. There is no explicit regularization that could simplify the representations, such as $L1$ regularization, and there is no sparsification or norm reduction of the weights (see supplementary). 

While both the small ($5\%$) and large ($85\%$) training sample sizes exhibit the same ERM phase, the compression phase drastically reduces the layers' label information for a small number of examples. However, with large sample sizes, the layers' label information increases. Those results seem very much to be caused by overfitting the small sample noise, a problem that can be eliminated by early stopping methods \citep{Larochelle:2009:EST:1577069.1577070}. This overfitting is mainly the consequence of the compression phase. It simplifies the layers' representations and loses relevant information.  Understanding what determines the convergence points of the layers in the information-plane for different training data sizes is an interesting theoretical goal.

\begin{figure}[t]
  \centering
\subfloat[\label{fig:gradients}
The change of weights, the means and standard deviations of the gradients for one layer, during the training (log-log scale). The grey line marks the transition between the \emph{drift} to the \emph{diffusion}]{\includegraphics[width =0.47\columnwidth]{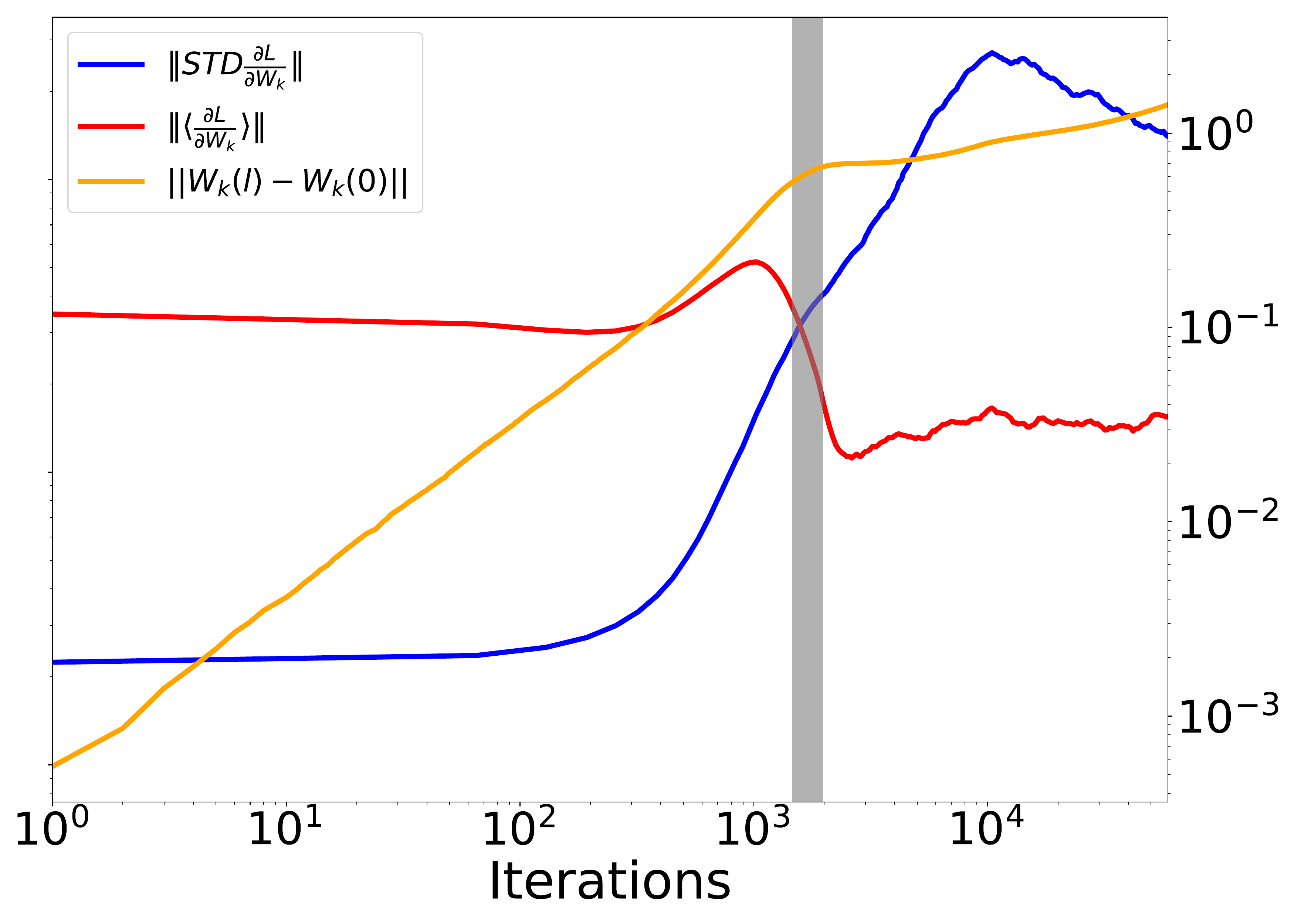}} 
\qquad
\subfloat[\label{fig:gradeints_batchs1}The transition point of the SNR ($Y$-axis) versus the beginning of the information compression ($X$-axis), for different mini-batch sizes] {\includegraphics[width=0.47\columnwidth]{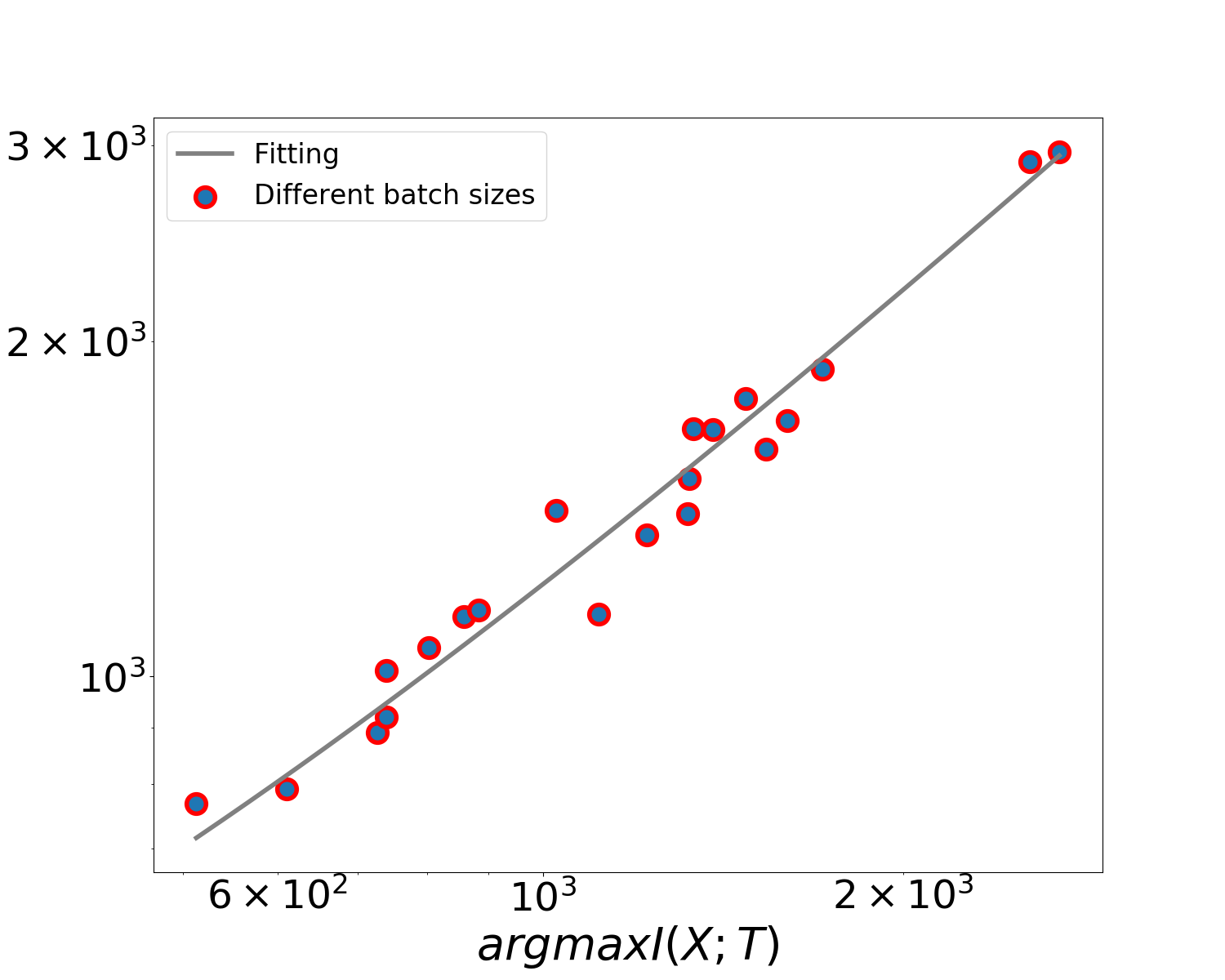}}
 \caption{ The Two Phases of SGD Optimization}
    \label{fig:mnist1}
\end{figure}
\subsection*{The two phases of SGD optimization}
\label{sub_sub_sec:grad_dynamics}
By examining the stochastic gradients' behavior along the epochs, a better understanding of the ERM and the representation-compression phases can be gained.  
For each layer, we look at the mean and standard deviations of the weights' stochastic gradients (in the sample batches). Namely,  $M_i=\left \Vert \braket{\frac{\partial L}{\partial W_k}} \right \Vert_F$ and $S_i= \left \Vert \text{STD} \left( \frac{\partial L}{\partial W_k} \right) \right \Vert_F,$ where $\braket{\cdot}$ and $\text{STD}(\cdot)$ are the mean and the element-wise standard deviation respectively, and $\left \Vert \cdot \right \Vert_F $ denotes the Frobenius norm. 
Figure \ref{fig:gradients} shows the mean and standard deviation of the weights' gradients for one layer in our network, as a function of the iterations (in log-log scale).
As can be seen, there is a transition between two distinct phases (the vertical line). The phases are defined by the ratio between the means of the gradients and their standard deviations. We refer to this ratio as the signal-to-noise ratio (SNR). The first phase is an \emph{drift phase}, when the gradient means are much larger than the standard deviations, indicating relatively small gradient stochasticity (high SNR). Gradient means are very small compared to batch-to-batch fluctuations (low SNR) in the second phase, which we call the \emph{diffusion phase}. This transition is generally expected when the empirical error becomes saturated, and SGD is dominated by its fluctuations \citep{bertsekas2011incremental}.
We claim that these distinct SGD phases (grey line in Figure \ref{fig:gradients}), correspond and explain the ERM and compression phases we observe in the information plane (marked green paths in Figure \ref{network_epochs}).  

This dynamic phase transition occurs in the same number of iterations as in the layers' trajectories in the information plane. To relate the transition phase in the information plane to the gradients, we examine the mini-batch size effect on them. For each mini-batch size, we find both the starting point of the information compression and the gradient phase transition (the iteration where the derivative of the SNR is maximal). This is given in Figure \ref{fig:gradeints_batchs1}. The $X$-axis is the iteration where compression is started, and $Y$-axis is the iteration at which a phase transition occurs in the gradient. There is a linear trend between the two.

In the drift phase, $I(T; Y)$ increases for every layer, as it rapidly reduces the empirical error. Conversely, the diffusion phase adds random noise to the weights, which evolve like Wiener processes under the training error or label information constraint. A Focker-Planck equation can describe such diffusion processes (see, e.g., \cite{risken1989fokker}), whose stationary distribution maximizes the entropy of the distribution of the weight under the training error constraint.  That, in turn, maximizes the conditional entropy, $H(X|T_i)$, or minimizes the mutual information $I(X;T_i)=H(X)-H(X|T_i)$, because the input entropy, $H(X)$ is constant.  This entropy maximization by additive noise, also known as stochastic relaxation, is constrained by the empirical error or equivalently (for small errors) by $I(T; Y)$.  

It remains unclear why different hidden layers converge to different points in the information-plane. Our analysis suggests that different layers have different noise levels in the gradients during the compression phase, explaining why they end up in different maximum entropy distributions. However, the gradient noises appear to vary and eventually decrease when the layers converge, suggesting that the convergence points are related to the \emph{critical slowing down}  of stochastic relaxation near phase transitions on the IB curve. 

An interesting outcome of \emph{compression by diffusion} is the randomized nature of the network's final weights.
 The correlations between the in-weights of different neurons in the same layer, which converge to essentially the same point in the plane, are very small. This indicates that there are many different networks with essentially optimal performance, and attempts to interpret single weights or even single neurons in such networks are meaningless.
\subsection*{The benefit of the hidden layers}
\label{sub_sub_sec:benefit_hidden}
\begin{figure}[t]
\centering     
\subfloat[information-planes for different number \\ of hidden layers.]{
  \includegraphics[height=5.5cm,width=0.5\columnwidth]{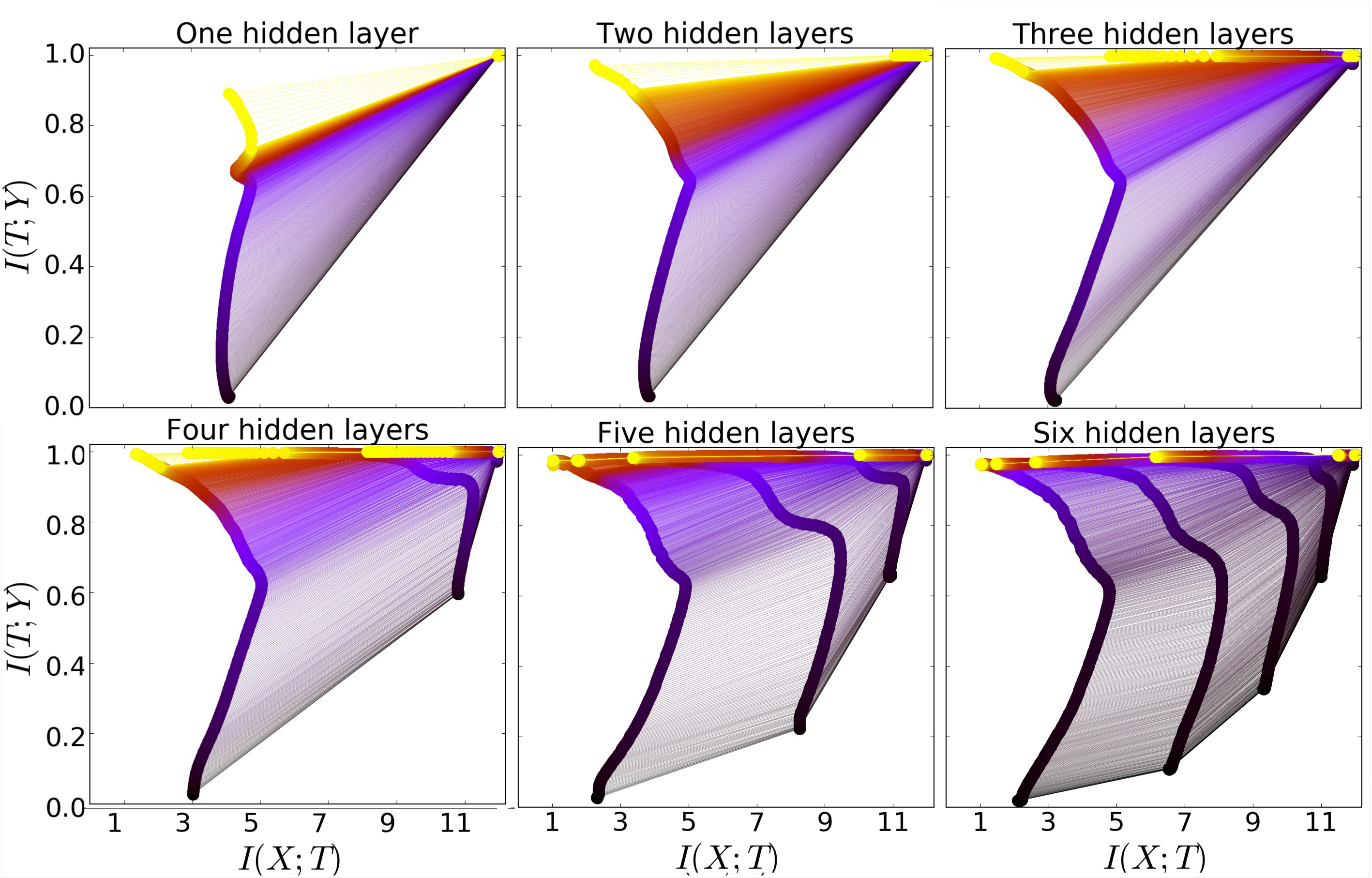}\label{layers_inf}}
\subfloat[The converged iteration as a function of the number of layers - loglog scale]{        
  \includegraphics[ width=0.5\columnwidth]{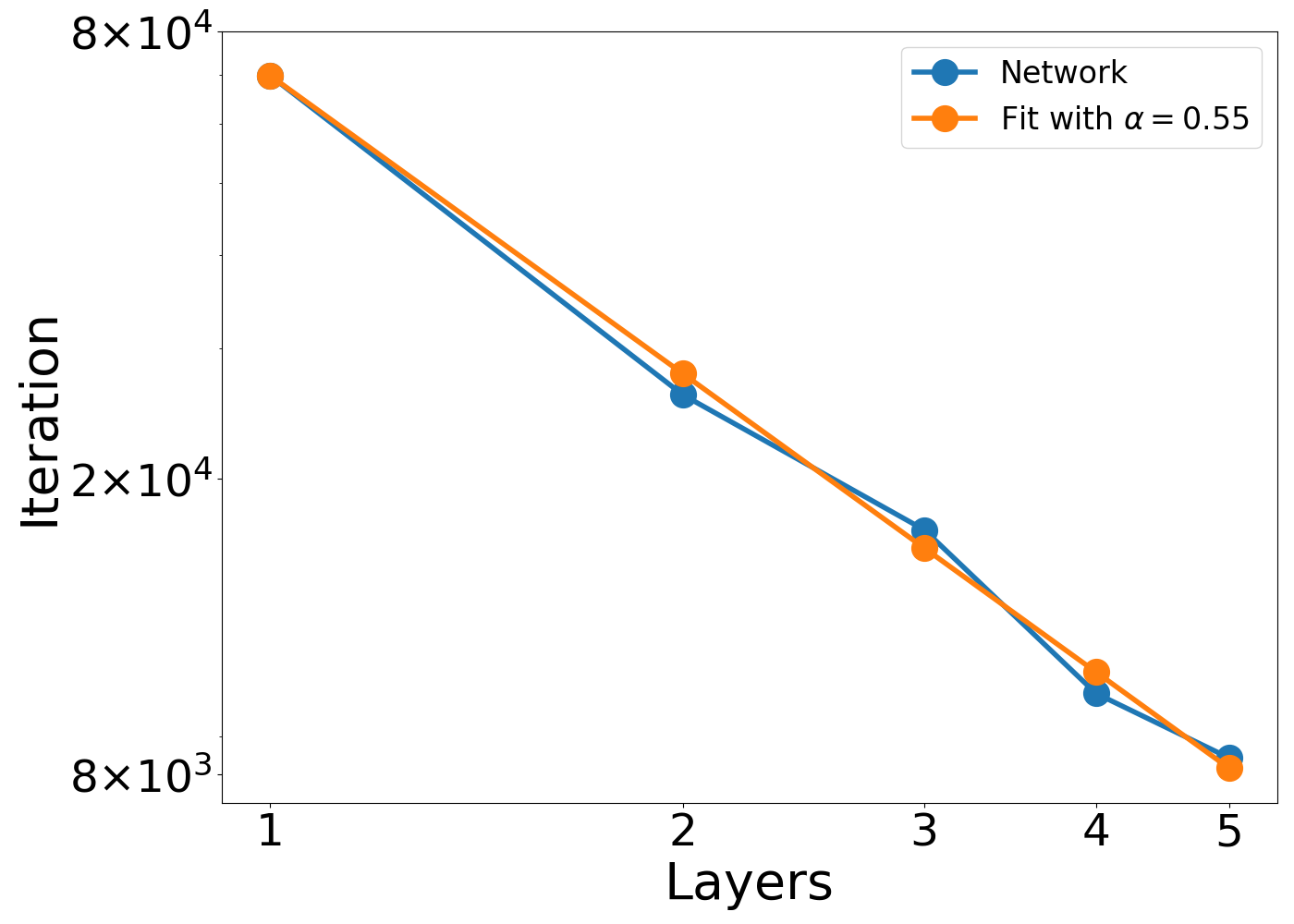}\label{fig:all_layers_time}}
\caption{\textbf{The computational benefit of the layers}}
\end{figure}

What is the benefit of the hidden layers? This question is one of the fundamentals of deep learning. For this reason, we train networks with a different number of layers ($1-5$) and determine the minimum iteration for which the network converges.
As before, we repeat each training $50$ times with randomized initial weights and training samples. 
Figure \ref{layers_inf} shows the information-plane paths for these six architectures during the training epochs, while Figure \ref{fig:all_layers_time} shows the iteration number that the different networks reached 98\% accuracy.
  
There are several important results from this experiment:
\begin{enumerate}

\item \emph{The addition of hidden layers dramatically reduces the number of training iterations required for good generalization.} As more layers are added, the convergence time decreases by a factor of $k^{\frac{1}{\alpha}}$ where $alpha=0.55$. 

\item \emph{When starting from previously compressed layers, each compression phase will be shorter.}
By comparing the time to good generalization with four hidden layers and five hidden layers, we can see that convergence with four layers is much slower than with five or six hidden layers, where it takes half the time to reach the endpoints. 
\item \emph{The compression rate is faster for the deeper (narrower) layers, which are located closer to the output.}
While in the drift phase, the lower layers move first (as a result of DPI), in the diffusion phase, the top layers compress first and pull the lower ones after them. Adding more layers seems to accelerate the compression. \end{enumerate} 

\subsubsection*{The IB optimality of the layers}
\label{IB.optimal}
Finally, to quantify the IB optimality of the layers, we test whether the converged layers satisfy the encoder-decoder relations of Eq. (\ref{eqn:IB}), for some value of the Lagrange multiplier $\beta$. With the encoder and decoder distributions based on quantized values of layer neurons, $p_{i}\left(t|x\right)$ and $p_{i}\left(y|t\right)$, respectively, we compute the information values for each converged layer.

\begin{figure}[t]
\centering
\includegraphics[width=0.5\columnwidth]{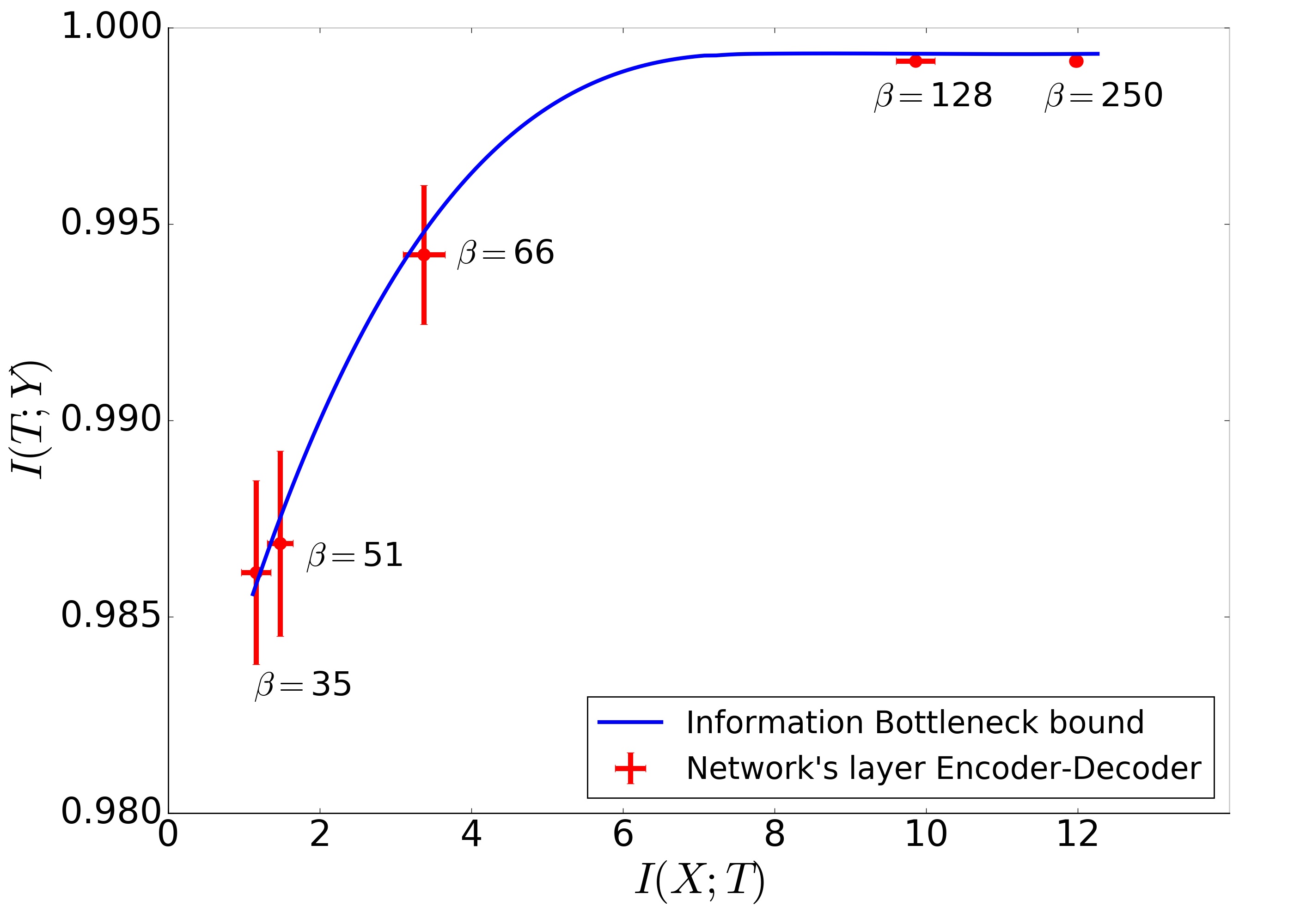}
\caption{\textbf{\label{IB}The DNN layers are fixed-points of the
IB equations}. The error bars represent standard error measures for $N=50$. In each line, there are $5$ points for the different layers. For each point, $\beta$ is the optimal value tof that layer.}
\vskip -0.1in 
\end{figure}

To test the IB optimality of the layers, we calculate the optimal IB encoder, 
$p_{i,\beta}^{IB}\left(t|x\right)$
using the $i^{th}$ layer decoder, $p_{i}\left(y|t\right)$, through Eq.(\ref{eqn:IB}). 
This can be done for any value of $\beta$, with the known $P(X,Y)$.  Then, we find the optimal $\beta_i$ for each layer, by minimizing the average KL divergence between the IB and the layer's encoders,
\[
\beta_{i}^{\star}=\arg \min_{\beta} \mathbb{E}_{x}  { D_{KL}\left[p_{i}\left(t|x\right)||p_{\beta}^{IB}\left(t|x\right)\right]} ~.
\]
In figure \ref{IB} we see the information plotted over the layers together with the IB curve (blue line). The five empirical layers (trained with SGD) lie remarkably close to the theoretical IB limit. Furthermore,  the slope of the curve, $\beta^{-1}$, matches their estimated optimal $\beta_{i}^{\star}$.  

Hence, the DNN layers' encoder-decoder distributions satisfy the IB self-consistent equations within our numerical precision, decreasing  $\beta$ as we go deeper in the network.  
The error bars are calculated over $50$ randomized networks. 
As predicted by the IB equations, near the information curve $\Delta I_Y \sim \beta^{-1} \Delta I_X$. 

\subsubsection*{The evolution of the layers with training size}
\label{sub_sub_sec:train_size}
\begin{figure}[ht]
\begin{centering}
\includegraphics[width=0.5\columnwidth]{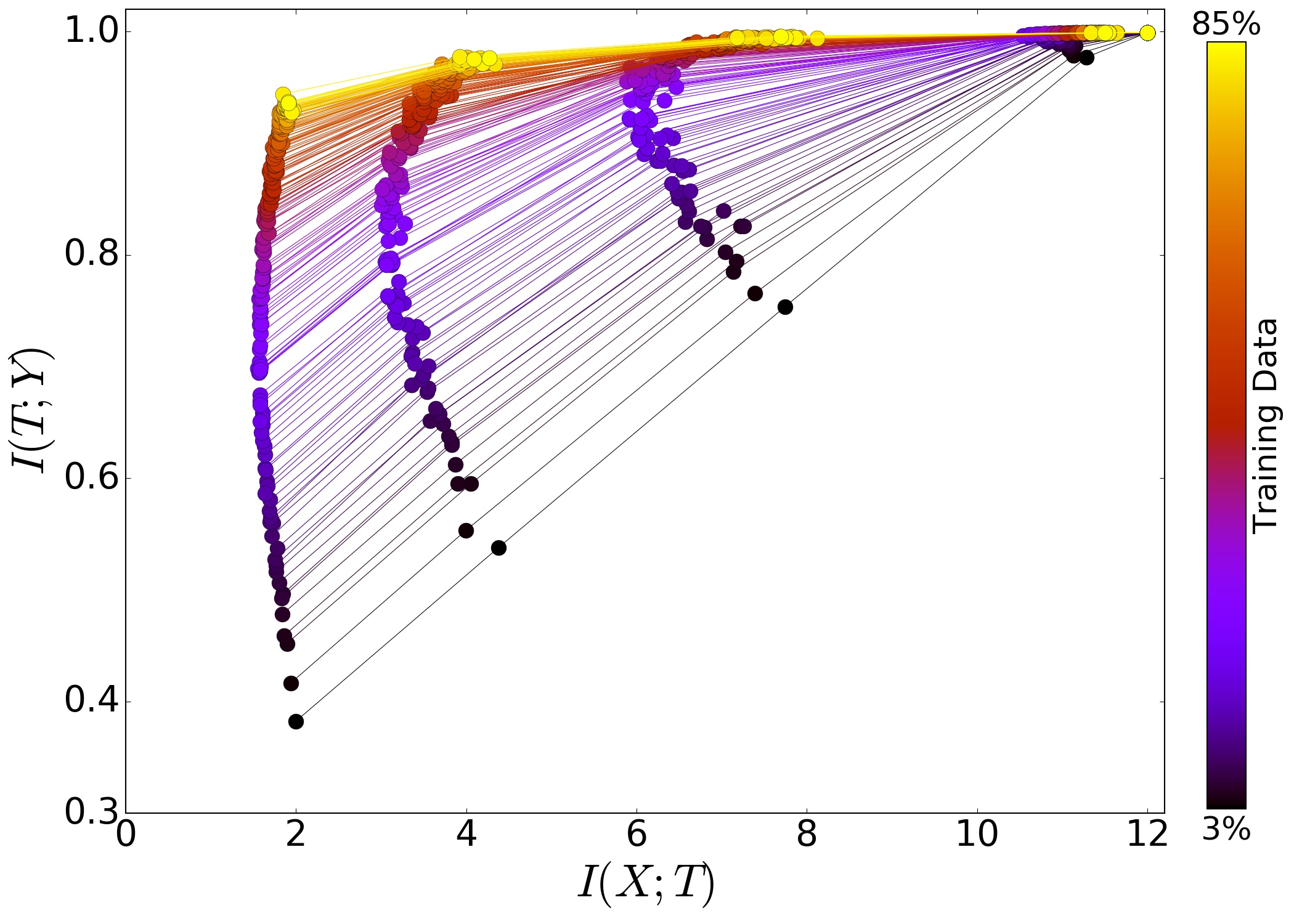}
\par\end{centering}
\caption{\label{fig:learn-inf-plane}\textbf{The effects of different training data sizes on the layers in the information-plane}. Each line represents a converged network that had different training data sizes.}
\end{figure}

Another problem in machine learning, which we just briefly discuss in this paper, is the importance of the training data size \citep{cho2015much}.
It is useful to visualize the hidden layers' converged locations for different training data sizes in the information-plane (Figure \ref{fig:learn-inf-plane}).

As before, we train the network using six hidden layers, with different sample sizes ranging from 3 percent to 85 percent of the data.  
As expected, the layer's true label information, $I(T; Y)$ pushed up and gets closer to the theoretical IB bound on the rule distribution with increasing training size. 

Despite the randomizations, the converged layers for different training sizes line up on a smooth line with remarkable regularity for each layer. We claim that the layers converge to specific points on the finite sample information curves, which can be calculated using the IB self-consistent equations (Eq. (\ref{eqn:IB})), with the decoder replaced by the empirical distribution. This finite sample IB bound also explains the bounding shape on the left of Figure \ref{network_epochs}.   
Since the IB information curves are convex for any distribution, the layers converge to a convex curve in the plane even with very small samples. 

However, the layers' training size effect is different for $I(T; Y)$ and $I(X; T)$.  The training size hardly changes the information in the lower layers since even random weights keep most of the mutual information on both $X$ and $Y$. However, for deeper layers, the network learns to preserve the relevant information about $Y$ and  compress the irrelevant information in $X$. 
More details on $X$ are relevant for $Y$ with a larger training set, and $I(X; T)$ increases in the middle layers. 
\section*{Related Work}
\addcontentsline{toc}{section}{Related Work}

\label{sec:realted_work}
Recently, the IB theory of deep learning has received much attention, including criticism of its rationale. Follow-up works attempt to address several points:
\subsubsection*{Information compression}
{
\cite{saxe2018information} constructed several experiments to explore and refute the IB interpretation. They claim that the compression phase is an artifact of the quantization used to approximate $I(X; T)$ and the activation function (saturating nonlinearity). They observed no compression for the nonsaturating ReLU activation function.

Getting a reasonable estimate of mutual information between joint distributions in high-dimensional contexts is one of the main challenges in applying information-theoretic measures to real-world data. This problem has been extensively studied over the years (e.g., \cite{Paninski:2003:EEM:795523.795524}), showing that there is no ``efficient'' solution when the dimension of the problem is large and known approximations do not scale well with dimension and sample size \citep{gao2015efficient}. 

Several recent works have attempted to develop novel and efficient methods for estimating mutual information for DNNs.
One line of works uses a generative decoder network (PixelCNN++) to estimate a lower bound on the mutual information \citep{darlow2020information,nash2018inverting}. In \cite{darlow2020information}, the authors observed compression in the hidden layers during learning on the ImageNet dataset, using both ResNet and autoencoder architectures. This work confirms our two-stage learning for both classification and autoencoding tasks, characterized by (1) an initial short increase phase (2) following by a longer decrease in the mutual information with the input. Further, they observe that when the information is maximally compressed, the input images' class-irrelevant features are discarded; Namely, conditionally generated samples vary more while retaining information relevant to classification.
\cite{nash2018inverting} reproduced our work using a similar approach. They observed a compression phase when using a convolution network with ReLU activation on the MNIST dataset.
A two-phase behavior including a compression phase on the MNIST dataset with both fully-connected networks and convolutional networks was also reported by Noshad \& Alfred \citep{noshad2018scalable}. The authors introduced a new mutual information estimator, the ensemble dependency graph estimator (EDGE), which combines locality-sensitive hashing with dependency graphs and ensemble bias-reduction methods.  The authors of \cite{chelombiev2019adaptive} proposed adaptive approaches to estimating mutual information. These adaptive approaches compared the behavior of different activation functions and observed compression in DNNs with nonsaturating activation functions. It was observed that unlike saturating activation functions, compression does not always happen, and it is sensitive to initialization. They also observed that DNNs with L2 regularization strongly compress information.

In \cite{2018Estimating}, the authors propose a new theoretical noisy entropy estimator to estimate mutual information.  With both ReLU and linear activations, they observed the compression phase on our dataset and the MNIST dataset and related this compression behavior to geometric clustering. They also found that the behavior of $I(X; T)$ is strongly influenced by the "binning size," which is used for estimating the mutual information. 

In \cite{elad2018effectiveness}, the authors utilized the mutual information neural estimator (MINE) \citep{Belghazi2018MutualIN}, which estimates the KL divergence through the maximization of the dual representation of Donsker \& Varadhan \citep{donsker1975asymptotic}. They showed that for the MNIST dataset with ReLU activation, the compression phase did not appear for ``vanilla'' cross-entropy training but did appear for training with weight decay regularization. 

The authors in \cite{achille2017critical} used Fisher information on the weights to demonstrate a two-phase learning process, involving an initial short increase, followed by a longer phase of decreasing information. Their paper described the reduction in information in the weights as implying a reduction in information in the activation. 
An additional confirmation on the two-phase behavior of DNNs can found in \cite{li2017convergence}, who investigated shallow neural networks with residual connections and normal input distribution and showed that the SGD  has two phases; (1) search and (2) convergence. Additionally, in \cite{dieuleveut2017bridging} they presented transient and stationary phases by looking at the inner product between successive mini-batch gradients in the network.

}
\subsubsection*{Generalization and compression}
{
Another claim of\cite{saxe2018information} is that the generalization does not require compression. They constructed a deep linear network toy example to illustrate generalization without compression. 

The authors in \cite{chelombiev2019adaptive} found that generalization accuracy was positively correlated with the degree of compression of the last layer.  The connection of the generalization to compression is also discussed in \cite{shwartz2018representation}. They showed that the generalization error depends exponentially on $I(X; T)$,  once $I(T; X)$ becomes smaller than $\log2n$ - the query sample complexity. Moreover, They showed that M bits of compression of $X$ are equivalent to an exponential factor of $2^M$ training examples. Furthermore, \cite{achille2017emergence} proved that flat minima, which have better generalization properties, bound information with the weights, and the information in the weights bound information in the activations.
}
\subsubsection*{Information in deterministic networks}{
Several works \citep{saxe2018information, amjad2018not, 2018Estimating}, state that the term $I(X; T)$ in the IB functional is theoretically either infinite or a constant for deterministic DNNs with continuous input. Thus, they wonder about the meaning of measuring it.

This problem was addressed in the literature by (1) training stochastic DNNs, which ensure the information is finite, and by (2) using a tractable variational
approximation of the mutual information \citep{alemi2016deep, kolchinsky2017estimating, chalk2016relevant, achille2018information, belghazi2018mine}.
These works inject stochasticity by adding noise (or quantizing the input), only for quantifying the mutual information between the input and the hidden layer and not into the DNN itself. \cite{2018Estimating} introduced an auxiliary (noisy) DNN framework and showed that it is a good proxy for the original (deterministic) DNN both regarding performance and the learned representations.

To add noise, we can inject it directly into the representation (the activation values) and get a noisy variable we can measure. \cite{saxe2018information} used a non-parametric KDE estimator outlined by \cite{kolchinsky2017estimating}, which directly adds small Gaussian noise to the data. However, all the training processes had a fixed level of noise, which led to failure due to the huge variation in the activation values (See \citep{chelombiev2019adaptive} for detailed explanation).

It is also possible to add noise by discretizing the continuous variables into bins and approximating the representation as a discrete variable. In \cite{2018Estimating}, they claimed that this estimation injects noise that is not present in the actual network, which is highly sensitive to selecting bin size and does not track $I(X; T)$ for different choices of the bin size. They developed a noisy DNNs framework and a rigorous estimator for $I(X;T)$. Using this estimator, they observed compression in various models. By relating $I(X; T)$ in the noisy DNN to an information-theoretic communication problem, they showed that compression is driven by the progressive clustering of hidden representations of inputs from the same class. They also proved that the estimator of $I(X;T)$ using binning is a measure for clustering.

Binning saturating activation functions facilitates mutual information estimation since all hidden activity is bounded within a predetermined range. However, with nonsaturating functions, the estimation procedure's noise level must be adapted for every layer of the network each time.  
In \cite{saxe2018information}, the activation values are binned using a single range for the whole training, which is terminated by the maximum value across all epochs and all layers.
However, since the network at every epoch is different, by this binning procedure, the network's estimation mixes unrelated factors \citep{chelombiev2019adaptive}.

\cite{achille2018information} stated that minimizing a stochastic network with an approximate compression term from the IB functional as a regularizer is equivalent to minimizing cross-entropy over deterministic DNNs with multiplicative noise (information dropout). Moreover, they proved that the special case of a Bernoulli noise results in the dropout method.  In \cite{elad2018effectiveness}, the authors showed that the binned information could be interpreted as a weight decay penalty, which aligns with common practice in DNN training. 
}
\subsubsection*{Invertible convolutional neural networks}{
 The invertible DNN  \citep{jacobsen2018irevnet} is a network's architecture that achieves state-of-the-art performance. The authors claimed that in such networks, $I(X; T_k)$ is never discarded regardless of the network parameters, and it is possible to reconstruct the input from each layer. However, although reversible networks do not discard the input's irrelevant components, we hypothesize that these networks progressively separate the irrelevant components from the relevant, allowing the final classification mapping to discard this information. 

}
\section*{Discussion and Conclusions}
\addcontentsline{toc}{section}{Discussion and Conclusions}

Motivated by the IB framework, our numerical experiments demonstrate that the visualization of the layers in the information-plane reveals many - so far unknown, details about the inner working of DNNs. They reveal the distinct phases of the SGD optimization, drift, and diffusion, explaining the ERM and the representation compression trajectories of the layers' information. The stochasticity of SGD methods is usually motivated by escaping local minima of the training error. In this paper, we give it a new, perhaps much more important role: It generates highly efficient internal representations through \emph{compression by diffusion}. This is consistent with other recent suggestions on the role of noise in DNNs \citep{achille2018information,Kadmon2016OptimalAI}.

We also argue that SGD seems an overkill during the diffusion phase, which consumes most of the training epochs, and that much simpler optimization algorithms, such as Monte-Carlo relaxations \citep{geman1988stochastic}, can be more efficient. However, the IB framework may provide even more. If the layers converge to the IB theoretical bounds, there is an analytic connection between the encoder and decoder distributions for each layer, which can be exploited during training. Combining the IB iterations with stochastic relaxation methods may significantly boost DNN training.  
  
To conclude, our analysis suggests that SGD with DNNs is, in essence, learning algorithms that effectively find efficient representations approximate minimal sufficient statistics in the IB sense.  

\subsection*{Acknowledgments}
This work is partially supported by the Gatsby Charitable
Foundation, The Israel Science Foundation, and Intel ICRI-CI
center. 

\bibliographystyle{dcu}

\bibliography{main}

%% file: Chapters/rep_compression.tex
\chapter*{Representation Compression  \\ and Generalization in Deep Neural Networks} 
\addcontentsline{toc}{chapter}{3: Representation Compression and Generalization in Deep Neural Networks}

\textbf{Unpublished} \\
Ravid Shwartz-Ziv, Amichai Painsky and Naftali Tishby (2019)
\newpage

\begin{center}
        \vspace*{0.5cm}
        \LARGE
        \textbf{Representation Compression and Generalization in Deep Neural Networks} \\
        \vspace{0.8cm}
        \normalsize
    Ravid Shwartz-Ziv \textsuperscript{1}
    Amichai Painsky \textsuperscript{2}
    Naftali Tishby\textsuperscript{1,2} \\
            \vspace{2.cm}
    \textsuperscript{1} The Edmond and Lilly Safra Center for Brain Sciences, The Hebrew University, \\
  Jerusalem, Israel.\\
    \textsuperscript{2} School of Computer Science and Engineering, \\
    The Hebrew University, \\
  Jerusalem, Israel.\\
    \end{center}
\begin{center}
  \vspace*{0.5cm}
         \normalsize
        \textbf{Abstract} \\
\end{center}

Understanding the groundbreaking performance of Deep Neural Networks (DNNs) is one of the most significant challenges to the scientific community today. In this work, we introduce an information-theoretic viewpoint on the behavior of deep network optimization processes and their generalization abilities. Specifically, we study DNNs on the information plane, the plane of the MI between each layer with the input variable, and the desired label, during the training dynamics. We show that we can characterize the network's training by a rapid increase in the mutual information (MI) between the layers and the target label, followed by a longer decrease in the MI between the layers and the input variable.

    Furthermore, we explicitly show that these two fundamental information-theoretic quantities govern the network's generalization error by introducing a new generalization gap bound that is exponential in the input representation compression. 
    The analysis focuses on typical patterns of large-scale problems. For this purpose, we introduce a novel analytic bound on the MI between consecutive layers in the network. An important consequence of our analysis is a superlinear boost in training time with the number of non-degenerate hidden layers, demonstrating the hidden layers' computational benefit.
\vspace*{0.8cm}

\section*{Introduction}
\addcontentsline{toc}{section}{Introduction}

\label{Introduction}

Deep Neural Networks (DNNs) heralded a new era in predictive modeling and machine learning. Their ability to learn and generalize has set a new bar on performance compared to state-of-the-art methods. This improvement is evident across almost every application domain, especially in areas involving complicated dependencies between the input variable and the target label \citep{natureDeepLeraning}. However, despite their great empirical success, there is still no comprehensive understanding of their optimization process and its relationship to their (remarkable) generalization abilities. 

This work examines DNNs from an information-theoretic viewpoint. For this purpose, we utilize the Information Bottleneck (IB) principle \citep{DBLP:journals/corr/Tishby1999}. The IB is a computational framework for extracting the most compact yet informative representation of the input variable ($X$) with respect to a target label variable ($Y$). The IB bound defines the optimal tradeoff between representation complexity and its predictive power. Specifically, it is achieved by minimizing the mutual information (MI) between the representation and the input, subject to the level of MI between the representation and the target label.

Recent results \citep{shwartz2017opening}, demonstrated that the layers of DNNs tend to converge to the IB optimal bound. The results pointed to a distinction between the two phases of the training process. The first phase is characterized by an increase in the MI with the label (i.e., fitting the training data), whereas in the second and most important phase, the training error slowly reduces the MI between the layers and the input (i.e., representation compression). These two phases appear to correspond to fast convergence to a flat minimum (drift) following a random walk, or diffusion, in the vicinity of the training error's flat minimum, as reported in other studies (e.g., \citep{DBLP:journals/corr/abs-1801-02254}).

These observations raised several interesting questions: (a) which properties of the SGD optimization cause these two training phases? (b) how can the diffusion phase improve the generalization performance? (c) can the representation compression explain the convergence of the layers to the optimal IB bound? (d) can this diffusion phase explain the benefit of many hidden layers?

In this work, we attempt to answer these questions. Specifically, we draw important connections between recent results inspired by statistical mechanics and information-theoretic principles. We show that the layers of a DNN indeed follow the behavior described by \cite{shwartz2017opening}. We claim that the reason is the Stochastic Gradient Descent (SGD) optimization mechanism.  We show that the first phase of the SGD is characterized by a rapid decrease in the training error, which corresponds to an increase in the MI with the labels. Then, the SGD behaves like a nonhomogeneous Brownian motion in the weights space, in the proximity of a flat error minimum. This nonhomogeneous diffusion corresponds to a decrease in MI between the layers and the input variable, in ``directions'' that are irrelevant to the target label.  

One of the main challenges in applying information-theoretic measures to real-world data is a reasonable estimation of high-dimensional joint distributions. This problem has been extensively studied over the years (e.g., \citep{Paninski:2003:EEM:795523.795524}) and has led to the conclusion that there is no ``efficient'' solution when the dimension of the problem is large. Recently, several studies have focused on calculating the MI in DNNs using statistical mechanics. These methods have generated promising results in various special cases \citep{gabrie2018entropy}.

In this work, we provide an analytic bound on the MI between consecutive layers, which is valid for any nonlinearity of the units and directly demonstrates the representation's compression during the diffusion phase. Specifically, we derive a Gaussian bound that only depends on the linear part of the layers. This bound gives a superlinear dependence of the layers' convergence time, which enables us to prove the superlinear computational benefit of the hidden layers. Furthermore, the Gaussian bound allows us to study the MI in DNNs in real-world data without estimating it directly.  


\subsection*{Preliminaries and Notations}
\label{notations}
Let $X \in \mathcal{X}$ and $Y \in \mathcal{Y}$ be a pair of random variables of the input patterns and their target label (respectively). We consider the practical setting where $X$ and $Y$ are continuous random variables represented in a finite precision machine throughout this work. This means that both $X$ and $Y$ are practically binned (quantized) into a finite number of discrete values. Alternatively, $X, Y$, may be considered as continuous random variables measured in the presence of small independent additive (Gaussian) noise, corresponding to their numerical precision. We use these two interpretations interchangeably, at the limit of infinite precision, where the limit is applied at the final stage of our analysis. 

We denote the joint probability of $X$ and $Y$ as $p(x,y)$, whereas their corresponding MI is defined as $I(X;Y)=D\left[p(y|x)||p(y)\right]=D\left[p(x|y)||p(x)\right].$ 
We use the standard notation $D[p||q]$ for the Kullback-Liebler (KL) divergence between the probability distributions $p$ and $q$. Let $f_{W^K}(x)$ denote a DNN, with $K$ hidden layers, where each layer consists of $d_k$ neurons, each with some activation function $\sigma_k(x)$, for $k=1,\dots,K$. 
We denote the values of the $k^{th}$ layer by the random vector $T_k$. The DNN mapping between two consecutive layers is defined as $T_k=\sigma_k\left(W_kT_{k-1}\right)$,
where $W_k$ is a $d_k\times d_{k-1}$ real weight matrix. Note that we consider both the weights, $W_k$ and the layer representations, $T_k$, as stochastic entities because they depend on the network's stochastic training rule and the random input pattern (as described in the next section). However, when the network weights are given, the weights are fixed realizations of the random training process (i.e., they are ``quenched''). Note that given the weights, the layers form a Markov chain of successive internal representations of the input variable $X$:  $Y\rightarrow X\rightarrow T_1 \rightarrow...\rightarrow T_K$, and their MI values obey a chain of Data Processing Inequalities (DPI), as discussed by \cite{shwartz2017opening}.

We denote the set of all $K$ layers weight matrices as $W^K=\{W_1,\dots,W_K\}$. Let the {\emph{training sample}}, $S^n =\{ \left(x_1, y_1\right),\dots,$ $\left(x_n,y_n\right) \}$ be a collection of $n$ independent samples from $p(x,y)$. Let $\ell_{W^K} \left(x_i,y_i \right)$ be a (differentiable) loss function that measures the discrepancy between a prediction of the network $f_{W^K}(x_i)$ and the corresponding true target value $y_i$, for a given set of weights ${W^K}$. Then, the empirical error is defined as  
$    \mathcal{L}_{W^K}\left(S^n \right)=\frac{1}{n}\sum_{i=1}^n \ell_{W^K} \left(x_i,y_i \right)~.$
The corresponding error gradients (with respect to the weights) are denoted as $\nabla_{W^K}\mathcal{L}_{W^K}\left(S^n \right)$.
\subsection*{Deep Neural Networks}
\ignore{
    In \citep{shwartz2017opening}, the authors noted that layer's structure in neural networks form a Markov chain of successive representations of the input layer and empirically studied them in the \textit{information plane}  - 
    the plane of the MI values of any other variable with the input variable $X$ and desired output variable $Y$.
    This analysis's rationale was based on two proprieties -  invariance of the MI to invertible re-parameterization and the layer's structure that forms a Markov chain of successive representations. In their work, they showed that DNN's layers converged to the Information Bottleneck (IB) optimal bound \citep{DBLP:journals/corr/Tishby1999}] of the optimal achievable representations of the input $X$. Moreover, they showed that the training optimization has two distinct phases: empirical error minimization (ERM) and representation compression. 
    These phases are characterized by very different signal-to-noise ratios of the stochastic gradients in every layer. In the ERM phase, the gradient norms are much larger than their stochastic fluctuations, resulting in a rapid increase in the MI on the label variable $Y$. In the compression phase, the gradients' fluctuations are much larger than their means, with a minimal influence of the error gradients. 
    This phase is marked by a slow representation's compression or reduction of the MI on the input variable $X$.
}
DNNs opened a new era of machine learning capabilities. They dramatically improved predictive abilities as they outperform state-of-the-art methods in a wide variety of fields, ranging from visual object recognition, speech recognition to drug discovery, genomics, and automatic game playing \citep{DBLP:journals/corr/abs-1303-5778}.
DNNs are multilayer data structures, where each layer consists of multiple simple processing units called neurons (as an analogy to the structure of a biological brain). Each layer's neurons are defined as a linear function of the neurons of the previous layer, followed by a nonlinear activation function.

\subsection*{Training the Network -- the SGD Algorithm}
\label{training_SGD}

Training a DNN corresponds to setting the weights $W^K$ from a given set of samples $S^n$. It is typically done by minimizing the empirical error, which approximates the expected loss. The SGD algorithm is a standard optimization method for this purpose \citep{robbins1951stochastic}.  

Let $S^{(m)}$ be a random set of $m$ samples drawn (uniformly, with replacement) from $S^n$, where $m<n$. We refer to $S^{(m)}$ as a \textit{mini-batch} of $S^n$. Define the corresponding empirical error and gradient of the minibatch as
    ${\mathcal{L}}_{W^K}\left(S^{(m)}\right) = \frac{1}{m}\sum_{\{x_i,y_i\}\in S^{(m)}} \ell_{W^K} \left(x_i,y_i \right)$
    and 
    $\nabla_{W^K} {\mathcal{L}}_{W^K}\left(S^{(m)}\right) = \frac{1}{m}\sum_{\{x_i,y_i\}\in S^{(m)}} \nabla_{W^K} \ell_{W^K} \left(x_i,y_i \right)$ respectively.  
Then, the SGD algorithm  is defined by the update rule:
$    W^{K}(l)=W^{K}(l-1)-\eta \nabla_{W^K(l-1)} {\mathcal{L}}_{W^K(l-1)}\left(S^{(m)}\right)
    \label{eq:SGD}$,
where $W^{K}(l)$ are the weights after $l$ iterations of the SGD algorithm and $\eta\in\mathbb{R}_+$  is the learning rate.

\subsection*{The Different Phases of SGD Optimization}
\label{two_phases_of_SGD}
The SGD algorithm plays a key role in the astonishing performance of DNNs. As a result, it has been extensively studied in recent years, especially in the context of flexibility and generalization \citep{chee2017convergence}. Here, we examine the SGD as a stochastic process that can be decomposed into two separate phases. This idea has been studied in several works \citep{murata1998statistical, jin2017escape, hardt2015train}. Murata argued that stochastic iterative procedures are initiated at some starting state and then move through a fast \textit{transient phase} towards a \textit{stationary phase}, where the distribution of the weights becomes time-independent. However, this may not be when the SGD induces non-isotropic state-dependent noise, as argued, for example,  by \cite{DBLP:journals/corr/abs-1710-11029}.
\ignore{Specifically, the SGD process consists of a short transient phase in which it locates a neighborhood of (local) minimum. Then, it oscillates around that point during a longer second phase. The gradient of the process characterizes these phases; in the transient phase, the gradients are typically in the same direction with low variance, while in the second phase, they point to different directions with a corresponding high variance \citep{chee2017convergence}.
} 

In contrast, \cite{shwartz2017opening} described the transient phase of the SGD as having two very distinct dynamic phases. The first is a \textit{drift} phase, where the means of the error gradients in every layer are large compared to their batch-to-batch fluctuations. This behavior indicates small variations in the gradient directions, or \emph{high-SNR gradients}. In the second part of the transient phase, which they refer to as \textit{diffusion}, the gradient means become significantly smaller than their batch-to-batch fluctuations -- \emph{low-SNR gradients}. The transition between the two phases occurs when the training error saturates and the weights growth is dominated by the gradient batch-to-batch fluctuations. Typically, most SGD updates are expended in the diffusion phase before reaching Murata's stationary phase. In this work, we rigorously argue that this diffusion phase causes the representation compression, the observed reduction in $I(T_k; X)$, for most hidden layers.    

\ignore{

\begin{figure}[t]
  \centering
  \subfloat[\label{fig:gradients}The norms' means, the STD of the gradients and the norm weights, during the training process (in a log-log scale)] 
  {\includegraphics[width=0.47\columnwidth]{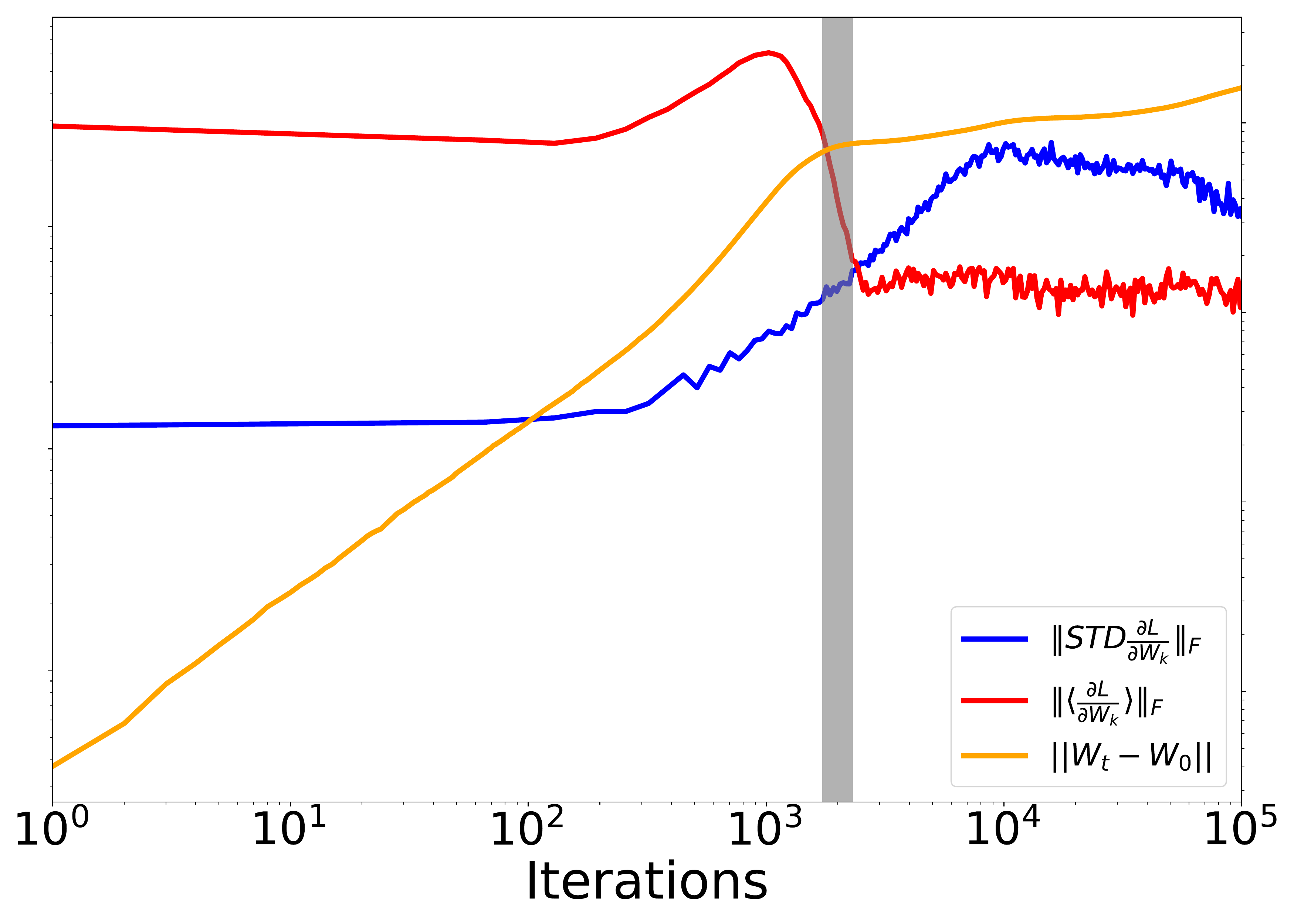}}
\qquad
\subfloat[The evolution of layers during the training on the information plane]
{\includegraphics[width=0.47\columnwidth]{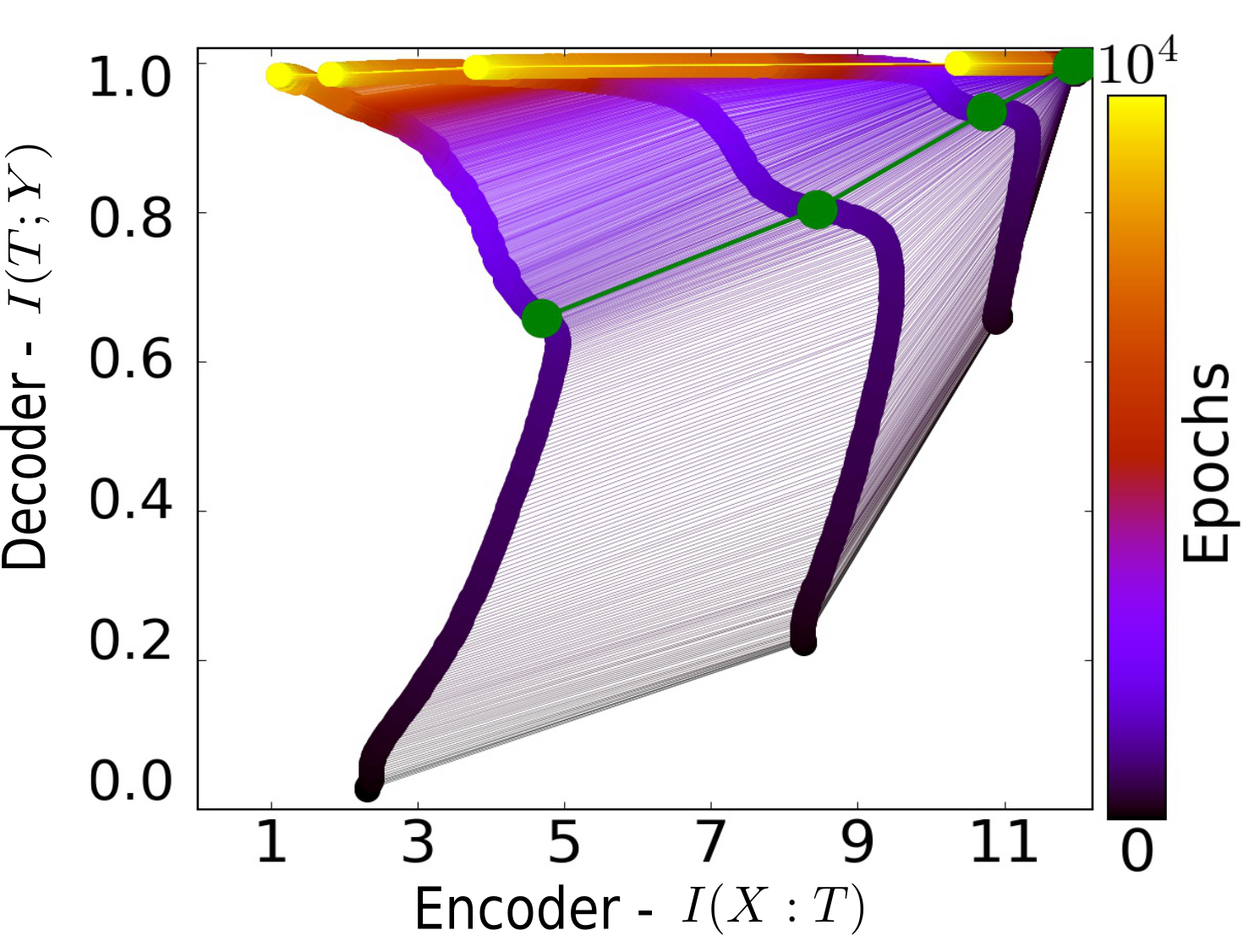}}
 \caption{The two phases of the training process -  transition between the drift phase to the diffusion.}
    \label{fig:weights_norm}
\end{figure}
}
\subsection*{Drift and Diffusion with SGD}
\label{modeling_the_diffusion}
It is well known that the discrete-time SGD can be considered as an approximation of a continuous-time stochastic gradient flow if the discrete-time iteration parameter $l$ is replaced by a continuous parameter $\tau$. 
\cite{li2015stochastic}  showed that when the minibatch gradients are unbiased with bounded variance, the discrete-time SGD is an approximation of a continuous-time Langevin dynamics,
\ignore{
Lemma \ref{cont-sgd-lemma} introduces an important property that results from this approximation.      
}
\ignore{Gradient over the minibatch is unbiased with respect to the full gradient and has bounded variance.
    If the examples in a minibatch are sampled with replacement,  the variance of minibatch gradients is 
    $var\left(\nabla_W\hat{\mathcal{L}}^{S}\left(w_{1\dots K}\right)\right)=\frac{D\left(w_{1\dots K}\right)}{|S|}$
    where 
    \begin{align}
        D\left(w_{1,\dots , K}\right) = \left(\frac{1}{m}\sum_{i=1}^{m}{\nabla_W\ell_i\left(w_{1\dots K}\right){\nabla_W\ell_i\left(w_{1\dots K}\right)}}^T\right) - {\nabla_W\mathcal{L} \left(w_{1\dots K}\right){\nabla_W\mathcal{L}\left(w_{1\dots K}\right)}}^T\succeq 0
    \end{align}
    is the sample covariance matrix.
    
    Note that $D(w_{1 \dots K})$ is independent of the learning rate  and the batch size. It only depends on the weights, architecture and loss.}
\ignore{
\begin{lemma}{\citep{li2015stochastic}}
    \label{cont-sgd-lemma}
    
    Assume that the mini-batches are sampled with replacement. Then, the update rule of the SGD processes (\ref{eq:SGD}) may be viewed as a discretization  of the following continuous-time equation 
    \end{lemma}
    }
    \begin{align}
        \label{eq:sgd_cont}
        dW^K(\tau)=- \nabla_{W^K(\tau)} \mathcal{L}_{W^K(\tau)}\left(S_n\right)d\tau+\sqrt{2 \beta^{-1}C\left(W^K(\tau)\right)}dB\left(\tau\right)
    \end{align}
    
    \ignore{
        \begin{align}
            \label{eq:sgd_cont}
            dw_{1\dots K}\left(\tau\right)=- \nabla \mathcal{L}\left(w_{1\dots K}\left(\tau\right)\right)d\tau+\sqrt{2 \beta^{-1}D\left(w_{1,\dots K}\right)}dB\left(\tau\right)
    \end{align}}
    \noindent where $C\left(W^K(\tau)\right)$ is the sample covariance matrix of the weights, 
    $B(\tau)$ is a standard Brownian motion (Wiener process) and $\beta$  is the Langevin temperature constant. 
The first term in (\ref{eq:sgd_cont}) is called the gradient flow or drift component, and the second term corresponds to random diffusion.  Although this stochastic dynamics holds for the entire SGD training process, the first term dominates the process during the high SNR gradient phase, while the second term becomes dominant when the gradients are small, due to the saturation of the training error in the low SNR gradient phase. Hence, these two SGD phases are referred to as  drift and diffusion.  

The \textit{mean $L_2$ displacement} (MSD) measures the Euclidean distance from a reference position over time, which is used to characterize a diffusion process. Normal diffusion processes are known to exhibit a power-law MSD in time, $\mathbb{E}\left[\left\Vert W^K(\tau)-W^K(0)\right\Vert _2\right]\sim \gamma t^\alpha$, where $t$ is the diffusion time, $\gamma$ is related to the diffusion coefficient, and $0<\alpha\leq0.5$ is the diffusion exponent. For a standard flat space diffusion, the MSD increases as a square root of time ($\alpha=0.5$). \cite{hu2017diffusion} showed (empirically) that the weights' MSD in a DNN trained with SGD, indeed behaves (asymptotically) like a normal diffusion, where the diffusion coefficient $\gamma$ depends on the batch size and learning rate.
In contrast, \cite{hoffer2017train} showed that the weights' MSD demonstrates a much slower logarithmic increase. This type of dynamics is also called "ultra-slow" diffusion.

\section*{Information Plane Analysis}
\addcontentsline{toc}{section}{Information Plane Analysis}

Following \cite{DBLP:journals/corr/TishbyZ15} and  \cite{shwartz2017opening}, we study the layer representation dynamics in the two-dimensional plane $(I(X;T_k), I(T_k;Y))$. Specifically, 
for any input and target variables, $X, Y$, let $T\triangleq T(X)$ denote a representation, or an encoding (not necessarily deterministic), of $X$. Clearly, $T$ is fully characterized by its \emph{encoder}, the conditional distribution $p(t|x)$. Similarly, let $p(y|t)$ denote any (possibly stochastic) \emph{decoder} of $Y$ from $T$. 
Given a joint probability function $p(x,y)$, the \textit{information plane} is defined the set of all possible pairs $I(X;T)$ and $I(T;Y)$ for any possible representation, $p(T|X)$. 

It is evident that not all points on the plane are feasible (achievable), as there is a tradeoff between these quantities; the more we compress $X$ (reduce $I(X; T)$), the less information can be maintained about the target, $I(T; Y)$. 

Our analysis is based on the fundamental role of these two MI quantities. We argue that for large-scale (high dimensional $X$) learning, for almost all (\emph{typical}) input patterns, with mild assumptions (ergodic Markovian input patterns): (i) the MI values concentrate with the input dimension, (ii) the minimal sample complexity for a given generalization gap is controlled by $I(X; T)$, and (iii) the accuracy - the generalization error - is governed by $I(T; Y)$, with the Bayes optimal decoder representation. 

We argue that these two MI quantities characterize the sample-size and the accuracy tradeoff of large-scale representation learning. For DNNs, this amounts to a dramatic reduction in the complexity of the analysis of the problem. 
We discuss these ideas in the following sections and prove the connection between the input representation compression, the generalization gap (the difference between training and generalization errors), and the minimal sample complexity (Theorem \ref{compression_thm} below).

\subsection*{Label Information and Generalization Error}
The optimization of MI quantities is not a new concept in supervised or unsupervised learning \citep{deco2012information,linsker1988self,painsky2016generalized}. This is not surprising, as it can be shown that $I(T;Y)$ corresponds to the irreducible error when minimizing the logarithmic loss \citep{painsky2018universality,harremoes2007information}. 
Here, we emphasize that $I(T;Y)$, for the optimal decoder of the representation $T$, governs all reasonable generalization errors (under the mild assumption that label $y$ is not entirely deterministic; $p(y|x)$ is in the interior of the simplex, $\Delta(Y)$, for all typical $x\in X$). 
First, note that for the Markov chain $Y-X-T$, 
$I(T;Y)=I(X;Y)-\mathbb{E}_{X,T}D\left[p(y|x)||p(y|t)\right]$.
By using the Pinsker inequality \citep{cover2012elements}, the variation distance between the optimal and the representation decoders can be bounded by their KL divergence,
\begin{align}
   D\left[p(y|x)||p(y|t)\right] \geq \frac{1}{2\ln{2}}\big|p(y|x)-p(y|t)\big|_1^2.
\end{align}
Hence, by maximizing $I(T;Y)$ we minimize the expected \textit{variation risk} between the representation decoder $p(y|t)$ and $p(y|x)$. For more similar bounds on the error measures see \citep{painsky2018bregman}. 

\subsection*{Representation Compression and Sample Complexity}
The \textit{Minimum Description Length} (MDL) principle \citep{rissanen1978modeling} suggests that the best representation for a given set of data is the one that leads to the minimal code length needed to represent the data. This idea has inspired the use of $I(X; T)$ as a regularization term in many learning problems (e.g., \cite{chigirev2004optimal,painsky2018information}). Here, we argue that $I(X; T)$ plays a much more fundamental role; we show that for large scale learning  (high dimensional $X$) and typical input patterns, $I(X; T)$ controls the sample complexity of the problem, given a generalization error gap. 

\begin{thm}[Input Compression bound]
\label{compression_thm}
Let $X$ be a $d$-dimensional random variable that obeys an ergodic Markov random field probability distribution, asymptotically in $d$. Let $T\triangleq T(X)$ be a representation of $X$ and denote by $T_m=\{(t_1,y_1),\dots,(t_m,y_m)\}$ an $m$-sample vector of $T$ and $Y$, generated with $m$ independent samples of $x_i$, with $p(y|x_i)$ and $p(t|x_i)$.
Assume that $p(x,y)$ is bounded away from 0 and 1 (strictly inside the simplex interior). Then, for large enough $d$, with probability $1-\delta$, the \textit{typical expected} squared generalization gap satisfies 
\begin{equation}
    \big|\mathcal{L}\left(T_m \right)-\mathbb{E}_{T_m}\left[\mathcal{L}\left(T_m \right)\right]\big|^2 \leq \frac{2^{I(X;T)}+\log{\frac{2}{\delta}}}{2m}.
\end{equation}
where the typicality follows the standard \textit{Asympthotic Equipartition Property (AEP)} \citep{cover2012elements}. 
\end{thm}
The proof of this theorem is given in Appendix A. This theorem is also related to the bound proved by \cite{SHAMIR20102696}, with the typical representation cardinality, $|T(X)|\approx 2^{I(T;X)}$. 
The ergodic Markovian assumption is common in many large-scale learning problems. It means that $p(x) \approx \prod_{i=1:d} p(x_i|Pa(x_i))$, where $Pa(x_i)$ is a finite set of adjacent "parents" of $x_i$ in the $d$ dimensional pattern $X$.

The consequences of this input-compression bound are quite striking: the generalization error decreases exponentially with $I(X; T)$, once $I(T; X)$ becomes smaller than $\log{2m}$ - the query sample-complexity. Moreover, it means that $M$ bits of representation compression, beyond $\log{2m}$, are equivalent to a factor of $2^M$ training examples. The tightest bound on the generalization bound is obtained for the most compressed representation or the last hidden layer of the DNN. The input-compression bound can yield a tighter and more realistic sample complexity than any of the worst-case PAC bounds with any reasonable estimate of the DNN class dimensionality, as typically, the final hidden layers are compressed to a few bits.

Nevertheless, two important caveats are in order. First, the layer representation in deep learning is learned from the training data; hence, the encoder, the partition of the typical patterns $X$, and the \emph{effective "hypothesis class"}, depend on the training data. This can lead to considerable overfitting. Training with SGD avoids this potential overfitting because of the way the diffusion phase works. Second, for low $I(T; Y)$, there are exponentially (in $d$) many random encoders (or soft partitions of $X$) with the same value of $I(T; X)$. This seems to suggest that there is a missing exponential factor in our estimate of the hypothesis class cardinality. However, note that the vast majority (almost all) of these possible encoders are never encountered during a typical SGD optimization. Moreover, as $I(T; Y)$ increases, the number of such random encoders rapidly collapses to $O(1)$ when $I(T; Y)$ approaches the optimal IB limit, as we show next.                

\subsection*{The Information Bottleneck Limit}

As presented above, we are interested in the boundary of the achievable region in the information plane, or encoder-decoder pairs that minimize the sample complexity (minimize $I(X;T)$) and generalize well (maximize $I(T;Y)$).

These optimal encoder-decoder pairs are given precisely by the IB framework \citep{DBLP:journals/corr/Tishby1999}, which is formulated by the following optimization problem: $\min_{p(t|x)} I\left(X;T\right)-\beta I\left(T;Y\right), $
over all possible encoders-decoders pairs that satisfy the Markov condition $Y-X-T$.  Here, $\beta$ is a positive Lagrange multiplier associated with the decoder information on $I(T; Y)$, which determines the representation's complexity.

The IB limit defines the set of optimal encoder-decoder pairs for the joint distribution $p(x,y)$. Furthermore, it characterizes the achievable region in the information plane, similar to Shannon's rate-distortion theory \citep{cover2012elements}. This analysis, also determines the optimal tradeoff between sample complexity and generalization error. The IB can only be solved analytically in exceptional cases (e.g., jointly Gaussian $X, Y$ \citep{chechik2005information}). In general, a (locally optimal) solution can be found by iterating the self-consistent equations, similar to the Blahut-Arimoto algorithm in rate-distortion theory \citep{DBLP:journals/corr/Tishby1999}. For general distributions, no efficient algorithm for solving the IB is known, though there are several approximation schemes \citep{chalk2016relevant,painsky2017gaussian}.
\ignore{
Besides, several important approximation techniques were introduced in recent years \citep{chalk2016relevant,painsky2017gaussian}.
}

The self-consistent equations are satisfied along the \textit{information curve}. This monotonic curve separates between the achievable and non-achievable regions on the information plane. Notice that for smooth joint distributions $p(x,y)$, the information curve is strictly concave with a unique slope, $\beta^{-1}$, at every point, and a finite slope at the origin. In these cases of interest (where $Y$ is not a deterministic function of $X$), every value of $\beta$ corresponds to a single point on the information curve with a corresponding optimal encoder-decoder pair.

\ignore{

    Two properties of the MI are fundamental in the context of DNNs. 
    The first is its invariance to invertible transformations:
    \begin{equation}
        \label{eqn:invariance}
        I\left(X;Y\right)=I\left(\psi(X);\phi(Y)\right)
    \end{equation}
    for any invertible functions $\phi$ and $\psi$.
    
    The second is the Data Processing Inequality (DPI) [\citep{Cover:2006}]: 
    for any 3 variables that form a Markov chain $X\rightarrow Y \rightarrow Z$,
    \[
    I\left(X;Y\right) \ge I(X;Z)
    \]}


\section*{The Information Plane and SGD Dynamics for DNNs}
\addcontentsline{toc}{section}{The Information Plane and SGD Dynamics for DNNs}

\label{information plane and Deep Neural Networks}
By applying the DPI to the Markov chain of the DNN's layers, we obtain the following chains: 
\begin{align*}
I(X;Y) & \ge I(T_1;Y) \ge I(T_2;Y) \ge ... I(T_k;Y) \ge I(\hat{Y};Y)\\
H(X) & \ge I(X;T_1) \ge I(X;T_2) \ge ... I(X;T_k) \ge I(X;\hat{Y}) .
 \label{DPI-1}
\end{align*}
where $\hat{Y}$ is the output of the network. 

The pairs $(I(X;T_k),I(T_k,Y))$, for each SGD update, form a unique concentrated \textit{information path} for each layer of a DNN, as demonstrated by \cite{shwartz2017opening}.   


As we are only interested in the information that flows through the network, invertible transformations of the representations that preserve information generate equivalent representations even if the individual neurons encode entirely different input features. Therefore, we quantify the representations by two numbers, or order parameters for each layer - the MI of $T$ with the input $X$ (the decoder's information) and with the desired output $Y$ (the information of the encoder). These quantities are invariant to any invertible re-parameterization of $T$. 

For any fixed realization of the weights, the network is, in principle, a deterministic map. This does not imply that information is not lost between the layers; the layers' inherent finite precision, with possible saturation of the nonlinear activation function $\sigma_k$, can result in non-invertible mapping between the layers. Moreover, we argue below that for large networks, this mapping becomes effectively stochastic due to the diffusion phase of the SGD. 

On the other hand, the  paths of the layers in the information plane are invariant to invertible transformations of the representation $T_k$. Thus, the same paths are shared by very different weights and architectures and possibly different encoder-decoder pairs. This freedom is drastically reduced when the target information, $I(T_k,Y)$, increases, and the layers approach the IB limit. Minimizing the training error, together with standard uniform convergence arguments, clearly increases $I(T; Y)$. This raised the question what in the SGD dynamics can lead to the observed representation compression, which further improves the generalization?  
Moreover, can the SGD dynamics push the layer representations to the IB limit, as claimed in \cite{shwartz2017opening}? 

We provide affirmative answers to both questions, using the properties of the drift and diffusion phases of the SGD dynamics.

\ignore{
\begin{figure}[t]
    \vskip 0.1in
    \centerline{\includegraphics[width=0.55\columnwidth]{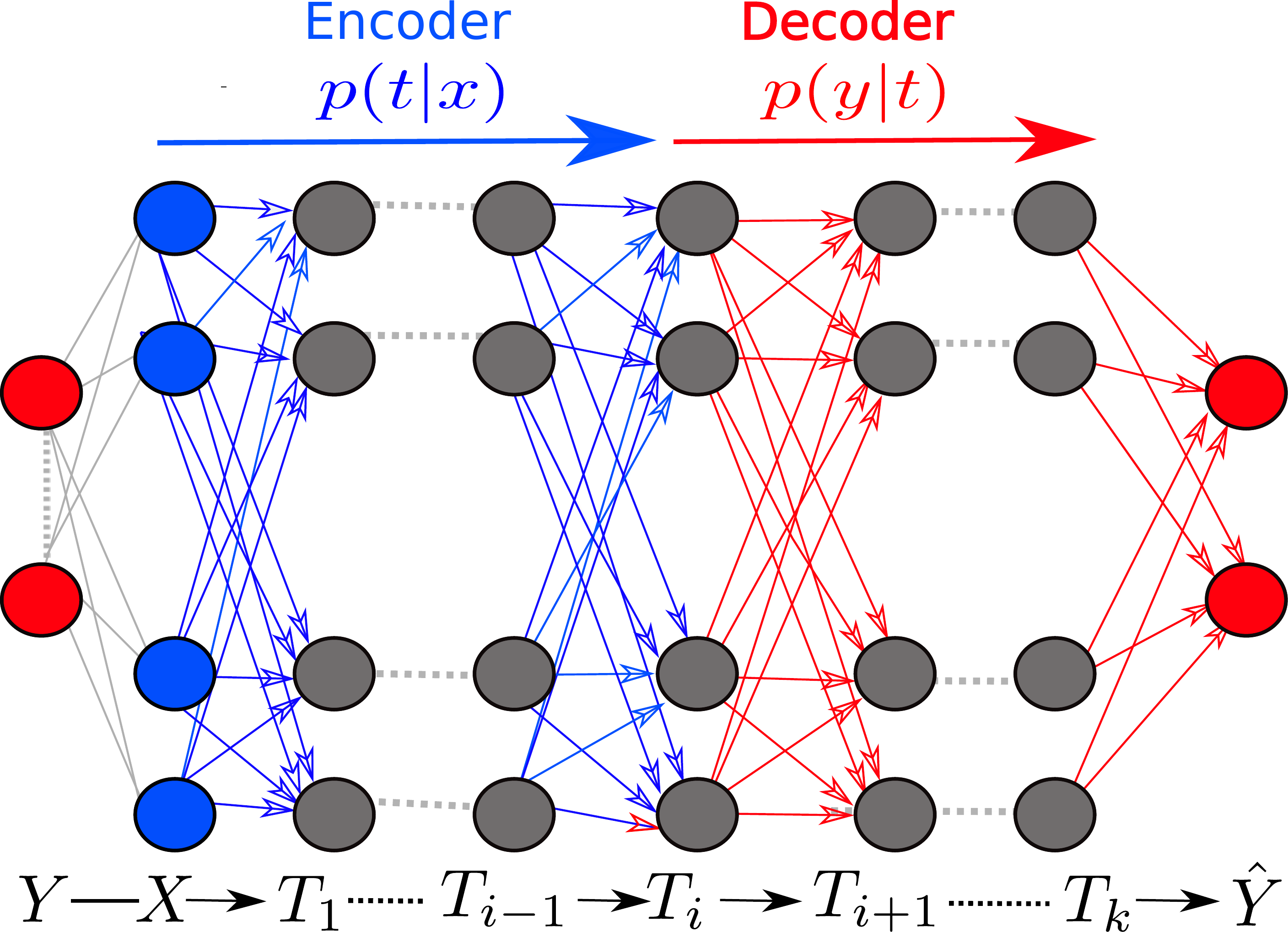}}
    \caption{The DNN layers form a Markov chain of successive representations of the input layer $X$. Every representation of the input, $T$, is defined by an encoder, $p(t|x)$, and a decoder $p(y|t)$, and may be characterized on the information plane as a pair of coordinates: $I(X;T)$ and $I(T;Y)$.} 
    \label{DNN-layers}
    \vskip -0.1in
\end{figure}
}
\ignore{
    \subsection{Successive refinement approximation} 
    A DNN forms a Markov chain of representations. Therefore, it is natural to approximate the original IB bound as a sequence of smaller IB problems, where each problem corresponds to a single layer. We refer to such an approximation as a successive refinement (as in \citep{cover2012elements}). Specifically, assume that the layer $T_{k-1}$ is fixed. Then, the successive refinement approximation seeks a pair of optimal encoder-decoder with respect to $T_{k-1}$ and $Y$ (as opposed to $X$ and $Y$). The quality of the approximation depends on the quality of the previous layer. In other words, the optimal Information Curve induced by sequentially solving the IB problem for each layer separately is a lower bound to the original IB curve. When this bound equal to the IB curve, we say that the problem is successively refinable \citep{cover2012elements}.   
    
    

    In a standard training procedure, we would like to minimize the loss function of the network. However, our goal is to show the converges of all the layers to the information curve. 
    
    To do so, we prove a new lower bound on the standard cross-entropy loss function, which bounds the information between consecutive layers. Using this bound, the minimization of the loss function will minimize the $D$ of the individual layers. We also know that the minimization of the cross-entropy is equivalent to increasing the MI. Therefore, each layer's information will increase during the learning and converge to the optimal information curve with its constraint on the information with $X$. 
    
    Consider the cross-entropy loss function, $D\left(p(y|x)||p\left(y|t_K\right)\right)$ and denote the $Q_k$ as the set of distributions with linear constrain on the information with X, namely $Q_k=\{p\left(y|\bar{t}_k\right)|I\left(X;\bar{T_k}\right)=I\left(X;T_k\right)\}, 1\leq k\leq K$ 
    Denote the reverse I projections of a distribution $p(y|x)$ onto a set of distributions $Q_k$ , $1\leq k \leq K$ in $p(y|t_k)^\ast$, namely  
    \[p(y|t_k)^\ast=\argmin_{q\in Q_k}D\left[p(y|x)||q \right]
    \]
    Using Pythagorean theorem \citep{csiszar2003information}, if $Q_k$ is log-convex set, the following inequality holds 
    \begin{align}
        D\left(p\left(y|x\right)||p\left(y|t_K\right)\right) \geq 
        D\left(p\left(y|x\right)||p\left(y|t_k\right)^\ast\right)  +
        D\left(p\left(y|t_k\right)^\ast||p\left(y|t_K\right)\right) 
    \end{align} Due to the minimization of the right side in the optimization process, we get that the right side above will also decrease, and that the cross entropy of the k-th layer -  $D\left(p\left(y|x\right)||p\left(y|t_k\right)^\ast\right)$ can only decrease during the learning as requested.
    }
\ignore{
\subsection*{SGD Dynamics in the Information Plane}
\cite{shwartz2017opening} studied the Information Paths of DNNs on the information plane;  they evaluated and visualized the pairs of $I(X; T_k)$ and $I(T_k; Y)$, along the learning process and compared different network architectures. Specifically, they focused on the Information Curve dynamics (the change of its location on the plane) during the SGD process. 
Interestingly, they showed that the SGD consists of two distinct phases on the information plane. First, they observe a short phase in which $I(T_k; Y)$ increases quite rapidly. They then observed a significantly longer phase (which we called representation compression), in which $I(X; T_k)$ drops. 
In other words, they claimed that the SGD first learns the target and then compresses the representation. This behavior seems to be related to the drift and diffusion phases, discussed in Section \ref{two_phases_of_SGD}. It is not surprising that $I(T_k; Y)$ increases during the drift phase; the drift phase is characterized by a rapid decrease in the cross-entropy loss function. This is equivalent to an increase in the (empirical) MI. On the other hand, the correspondence between diffusion and compression appears to be quite vague. We study this phenomenon in the following section. 
}


\ignore{
\begin{figure}[ht]
    \centerline{\includegraphics[width=\columnwidth]{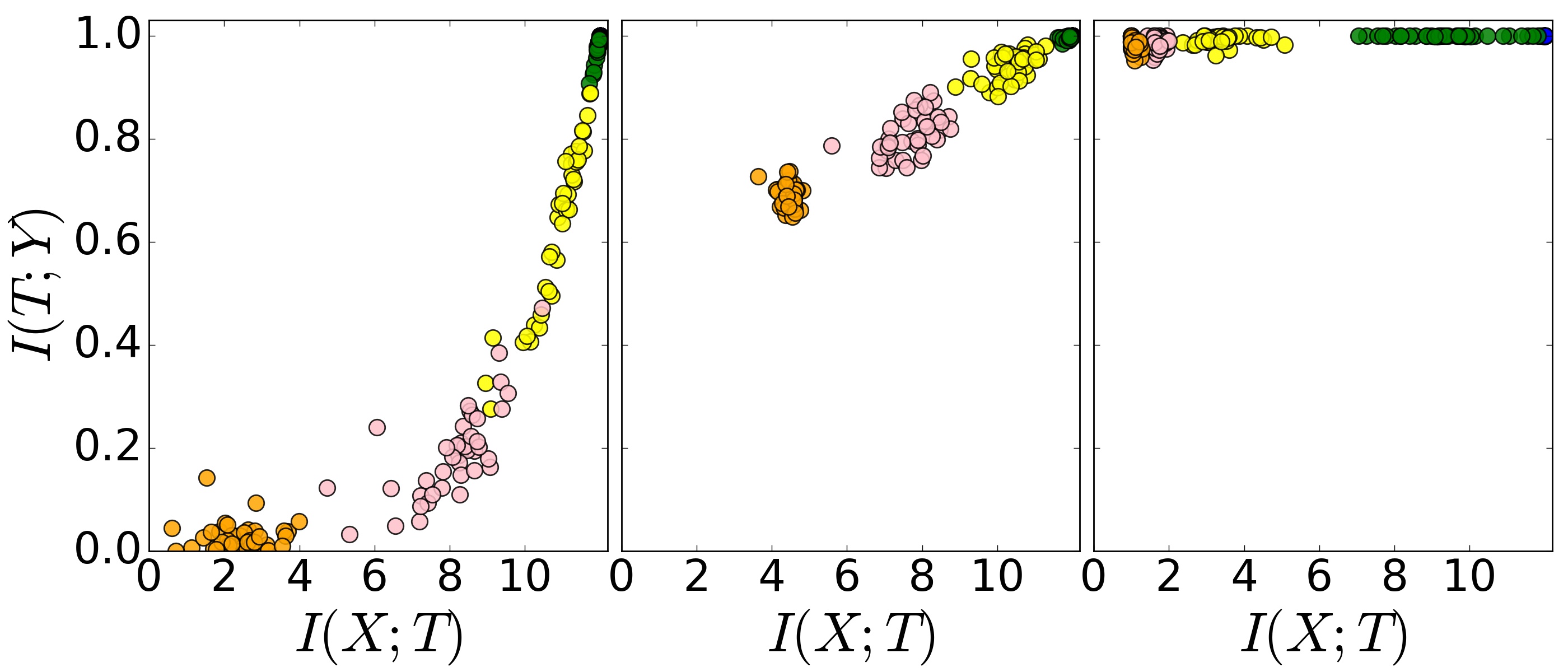}}
    \caption{Snapshots of layers (different colors) of $50$ randomized networks during the SGD optimization process on the information plane (in bits).  Left - initial weights. Center -  at $400$ epochs. Right -  after $9000$ epochs}
    \label{opt_process}
\end{figure}
\ignore}

\subsection*{Representation Compression by Diffusion}
\label{compression_and_diffusion}
This section quantifies the roles of the drift and diffusion SGD phases and their influence on the MI between consecutive layers. Specifically, we show that the drift phase corresponds to an increase in information with the target label $I(T_k; Y)$, whereas the diffusion phase corresponds to representation compression or reduction of the $I(X; T_k)$. The representation compression is accompanied by further improvement in the generalization.

The general idea is as follows: the drift phase increases $I(T_k; Y)$ as it reduces the cross-entropy empirical error. On the other hand, the diffusion phase in high-dimensional weight space effectively adds an independent nonuniform random component to the weights, mainly in the directions that do not influence the loss - i.e., \emph{irrelevant directions}. This results in a reduction of the SNR of the patterns' irrelevant features, which leads to a reduction in $I(X; T_k)$, or representation compression. We further argue that different layers filter out different irrelevant features, resulting in their convergence to different information plane locations.     

\subsection*{The SGD Compression Mechanism}
\ignore{In this section, we introduce a novel asymptotic result which shows that under several assumptions, a DNN compresses the original representation $X$ during the diffusion phase, in the sense that $I(X;T_k)$ reduces with the iterations of the SGD along this phase.}

First, DPI implies that $I(X;T_k)\leq I(T_{k-1};T_k)$. We focus on the second term during the diffusion phase and prove an asymptotic upper bound for $I(T_{k-1};T_k)$, which reduces sub-linearly with the number of SGD updates.

For clarity, we describe the case where $T_{k} \in \mathbb{R}^{d_k}$ is a vector and $T_{k+1} \in \mathbb{R}$ is a scalar. The generalization to higher $d_{k+1}$ is straightforward. We examine the network during the diffusion phase, after $\tau$ iterations of the SGD beyond the drift-diffusion transition.
For each layer,$k$, the weights matrix, $W^k(\tau)$ can be decomposed as follows, 
\begin{equation}
W^k(\tau) = {W^k}^{\star}+\delta W^k(\tau). 
\end{equation}
The first term, ${W^k}^{\star}$, denotes the weights at the end of the drift phase ($\tau_0=0$) and remains constant with increasing $\tau$. As we assume that the weights converge to a (local, flat) optimum during the drift phase, ${W^k}^{\star}$ is close to the weights at this local optimum. The second term, $\delta W^k(\tau)$, is the accumulated Brownian motion in $\tau$ steps due to the batch-to-batch fluctuations of the gradients near the optimum. For large $\tau$ we know that $\delta W^k(\tau) \sim \mathcal{N}(0,\tau C(W^k(\tau_0)))$ where $\tau_0$ is the time of the beginning of the diffusion phase. Note that at any given $\tau$, we can treat the weights as a fixed (quenched) realization, $w^k(\tau)$, of the random Brownian process $W^k(\tau)$. We can now model the mapping between the layers $T_k$ and $T_{k+1}$ at that time as 
\begin{equation}
	\label{channel}
	T_{k+1}=\sigma_k \left({w^*}^T T_k+{\delta w^k(\tau)}^T T_k + Z\right)
\end{equation}
where $w^* \in \mathbb{R}^{d_k}$ is the SGD's empirical minimizer, and $\delta w \in \mathbb{R}^{d_k}$ is a realization from a Gaussian vector $\delta w \sim \mathcal{N}(0,C_{\delta w})$, of the Brownian process. In addition, we consider $Z \sim \mathcal{N}(0,\sigma^2_z)$ to be the small Gaussian measurement noise, or quantization, independent of $\delta w^k$ and $T_k$. This standard additive noise allows us to treat all the random variables as continuous. 

For simplicity, we assume that the $d_k$ components of $T_k$ have zero mean and are asymptotically independent for $d_k \rightarrow \infty$, and that $\lim_{ d_k \rightarrow \infty}{w^*}^T\delta w =0$ almost surely. 
\begin{prop1}
	\label{CLT_prop}
	Assume that the moments of $T_k$ are finite. Further assume that the components of $w^*$ and $\delta w(\tau)$ are \textit{in-general-positions}, satisfying $\lim_{d_k \rightarrow \infty}{\sum_{i=1}^{d_k} {w^*_i}^4}\big/{\left(\sum_{i=1}^{d_k} {w^*_i}^2\right)^2}=0$ and $\lim_{d_k \rightarrow \infty}{\sum_{i=1}^{d_k} {\delta w_i}^4}\big/{\left(\sum_{i=1}^{d_k} {\delta w_i}^2\right)^2}=0$ almost surely.  Then, 
	\begin{equation}
		\frac{1}{\sqrt{\sigma_{T_k}^2}}\left[\frac{{w^*}^T T_k}{||w^*||_2} \quad \frac{{\delta w}^T T_k}{||\delta w||_2}   \right]^T \xrightarrow[d_k\rightarrow \infty]{\mathcal{D}} \mathcal{N}(0,I)
	\end{equation}
	almost surely, where $\sigma_{T_k}^2$ is the variance of the  components of $T_k$. \end{prop1}
\noindent A proof for this CLT proposition is given in Appendix B. 

Proposition \ref{CLT_prop} shows that under the standard conditions above, 
${w^*}^T T_k $ and ${\delta w}^T T_k$ are asymptotically jointly Gaussian and independent, almost surely. We stress that the components of $T_k$ do not have to be identically distributed to satisfy this property; Proposition \ref{CLT_prop} may be adjusted for this case with different normalization factors. 
Similarily, the i.i.d. assumption on $T_k$ can be relaxed to Markovian ergodic.
It is easy to verify that Proposition \ref{CLT_prop} can be extended to the general case where $w^*, \delta w \in \mathbb{R}^{d_k \times d_{k+1}}$, under similar  general position conditions, with almost sure orthogonality of $w^*$ and $\delta w$. 
\ignore{we have that  

\begin{equation}
	\frac{1}{\sqrt{\sigma^2_{T_k}}}\left[\left({w^*}^Tw^*\right)^{-1/2}{w^*}^T T_k \quad \left(\delta w^T\delta w\right)^{-1/2} \delta w^T T_k\right]   \xrightarrow[d_k\rightarrow \infty]{\mathcal{D}} \mathcal{N}(0,I) \end{equation}
almost surely. This means that under the conditions stated above, $w^*T_k$ and $\delta w T_k$ are asymptotically jointly Gaussian and statistically independent. 
}

We can now bound the mutual information between $T_{k+1}$ and the linear projection of the previous layer $W^*T_k$, during the diffusion phase, for sufficiently high dimensions $d_k, d_{k+1}$, under the above conditions. Note that in this case, Equation \ref{channel} behaves like an additive Gaussian channel where ${w^*}^T T_k$ is the signal and $\delta w^T T_k + Z$ is an independent additive Gaussian noise (i.e., independent of signal and normally distributed). Hence, for sufficiently large $d_k$ and $d_{k+1}$, we can write
\begin{align}
	\label{upper_bound}
	I(T_{k+1}; T_k | w^*)\leq & I(T_{k+1}; {w^*}^T T_k | w^*)\\\nonumber \leq& I\left({w^*}^T T_k+{\delta w}^T T_k + Z; {w^*}^T T_k |{w^*}   \right) \\\nonumber 
	=&\frac{1}{2}\log\left(\frac{{\left|\sigma_{T_k}^2 {w^*}^T{w^*} +\sigma_{T_k}^2{\delta w}^T {\delta w} + \sigma^2_z I\right|}}{{\left|\sigma_{T_k}^2{\delta w}^T {\delta w} + \sigma^2_z I\right|}}  \right)  
\end{align}
almost surely, where the first inequality is due to DPI for the Markov chain $T_k - {w^*}^T T_k - T_{k+1}$. Finally, we apply an orthogonal eigenvalue decomposition to the multivariate Gaussian channel in Equation \ref{upper_bound}. Let ${\delta w}^T {\delta w}=Q\Lambda Q^T$ where $QQ^T=I$ and $\Lambda$ is a diagonal matrix whose diagonal elements are the corresponding eigenvalues, $\lambda_i$, of ${\delta w}^T {\delta w}$. Then, we have that
\begin{align}
	\label{SVD}
	{\left|\sigma_{T_k}^2 {w^*}^T{w^*} +\sigma_{T_k}^2{\delta w}^T {\delta w} + Z\right|}=&\sigma^2_{T_k} |Q|\cdot|Q^T {w^*}^T{w^*} Q + \Lambda + \frac{\sigma^2_z}{\sigma^{2}_{T_k}} Q^T Q|\cdot|Q^T|\\\nonumber =
	&\sigma^2_{T_k} |Q^T {w^*}^T{w^*} Q + \Lambda + \frac{\sigma^2_z}{\sigma^{2}_{T_k}}I| \\\nonumber  \leq &\sigma^2_{T_k} \prod_{i=1}^{d_{k+1}} \left(A_{ii}+\lambda_i+\frac{\sigma^2_z}{\sigma^{2}_{T_k}}\right)
\end{align}
where $A \triangleq Q^T {W^*}^T{W^*} Q$. The last inequality is due to the Hadamard inequality. Plugging Equation \ref{SVD} into Equation \ref{upper_bound} yields that for sufficiently large $d_k$ and $d_{k+1}$,
\begin{align}
	\label{upper_bound_final}
	I(T_{k+1}; T_k | w^*)\leq& \frac{1}{2}\log\left(\frac{\prod_{i=1}^{d_{k+1}} \left(A_{ii}+\lambda_i+\frac{\sigma^2_z}{\sigma^{2}_{T_k}}\right)}{\prod_{i=1}^{d_{k+1}} \left(\lambda_i+\frac{\sigma^2_z}{\sigma^{2}_{T_k}}\right)}  \right)  \\\nonumber=
	&\frac{1}{2} \sum_{i=1}^{d_{k+1}}\log  \left( 1+ \frac{A_{ii}}{\lambda_i+\frac{\sigma^2_z}{\sigma^{2}_{T_k}}} \right) \xrightarrow[\sigma^2_z\rightarrow 0]{} \frac{1}{2}\sum_{i=1}^{d_{k+1}}  \log  \left( 1+ \frac{A_{ii}}{\lambda_i} \right). 
\end{align}
As previously established, $\delta w$ is a Brownian motion along the SGD iterations during the diffusion phase. This process is characterized by a low (and fixed) variance of the informative gradients (relevant dimensions), whereas the remaining irrelevant directions suffer from increasing variances as the diffusion proceeds (see, e.g. \cite{sagun2017empirical,zho2018,jastrzkebski2017three}). In other words, we expect the ``informative'' $\lambda_i$  to remain fixed, while the irrelevant consistently grow as sub-linearly with time. Denote the set of ``informativ'' directions as $\Lambda^*$ and the set of ``non-informative'' as $\Lambda_{NI}$. Then our final limit, as the number of SGD steps grows, is 
$$I(T_{k+1}; T_k | w^*)\leq \frac{1}{2}\sum_{\lambda^*_i \in \Lambda^*}  \log  \left( 1+ \frac{A_{ii}}{\lambda^*_i}\right).$$ Note that for real problems the distinction between informative and non-informative directions may not be that sharp and we can expect a gradual (exponential asymptotically) decrease of $A_{ii}$ with $i\rightarrow \infty$. Which directions are compressed and which are preserved depend on the required compression level. This is why different layers converge to different values of $I(T_k;X)$. 
\subsection*{Relation to Other Works}
The analysis above suggests that the SGD compresses during the diffusion phase in many directions of the gradients. We argue that these directions are the ones in which the gradients' variance is increasing (non-informative), whereas the information is preserved in the directions where the variance of the gradients remains small.

This statement is consistent with recent work on the statistical properties of gradients and generalization \citep{sagun2017empirical, zho2018,zhang2018energy}. These works showed that the gradients' covariance matrix is typically highly non-isotropic and that this is crucial for generalization by SGD. They suggested that the reason lies in the gradients' covariance matrix's proximity to the Hessian of the loss approximation.
Furthermore,  \cite{zhang2018energy,keskar2016large, jastrzkebski2017three}  argued that SGD tends to converge to flat minima, which often results in a better generalization. \cite{zhang2018energy} emphasized that SGD converges to flat minima values characterized by high entropy due to the non-isotropic nature of the gradients' covariance and its alignment with the error Hessian at the minima. 
In other words, the findings above suggest that non-isotropic gradients and Hessian typically characterize good generalization performance in orthogonal directions to the flat minimum of the training error objective.
\section*{The Computational Benefit of the Hidden Layers}
\addcontentsline{toc}{section}{The Computational Benefit of the Hidden Layers}

Our Gaussian bound on the representation compression (Equation \ref{upper_bound_final}) allows us to relate the convergence time of the layer representation information, $I(T_k;X)$, to the diffusion exponent $\alpha$, defined above.
Denote the representation information at the diffusion time $\tau$ as $I(X;T_k)(\tau)$. It follows from  Eqaution \ref{upper_bound_final} that
\begin{align}
    \label{upper_bound_by_time}
    I(X;T_k)(\tau)  \leq
    C+\frac{1}{2}\sum_{\lambda_i \in \Lambda^{NI}}  \log  \left( 1+ \frac{A_{ii}}{\lambda_i(\tau)} \right) \leq
    C+\frac{1}{2}\sum_{\lambda_i \in \Lambda^{NI}}
    \left(\frac{A_{ii}}{\lambda_i(\tau)} \right)
\end{align}
 where $C$  depends on the informative for this layer, but not on $\tau$.
 
 Notice that $\lambda_i(\tau)$ are the singular values of the weights of a diffusion process, which grow as $\tau^\alpha$, where $\alpha$ is the diffusion exponent.
 Hence, $\lambda_i (\tau)=\lambda^0_i\cdot \tau^\alpha$. Therefore,  
 \[I(X;T_k)(\tau)\leq  C+\frac{1}{\tau^\alpha}{\sum_{\lambda_i \in \Lambda^{NI}}
    \left(\frac{A_{ii}}{\lambda_i^0} \right)}\]
    
Inverting this relation, the time to compress the representation $T_k$ by 
$\Delta I(X;T_k) = \Delta I_k$ scales as:   
$    \tau(\Delta I_k) \propto \left(\frac{-R}{\Delta I(X;T)}\right)^\frac{1}{\alpha}$,
where $R=\frac{1}{2}\sum_{\lambda_i \in \Lambda^{NI}}
    \left(\frac{A_{ii}}{\lambda_i^0} \right)$. Note that $R$ depends solely on the problem, $f(x)$ or $p(y,x)$, and not on the architecture. The idea behind this argument is as follows - one can expand the function in any orthogonal basis (e.g. Fourier transform). The expansion coefficients determine both the dimensionality of the relevant/informative dimensions and the total trace of the irrelevant directions. Since these traces are invariant to the specific function basis, these traces remain the same when expanding the function in the network functions using the weights.

With $K$ hidden layers, each layer only needs to compress from the previous (compressed) layer, by $\Delta I_k$, and the total compression is $\Delta I_X=\sum_{k}{\Delta I_k}$. 
Under these assumptions, even if the layers compress one after the other, the total compression time can be broken down into $k$ smaller steps, as at  
$$    \left(\frac{R}{\sum_k {\Delta I_k}}\right)^\frac{1}{\alpha} \ll 
    \sum_k{ \left(\frac{R}{\Delta I_k}\right)^\frac{1}{\alpha}}$$
If the $\Delta I_k$ are similar, we obtain a super-linear boost in the computational time by a factor $K^{\frac{1}{\alpha}}$. Since $\alpha \le 0.5$ this is at least a quadratic boost in $K$.
For ultra-slow diffusion, we obtain an exponential boost (in $K$) in the convergence time to a good generalization.  
\section*{Experiments}
\addcontentsline{toc}{section}{Experiments}

\ignore{The two main predictions that emerge from our theory are: 
\begin{itemize}
\item The diffusion phase corresponds to a reduction of MI with the input. 
\item There is a direct connection between the distinct phases in the gradients, the transfer to the diffusion, and the information's compression.
\end{itemize}
}  

We now illustrate our results in a series of experiments. We examine several different setups. 
\ignore{
In this group, we have a setup that closely follows the experiment described in \cite{shwartz2017opening}. Here, we train a fully-connected neural network with six hidden layers of width $12-10-7-7-5-4-3$ and a hyperbolic tangent (tanh) activation functions to produce a binary classification from a $12$-dimensional input (for more information about the dataset, see appendix \todo{Add it to the appendix})
}

\begin{figure}[t]
  \centering
\subfloat[\label{fig:mnist_grad}The change of weights, the SNR of the gradients, the MI and the Gaussian bound during the training for one layer. In log-log scale]{\includegraphics[width =0.45\columnwidth]{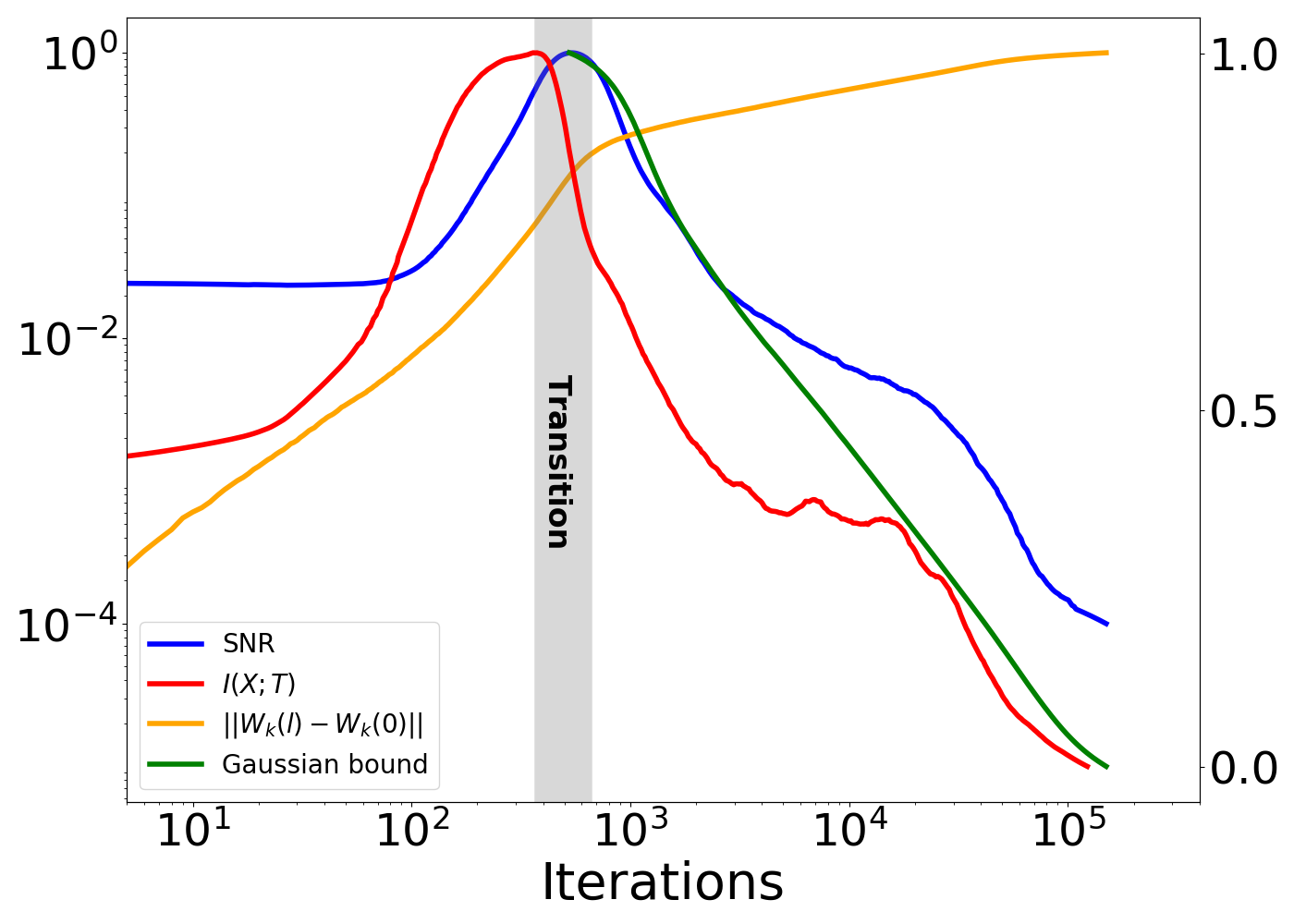}} 
\qquad
\subfloat[\label{fig:gradeints_batchs}The transition point of the SNR ($Y$-axis) versus the beginning of the information compression ($X$-axis), for different mini-batch sizes] {\includegraphics[width=0.45\columnwidth]{Figures/batch.png}}
 \caption{MNIST data-set}
    \label{fig:mnist}
\end{figure}
\textbf{MNIST dataset} -- In the first experiment, we evaluate the MNIST handwritten digit recognition task \citep{lecun1990handwritten}. For this data set, we use a fully connected network with $5$ hidden layers of width $500-250-100-50-20$, with an hyperbolic tangent (tanh) activation function.
The relatively low dimension of the network and the bounded activation function allow us to empirically measure the MI in the network.
The MI is estimate by binning the neurons' output into the interval $[-1,1]$. The discretized values are then used to estimate the joint distributions and the corresponding MI, as described by \cite{shwartz2017opening}.

Figure \ref{fig:mnist_grad}  depicts the norms of the weights, the signal-to-noise ratio (the ratio between the means of the gradients and their standard deviations), the compression rate $I(X; T)$ and the Gaussian upper bound on $I(X; T)$, as defined in Equation \ref{upper_bound_final}. As expected, the two distinct phases correspond to the drift and diffusion phases. Furthermore, these two phases are evident by independently observing the SNR, the change of the weights $||W(l)-W(0)||$, the MI, and the upper bound.
In the first phase, the weights grow almost linearly with the iterations, the SNR of the gradients is high, and there is almost no change in the MI. Then, after the transition point (that accrues almost at the same iteration for all the measures above), the weights behave as a diffusion process. In this phase, the SNR and MI decrease remarkably. In this phase, there is also a clear-cut reduction of the bound.

\begin{figure}[t]
  \centering
\subfloat[CIFAR-10]{\includegraphics[width =0.5\columnwidth]{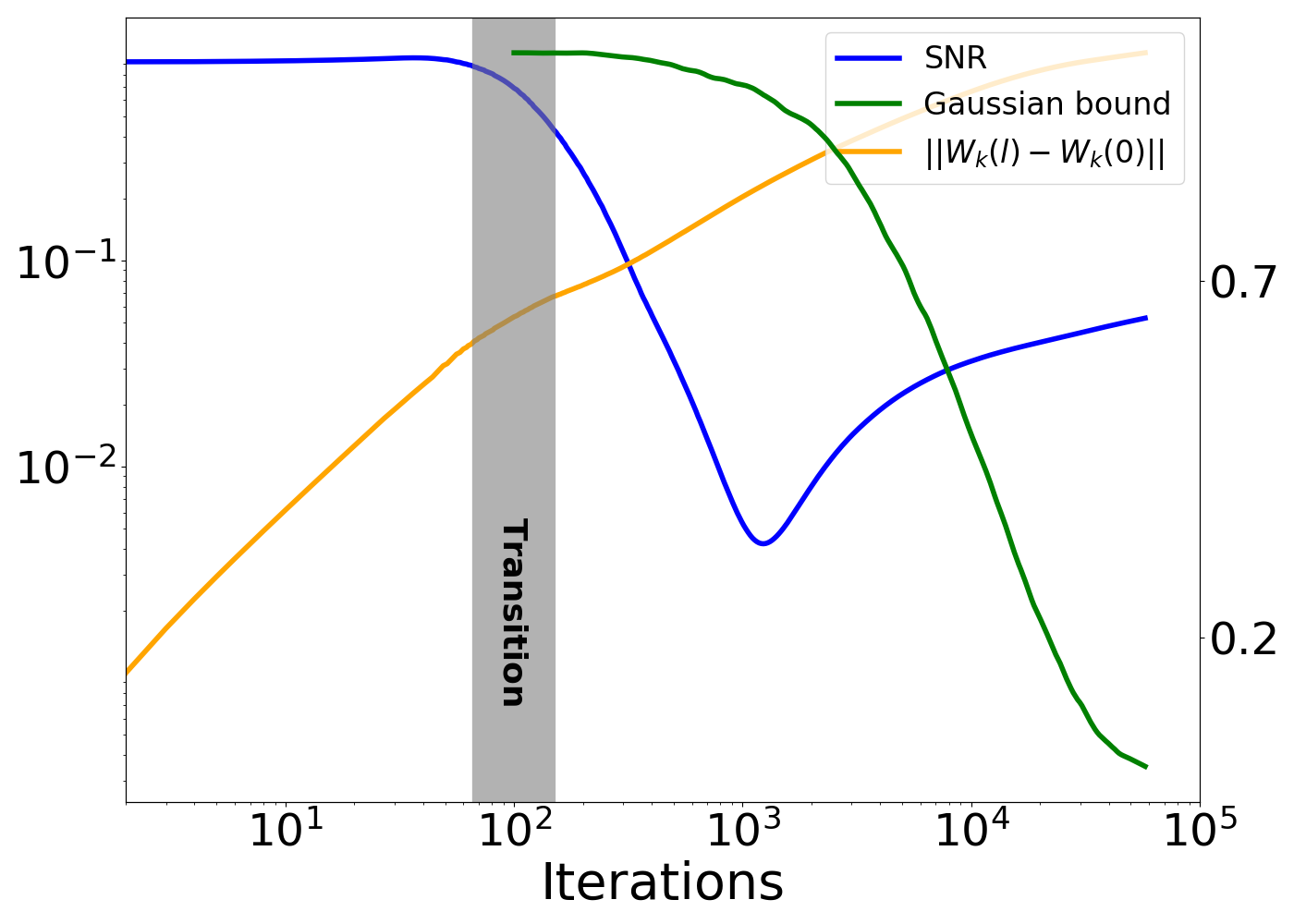}} 
\subfloat[CIFAR-100] {\includegraphics[width=0.5\columnwidth]{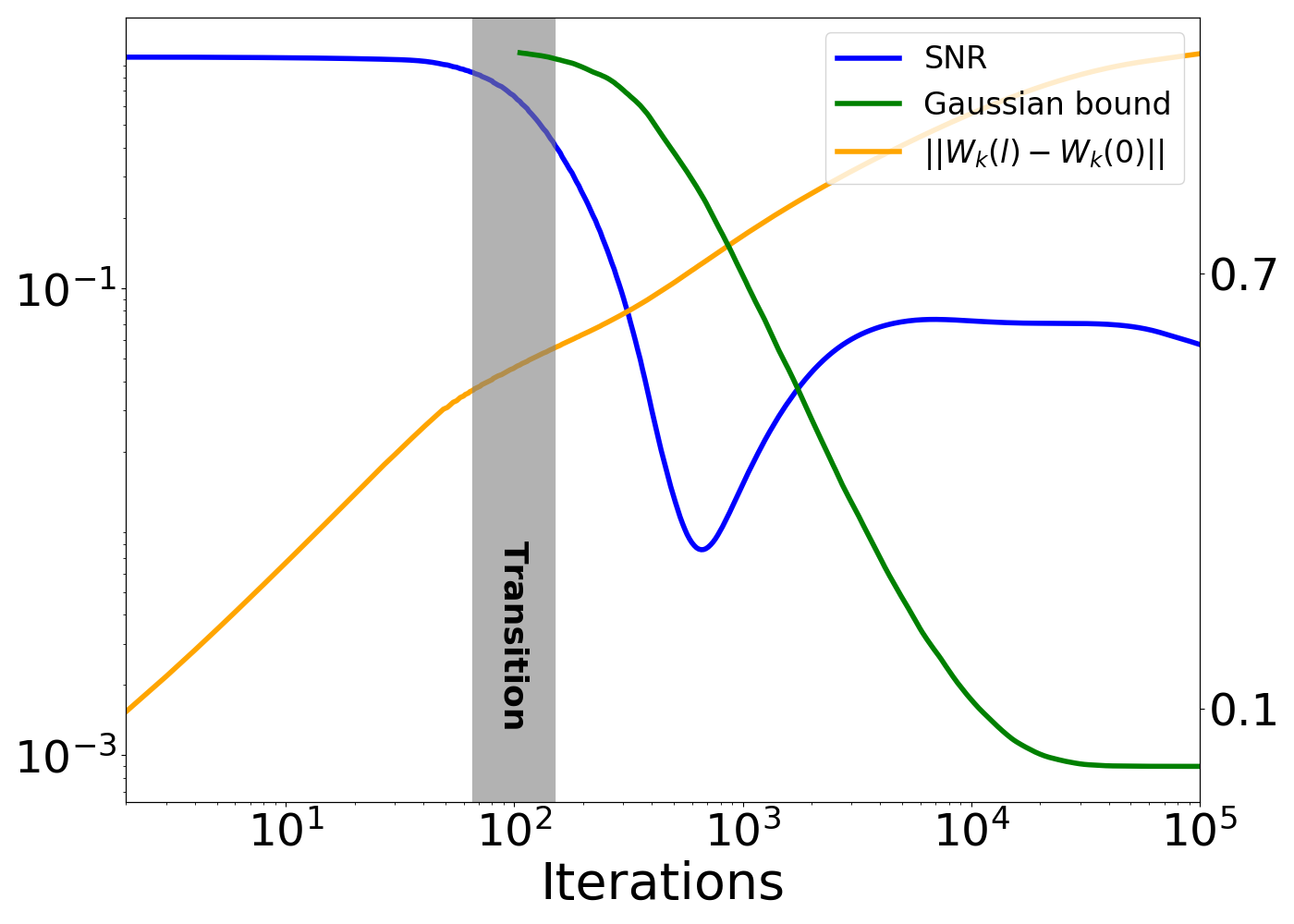}}
 \caption{Change in the SNR of the gradients and the Gaussian bound on the MI during the training of the network for one layer on ResNet-32, in log-log scale.}
    \label{fig:cifar_figure}
\end{figure}

\textbf{CIFAR-10 and CIFAR-100} -- Next, we validate our theory on large-scale modern networks. 
In the second experiment, we consider two data sets, CIFAR-10 and CIFAR-100. Here, we train a ResNet-32 network, using a standard architecture (including ReLU activation functions as described in \citep{he2016deep}. In this experiment, we do not estimate the MI directly due to the problem's large scale. Figure \ref{fig:cifar_figure} shows the SNR of the gradients and the Gaussian bound for one layer in CIFAR-10 and CIFAR-100 on the ResNet-32 network averaged over 50 runs. Here, we observed similar behavior, as reported in the MNIST experiment. Specifically, there is a clear distinction between the two phases and a reduction of the MI bound along with the diffusion phase. Note that the same behavior was observed in most of the $32$ layers in the network.

Recently, several attempts characterize the correspondence between the diffusion rate of the SGD and the size of the mini-batch (\cite{hu2017diffusion, hoffer2017train}). In these articles, the authors claimed that a larger mini-batch size corresponds to a lower diffusion rate. Here, we examine the effect of the mini-batch size on the transition phase in the information plane. For each mini-batch size, we find both the starting point of the information compression and the gradient phase transition (the iteration where the derivative of the SNR is maximal). Figure \ref{fig:gradeints_batchs} illustrates the results. The $X$-axis is the iteration where the compression started, and the $Y$-axis is the iteration where the phase transition in the gradients accrued for different minibatch sizes.
There is a clear linear trend between the two. This further justifies our suggested model since the two measures are strongly related.

\begin{figure}[t]
  \centering
\subfloat[Symmetric dataset]{\includegraphics[width =0.5\columnwidth]{Figures/symatric_time.png}} 
\subfloat[MNIST] {\includegraphics[width=0.5\columnwidth]{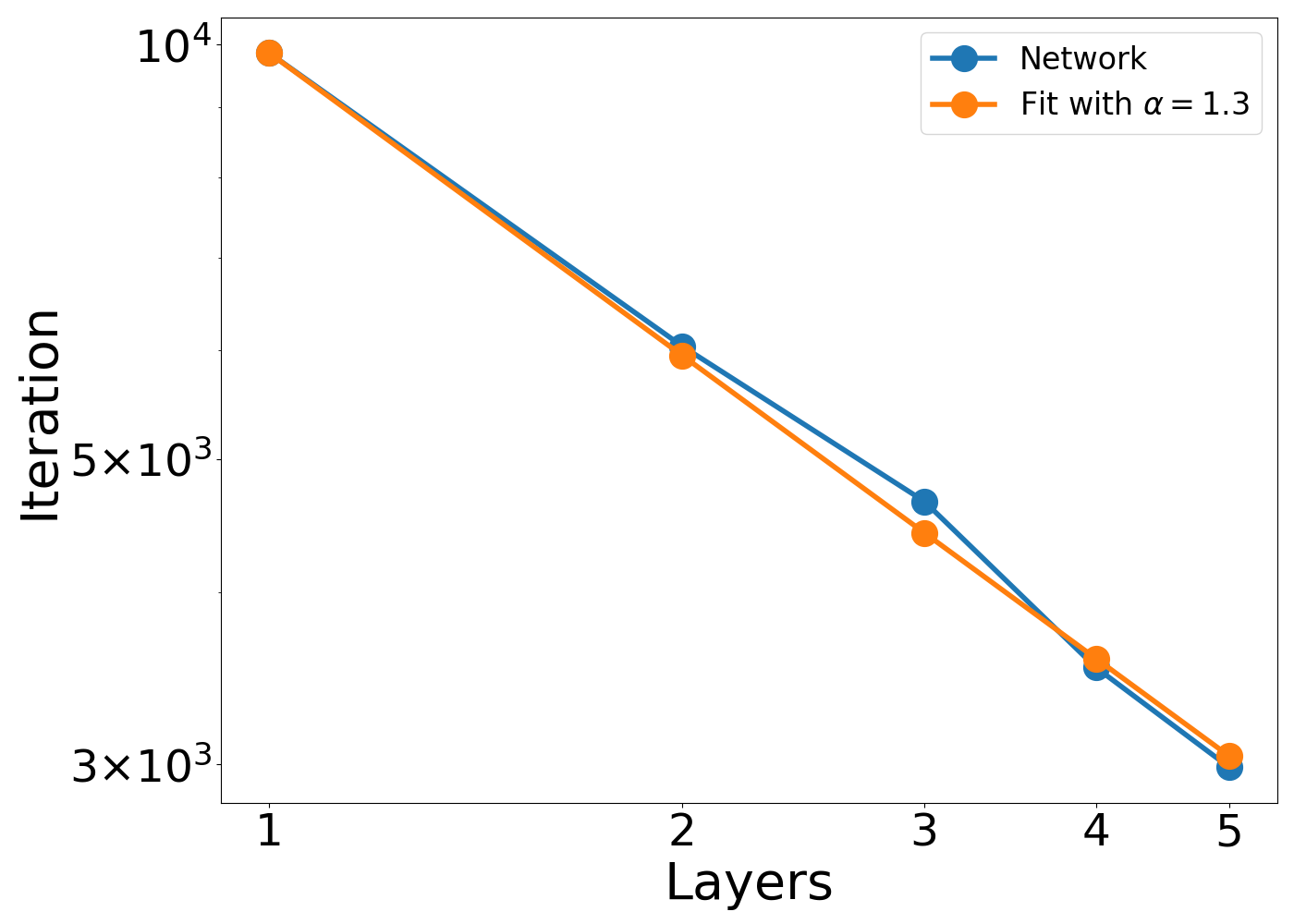}}
 \caption{The computational benefit of the layers - The converged iteration as function of the number of layers in the network}
    \label{fig:time_results}
\end{figure}

    NNext, we validate our results on the computational benefit of the layers. We train networks with different number of layers (1-5 layers) and examine the number of iterations a network takes to converge. Then, we find the $\alpha$ which fits the best trend $K^{\frac{1}{\alpha}}$, where $K$ is the number of layers. Figure \ref{fig:time_results} shows the results for two data-sets - MNIST and the symmetric dataset from \cite{shwartz2017opening}.
    As our theory suggest, as we increase the number of layers, the convergence time decreases with a factor of $k^{\frac{1}{\alpha}}$ for different values of $\alpha$.   

\ignore{
Figure \ref{fig:ib_bound} demonstrates the Gaussian upper bound on $I(X;T)$, as defined in  (\ref{upper_bound_final}), for all the layers in the network, starting from the diffusion phase. As we can see, there is an evident reduction in all the layers. 
\begin{figure}[ht]
  \centering
    \includegraphics[width=0.7\columnwidth]{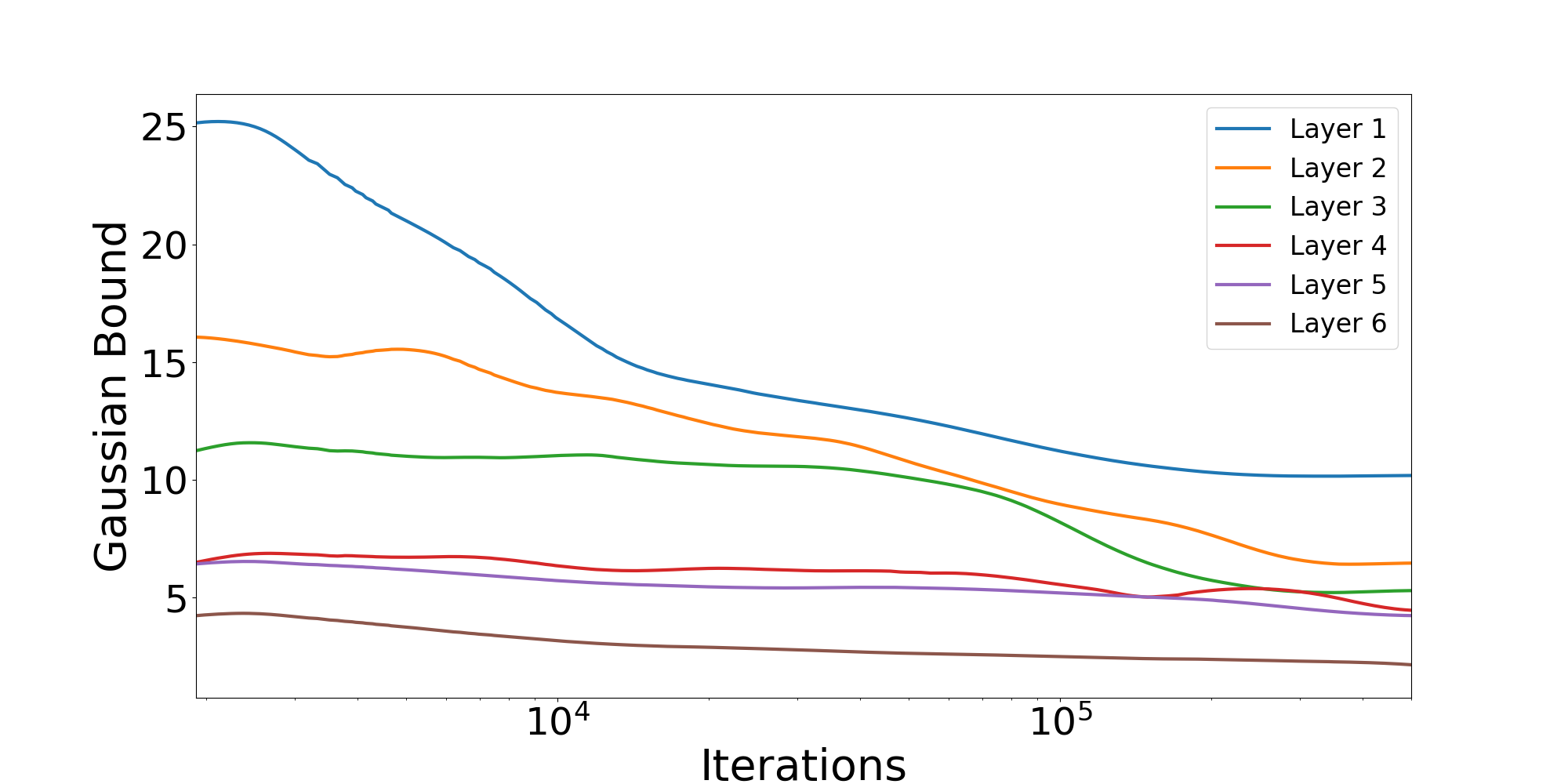}
  \caption{The Gaussian bound in each layer during the training of the network. The iterations start from the beginning of the diffusion phase}
    \label{fig:ib_bound}
\end{figure}
}

\ignore{
\begin{figure}[ht]
  \centering
    \includegraphics[width=0.7\columnwidth]{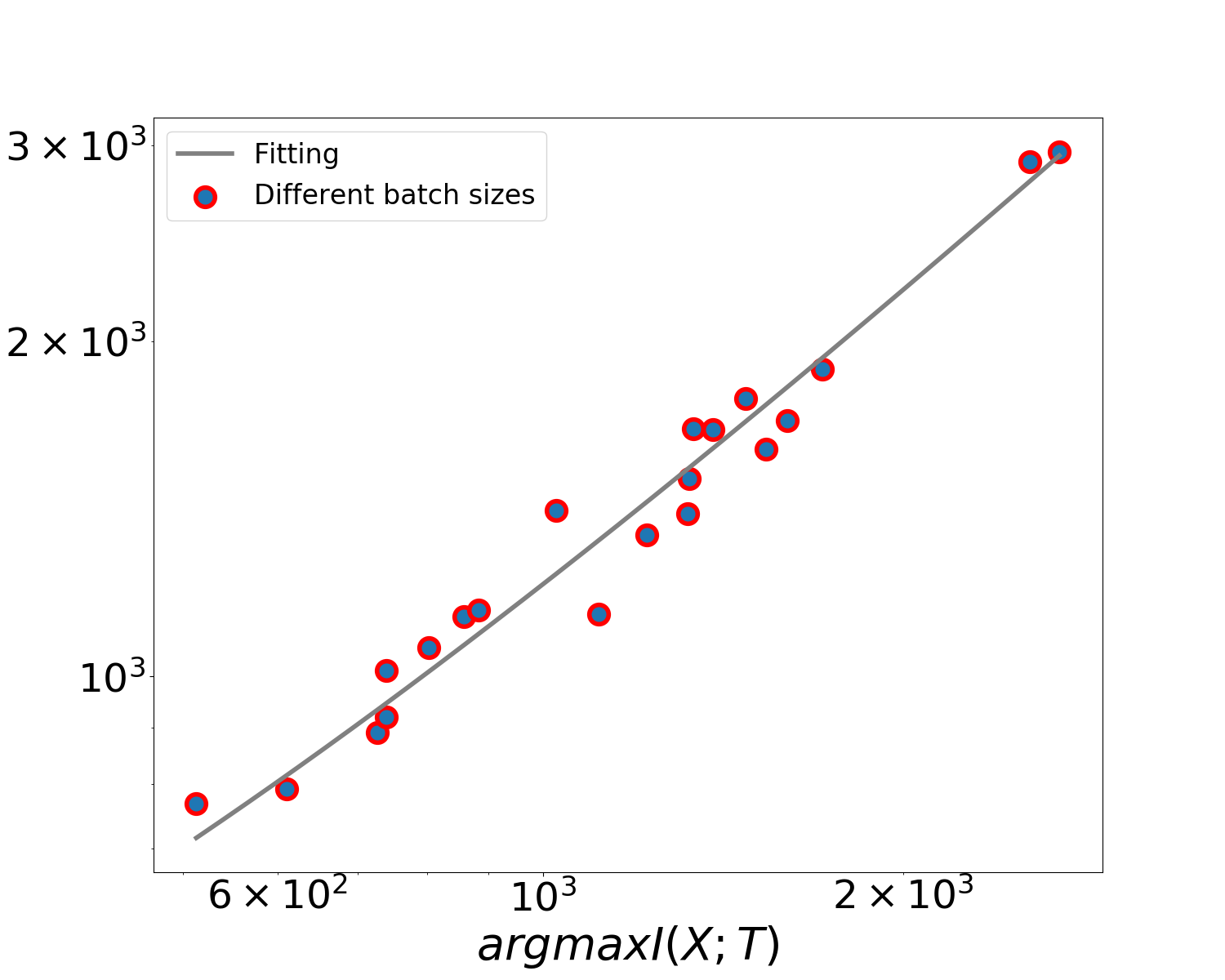}
  \caption{The transition point of the SNR of gradient versus the starting point of the compression in the information plane, for different mini-batch sizes. Each point is a different batch size. The $X$-axis is the iteration with the maximal $I(X;T_k)$, while the $Y$-axis describes the iteration with the minimum value of the gradient's derivative (the phase transition)}
    \label{fig:gradeints_batchs}
\end{figure}
}
\section*{Discussion and Conclusions}
\addcontentsline{toc}{section}{Discussion and Conclusions}

In this work, we study DNNs using information-theoretic principles. We describe the network's training process as two separate phases, as has been previously done by others. In the first phase (drift), we show that $I(T_k; Y)$ increases, corresponding to an improved generalization with ERM. In the second phase (diffusion), the representation information, $I(X;T_k)$ slowly decreases, while $I(T_K;Y)$ continues to increase. We rigorously prove that the representation compression is a direct consequence of the diffusion phase, independent of the nonlinearity of the activation function. We provide a new Gaussian bound on the representation compression and then relate the diffusion exponent to the compression time. One key outcome of this analysis is a new proof of the computational benefit of the hidden layers, where we show that they boost the overall convergence time of the network by at least a factor of $K^2$, where $K$ is the number of non-degenerate hidden layers. This boost can be exponential in the number of hidden layers if the diffusion is ``ultra-slow'', as recently reported.     
\ignore{Interestingly, an equivalent distinction to two separate phases was also observed by others (beginning with \citep{murata1998statistical}) in independent studies of the SGD process. There, it is claimed that the SGD first seeks a minimizer of the objective function and then lingers around it. 
We claim that this is not by chance; the SGD algorithm is the key to understanding these phenomena. In other words, we show that minimizing the objective corresponds to maximizing the MI with the target, while the random walk around the minimum results in a reduction of MI with the input patterns, which can further improve the generalization. This understanding allows us to characterize the behavior of the DNN on the information plane and claim that there is a fundamental limit for its performance -- the Information Bottleneck curve.}

\bibliographystyle{dcu}
\bibliography{main}
\clearpage
\section*{Appendix}
\addcontentsline{toc}{section}{Appendix}
\section*{Appendix A\addtocontents{toc}{\protect\setcounter{tocdepth}{1}}
 - proof of Theorem \ref{compression_thm}}

We first revisit the well-known \textit{Probably Approximately Correct} (PAC) bound. Let $\mathcal{H}$ be a finite set of hypotheses. Let $\ell_{h} \left(x_i,y_i \right)$ be a bounded loss function, for every $h \in \mathcal{H}$. For example, $\ell_{h} \left(x_i,y_i \right)=(y_i-h(x_i))^2$ is the squared loss while $\ell_{h} \left(x_i,y_i \right)=-y_i\log h(x_i)$ is the logarithmic loss (which may be treated as bounded, assuming that the underlying distribution is bounded away from zero and one). Let $\mathcal{L}_{h}\left(S_m \right)=\frac{1}{m}\sum_{i=1}^m \ell_{h} \left(x_i,y_i \right)$ be the empirical error. Hoeffding's inequality \cite{hoeffding1963probability} shows that for every $h \in \mathcal{H}$ 
\begin{equation}
\mathbb{P}\left[\big|\mathcal{L}_{h}\left(S_m \right)-\mathbb{E}_{S_m}\left[\mathcal{L}_{h}\left(S_m \right)\right]\big|\geq \epsilon\right]\leq2\exp{\left(-2\epsilon^2m\right)}.    
\end{equation}
Then, we can apply the union bound and conclude that
$$\mathbb{P}\left[ \exists h \in \mathcal{H} \bigg| \big|\mathcal{L}_{h}\left(S_m \right)-\mathbb{E}_{S_m}\left[\mathcal{L}_{h}\left(S_m \right)\right]\big|\geq \epsilon\right]\leq 2\big|\mathcal{H}\big|\exp{\left(-2\epsilon^2m\right)}.$$
We want to control the above probability  with a confidence level of $\delta$. Therefore, we ask that $2\big|\mathcal{H}\big|\exp{\left(-2\epsilon^2m\right)}\leq \delta$. This leads to a PAC bound, which states that for a fixed $m$ and for every $h \in \mathcal{H}$, we have with probability $1-\delta$ that 
\begin{equation}
\big|\mathcal{L}_{h}\left(S_n \right)-\mathbb{E}_{Sm}\left[\mathcal{L}_{h}\left(S_m \right)\right]\big|^2 \leq \frac{\log\big| \mathcal{H}\big|+\log{\frac{2}{\delta}}}{2m}.
\end{equation}
Note that under the definitions stated above, we have that $|\mathcal{H}|\leq 2^{\mathcal{X}}$. However, the PAC bound above also holds for a infinite hypotheses class, where $\log|\mathcal{H}|$ is replaced with the VC dimension of the problem, with several additional constants  \citep{vapnik1968uniform,shelah1972combinatorial,sauer1972density}.

Let us now assume that $X$ is a $d$-dimensional random vector that follows a Markov random field structure. As stated above, this means that $p(x_i)=\prod_i p(x_i|Pa(x_i))$ where $Pa(X_i)$ is a set of components in the vector $X$ that are adjacent to $X_i$. 
Assuming that the Markov random field is ergodic, we can define a \textit{typical set} of realizations from $X$ as a set that satisfies the \textit{Asymptotic Equipartition Property} (AEP) \citep{cover2012elements}. 
Therefore, for every 
$\epsilon>0$, the probability of a sequence drawn from $X$ to be in the typical set $A_{\epsilon}$ is greater than $1 - \epsilon$ and $|A_{\epsilon}|\leq 2^{H(X)+\epsilon}$.
Hence, if we only consider a typical realization of $X$ (as opposed to every possible realization), we have that asymptotically $\big|\mathcal{H}\big| \leq 2^{H(X)}$. Finally, let $T$ be a  mapping of $X$. Then, $2^{H(X|T)}$ is the number of typical realizations of $X$ that are mapped to $T$. This means that the size of the typical set of $T$ is bounded from above by ${2^{H(X)}}\big/{2^{H(X|T)}}=2^{I(X;T)}$. Plugging this into the PAC bound above yields that with probability $1-\delta$, the typical squared generalization error of $T$ ,$\epsilon_{T}^2$ satisfies 
\begin{equation}
    \epsilon_{T}^2 \leq \frac{2^{I(X;T)}+\log{\frac{2}{\delta}}}{2m}.
\end{equation}
\section*{Appendix B - Proof of Proposition \ref{CLT_prop}}
 We make the following technical assumptions:
 
 \begin{enumerate}
     \item ${w^*}$ and $\delta w$ satisfy      $\lim_{ d_k \rightarrow \infty}{w^*}^T\delta w =0$ almost surely.
     
     \item The moments of $T_k$ are finite.
     
     \item The components of $w^*$ and $\delta w$ are \textit{in-general-positions}, satisfying $$\lim_{d_k \rightarrow \infty}{\sum_{i=1}^{d_k} {w^*_i}^4}\big/{\left(\sum_{i=1}^{d_k} {w^*_i}^2\right)^2}=0$$ and $$\lim_{d_k \rightarrow \infty}{\sum_{i=1}^{d_k} {\delta w_i}^4}\big/{\left(\sum_{i=1}^{d_k} {\delta w_i}^2\right)^2}=0$$ almost surely. 
 \end{enumerate}

Consider a sequence of i.i.d. random variables, $\left\{X_i\right\}_{i=1}^d$ with zero mean and finite moments, $\mathbb{E}\left[X_i^r\right]< \infty$ for every $r\geq 1$. 

Let $\{a_i\}_{i=1}^d$ be a sequence of constants. Denote $Y_i=a_i X_i$, so that $\left\{Y_i \right\}_{i=1}^d$ are independent with zero mean and $\text{Var}(Y_i)=a^2_i \mathbb{E}\left[X^2\right]$. Let $S=\sum_{i=1}^d a_i X_i=\sum_{i=1}^d Y_i$ and denote $U_d^2=\sum_{i=1}^d \text{Var}(Y_i)=\mathbb{E}[X^2]\sum_{i=1}^d a_i^2 $. 

The Lyapunov Central Limit Theorem (CLT) \cite{billingsley2008probability} 
states that if there exists some $\delta>0$ for which 

\begin{equation}
\label{CLT}
\lim_{d\rightarrow \infty} \frac{1}{U_d^{2+\delta}}\sum_{i=1}^d \mathbb{E}\left[|Y_i|^{2+\delta}\right]=0
\end{equation}
then
\begin{equation}
\label{CLT2}
\frac{1}{U_d}\sum_{i=1}^d Y_i \xrightarrow[d\rightarrow \infty]{\mathcal{D}} \mathcal{N}(0,1).
\end{equation}
Plugging $\delta=2$ yields the following sufficient condition, 
\begin{equation}
    \label{cond}
    \lim_{d\rightarrow \infty}\frac{1}{U_d^{4}}\sum_{i=1}^d \mathbb{E}\left[Y_i^{4}\right]= \frac{\sum_{i=1}^d a_i^4}{\left(\sum_{i=1}^d a_i^2\right)^2}\frac{\mathbb{E}[X^4]}{\mathbb{E}^2[X^2]}=0
\end{equation} 

Let us apply the Lyapunov CLT to our problem. Here, the components of $T_k$ are i.i.d. for sufficiently large $d_k$, with zero mean and finite $r^{th}$ moments for every $r\geq1$. 
Furthermore, we assume that the components of $w^*$ and $\delta w$ are in-general-positions. 
This means that Lyapunov condition (\ref{cond}) is satisfied for both  ${w^*}^T T_k$ and ${\delta w}^T T_k$ almost surely, which means that 

\begin{equation}
\label{CLT3}
\frac{1}{\sqrt{\sigma_{T_k}^2}||w^*||_2}{w^*}^T T_k \xrightarrow[d_k\rightarrow \infty]{\mathcal{D}} \mathcal{N}(0,1)
\end{equation}
and 
\begin{equation}
\label{CLT4}
\frac{1}{\sqrt{\sigma_{T_k}^2}||\delta w||_2}{\delta w}^T T_k \xrightarrow[d_k\rightarrow \infty]{\mathcal{D}} \mathcal{N}(0,1).
\end{equation}
almost surely, where $\sigma_{T_k}^2$ is the variance of the components of $T_k$. 

Furthermore, for every pair of constants $a$ and $b$, the linear combination $\left(a{w^*}+b\delta w\right)^T T_k$ also satisfies Lyapunov's condition almost surely, which means that ${w^*}^T T_k$ and ${\delta w}^T T_k$ are asymptotically jointly  Gaussian, with 

$$\mathbb{E}\left[   {w^*}^T T_k \left({\delta w}^T T_k   \right)^T \right]=\sigma_{T_k}^2{w^*}^T \delta w\xrightarrow[d_k\rightarrow \infty]{}0$$
almost surely. \hfill $\square$
\ignore{
\section*{Appendix C - The benefit of the hidden layers
To illustrate the layers' computational benefit, we trained six different architectures with $1-6$ hidden layers, with the same settings as described in \cite{shwartz2017opening}. We train a fully connected network, where the activation function was hyperbolic tangent function. We train it on 80\% of the patterns and repeated each training 50 times with randomized initial weights and training samples. Figure \ref{layers_inf} demonstrates the  Information Paths for these six architectures during the training epochs, averaged over different networks. Three main conclusions arise from our experiments.

1. \emph{Adding hidden layers dramatically reduces the number of training epochs to attain good generalization.}
Compare the color of the paths in the left panels of Figure \ref{layers_inf} (which correspond to $1$ and $2$ hidden layers) with the colors of the right panels (which correspond to $5$ and $6$ hidden layers). While a single-layer network cannot achieve a high value of $I(T; Y)$, even after $10^4$ epochs, a $6$ layers network reaches the full relevant information at the output layer after only $400$ epochs.
  
2. \emph{The diffusion phase of each layer is shorter when it starts from a previously compressed layer.}
This can be seen by comparing the elapsed time required to attain good generalization with $4$ and $5$ hidden layers. The yellow color at the left indicates a much slower convergence with $4$ layers than networks with $5$ or $6$ layers.

3. \emph{The compression is faster for the deeper (closer to the output) layers.}
During the drift phase, the lower layers move first due to the data processing inequality. Then, during the diffusion phase, the top layers compress first and" pull" the lower layers with them. Adding more layers seems to add intermediate representations, which accelerates the compression.   
}
\begin{figure}[ht]
\centering
\includegraphics[width=1\columnwidth, left]{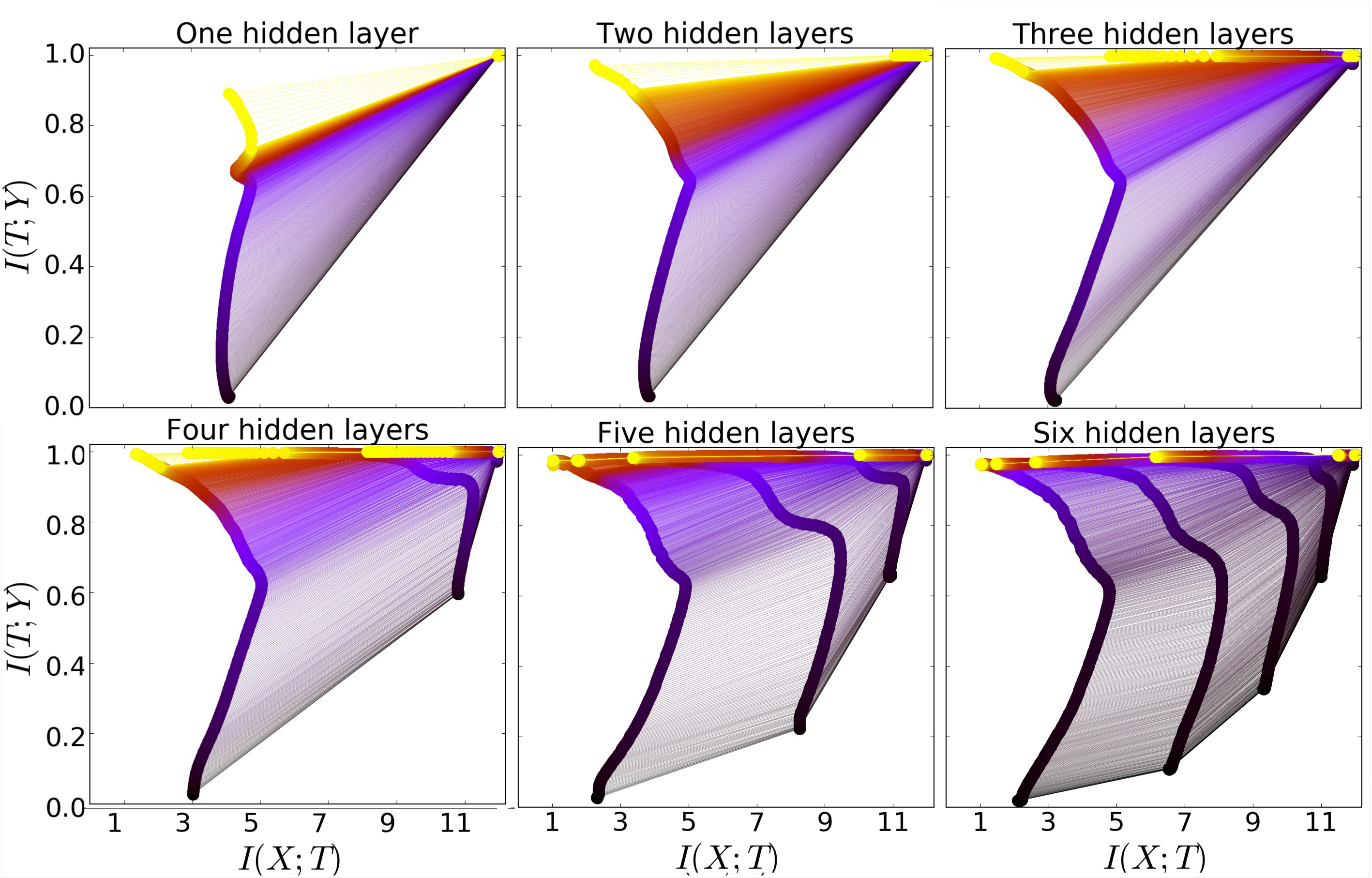}
\caption{{\label{fig:and-inf_4_per-1}The network Information Paths during the SGD
optimization when adding hidden layers. } Each panel is
the information plane for a network with a different number of hidden
layers. The width of the hidden layers start with $12$, and each additional layer has $2$ fewer neurons. 
The last layer with $2$ neurons is shown in all panels.  
The line colors correspond to the number of training epochs. 
}
\label{layers_inf}
\end{figure}
}

%% file: Chapters/abbi.tex
\chapter*{Information in Infinite Ensembles of Infinitely-Wide Neural Networks}
\addcontentsline{toc}{chapter}{4:   Information in Infinite Ensembles of Infinitely-Wide Neural Networks}

\textbf{Proceedings of the Symposium on Advances in Approximate Bayesian Inference, PMLR}, 2020 \\
Ravid Shwartz-Ziv and Alexander A. Alemi.
\newpage

\begin{center}
        \vspace*{0.5cm}
        \LARGE
        \textbf{Information in Infinite Ensembles of Infinitely-Wide Neural Networks} \\
        \vspace{0.8cm}
        \normalsize
    Ravid Shwartz-Ziv \textsuperscript{1}
    Alexander A. Alemi\textsuperscript{2} \\
            \vspace{2.cm}
    \textsuperscript{1} The Edmond and Lilly Safra Center for Brain Sciences, The Hebrew University, \\
  Jerusalem, Israel.\\
    \textsuperscript{2} Google Research \\
    USA
    \end{center}
\begin{center}
  \vspace*{0.5cm}
         \normalsize
        \textbf{Abstract} \\
\end{center}

In this work, we study the generalization properties
    of infinite ensembles of infinitely-wide neural networks.
    Amazingly, this model family admits tractable calculations 
    for many information-theoretic quantities.
    We report analytical and empirical investigations in the search for signals that correlate with generalization.


\section*{Introduction}
\addcontentsline{toc}{section}{Introduction}

According to the statistical learning theory \citep{boucheron2005theory}, models with many parameters tend to overfit by representing the learned data too accurately, therefore diminishing their ability to generalize to unseen data. However, in DNNs, we see that the 'generalization gap; i.e., the difference between 'training error' and 'test error' is minimal. 
One promising research direction is to view deep neural networks through the lens of information theory\citep{tishbydeep}. Abstractly, deep connections exist between the information
a learning algorithm extracts and its generalization
capabilities~\citep{littlebits, bayesianbounds}.  Inspired by these general results,
recent papers have attempted to measure information-theoretic quantities in ordinary
deterministic neural networks~\citep{blackbox,emergence,whereinfo}.  

Both practical and theoretical problems arise 
in the deterministic case~\citep{hownot, saxe, brendan}.
These difficulties stem from the fact that 
mutual information (MI) is reparameterization independent~\citep{coverthomas}.\footnote{
This implies
that if we send a random variable through an invertible function, its MI
with respect to any other variable remains unchanged.} 
One workaround is to make a network explicitly stochastic, either in its activations~\citep{vib} or
its weights~\citep{emergence}.
Here we take an alternative approach, harnessing the stochasticity in our choice of initial parameters.
We consider
an \emph{ensemble} of neural networks, all trained with the same training procedure and data.
This generates an ensemble of predictions. In other words, we consider an infinite ensemble of neural networks that describes a distribution over the output space when we marginalize out our choice of initial parameters.  
Characterizing the generalization properties of the ensemble should characterize the generalization of individual draws from this ensemble.
However, the challenge is in describing this ordinarily intractable distribution.

%

Infinitely-wide neural networks behave as if
they are linear in their parameters~\citep{widelinear}.
Their evolution is fully described by the 
\emph{neural tangent kernel} (NTK). The NTK
is constant in time and can be tractably computed~\citep{neuraltangents} as a function of the network's architecture, e.g., the number and the structure of layers, nonlinearity, initial parameters' distributions, etc.

~\citep{widelinear} showed that the output of an infinite ensemble of
infinitely-wide neural networks initialized with Gaussian weights and biases
and trained with gradient flow to minimize a square loss is
simply a conditional Gaussian distribution:
\begin{equation}
    p(z|x) \sim \N(\mu(x,\tau), \Sigma(x,\tau)) ,
    \label{eqn:rep1}
\end{equation}
where $z$ is the output of the network and $x$ is its input.
The mean $\mu(x,\tau)$ and covariance $\Sigma(x, \tau)$ functions
can be computed~\citep{neuraltangents}.

Recently, there has been much interest in understanding the importance of implicit regularization as a tool for explaining the generalization gap of DNNs.
Numerical experiments demonstrate that network size may not be the main form of capacity control, and hence, some other
unknown form plays a central role in learning multi-layer network \citep{neyshabur2014search,neyshabur2015norm}.  A line of works have derived interesting results proving that some information-theoretic quantities provide concise bounds on the generalization gap, hence behave as implicit regularization \citep{russo2019much, xu2017information,pensia2018generalization,NIPS2019_9282,NIPS2018_7954, russo2016controlling, steinke2020reasoning, whereinfo, bayesianbounds, littlebits}. However there is not direct evidence what are the most important quantities which determinate the generalization ability of the network.

In this work, the simple structure of the NTK allows us to bound
several interesting information-theoretic quantities, including:
the MI between the representation and the targets, $I(Z;Y)$,
the MI between the representation and the inputs after training, $I(Z;X|D)$,
and the MI between the representations and the training set, conditioned on the input, $I(Z;D|X)$.
We are also able to compute in closed form:
the Fisher information metric,  the distance the parameters move
and the MI between the parameters and the data, $I(\Theta;D)$.
All derivation of all these information-theoretic quantities allow us to explore what are the important factors in the generalization ability of DNNs.
\section*{Background}
\addcontentsline{toc}{section}{Background}

\subsection{Neural Tangent Kernel}
\label{sec:ntk}
In this section we describe fully-connected deep neural net architecture and its infinite width limit, and how training it with respect to the $l_2$ loss gives rise to a kernel regression problem involving the NTK. We denote by $f_t(\theta, x) \in \mathbb{R}$ the output of a neural network at time $t$ where $\theta \in \mathbb{R}^N$ is all the parameters in the network and $x \in \mathcal{R}^d$ is the input. Let $\ell(\hat{y}, y): \mathbb{R}\times \mathbb{R} \to \mathbb{R}$ denote the loss function where the first argument is the prediction and the
second argument the true label. Given a training dataset $(x_i,y_i)_{i=1}^N \subseteq \mathbb{R}^d \times \mathbb{R}$,  consider training the neural network by minimizing the empirical loss over training data:
$\Li=\sum_{i=1}^n\ell (f_t(x_i, \theta), y)$.
Let $\eta$ be the learning rate. Via continuous time gradient descent, the evolution of the parameters $\theta$ and the logits $f$ can be written as
\begin{lemma}

\begin{align*} 
    &\frac{d \theta_t}{dt}=-\eta \nabla_\theta f_t(\X)^T \nabla_{f_t(\X)}\Li \\
     &\frac{d f_t(\X)}{dt}=\nabla_\theta f_t(\X)\frac{d\theta_t}{dt}=-\eta \hat{\Theta}_t\left(\X,\X\right)\nabla_{f_t(\X)}\Li
\end{align*} 
\end{lemma}

where $f_t(\X) = vec\left(\left[f_t(x)\right]x\in \X\right)$ , the $k|D| \times 1$ vector of concatenated logits for all example, and $\hat{\Theta}_t\equiv\hat{\Theta_t(\X, \X)}$ is the 
tangent kernel at time t - an $n \times n$ positive semidefinite matrix whose $(i, j)$-th entry is 
\[\left< \frac{\partial f_t(x_i)}{\partial \theta}, \frac{\partial f_t( x_j)}{\partial \theta} \right>\]
The shorthand $\Theta$ denotes the kernel function evaluated on the train data ($\Theta \equiv \Theta(\X, \X)$).
For a finite width network, the NTK corresponds to $JJ^T$, the gram matrix of neural network gradients.  As the width of a network increases to infinity, this kernel converges in probability to a fixed value.
There exist tractable ways to calculate
the exact infinite-width kernel for wide classes of neural networks~\citep{neuraltangents}.

\subsubsection{ Linearized network}
Infinitely-wide neural networks behave as though they were linear in their parameters~\citep{widelinear}:
\begin{equation}
    f_t(x) = f_0(x) + \frac{\partial f_0(X)}{\partial \theta} (\theta - \theta_0)
\end{equation}
This makes them particularly analytically tractable.  
An infinitely-wide neural network, trained by gradient flow to minimize squared loss admits a closed form expression for evolution of its predictions as a function of time. For an arbitrary point $x$
\begin{equation}
    f_t(x) =  f_0(x) - \hat{\Theta}(x, \X) \hat{\Theta}^{-1} \left( I - e^{-\tau \hat{\Theta}} \right)(f_0(\X) - \Y) .
\end{equation}

Notice that the behavior of infinitely-wide neural networks trained with gradient flow and squared loss is just a time-dependent affine transformation of their initial predictions.
As such, if we now imagine forming an infinite ensemble of such networks as we vary
their initial weight configurations, if those weights are sampled from a Gaussian distribution,
the law of large numbers enforces that the distribution of outputs
of the ensemble of networks at initialization is Gaussian, conditioned on its input.  Since the
evolution is an affine transformation of the initial predictions, the predictions remain Gaussian
at all times.
\begin{corollary}
For every test points $x \in X$ and $t\leq 0$, $z=f_t(x)$ converges in distribution as width goes to infinity to 
    \begin{align*}
    p(z|x) &\sim \N(\mu(x,t), \Sigma(x,t)) 
    \label{eqn:rep1} \\
    \mu(x,t) &= \Theta(x, \X) \Theta^{-1} \left( I - e^{-t \Theta} \right) \Y\\
    \Sigma(x,t) &= \\ &\K(x, x) + \Theta(x, \X) \Theta^{-1} \left(I - e^{-t \Theta} \right) \left( \K \Theta^{-1} \left( I - e^{-\tau \Theta} \right) \Theta(\X, x) - 2 \K (\X, x) \right)
    \end{align*}
\end{corollary}
For more details see~\citet{widelinear}.
 
Here, $\K$ denotes the \emph{neural network gaussian process} kernel (NNGP). For a finite width network, the NNGP corresponds to the 
expected gram matrix of the outputs: $\K^{i,j}(x,x^\prime)=\mathbb{E}\left[f_t^i(x)f_t^j(x^\prime) \right]$. In the infinite width limit, this concentrates on a fixed value.
Just as for the NTK, the NNGP can be tractably computed~\citep{neuraltangents}, and should be considered just a function of the neural network architecture.
For this family of models, we would like to derive information-theoretic
characterizations of their performance.
\subsection{The Information Bottleneck Optimal Bound}
\label{sec:IB1}

What characterizes the optimal representations of $X$ w.r.t. $Y$? 
One of the candidate to this question is the classical notion of minimal sufficient statistics. Sufficient statistics, are maps or partitions of $X$, $S(X)$, capturing all the information that $X$ has on $Y$. Namely, $$I(S(X);Y)=I(X;Y)$$ \citep{coverthomas}. 

Minimal sufficient statistics, $T(X)$, are the simplest sufficient statistics and induce the coarsest sufficient partition on $X$. In other words, they are functions of any other sufficient statistic. A simple way of formulating this is through the Markov chain: $Y\rightarrow X \rightarrow S(X) \rightarrow T(X)$, which should hold for a minimal sufficient statistics $T(X)$ with any other sufficient statistics $S(X)$. Using  DPI we can cast it into a constrained optimization problem:
$$
T(X) = \arg \min_{S(X): I(S(X);Y)=I(X;Y)} I(S(X);X) ~.
$$

Since exact minimal sufficient statistics only exist for very special distributions, (i.e., exponential families), \citep{Tishby1999} relaxed this optimization problem by first allowing the map to be stochastic, defined as an encoder $P(T|X)$, and then, by enabling the map to capture \emph{as much as possible} of $I(X;Y)$, not necessarily all of it.

This leads to the \textit{Information Bottleneck} (IB) trade off \citep{Tishby1999}, which provides a computational framework for finding approximate minimal sufficient statistics, or the optimal trade off between compression of $X$ and prediction of $Y$.

If we define $t\in T$ as the compressed representations of $x\in X$, the
representation of $x$ is now defined by the mapping $p\left(t|x\right)$.
This IB trade off is formulated by the following optimization problem, carried independently for the distributions, $p(t|x), p(t), p(y|t)$, with the Markov chain: $Y\rightarrow X \rightarrow T$,
\begin{equation}
\label{eq:ib}
\min_{p\left(t|x\right),p\left(y|t\right),p\left(t\right)}\left\{ I\left(X;T\right)-\beta I\left(T;Y\right)\right\} ~.
\end{equation}

The Lagrange multiplier $\beta$ determines the level of relevant information captured by the representation $T$, $I(T;Y)$.

If we denote $I_{X}^{\beta}=I_{\beta}\left(T;X\right)$ and $I_{Y}^{\beta}=I_{\beta}\left(T;Y\right)$
for some $\beta$, the optimal information curve is then defined as the optimal values of the trade-off $\left(I_{X}^{\beta},I_{Y}^{\beta}\right)$
for each $\beta$. The two-dimensional plane in which the IB curve resides is coined as the information plane. The information curve is a monotonic concave line of optimal representations that
separates the achievable and unachievable regions in the information-plane.

\subsubsection{Gaussian Information Bottleneck}
Generally speaking, solving the IB problem \cref{eq:ib} for an arbitrary joint distribution is an hard task. \cite{Tishby1999} defined a set of
self-consistent equations which formulate the necessary conditions for the optimal solution
of \cref{eq:ib}. In general, these equations do not hold a tractable solution and
are usually approximated by different means \citep{slonim2002information}.

A special exception is the Gaussian case, where $X$ and $Y$ are follow a jointly normal
distribution. Namely -
 \[ p(x,y) = \mathcal{N}\left( 0, \begin{pmatrix} \Sigma_{xx} & \Sigma_{xy} \\ \Sigma_{yx} & \Sigma_{yy} \end{pmatrix}  \right) \]

In this case, the Gaussian IB problem is analytically solved by linear projections
to the canonical correlation vector space \citep{gaussib}. In this case, the Gaussian IB problemis solved by a noisy linear projection, $T = AX + \epsilon$ and everything is governed by the eigenspectrum of:
 \[ \Sigma_{x|y}\Sigma_{xx}^{-1} =  I - \Sigma_{xy}\Sigma_{yy}^{-1}\Sigma_{yx}\Sigma_{xx}^{-1} \]
 
\subsection{Information Bounds on the Generalization Gap}
Many works tried to understand the importance of implicit regularization in the small generalization gap of DNNs. Numerical
experiments in \citep{neyshabur2014search,neyshabur2015norm} demonstrate that network size may not be the main factor to explain the capacity of the network, and hence,
some other unknown form plays a central role in the learning. Further work in \cite{zhang2016understanding}, found that in contrast with classical convex empirical risk minimization, regularization plays a rather different
role in deep learning. From the theoretical viewpoint, regularization seems to be an indispensable component, while convincing experiments support the idea that the absence of explicit regularization does not necessarily induce poor generalization. A line of works have derived interesting results proving that the mutual information between the training inputs and the inferred parameters provides a concise bound on the gap, which crucially depends on a mapping of the training set into the network parameters, whose characterization is not an easy task \citep{russo2019much, xu2017information,pensia2018generalization,NIPS2019_9282,NIPS2018_7954, russo2016controlling, steinke2020reasoning, whereinfo}. \citep{emergence} explored how the use of an IB objective on the network parameters may help avoid overfitting while enforcing invariant representations. On the other side, direct use of statistical learning theory, such as Rademacher complexity \citep{bartlett2002rademacher}, VC-dimension \citep{stavac}, and uniform stability \citep{bousquet2002stability} seem to be inadequate to explain the unexpected numerical observations on the generalization gap.

\section*{Family of Trained Models}
\addcontentsline{toc}{section}{Family of Trained Models}

In the recent years there are two lines of works which investigate generalization in DNNs; (1) Some works try to use information quantities to train DNNs better, while
others (2) try to explain the current DNNs generalization ability.
In this work, we would like to follow the second line of works.

Based on the NTK framework, we derive several information-theoretic quantities that are good candidates to explain the performance of DNNs. Following this, we empirically investigate these measures by varying the hyperparameters of the networks. Specifically, we check three types of hyperparameters relevant to the NTK framework -- 
\begin{itemize}
    \item  Network architecture (number of layers, activation's function, etc.).
    \item  The initial noise of the weights and biases.
    \item The dataset (type, number of training examples, etc.) -- We train the networks on three different datasets - MNIST, CIFAR-10, and a jointly Gaussian dataset where we can compare the network performance to that of an optimal analytical solution
\end{itemize}

The NTK framework is full batch training, so there is no effect of the learning rate or the learning algorithm on dynamics.

More train examples, for example, should in principle store more information about the data. However, the usual PAC-Bayes generalization bounds for information suggest that a high-performance network should be compressed\citep{bayesianbounds}.

\section*{Derive Information Metrics}
\addcontentsline{toc}{section}{Derive Information Metrics}

\label{sec:info}

We can compute several information-theoretic quantities in the presence of a tractable form for representing an infinitely-wide ensemble of networks. As a result, we can shed light on previous attempts to explain generalization in neural networks and identify candidates for empirical investigations into quantities that can predict generalization.
\subsection*{Loss}
\label{sec:loss}

To compute our ensemble's expected loss, we need to marginalize the stochasticity in the output
of the network.  Training with squared loss is equivalent
to assuming a Gaussian observation model $p(y|z) \sim \N(0, 1)$. 
We can marginalize out our representation to obtain
\begin{equation}
    q(y|x) = \int dz\, q(y|z)p(z|x) \sim \N( \mu(x, \tau), I + \Sigma(x, \tau) ).
\end{equation}

The expected log loss has contributions both from the square loss of the mean prediction, as well
as a term which couples to the trace of the covariance:
\begin{equation}
    \mathbb{E}\left[ \log q(y|z) \right] = \frac 12 \mathbb{E}\left[ ( y - z(x, \tau ))^2  \right]= \frac 12 ( y - \mu( x, \tau))^2 + \frac 12 \Tr \Sigma(x,\tau) - \frac k 2 \log 2 \pi
\end{equation}

here $k$ is the dimensionality of $y$.
\subsection*{\texorpdfstring{$H(Y|Z)$}{H(Y|Z)}}
\label{sec:hy_z}
The informativeness, or accuracy, of the representation is measured by $I(Z;Y)$, which is the amount of relevant information about $Y$ preserved by the representation.
It measures how much of the predictive features in $X$ for $Y$ is captured by our model. Because we know that $I(X:Y)=H(Y)-H(Y|Z)$, we would like to estimating the conditional entropy. To calculate an lower bound, similarly to what we did before, we need to marginalize the stochasticity in the output by finding the observation model which gives us the lowest conditional entropy. However, in this case, we are not assuming a model with Gaussian's variance one, that could be sub-optimal.  
\begin{align*}
    \mathbb{E}\left[ \log q(y|z) \right] =  \frac 12 \mathbb{E}\left[ ( y - z(x, \tau ))^2  \right]= \\ \frac {1}{2}\left( (y - \mu( x, \tau))\Sigma_r^{-1}(y - \mu( x, \tau)) + \frac{\Tr \Sigma(x,\tau)*d}{n} \right) \\- \frac k 2 \log 2 \pi \Sigma_r
\end{align*}

\subsection*{\texorpdfstring{$I(Z;Y)$}{I(Z;Y)}}
\label{sec:izy}
While the MI between the network's output and the targets is intractable in
general, we can obtain a tractable variational lower bound:~\citep{vmibounds}
\begin{equation}
    I(Z; Y) = \mathbb{E}\left[ \log \frac{p(y|z)}{p(y)} \right] 
    \leq \mathbb{E}\left[ \log \frac{q(y|z)}{p(y)} \right] = H(Y) + \mathbb{E}\left[ \log q(y|z) \right]
\end{equation}

\subsection*{\texorpdfstring{$I(Z;X|D)$}{I(Z;X|D)}}
\label{sec:izx}
According to IB, the complexity of the representation is measured by $I(X; Z|D)$,
which is roughly the number of bits that are required for representing the input ($X$) using the network's output ($Z$) conditioned on the dataset ($D$):
\begin{equation}
    I(Z;X | D) = \mathbb{E}\left[ \log \frac{p(z|x,D)}{p(z|D)} \right].
\end{equation}
This requires knowledge of the marginal distribution $p(z|D)$.  Without knowledge of 
$p(x)$, this is in general intractable, but
there exist simple tractable multi sample 
upper and lower bounds~\citep{vmibounds}:
\begin{equation}
    \frac 1 N \sum_i \log \frac{p(z_i|x_i, D)}{\frac 1 {N} \sum_{j} p(z_i| x_j, D)} \leq
    I(Z;X | D) 
    \leq \frac 1 N \sum_i \log \frac{p(z_i|x_i, D)}{\frac 1 {N-1} \sum_{j \neq i} p(z_i| x_j, D)}.
\end{equation}
In this work, we show the minibatch lower bound
estimates, which are upper bounded themselves by
the log of the batch size.

\subsection*{\texorpdfstring{$I(Z;D|X)$}{I(Z;D|X)}}
\label{sec:izd}

We can also estimate a variational upper bound on the
MI between the representation 
of our networks and the training dataset.
\begin{equation}
    I(Z;D|X) = \mathbb{E}\left[ \log \frac{p(z|x,D)}{p(z|x)} \right] \leq \mathbb{E}\left[ \log \frac{p(z|x,D)}{p_0(z|x)} \right].
\end{equation}
Here, the MI we extract from the dataset 
involves the expected log ratio of our posterior distribution
of outputs to the marginal over all possible datasets.
Not knowing the data distribution, this is intractable in
general, but we can variationally upper bound it with an
approximate marginal.  A natural candidate is the prior
distribution of outputs, for which we have a tractable estimate.
\subsection*{Fisher information }
\label{sec:fisher}
It is usually assumed that the Fisher matrix approximates the Hessian spectrum, which can be used to estimate the objective function shape. 
In the literature on deep learning, it has been shown that eigenvalues close to zero locally form flat minima, leading to better generalization empirically \citep{keskar2016large, liang2019fisher}.
\cite{whereinfo} connected flat minima (low Fisher information) to path stability of
SGD \citep{hardt2016train} and information stability \citep{xu2017information}, showing that optimization algorithms that converge to flat minima and are path stable also satisfy a
form of information stability, and hence generalization, by 
PAC-Bayes bound \citep{xu2017information}.

Infinitely-wide networks behave as though they were linear in their parameters with a fixed Jacobian.  This leads to trivially flat information geometry.  For squared loss, the true Fisher can be computed simply as $F = J^TJ$~\citep{fisher}.   While the trace  of the Fisher information has recently been proposed as an important quantity for controlling generalization in neural networks~\citep{whereinfo}, for infinitely-wide networks we can see that the trace of the Fisher is the same as the trace of the
NTK, which is a constant and does not evolve with time $$\Tr F = \Tr J^T J =\Tr J J^T = \Tr \Theta$$.
In so much as infinite ensembles of infinitely-wide neural networks generalize, the degree to which
they cannot be explained by the time evolution of the trace of the
Fisher, given that the trace of the Fisher does not evolve.

\subsection*{Parameter distance}
\label{sec:dist}

How much do the parameters of an infinitely-wide network change?  
~\cite{widelinear} emphasizes that the relative Frobenius norm change of the 
parameters throughout training vanish in the limit of infinite width.
This is a justification for the linearization becoming more
accurate as the network becomes wider. 
But is it thus fair to say the parameters are not
changing?  Instead of looking at the Frobenius norm we can investigate the
\emph{length} of the parameters path over the course of training.  This 
reparameterization independent notion of distance utilizes the 
information geometric metric provided by
the Fisher information:
\begin{equation}
    L(\tau) = \int_0^\tau ds =
    \int_0^\tau d\tau\, \sqrt{\dot\theta_\alpha(\tau) g_{\alpha \beta} \dot\theta_\beta(\tau) }  
    =  \int_0^\tau d\tau\, \left\lVert \Theta  e^{-\tau \Theta} (z_0(\X) -\Y) \right\rVert
\end{equation}
The length of the trajectory in parameter space is the integral of a norm of our
residual at initialization projected along $\Theta e^{-\tau \Theta}$.  This integral is both 
positive and finite even as $t \to \infty$.  To get additional understanding into the
structure of this term, we can consider its expectation over the ensemble, where
we can use Jensen's inequality to bound the expectation of trajectory lengths.  Since we
know that at initialization $z_0(\X) \sim \N(0, \K)$ we obtain further simplifications:
\begin{align}
    \mathbb{E}[L(\tau)]^2 &\leq \mathbb{E}[L^2(\tau)] =
    \int_0^\tau d\tau\, \mathbb{E}\left[ (z_0(\X) - \Y)^T \Theta^2 e^{-2\tau \Theta} (z_0(\X) - \Y) \right]  \\
     &=  \frac 12 \mathbb{E}\left[ (z_0(\X) - \Y)^T \Theta \left( 1 - e^{-2\tau \Theta} \right) (z_0(\X) - \Y) \right]  \\
     &= \frac{1}{2} \left[ \Tr{\left( \K \Theta \left( 1 - e^{-2\tau \Theta}\right)\right)} + \Y^T\Theta \left(1 - e^{-2\tau \Theta}\right) \Y \right].
\end{align}

\subsection*{\texorpdfstring{$\kl{p(\theta|D)}{p_0(\theta)}$}{KL[p(θ|D), p0(θ)]}}
\label{sec:itd}

The MI between the parameters and the dataset $I(\theta; D)$ has been shown to control for overfitting~\citep{littlebits}. We can generate a variational upperbound on this 
quantity by consider the KL divergence between the posterior distribution of our parameters
and the prior distribution $\kl{p(\theta|D)}{p_0(\theta)}$, a quantity that itself has been
shown to provide generalization bounds in PAC Bayes frameworks~\citep{emergence}. For our networks, 
the prior distribution is known and simple, but the posterior distribution can be quite rich. 
However, we can use the \emph{instantaneous change of variables} formula~\citep{neuralode}
\begin{equation}
   \log p(\theta_\tau) = \log p(\theta_0) - \int_0^\tau d\tau\, \Tr \left( \frac{\partial \dot \theta}{\partial \theta} \right),
\end{equation}
which gives us a value for the log likelihood the parameters of a trained model at any point in time
in terms of its initial log likelihood and the integral of the trace of the kernel governing its
time evolution.  For our infinitely-wide neural networks this is tractable:
%
%
\begin{align}
    I(\theta; D) &\leq 
    \mathbb{E}_{p(\theta_\tau)}\left[ \log p(\theta_\tau) - \log p_0(\theta_\tau)  \right]\\
    &=  \mathbb{E}_{p(\theta_0)}\left[ \log p(\theta_0) - \log p_0(\theta_\tau) - \int_0^\tau d\tau\, \Tr\left( \frac{\partial \dot\theta}{\partial \theta}\right) \right]\\
    &=  \mathbb{E}_{p(\theta_0)}\left[ \log p(\theta_0) - \log p_0(\theta_0 + \Delta\theta_\tau) + \tau \Tr \Theta \right]\\
    &=  \mathbb{E}_{p(\theta_0)}\left[ -\frac 12 \theta_0^2 + \frac 12 \left( \theta_0 + \Delta\theta_\tau \right)^2 \right] + \tau \Tr \Theta \\
    &=  \mathbb{E}_{p(\theta_0)}\left[  \theta_0^T \Delta\theta_\tau + (\Delta\theta_\tau   )^2 \right] + \tau \Tr \Theta \\
    &=  \mathbb{E}_{p(\theta_0)}\left[  \theta_\tau^T \Delta\theta_\tau \right] + \tau \Tr \Theta \\
    &=  \mathbb{E}_{p(\theta_0)}\left[  (z_0(\X) - \Y)^T (I-e^{-\tau\Theta})^2\Theta^{-1}(z_0(\X)-\Y) \right] + \tau \Tr \Theta \\
    &=  \Tr \left(\K \Theta^{-1} (I-e^{-\tau \Theta})^2 \right) + \Y^T \Theta^{-1} (I-e^{-\tau \Theta})^2 \Y  + \tau \Tr \Theta .
\end{align}
This tends to infinity as the time goes to infinity.  This renders the usual PAC-Bayes style
generalization bounds trivially vacuous for the generalization of infinitely wide neural networks
at late times.  Yet, infinite networks can generalize well~\citep{cando}.

\subsection*{WAIC}
\label{sec:waic_s}
Watanabe-Akaike Information Criterion (WAIC) first introduced in \citep{watanabe2010asymptotic}, and gives an asymptotically correct estimate of the gap between the training set and test set expectations. It is defined by the difference between Bayes' and Gibb's' errors;
\begin{align*}
    WAIC = \sum_i\left(\log(\mathbb{E}\left[p(y_i|\theta)\right])-\mathbb{E}[\left[\log(p(y_i|\theta))\right]\right)
\end{align*}
Training with squared loss is equivalent  to assuming a Gaussian observation model $p(y|z) \sim \N(z, I)$ and as we showed before that the expected log loss (Gibbs loss) has contributions both from the square loss of the mean prediction, as well
as a term which couples to the trace of the covariance:
\begin{flalign}
    \mathbb{E}\left[ \log q(y|z) \right] = \\
    -\frac 12 (y - \mu_t)^2 +\frac 12 Tr(\Sigma) 
    \\+\frac{k}{2}\log{\left((2\pi)|I+\Sigma_t|\right)}
\end{flalign}

For the Bayes loss 
\begin{align*}
    \log{\mathbb{E}\left[q(y|z)\right]}=-\frac 12 \left(y-\mu_t\right)^T\left(I+\Sigma_t)\right)^{-1}\left(y-\mu_t\right)
    \\+\frac{k}{2}\log{\left((2\pi)|I+\Sigma_t|\right)}
\end{align*}

The Woodbury matrix identity tells us --
\begin{align*}
    \left(A+UCV\right)^{-1}=A^{-1}-A^{-1}U\left(C^{-1}+VA^{-1}U\right)^{-1}VA^{-1}
\end{align*}
and if $A=I$, and $C=I$, and $U=I$ --
\begin{align*}
    \left(I+V\right)^{-1}=I-\left(I+V\right)^{-1}V
\end{align*}
we get:
\begin{align*}
    log{\mathbb{E}\left[q(y|z)\right]}=-\frac 12 \left(y-\mu_t\right)^2 +\frac 12 \left(y-\mu_t\right)\left(I+\Sigma_t)\right)^{-1}\Sigma\left(y-\mu_t\right)
    \\+\frac{k}{2}\log{\left((2\pi)|I+\Sigma_t|\right)}.    
\end{align*}
Combining all together
\begin{align*}
    WAIC =  \frac 12 \left(y-\mu_t\right)\left(I+\Sigma_t)\right)^{-1}\Sigma\left(y-\mu_t\right) 
   +\frac 12 Tr(\Sigma) 
\end{align*}

\section*{Experiments}
\addcontentsline{toc}{section}{Experiments}



The Gaussian Information Bottleneck ~\citep{gaussib}  gives the optimal trade-off betweenn $I(Z;X)$ and $I(Z;Y)$ for jointly Gaussian data, where $X$ is the input, $Y$ is the label, and $Z$ is a stochastic representation,  $p(z|x)$ of the input.

In the following, we fit infinite ensembles of infinitely-wide neural networks to jointly Gaussian data and calculate estimates of their mutual information. This allows us to determine how close these networks are to being optimal.

The Gaussian dataset we create has $|X|=30$ and $|Y|=1$ (for details, see Appendix A). We train 
a three-layer FC network 
with both \ReLU\ and \Erf\ activation functions. 

\begin{figure}[htb]
  \centering
  \subfloat[ReLU]{\includegraphics[width=0.45\textwidth]{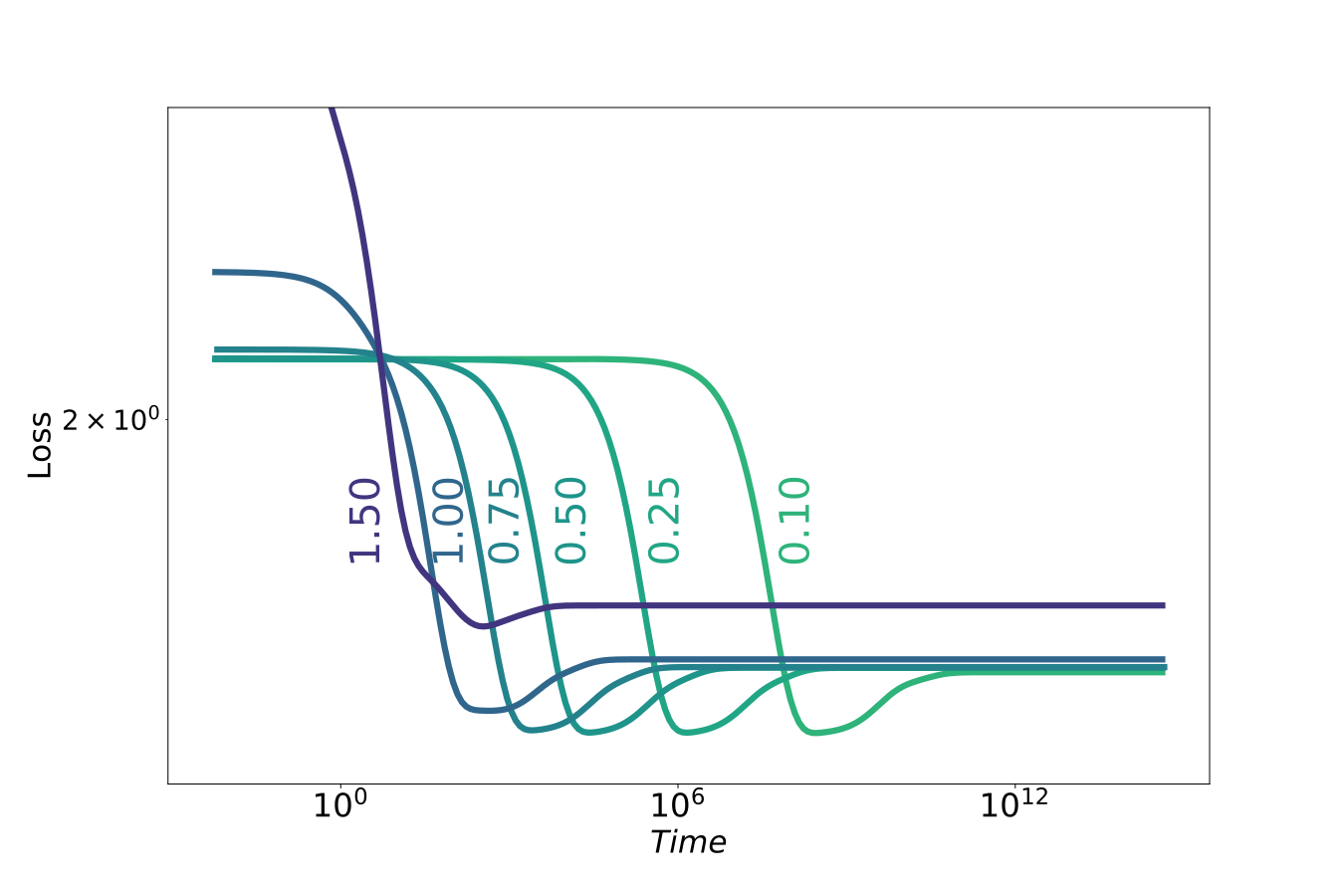} \label{fig:loss_vs_time_mnist_relu}}
  \subfloat[Erf]{\includegraphics[width=0.45\textwidth]{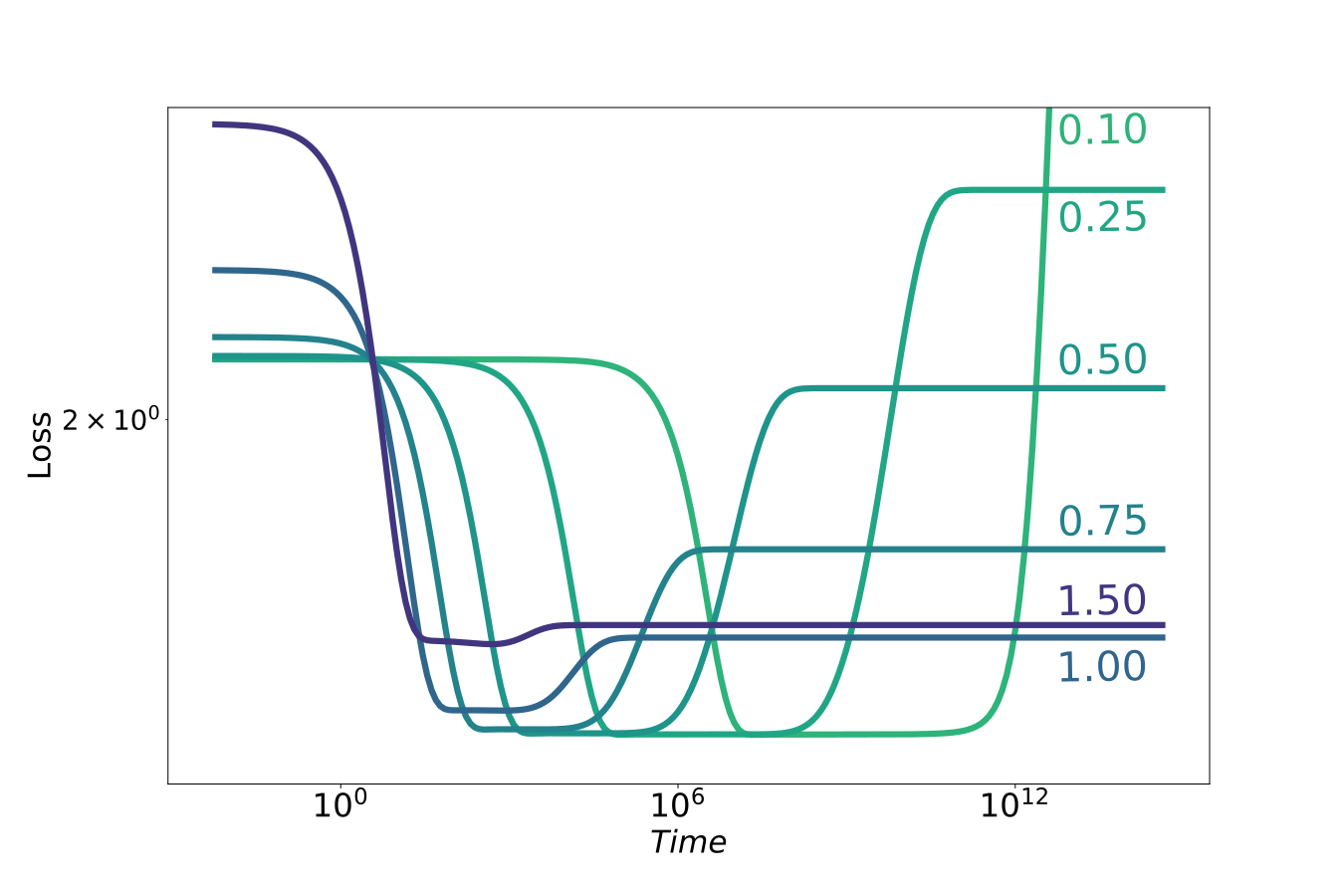}\label{ig:loss_vs_time_mnist_erf}}
        \caption{Loss as function of time for different initial weights' variances on the Gaussian dataset.}
         \label{fig:loss_vs_time_gauss}
\end{figure}

\Cref{fig:loss_vs_time_gauss} shows
the test set loss as a function of time for different choices of initial
weight variance ($\sigma_w^2$).  
For both the \ReLU\ and \Erf\ networks, at the highest $\sigma_w$ shown (darkest purple),
the networks \emph{underfit}. 
For lower initial weight variances, they all show signs of \emph{overfitting} in the sense that
the networks would benefit from early stopping.  This overfitting is worse
for the \Erf\ nonlinearity, 
where we see a divergence in the final test set loss as $\sigma_w$ decreases.
For all of these networks, the training loss goes to zero.

\begin{figure}[htb]
  \centering
  \subfloat[\ReLU]{\includegraphics[width=0.45\textwidth]{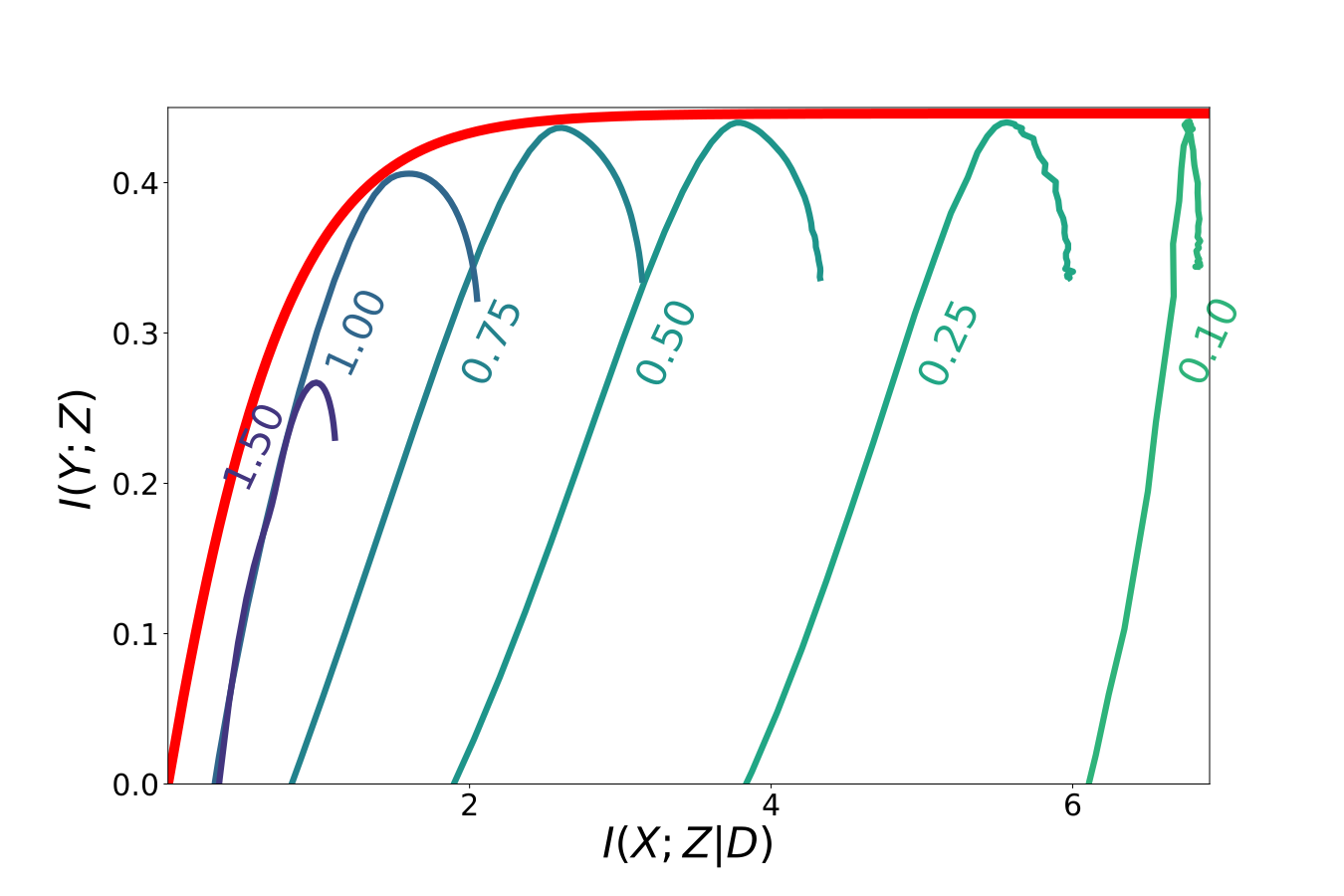} \label{fig:inf_plane_gauss_relu}}
  \subfloat[\Erf]{\includegraphics[width=0.45\textwidth]{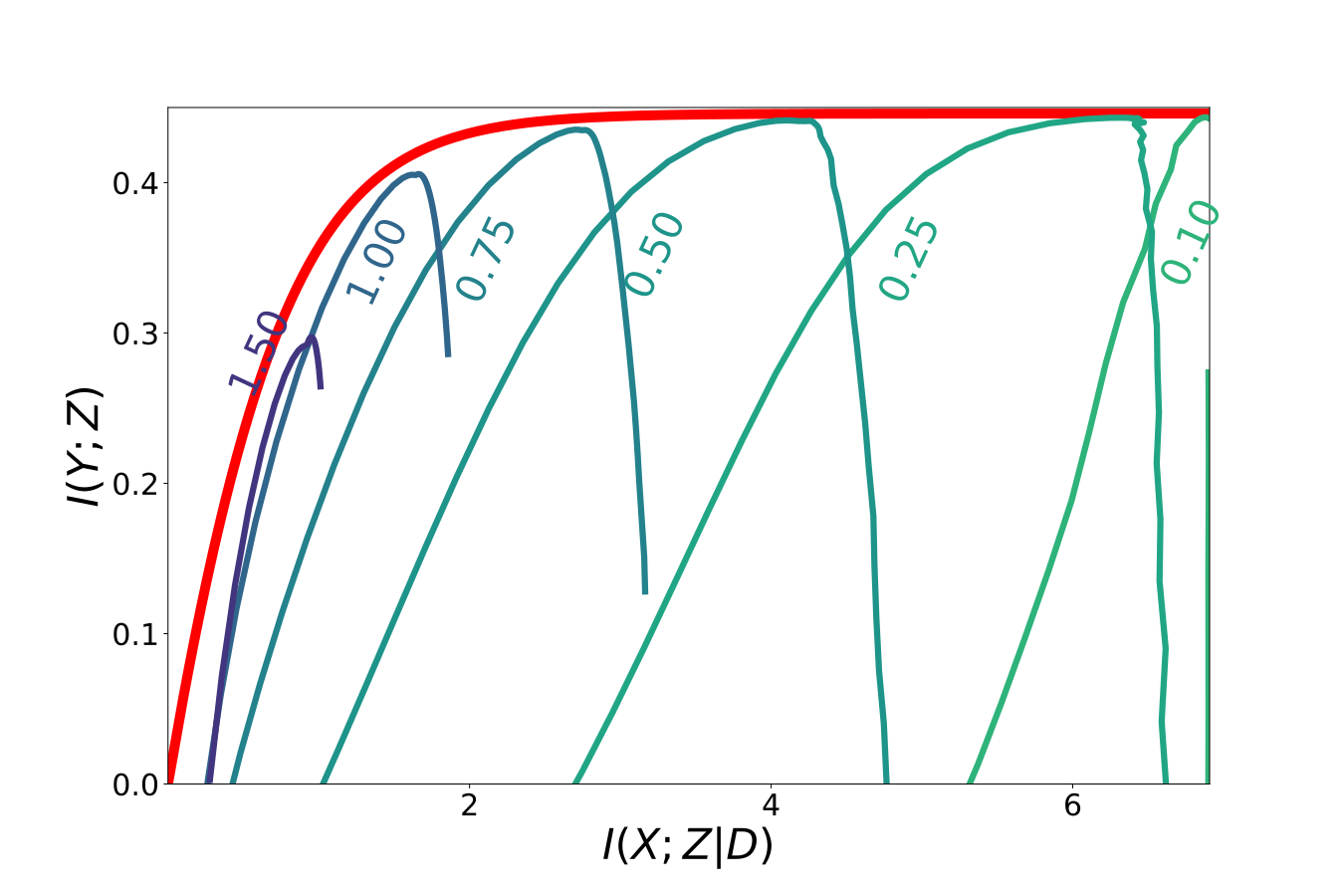}\label{fig:inf_plane_gauss_erf}}
        \caption{Trajectories of the (bounds on) MI between the representation $Z$ and the input $X$ versus time.
        Curves differ only in their initial weight variance.
        The red line is the optimal IB as predicted by theory.  
        Our estimate for $I(Z;X)$ is upper bounded by the log of the batch size ( $\log 1000 = 6.9$.)}
  \label{fig:inf_plane_gausssian}
\end{figure}

In~\cref{fig:inf_plane_gausssian} we show the performance of these networks on the information plane.
The $x$-axis shows a variational lower bound on the complexity of the learned representation: $I(Z;X|D)$.
The $y$-axis shows a variational lower bound on learned relevant information: $I(Y;Z)$.
For details on the calculation of the MI estimates see~\cref{sec:info}.
The curves show trajectories of the networks' representation
as time varies from $\tau = 10^{-2}$ to $\tau = 10^{10}$
for different weight variances (the bias-variance in all networks was fixed to 0.01).
The red line is the optimal theoretical IB bound.  

There are several features worth highlighting.  First, we emphasize the somewhat surprising
result that, as time goes to infinity, the MI between an infinite ensemble
of infinitely-wide neural networks output and their input is finite and quite
small.  Even though every individual network provides a seemingly rich deterministic
representation of the input, when we marginalize over the random initialization, the ensemble compresses
the input quite strongly.
The networks overfit at late times. 
For \Erf\ networks, the more complex representations ($I(Z;X|D)$) overfit more.
With optimal early stopping, over a wide range,
these models achieve a near-optimal tradeoff in prediction versus compression.
Varying the initial weight variance controls the amount of information the ensemble extracts.

Next, we repeat the result of the previous section on the MNIST dataset~\citep{mnist}. 
Unlike the typical setup, we turn MNIST into a binary regression task for the parity of the digit (even or odd).
This time, the network is a standard two-layer convolutional neural network with $5 \times 5$ filters
and either \ReLU\ or \Erf\ activation functions.

\begin{figure}[htb]
  \centering
  \subfloat[ReLU]{\includegraphics[width=0.45\linewidth]{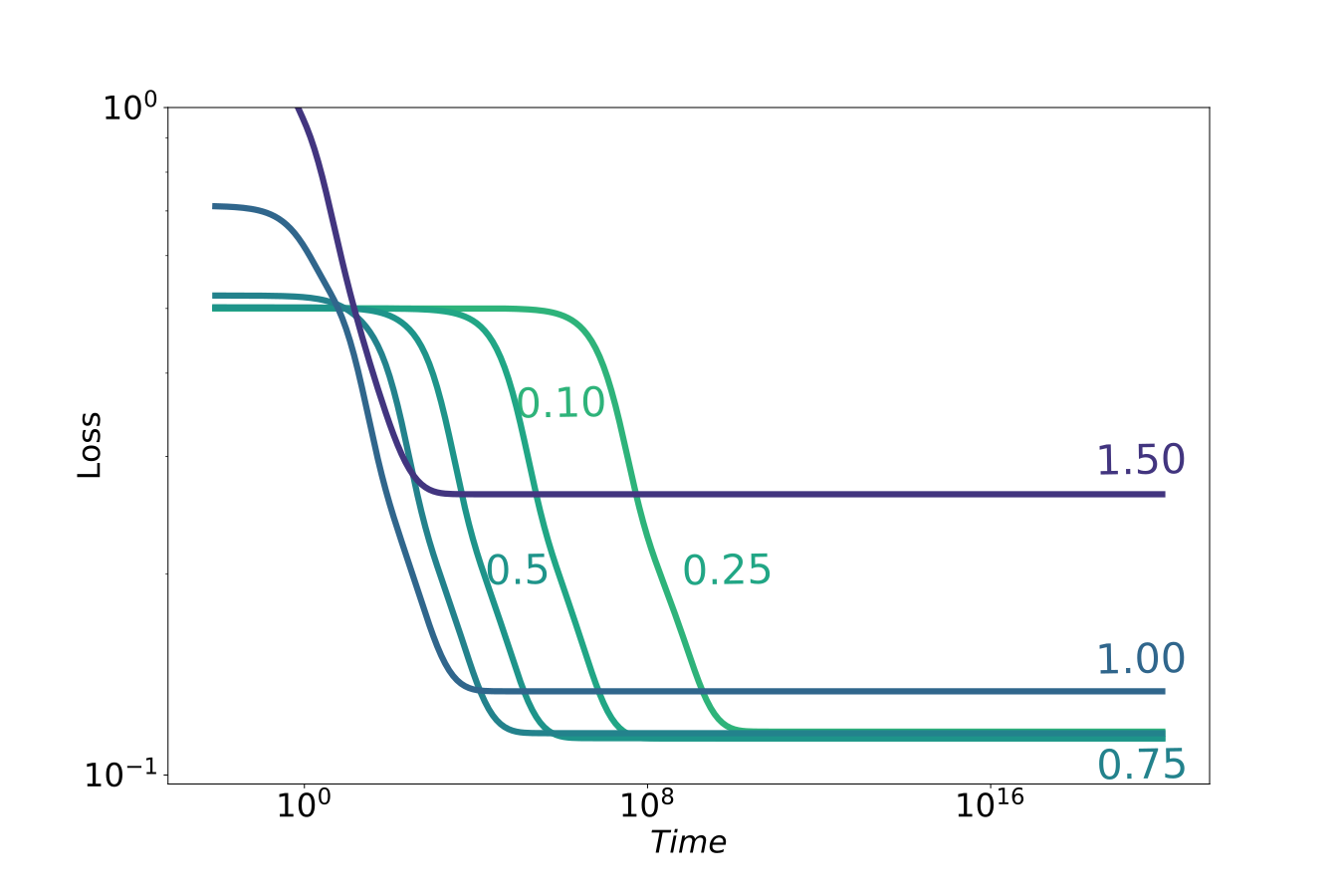} \label{fig:loss_vs_time_mnist_relu_2}}
  \subfloat[ERF]{\includegraphics[width=0.45\linewidth]{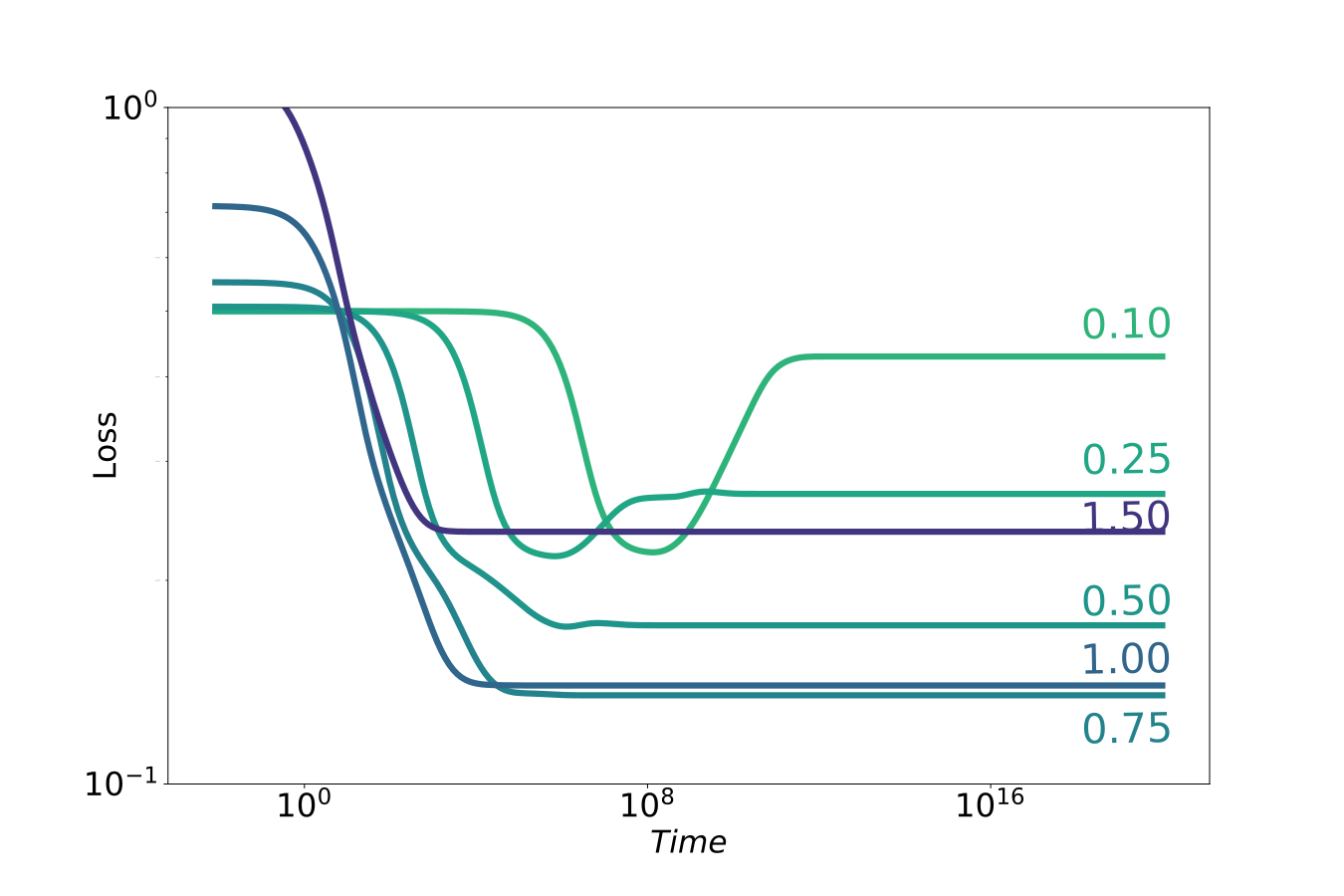}\label{fig:loss_vs_time_mnist_erf}} \\
  \subfloat[ReLU]{\includegraphics[width=0.45\textwidth]{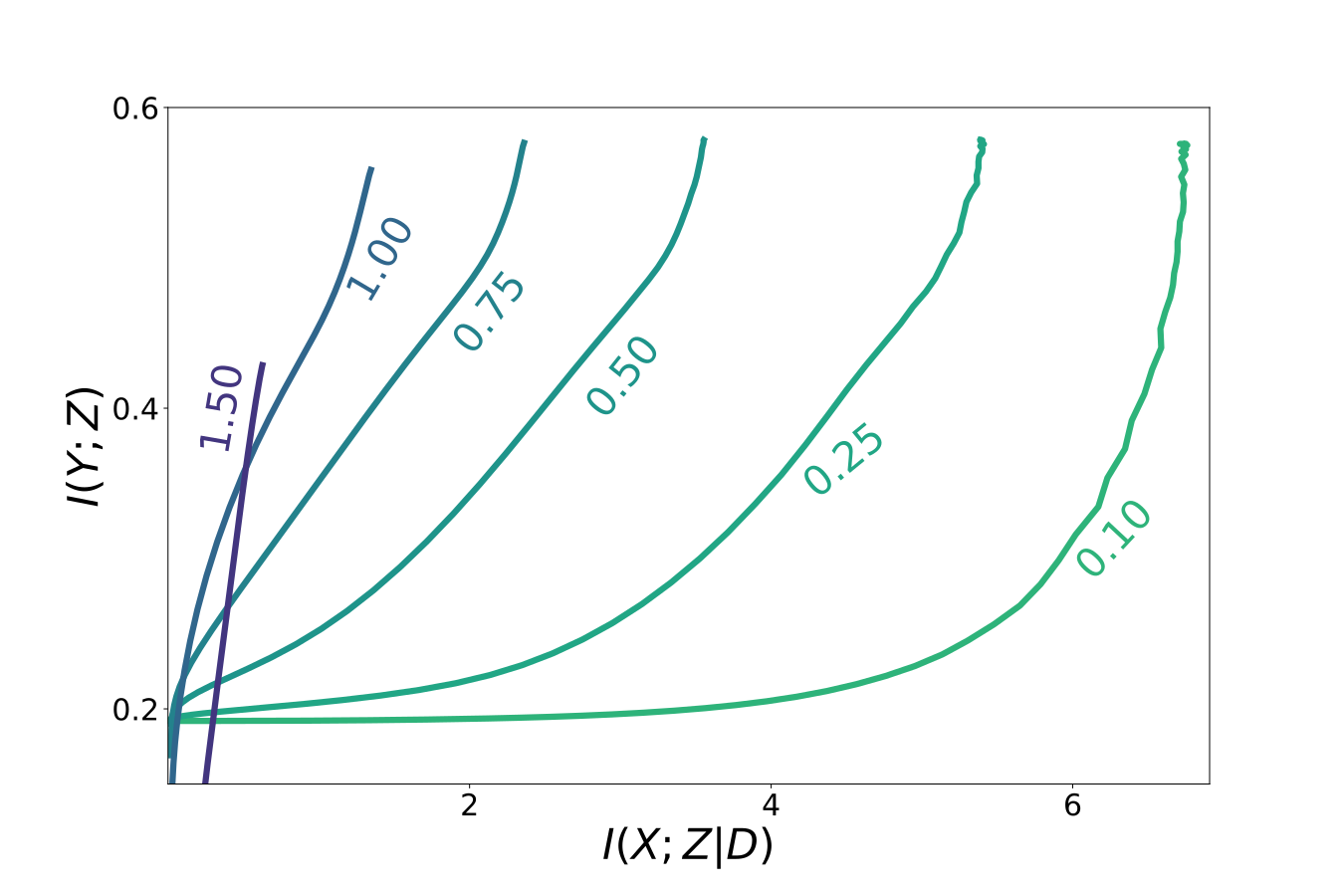} \label{fig:loss_vs_ws_mnist_relu}}
  \subfloat[ERF]{\includegraphics[width=0.45\textwidth]{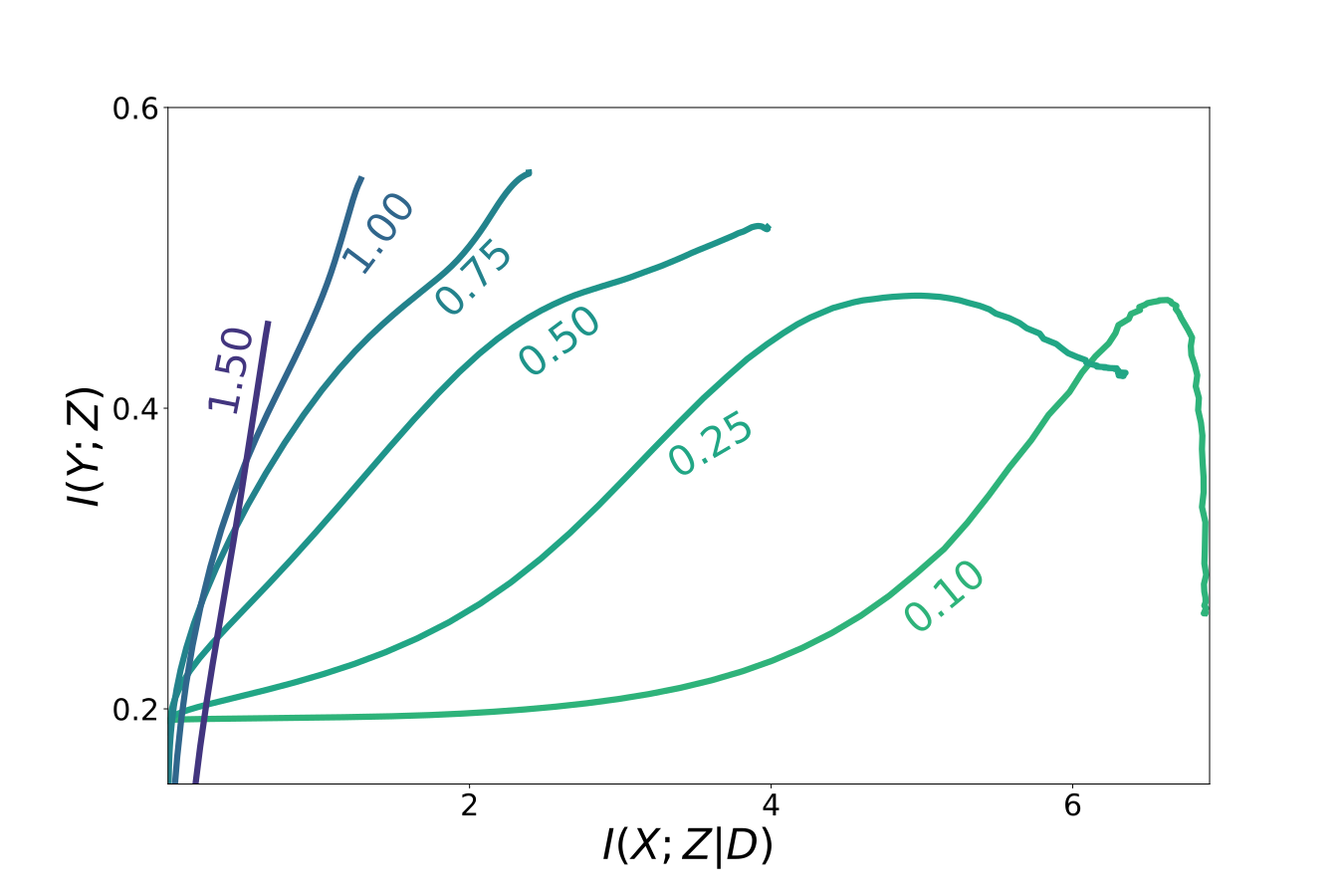}\label{fig:loss_vs_ws_mnist_erf}}
        \caption{Loss as function of time and information plan trajectories for different initial weights' variances on
        MNIST.}
         \label{fig:mnist2}

\end{figure}

\Cref{fig:mnist2} shows the results.
Unlike in the jointly Gaussian
dataset case, here both networks show some region of initial weight variances
that do not overfit in the sense of demonstrating any advantage from early stopping.  
The \Erf\ network at higher variances does show overfitting at low initial weight variances,
but the \ReLU\ network does not.
Notice that in the information plane, the \Erf\ network shows
overfitting at higher representational complexities ($I(Z;X)$ large),
while the \ReLU\ network does not.

\section*{Conclusions}
\addcontentsline{toc}{section}{Conclusion}

Infinite ensembles of infinitely-wide neural networks provide an interesting model family.
Being linear in their parameters, they permit a high number
of tractable calculations of information-theoretic quantities and their bounds.
Despite their simplicity, they still can achieve good generalization performance~\citep{cando}.
This challenges existing claims for the purported
connections between information theory and generalization in deep neural networks.
In this preliminary work, we laid the
groundwork for a larger-scale empirical and theoretical study of generalization
in this simple model family.  
Given that real networks approach this Family in their 
infinite width limit, we believe a better understanding of generalization in the NTK limit
will shed light on generalization in deep neural networks.
\clearpage
\bibliographystyle{dcu}

\bibliography{main}
\section*{Appendix}
\addcontentsline{toc}{section}{Appendix}

\section*{Appendix A - Gaussian Dataset}
\label{sec:gaussian}

For our experiments we used a jointly Gaussian dataset, for which there is an analytic solution
for the optimal representation~\citep{gaussib}.

Imagine a jointly Gaussian dataset, where we have $x_{ij} = L^x_{jk} \epsilon^x_{ik}$ with $\epsilon \sim \mathcal{N}(0, 1)$.  Make $y$ just an affine projection of $x$ with added noise.  
\begin{equation}
    y_{ij} = L^y_{jk}\epsilon_{ik} + A_{jk} x_{ik} = L^y_{jk}\epsilon^y_{ik} + A_{jk} L^x_{km}\epsilon_{im}^x.
\end{equation}
Both $x$ and $y$ will be mean zero.  We can compute their covariances.
\begin{equation} 
\Sigma^x_{jk} = \langle x_{ij} x_{ik} \rangle = \langle L_{jm}\epsilon_{im} L_{kl}\epsilon_{il} \rangle = L_{jm}L_{kl} \delta_{ml} = L_{jm}L_{km}
\end{equation}
Next look at the covariance of $y$.
\begin{align*}
    \Sigma^y_{jk} &= \langle y_{ij} y_{ik} \rangle \\ 
    &=  \left\langle \left(L^y_{jl}\epsilon^y_{il} + A_{jl} L^x_{lm}\epsilon_{im}^x\right)
    \left( L^y_{kn}\epsilon^y_{in} + A_{kn} L^x_{no}\epsilon_{io}^x\right) \right\rangle \\
 &=  L^y_{jl} L^y_{kn} \delta_{ln} + A_{jl}L^x_{lm}A_{kn}L_{no}\delta_{mo} \\
 &= L^y_{jn}L^y_{kn} + A_{jl}\Sigma^x_{ln} A_{kn}
\end{align*}

For the cross covariance:
\begin{align*}
    \Sigma^{xy}_{jk} &= \left\langle x_{ij} y_{ik} \right\rangle \\
    &=  \left\langle L^x_{jm} \epsilon^x_{im}
    \left( L^y_{kn}\epsilon^y_{in} + A_{kn} L^x_{no}\epsilon_{io}^x\right) \right\rangle \\
    &= L^x_{jm}A_{kn}L^x_{no} \delta_{mo} \\
    &= L^x_{jm}A_{kn} L^x_{nm} = \Sigma^x_{jn} A_{kn}
\end{align*}

So we have for our entropy of $x$:
\[ 
    H(X) = \frac {n_x}2 \log (2 \pi e) + n_x \log \sigma_x
\]
\[ 
    H(Y|X) = \frac{n_y}2 \log (2\pi e) + n_y \log \sigma_y 
\]

as for the marginal entropy, we will assume the SVD decomposition $A = U \Sigma V^T$
\[
    H(Y) = \frac{n_y}{2} \log (2\pi e) + \frac{n_y}{2} \log  \left| \sigma_y^2 I + \sigma_x^2 A A^T \right|=
    \frac{n_y}{2} \log (2\pi e) + \frac{1}{2} \sum_i \log \left( \sigma_y^2 + \sigma_x^2 \Sigma_i^2 \right)
\]

So, solving for the mutual information between $x$ and $y$ we obtain:

\[
    I(X;Y) = H(Y) - H(Y|X) = \frac 12 \sum_{i} \log \left( 1 + \frac{\sigma_x^2 \Sigma_i^2}{\sigma_y^2} \right)
\]

\section*{Appendix B - Additional Results}
 \label{sec:more_results}
 
shows $I(X;Z|D)$, $I(\theta;D)$, $\frac{dI(\theta;D)}{dt}$ and the loss as function of time for a fixed initial weight's variance ($\sigma_w = 0.25$).  (in log-log scale, notice that $y$-axes
are different for each measure). For both the \ReLU and \Erf networks,
we see clear features in each plot near the optimal test loss.

\subsection*{WAIC}
\label{sec:waic}
We know that the WAIC is the diffrence betweem the Bayes and the Gibbs errors, namely - 
\begin{align*}
    WAIC = \sum_i\left(\log(\mathbb{E}\left[p(y_i|\theta)\right])-\mathbb{E}[\left[\log(p(y_i|\theta))\right]\right)
\end{align*}
Training with squared loss is equivalent
to assuming a Gaussian observation model $p(y|z) \sim \N(z, I)$. 
We can marginalize out our representation to obtain
\begin{equation}
    \mathbb{E}\left[q(y|z)\right] = q(y|x) = \int dz\, q(y|z)p(z|x) \sim \N( \mu(x, \tau), I + \Sigma(x, \tau) ).
\end{equation}
The expected log loss (Gibbs loss) has contributions both from the square loss of the mean prediction, as well
as a term which couples to the trace of the covariance:
\begin{flalign}
    \mathbb{E}\left[ \log q(y|z) \right] = \\
    -\frac 12 \mathbb{E}\left[ ( y - z_t)^2  \right] +\frac{1}{2}\log{\left((2\pi)^k|I+\Sigma_t)|\right)}
    =\\
    -\frac 12 (y - \mu_t)^2 +\frac 12 Tr(\Sigma) 
    \\+\frac{k}{2}\log{\left((2\pi)|I+\Sigma_t|\right)}
\end{flalign}
here $k$ is the dimensionality of $y$.

For the Bayes loss 
\begin{align*}
    \log{\mathbb{E}\left[q(y|z)\right]}=-\frac 12 \left(y-\mu_t\right)^T\left(I+\Sigma_t)\right)^{-1}\left(y-\mu_t\right)
    \\+\frac{k}{2}\log{\left((2\pi)|I+\Sigma_t|\right)}
\end{align*}

and by the Woodbury matrix identity -
\begin{align*}
    \left(I+V\right)^{-1}=I-\left(I+V\right)^{-1}V
\end{align*}
and we get 
\begin{align*}
    log{\mathbb{E}\left[q(y|z)\right]}=-\frac 12 \left(y-\mu_t\right)^2 +\frac 12 \left(y-\mu_t\right)\left(I+\Sigma_t)\right)^{-1}\Sigma\left(y-\mu_t\right)
    \\+\frac{k}{2}\log{\left((2\pi)|I+\Sigma_t|\right)}    
\end{align*}
Combining all together
\begin{align*}
    WAIC =  \frac 12 \left(y-\mu_t\right)\left(I+\Sigma_t)\right)^{-1}\Sigma\left(y-\mu_t\right) 
   +\frac 12 Tr(\Sigma) 
\end{align*}
\subsection*{\texorpdfstring{$I(\theta;D$}{I(theta;D)}}
\begin{align*}
    I(\theta;D)= & \\ 
    &\mathbb{E}_{p(\theta_t)}\left[\log p(\theta_t|D)-\log p(\theta_t)\right] \\
    =& \mathbb{E}_{p(\theta_t)}\left[\log p(\theta_t|D)-\log \int dD^\prime p(D^\prime)p(\theta_t|D^\prime)\right] 
\end{align*}

%% file: Chapters/dualIB.tex
\title{The Dual Information Bottleneck}

%
\author{Zoe Piran\\
Racah Institute of Physics\\
The Hebrew University of Jerusalem\\
Jerusalem, Israel \\
\texttt{zoe.piran@mail.huji.ac.il} \\
\And
Ravid Shwartz-Ziv \\
School of Computer Science \\
The Hebrew University of Jerusalem\\
Jerusalem, Israel \\
\texttt{ravid.ziv@mail.huji.ac.il}
\And
Naftali Tishby \\
School of Computer Science\\
The Hebrew University  of Jerusalem\\
Jerusalem, Israel \\
\texttt{tishby@cs.huji.ac.il}
}

\chapter*{The Dual Information Bottleneck}
\addcontentsline{toc}{chapter}{5:   The Dual Information Bottleneck}

\textbf{Unpublished} \\
Zoe Piran, Ravid Shwartz-Ziv and Naftali Tishby (2020)
\newpage

\begin{center}
        \vspace*{0.5cm}
        \LARGE
        \textbf{The Dual Information Bottleneck} \\
        \vspace{0.8cm}
        \normalsize
    Zoe Piran \textsuperscript{1}
    Ravid Shwartz-Ziv \textsuperscript{2}
    Naftali Tishby\textsuperscript{2,3} \\
            \vspace{1.2cm}

    \textsuperscript{1} Racah Institute of Physics\\
                        The Hebrew University of Jerusalem\\
                        Jerusalem, Israel \\
    \textsuperscript{2} The Edmond and Lilly Safra Center for Brain Sciences, The Hebrew University, \\
  Jerusalem, Israel.\\
    \textsuperscript{3} School of Computer Science and Engineering, \\
    The Hebrew University, \\
  Jerusalem, Israel.\\
    \end{center}
\begin{center}
  \vspace*{0.5cm}
         \normalsize
        \textbf{Abstract} \\
\end{center}

The Information Bottleneck ({\ib}) framework is a general characterization of optimal representations obtained using a principled approach for balancing accuracy and complexity.
  Here we present a new framework, the Dual Information Bottleneck ({\dualib}), which resolves some of the known drawbacks of the {\ib}.
  We provide a theoretical analysis of the {\dualib} framework; (i) solving for the structure of its solutions (ii) unraveling its superiority in optimizing the \emph{mean prediction error exponent} and (iii) demonstrating its ability to preserve exponential forms of the original distribution.
  To approach large scale problems, we present a novel variational formulation of the {\dualib} for Deep Neural Networks. In experiments on several data-sets, we compare it to a variational form of the {\ib}. This exposes superior Information Plane properties of the {\dualib} and its potential in improvement of the error.

\section*{Introduction}
\addcontentsline{toc}{section}{Introduction}

\label{sec:intro}
The Information Bottleneck ({\ib}) method \cite{tishby99information}, is an information-theoretic framework for describing efficient representations of a ``input'' random variable $X$ that preserve the information on an ``output''  variable $Y$. In this setting the joint distribution of $X$ and $Y$, $p\brk*{x, y}$, defines the problem, or rule, and the training data are a finite sample from this distribution. The stochastic nature of the label is essential for the analytic regularity of the {\ib} problem. In the case of deterministic labels, we assume a {\emph{ noise model}} which induces a distribution. The representation variable $\hat{X}$  is in general a stochastic function of $X$ which forms a Markov chain $Y \rightarrow X \rightarrow \hat{X}$, and it depends on $Y$ only through the input $X$.   We call the map $p\brk*{\hx \mid x}$ the {\emph{encoder}} of the representation and denote by $p\brk*{y \mid \hx}$ the {\emph{Bayes optimal decoder}} for this representation; i.e., the best possible prediction of the \emph{desired label} $Y$ from the representation $\hX$.

The {\ib} has direct successful applications for representation learning in various domains, from vision and speech processing \cite{ma2019unpaired}, to neuroscience \cite{schneidman2001analyzing}, and Natural Language Processing \cite{li-eisner-2019}. 
Due to the notorious difficulty in estimating mutual information in high dimension, variational approximations to the {\ib} have been suggested and applied also to Deep Neural Networks (DNNs) \citep[e.g.,][]{Alemi2016DeepVI,achille2018emergence,Parbhoo2018CausalDI,poole2019variational}. Additionally, following \citep{shwartz2017}, several recent works tackled the problem of understanding DNNs using the {\ib} principle \citep{nash2018inverting, goldfeld2018estimating} 

Still, there are several drawbacks to the {\ib} framework which motivated this work. While the standard approach in representation learning is to
use the topology or a specific parametric  model over the input, the {\ib} principle and equations are completely non-parametric, and operate directly on the encoder and decoder probability distributions. Moreover, the original {\ib} formulation, as common in Information Theory, assumes full knowledge of the pattern-label distribution and does not relate directly to the task of prediction the label for unseen input patterns, when trained from finite samples. These issues were addressed before for general learning with the {\ib} \citep{slonim_MIB,DBLP:conf/alt/ShamirST08} and by extending it in the context of large DNNs by Achille et. al. \citep{achille2018emergence, achille2019information}.

Here, we address the above drawbacks by introducing a novel theoretical framework, the Dual Information Bottleneck ({\dualib}). The {\dualib} can account for 
known features of the data and use them to make better predictions over unseen examples, from small samples for large scale problems. Further, it emphasizes the prediction problem, inferring $\hY$, which wasn't present in the original {\ib} formulation due to the complete distributional knowledge assumption.

\subsection{Contributions of this work}
\label{sec:contribution}
We present here the Dual Information Bottleneck ({\dualib}) aiming to obtain optimal representations, which resolves the {\ib} drawbacks:
\begin{itemize}
\item We provide a theoretical analysis which obtains an analytical solution to the framework and compare its behaviour to the {\ib}.

\item For data which can be approximated by exponential families  we provide closed form solutions, {\expib}, which preserves the sufficient statistics of the original distribution.
\item We show that by accounting for the prediction variable, the {\dualib} formulation optimizes a bound over the error exponent. 

\item We present a novel variational form of the {\dualib} for Deep Neural Networks (DNNs) allowing its application to real world problems. Using it we empirically investigate the dynamics of the {\dualib} and validate the theoretical analysis.
\end{itemize}

\section*{Background} \label{sec:background}
\addcontentsline{toc}{section}{Background}

The Information Bottleneck ({\ib}) framework is defined as the trade off between the encoder and decoder mutual information values. It is defined by the minimization of the Lagrangian:
\begin{align}
\label{eq:IB_L}
    \mathcal{F}\brk[s]*{p_{\beta}\brk*{\hat{x} \mid x}; p_{\beta}\brk*{y \mid \hat{x} }}  =I(X;\hat{X}) - \beta I(Y; \hat{X})~,
\end{align}

independently over the convex sets of the normalized distributions, $\brk[c]*{p_{\beta}\brk*{\hat{x} \mid x}}$, $\brk[c]*{p_{\beta}\brk*{\hat{x}}}$ and $\brk[c]*{p_{\beta}\brk*{y \mid \hat{x}}}$, given a positive Lagrange multiplier $\beta$ constraining the information on $Y$, while preserving the Markov Chain $Y \rightarrow X \rightarrow \hat{X}$. Three self-consistent equations for the optimal encoder-decoder pairs, known as the \emph{{\ib} equations}, define the solutions to the problem. An important characteristic of the equations is the existence of critical points along the optimal line of solutions in the \emph{information plane} (presenting $I(Y;\hX)$ vs. $I(X;\hX)$) \citep{wu2020phase, parker}. 
The {\ib} optimization trade off can be considered as a generalized rate-distortion problem \cite{Cover:2006:EIT:1146355} with the distortion function, $d_{\ib}\brk*{x,\hx}=D\brk[s]*{p\brk*{y \mid x}||p_{\beta}\brk*{y|\hx}}$.
For more background on the {\ib} framework see \S \ref{app:ib}. 

\section*{The Dual Information Bottleneck}
\addcontentsline{toc}{section}{The Dual Information Bottleneck}

\label{sec:dualIB}

Supervised learning is generally separated into two phases: the training phase, in which 
 the internal representations are formed from the training data, and the prediction phase, in which  these representations are used to predict labels of new input patterns  \citep{shalev2014understanding}.
To explicitly address these different phases we add to the {\ib} Markov chain another variable, $\hY$, the \emph{predicted label} from the trained representation, which obtains the same values as $Y$ but is distributed differently: 
\begin{align} \label{eq:MC}
\rlap{$\overbrace{\phantom{Y \rightarrow X \rightarrow \hat{X}_{\beta}}}^{\textrm{training}}$}Y \rightarrow \underbrace{X \rightarrow
     \hat{X}_{\beta}\rightarrow \hat{Y}}_{\textrm{prediction}}.
\end{align}
The left-hand part of this chain describes the representation training, while the right-hand part is the Maximum Likelihood prediction using these representations \cite{slonim_MIB}. So far the prediction variable $\hY$ has not been a part of the $\ib$ optimization problem. 
It has been implicitly assumed that the \emph{Bayes optimal decoder}, $p_{\beta}\brk*{y \mid \hx}$, which maximizes the full representation-label information, $I(Y;\hX)$, for a given $\beta$, is also the best choice for making predictions. Namely, the prediction of the label, $\hY$, from the representation $\hX_{\beta}$ through the right-hand Markov chain by the mixture using the internal representations,
$ p_{\beta}\brk*{\hat{y} \mid x} \equiv \sum_{\hat{x}} p_{\beta}\brk*{y= \hat{y} \mid \hat{x}}  p_{\beta}\brk*{\hat{x} \mid x},$
is optimal when $p_{\beta}\brk*{y \mid \hx}$ is the \emph{Bayes optimal decoder}. However, this is not necessarily the case, for example when we train from finite samples \cite{DBLP:conf/alt/ShamirST08}.
 
 Focusing on the prediction problem, we define the {\dualib} distortion by switching the order of the arguments in the KL-divergence of the original {\ib} distortion, namely:
\begin{align}\label{eq:dist}
d_{\dualib}\brk*{x,\hx}&=D\brk[s]*{p_{\beta}\brk*{y\mid \hx} \| p\brk*{y \mid x}} 
=\sum_{y}p_{\beta}\brk*{y\mid \hx}\log \frac{p_{\beta}\brk*{y\mid \hx}}{p\brk*{y \mid x}}\ .    
\end{align}
In geometric terms this is known as the \emph{dual} distortion problem \cite{Ay2019}. The $\dualib$ optimization can then be written as the following rate-distortion problem:
\begin{align} \label{eq:min_func_dual_ib}
    \mathcal{F}^{*}\brk[s]*{p_{\beta}\brk*{\hat{x} \mid x} ;p_{\beta}\brk*{y \mid \hx }} &=  I(X;\hat{X})  
    + \beta \mathbb{E}_{p_{\beta}\brk*{x, \hx}}\brk[s]*{d_{\dualib}\brk*{x, \hx} }~.
\end{align}
As the decoder defines the prediction stage ($p_{\beta}\brk*{y=\hy \mid \hx}$) we can write (see proof in \S \ref{app:dual_ib}) the average distortion in terms of mutual information on $\hY$,
$I(\hX ; \hY)$ and $I(X ;\hY)$:
\begin{align} \label{eq:dual_dist_MI}
       \mathbb{E}_{p_{\beta}\brk*{x, \hx}}\brk[s]*{d_{\dualib}\brk*{x, \hx}} &=   \underbrace{I(\hX;\hY) - I( X;\hY)}_{(a)} 
       +  \underbrace{ \mathbb{E}_{p\brk*{x}}\brk[s]*{D\brk[s]*{p_{\beta}\brk*{\hy \mid x} \| p\brk*{y=\hy \mid x} }}}_{(b)} 
.\end{align}
This is similar to the known {\ib} relation: $ \mathbb{E}_{p_{\beta}\brk*{x, \hx}}\brk[s]*{d_{\ib}\brk*{x, \hx}} = I(Y;X) - I(Y;\hX) $
with an extra positive term $(b)$. Both terms, $(a)$ and $(b)$, vanish precisely when $\hX$ is a sufficient statistic for $X$ with respect to $\hY$. In such a case we can reverse the order of $X$ and $\hX$ in the Markov chain \eqref{eq:MC}. This replaces the roles of $Y$ and $\hY$ as the variable for which the representations, $\hX_{\beta}$, are approximately minimally sufficient statistics. 
In that sense the {\dualib}  shifts the emphasis from the training phase to the prediction phase.  This implies that minimizing the {\dualib} functional maximizes a lower bound on the mutual information $I(X;\hY)$.

\subsection{The {\dualib} equations}
\label{sec:dualIB_sols}
Solving the {\dualib} minimization problem \eqref{eq:min_func_dual_ib}, we obtain a set of self consistent equations. Generalized Blahut-Arimoto (BA) iterations between them converges to an optimal solution. The equations are similar to the original {\ib} equations with the following modifications: (i) Replacing the distortion by its dual in the encoder update; (ii) Updating the decoder by the encoder's geometric mean of the data distributions $p\brk*{y \mid x}$.
\begin{theorem}
\label{lm:dual_ib_eqs}
The {\dualib} equations are given by:
\begin{align} \label{eq:dIB}
	\begin{cases}
	\brk*{i}\ &p_{\beta}\brk*{\hx \mid x} = \frac{p_{\beta}\brk*{\hx} }{Z_{\hat{\rvx} \mid \rvx}\brk*{x;\beta} } e^{-\beta D\brk[s]*{ p_{\beta} \brk*{y \mid \hx} \| p\brk*{y\mid x} }} \\
	\brk*{ii}\ &p_{\beta}\brk*{\hat{x}} = \sum_{x} p_{\beta}\brk*{\hat{x} \mid x} p\brk*{x} \\
	\brk*{iii}\ &p_{\beta}\brk*{y \mid \hx } = \frac{ 1}{Z_{\rvy\mid \hat{\rvx}}\brk*{\hx; \beta}} \prod_{x}   p\brk*{y \mid x}^{p_{\beta}\brk*{x \mid \hx}} 
	\end{cases}
,\end{align}
where $Z_{\hat{\rvx} \mid \rvx}\brk*{x;\beta}, Z_{\rvy\mid \hat{\rvx}}\brk*{\hx; \beta}$ are normalization terms.
\end{theorem}
The proof is given in \S \ref{app:dual_ib_eq_proof}. 
It is evident that the basic structure of the equations of the {\dualib} and {\ib} is similar and 
they approach each other for large values of $\beta$. In the following sections we explore the implication of the differences on the properties of the new framework.

\subsection{The critical points of the {\dualib}}
\label{sec:bifurcation_points_dual}

As mentioned in \S \ref{sec:background} and \cite{wu2020phase}, the ``skeleton'' of the {\ib} optimal bound (the information curve) is constituted by the critical points in which the topology (cardinality) of the representation changes. 
Using perturbation analysis over the  {\dualib} optimal representations we find that small changes in the encoder and decoder that satisfy \eqref{eq:dIB} for a given $\beta$ are approximately determined through a nonlinear eigenvalues problem. \footnote{For simplicity we ignore here the possible interactions between the different representations.}

\begin{theorem} \label{th:stab_anl_dual}
The {\dualib} critical points are given by non-trivial solutions of the nonlinear eigenvalue problem:
\begin{align} \label{eq:stab_anl_dual}
	\brk[s]*{I - \beta 	C^{\dualib}_{xx'} \brk*{\hat{x}, \beta }} {\delta \log p_{\beta}\brk*{x' \mid \hat{x}}}&=0  
	~,~~~~  
	\brk[s]*{I - \beta C^{\dualib}_{y y'}\brk*{\hat{x}, \beta }}{\delta \log p_{\beta}\brk*{y' \mid \hat{x}}}=0
.\end{align}
The matrices $C^{\dualib}_{xx'}\brk*{\hat{x}; \beta}, C^{\dualib}_{yy'}\brk*{\hat{x}; \beta}$ have the same eigenvalues  $\brk[c]*{\lambda_{i}}$, with $\lambda_{1}(\hx) = 0$. With binary $y$, the critical points are obtained at $\lambda_{2}\brk*{\hx} =\beta^{-1}$.
\end{theorem}

\begin{figure}[htb!]
    \begin{center}
        \begin{subfigure}{.3\textwidth}
    \centering
    \includegraphics[width=\textwidth]{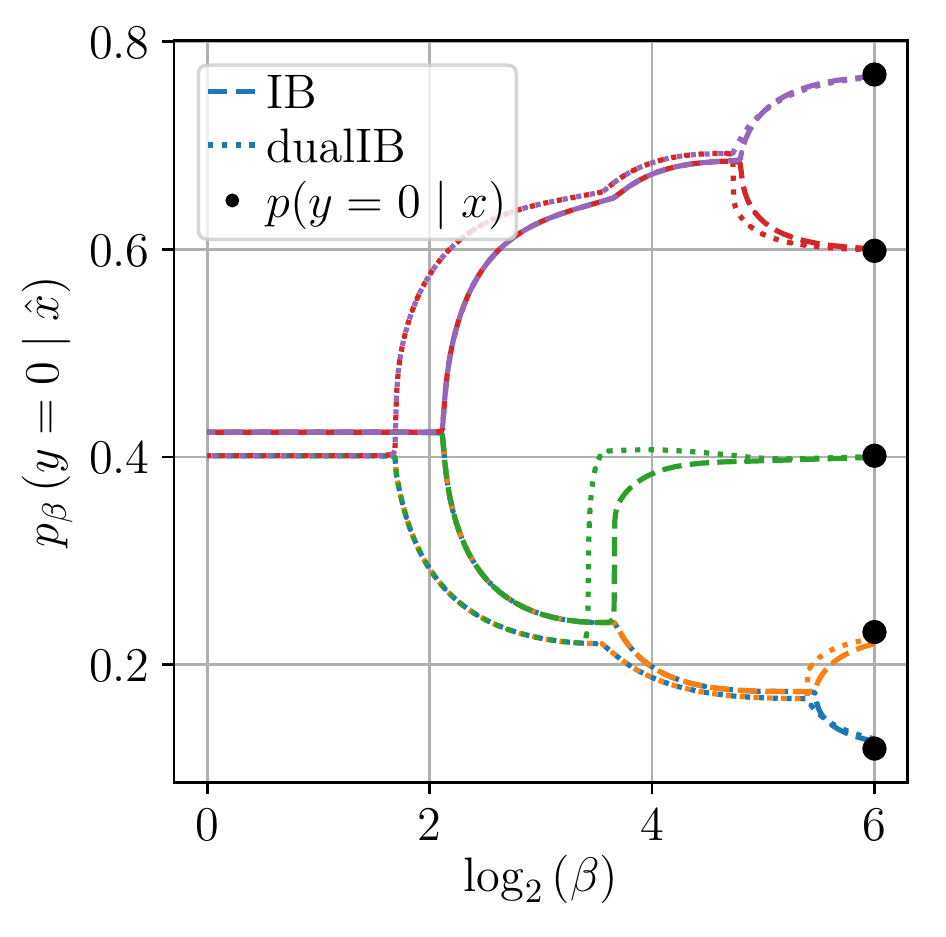}
    \caption{The bifurcation diagram}
    \label{fig:bif_a}
\end{subfigure}
            \begin{subfigure}{.3\textwidth}
    \centering
    \includegraphics[width=\textwidth]{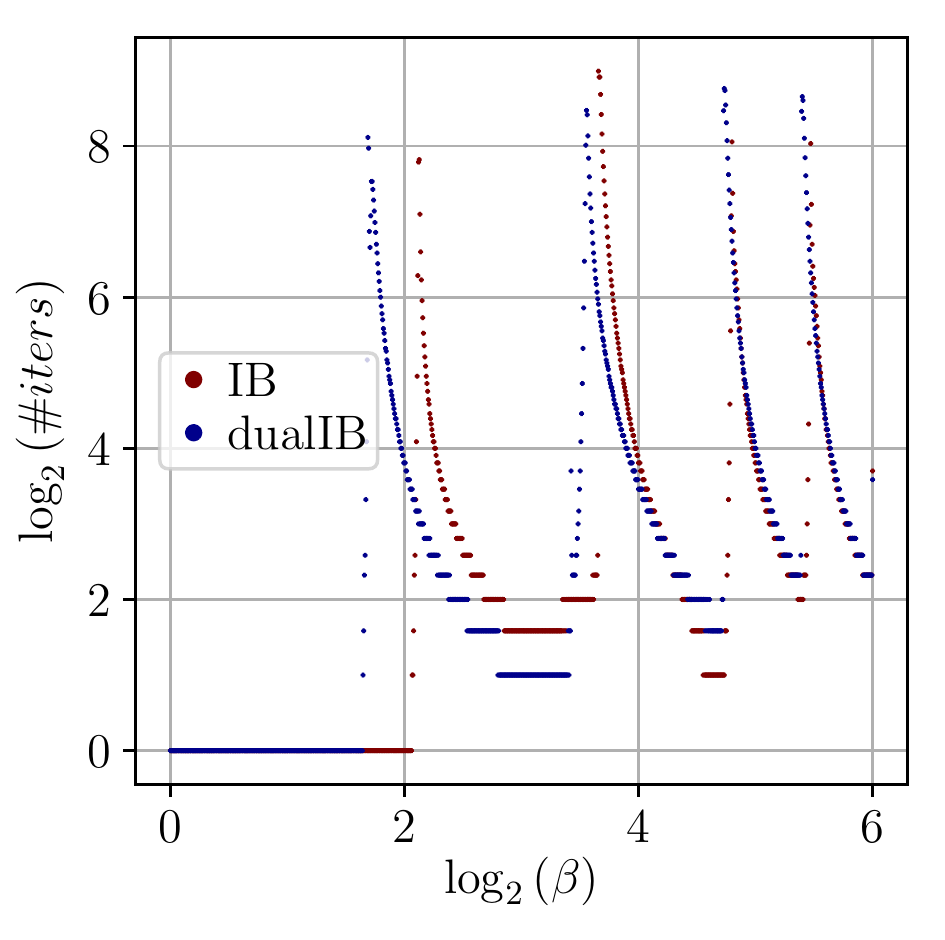}
    \caption{Convergence Time}
    \label{fig:bif_b}
\end{subfigure}
        \begin{subfigure}{.3\textwidth}
    \centering
    \includegraphics[width=\textwidth]{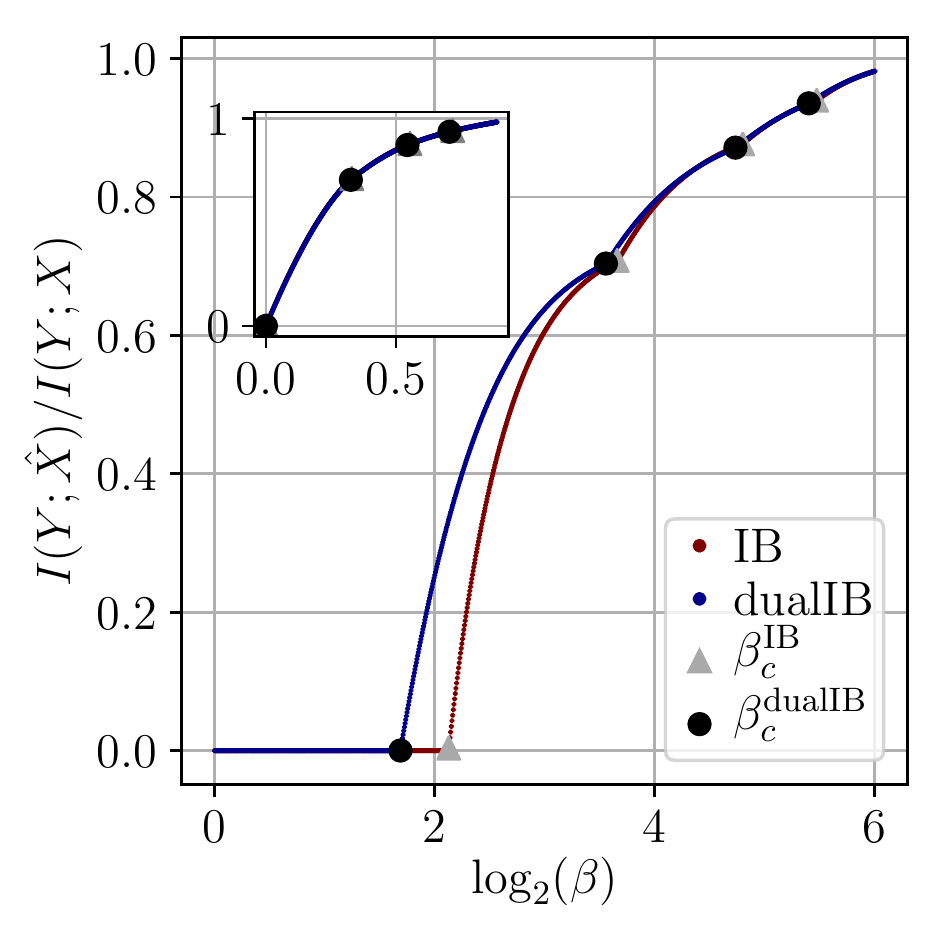}
    \caption{The information plane}
    \label{fig:bif_c}
\end{subfigure}
        \caption{$\brk{a}$ The \emph{bifurcation diagram}; each color corresponds to one component of the representation $\hx \in \hX$ and depicts the decoder $p_{\beta}\brk*{y = 0\mid \hat{x}}$. Dashed lines represent the {\ib}'s solution and dotted present the {\dualib}'s. The black dots denote the input distribution $p\brk*{y = 0 \mid x}$. $\brk*{b}$ Convergence time the BA algorithms as a function of $\beta$. $\brk*{c}$
         The desired label Information $I^{\ib}_{y}\brk*{\beta}$ and $I^{\dualib}_{y}\brk*{\beta}$ as functions of $\beta$. The inset shows the information plane, $I_{y}$  vs. $I_{x}$. The black dots are the {\dualib}'s critical points, $\beta^{\dualib}_c$, and the grey triangles are the {\ib}'s, $\beta^{\ib}_c$}
        \label{fig:bif_comb}
        \end{center}
  \end{figure}

The proof to Theorem \ref{th:stab_anl_dual} along with the structure of the matrices  $C^{\dualib}_{xx'}\brk*{\hat{x}; \beta}, C^{\dualib}_{yy'}\brk*{\hat{x}; \beta}$ is given in \S \ref{app:stability_anl}.
We found that this solution is similar to the nonlinear eigenvalues problem for the {\ib}, given in \S \ref{app:ib}.
As in the {\ib}, at the critical points we observe cusps 
with an undefined second derivative in the mutual information values as functions of $\beta$ along the optimal line.
That is the general structure of the solutions is preserved between the frameworks, as can be seen in \fref{fig:bif_c}. 

The \emph{Information Plane}, $I_{y}=I(\hX;Y)$  vs. $I_{x}=I(\hX;X)$, is the standard depiction of the compression-prediction trade-off of the {\ib} and has known analytic properties\citep{Gilad-bachrach}. First, we note that both curves obey similar constraints, as given in \lmref{lm:info_concave} below. 
\begin{lemma} \label{lm:info_concave}
  along the optimal lines of $I_{x}(\beta)$ and $I_{y}(\beta)$ the curves are non-decreasing
 piece-wise concave 
 functions of $\beta$. 
 When their second derivative (with respect to $\beta$) is defined, it is strictly negative.
\end{lemma} 

Next, comparing the {\dualib}'s and {\ib}'s information planes we find several interesting properties which are summarized in the following {Theorem} (see \S \ref{app:performance_anl} for the proof).

\begin{theorem} \label{th:dib_plane}
(i) The critical points of the two algorithms alternate, $\forall i, i+1$, $\beta_{c,i}^{\dualib} \leq \beta_{c,i}^{\ib} \leq \beta_{c,i+1}^{\dualib} \leq \beta_{c,i}^{\ib}$. (ii) The  distance between the two information curves is minimized at $\beta_{c}^{\dualib}$. (iii) The two curves approach each other as $\beta \rightarrow \infty$.
\end{theorem}

From Theorem \ref{th:dib_plane} we deduce that as the dimensionality of the problem increases (implying the number of critical points grows) the {\dualib}'s approximation of the {\ib}'s information plane becomes tighter. We illustrate the behavior of the  {\dualib}'s solutions in comparison to the {\ib}'s on a low-dimensional problem that is easy to analyze and visualize, with a binary $Y$ and only $5$ possible inputs $X$ (the complete definition is given in \S \ref{app:def_prob}). For any given $\beta$, the encoder-decoder iterations converge to stationary solutions of the \emph{{\dualib}} or \emph{{\ib} equations}. The evolution of the optimal decoder, $p_{\beta}\brk*{y=0 \mid \hx}$, $\forall \hx \in \hX$, as a function of $\beta$, forms a \emph{bifurcation diagram} (\fref{fig:bif_a}), 
 in which 
the critical points define the location of the bifurcation. At the critical points the number of iterations diverges (\fref{fig:bif_b}).
While the overall structure of the solutions is  similar, we see a ``shift'' in the appearance of the representation splits between the two frameworks. Specifically, as predicted by Theorem \ref{th:dib_plane} the {\dualib} bifurcations occur at lower $\beta$ values than  those of the {\ib}. The inset of \fref{fig:bif_c} depicts this comparison between the two information curves. While we know that $I^{\ib}_{y}\brk*{\beta}$ is always larger, we see that for this setting the two curves are almost indistinguishable.  Looking at $I_{y}$ as a function of  $\beta$ (\fref{fig:bif_c}) the importance  of the critical points is revealed as the corresponding cusps along these curves correspond to ``jumps'' in the accessible information. Furthermore, the distance between the curves is minimized precisely at the dual critical points, as predicted by the theory. 

\section*{The Exponential Family dualIB}
\addcontentsline{toc}{section}{The Exponential Family dualIB}

\label{sec:dualExpIb}

One of the major drawbacks of the {\ib} is that it fails to capture an existing parameterization of the data, that act as minimal sufficient statistics for it.
Exponential families are the class of parametric distributions for which minimal sufficient statistics exist,  forming an elegant theoretical core of parametric statistics and often emerge as maximum entropy \cite{Jaynes58} or stochastic equilibrium distributions, subject to observed expectation constraints. 
As the {\ib} ignores the structure of the distribution, given data from an exponential family it won't consider these known features. Contrarily, the {\dualib} accounts for this structure and its solution are given in terms of these features, defining the {\expib} equations.

We consider the case in which the rule distribution is of the form, $p\brk*{y \mid x} = e^{-\sum_{r=0}^{d} \lambda^r(y)A_r(x)} $,
where $A_r(x)$ are $d$ functions of the input $x$  and $\lambda^r(y)$ are functions of the label $y$, or the parameters of this exponential family \footnote{
The normalization factors, $Z_{\rvy \mid \rvx}\brk*{x}$, are written, for brevity, as $\lambda_{\rvx}^{0} \equiv \log ( \sum_{y} \prod_{r=1}^{d} e^{- {\lambda}^{r}\brk*{y} A_{r} \brk*{x} })$ with $A_{0} \brk*{x} \equiv 1 $.
We do not constrain the marginal $p\brk*{x}$.}. For exponential forms the mutual information, $I(X;Y)$, is fully captured by the $d$ conditional expectations. This implies that all the relevant information (in the training sample) is captured by $d$-dimensional empirical expectations which can lead to a reduction in computational complexity.

Next we show that for the {\dualib}, for all values of $\beta$, this dimension reduction is preserved or improved along the dual information curve. 
The complete derivations are given in \S \ref{app:dualexpIB}.

\begin{theorem} ({\expib})
\label{th:Exp-reduction}
For data from an exponential family the optimal encoder-decoder of the {\dualib} are given by:
\begin{align}  
\label{eq:dual_dec}
	 p_{\beta}\brk*{\hx\mid x} &= \frac{p_{\beta}\brk*{\hx}e^{\beta {\lambda}^{0}_{\beta}\brk*{\hx}}}{Z_{\hat{\rvx}\mid \rvx}\brk*{x;\beta} }  e^{-\beta  \sum_{r=1}^{d} {\lambda}_{\beta}^r\brk*{\hat{x}}   \brk[s]*{A_r \brk*{x} - A_{r, \beta}\brk*{\hat{x}}} } \nonumber \\
p_{\beta}\brk*{y\mid  \hx }  &= e^{-\sum_{r=1}^d \lambda^r \brk*{y} A_{r,\beta} \brk*{\hat{x}} -\lambda^0_{\beta}(\hx)} 
~, ~~~
 \lambda^0_{\beta}(\hx) = \log (\sum_{y} e^{-\sum_{r=1}^{d}  \lambda^r\brk*{y} A_{r,\beta} \brk*{\hx}})
,\end{align}
 with the constraints and multipliers expectations,
\begin{align}  
A_{r,\beta}\brk*{\hat{x}} &\equiv \sum_{x} p_{\beta}\brk*{x\mid \hat{x}} A_r \brk*{x}  
~,~
\lambda^r_{\beta}\brk*{\hat{x}} \equiv \sum_{y} p_{\beta}\brk*{y\mid \hat{x}} \lambda^r\brk*{y}~,~ 1\le r \le d ~.
\end{align}
\end{theorem}

This defines a simplified iterative algorithm  for solving
the {\expib} problem. Given the mapping of $x \in X$ to $\brk[c]*{A_{r}\brk*{x}}_{r=1}^{d}$ the problem is completely independent of $x$ and we can work in the lower dimensional embedding of the features, $A_{r}\brk*{x}$. 
Namely,   the update procedure 
is reduced to
 the dimensions of the sufficient statistics. Moreover, the representation is given in terms of the original features, a desirable feature for any model based problem.

\section*{Optimizing the Error Exponent}
\addcontentsline{toc}{section}{Optimizing the Error Exponent}

\label{sec:min_err_exp}
The {\dualib} optimizes an upper bound on the error exponent of the representation multi class testing problem. The error exponent accounts for the decay of the prediction error as a function of data size $n$. This implies the {\dualib} tends to  minimize the prediction error. For the classical binary hypothesis testing, the classification Bayes error, $P^{(n)}_e$, is the weighted sum of type 1 and type 2 errors. For large $n$, both errors decay exponentially with the test size $n$, and the best error exponent, $D^{*}$, is given by the Chernoff information. The Chernoff information is also a measure of distance  defined as, $ C\brk*{p_0, p_1}= \min_{0<\lambda <1 } \brk[c]*{\log \sum_{x} p_0^{\lambda}\brk*{x}p_1^{1-\lambda}\brk*{x}} $,  
 and we can understand it as an optimization on the $\log$-partition function of $p_{\lambda}$ to obtain $\lambda$ (for further information see \cite{Cover:2006:EIT:1146355} and \S \ref{app:err_exp}). 

The intuition behind the optimization of 
$D^{*}$ by the {\dualib} is in its distortion, the order of the prediction and the observation which implies the use of geometrical mean. The best achievable exponent (see \cite{Cover:2006:EIT:1146355}) in Bayesian probability of error is given by the KL-distortion between $p_{\lambda^{*}}$ ($ \propto p_0^{\lambda^*}\brk*{x}p_1^{1-\lambda^*}\brk*{x}$) to $p_0$ or $p_1$, such that $p_{\lambda^*}$ is the mid point between $p_0$ and $p_1$ on the geodesic of their geometric means.  
 By mapping the {\dualib} decoder to $\lambda$, it follows that the above minimization is proportional to the $\log$-partition function of $p_{\beta}\brk*{x \mid \hx}$, namely 
we obtain the mapping $p_{\beta}\brk*{x \mid \hx} = p_{\lambda}$. 

In the generalization to 
 multi class classification 
the error exponent is given by the pair of hypotheses with the minimal Chernoff information \cite{westover2008asymptotic}.
However, finding this value is generally 
difficult 
as it requires solving for each pair in the given classes.  Thus,
we consider an upper bound to it, the mean of the Chernoff information terms over classes. The representation variable adds a new dimension on which we average on and we obtain a bound on the optimal (in expectation over $\hx$) achievable exponent, $\hat{D}_{\beta}
 =\min_{p_{\beta}\brk*{y\mid \hx}, p_{\beta}\brk*{ \hx \mid x}}\mathbb{E}_{p_{\beta}\brk*{y, \hx}}\brk[s]*{D\brk[s]*{ p_{\beta}\brk*{ x \mid \hx} \mid p\brk*{x\mid y}}}$.
This expression is bounded from above by the {\dualib} minimization problem. 
Thus, the {\dualib} (on expectation) minimizes the prediction error for every $n$. A formal derivation of the above along with an analytical example of a multi class classification problem is given in \S \ref{app:err_exp}. In \S \ref{sec:experiment_setsize} we experimentally demonstrate that this also holds  for a variational {\dualib} framework using a DNN.

\section*{Variational Dual Information Bottleneck}
\addcontentsline{toc}{section}{Variational Dual Information Bottleneck}

The Variational Information Bottleneck (VIB) approach introduced by Achille and Soatto \citep{achille2018information} and Alemi et al. \citep{Alemi2016DeepVI} allows to parameterize the {\ib} model using Deep Neural Networks (DNNs). The variational bound of the {\ib} is obtained using DNNs for both the encoder and decoder. 
Since then, various extensions have been made \citep{strouse2017deterministic, elad2019direct} demonstrating promising attributes. 
Recently, along this line, the Conditional Entropy Bottleneck (CEB) \citep{fischer2018conditional} was proposed. The CEB 
provides variational optimizing bounds on $I(Y;\hX)$, $I(X;\hX)$ using  a variational decoder $q\brk*{y\mid \hx}$, variational conditional marginal, $q\brk*{\hx\mid y}$, and a variational encoder,  $p\brk*{\hx\mid x}$, all implemented by DNNs.

Here, we present the variational {\dualib} ({\vdib}), which optimizes the variational {\dualib} objective for using in DNNs. 
Following the CEB formalism, we bound the {\dualib} objective. We develop a variational form of the {\dualib} distortion and combine it with the bound for $I(X;\hX)$ (as in the CEB). This gives us the following Theorem (for the proof see \S \ref{app:vdib_obj}.).
\begin{theorem}
The {\vdib} objective is given by:
\begin{align}
    \min_{q\brk*{\hx \mid y}, p\brk*{\hx\mid x}}\brk[c]*{\mathbb{E}_{\tilde{p}\brk*{y\mid x}p\brk*{\hx\mid x}p\brk*{x}}\brk[s]*{
   \log\frac{p\brk*{\hx\mid x}}{q\brk*{\hx\mid y}} } + \beta \mathbb{E}_{p\brk*{y\mid\hx}p\brk*{\hx\mid x}}\brk[s]*{\log \frac{p\brk*{y\mid \hx}}{\tilde{p}\brk*{y\mid x}}}}
,\end{align}
where $\tilde{p}\brk*{y \mid x}$ is a distribution based on the given labels of the data-set, which we relate to as the noise model. Under the assumption that the noise model captures the distribution of the data the above provides a variational upper bound to the {\dualib} functional \eqref{eq:min_func_dual_ib}.
\end{theorem}
Due to the nature of its objective the {\dualib} requires a noise model.
 We must account for the contribution to the objective arising
 from $\tilde{p}\brk*{y \mid x}$
 which could be ignored in the {\vib} case. The noise model can be specified by its assumptions over the data-set. In  \S \ref{sec:cifar10_inf} we elaborate on possible noise models choices and their implications on the performance. Notice that the introduction of $\tilde{p}(y \mid x)$ implies that, unlike most machine learning models, the {\vdib} does not optimize directly the error between the predicted and desired labels in the training data. Instead, it does so implicitly with respect to the noisy training examples. This is not unique to the {\vdib}, as it is equivalent to training with noisy labels, often done to prevent over-fitting. For example, in \citep{muller2019does} the authors show that label noise can improve  generalization that results in a reduction in the mutual information between the input and the output. 

In practice, similarly to the CEB, for the stochastic encoder,  $p(\hx \mid x)$, we use the original architecture, replacing the final softmax layer with a linear dense layer with $d$ outputs. These outputs are taken as the means of a multivariate Gaussian distribution with unit diagonal covariance. For the variational decoder, $q(y \mid \hx)$, any classifier network can be used. We take a linear softmax classifier which takes the encoder as its input. The reverse decoder $q(\hx \mid y)$ is implemented by a network which maps a one-hot representation of the labels to the $d$-dimensional output interpreted as the mean of the corresponding Gaussian marginal.

\subsection{Experiments}
To investigate the {\vdib} on real-world data we compare it to the CEB model 
using a DNN over two data-sets, FasionMNIST and CIFAR10. 
For both, we use a Wide ResNet $28-10$ \citep{zagoruyko2016wide} as the encoder, a one layer Gaussian decoder and a single layer linear network for the reverse decoder (similarly to the setup 
in \citep{fischer2020ceb}). We use the same architecture to  train networks with {\vdib} and {\vib} objectives.
(See \S \ref{app:exp_setup} for details on the experimental setup). We note that in our attempts to train over the CIFAR100 data-set the results did not fully agree with the results on the above data-sets (for more information see \S \ref{app:cifar100}).
An open source implementation is available \href{https://github.com/ravidziv/dual_IB.git}{here}. 

\subsubsection{The variational information plane}
As mentioned, the information plane describes the compression-prediction trade-off. It enables us to compare 
different models and evaluate their ``best prediction level'' in terms of the desired label information, for each compression level. 
In \citep{fischer2020ceb} the authors provide empirical evidence that information bottlenecking techniques can improve both generalization and robustness. Other works \citep{fischer2018conditional, achille2018emergence, achille2018information} provide both theoretical and conceptual  insights into why these improvements occur.

\begin{figure}
\begin{subfigure}{.3\textwidth}
    \centering
    \includegraphics[width=\textwidth]{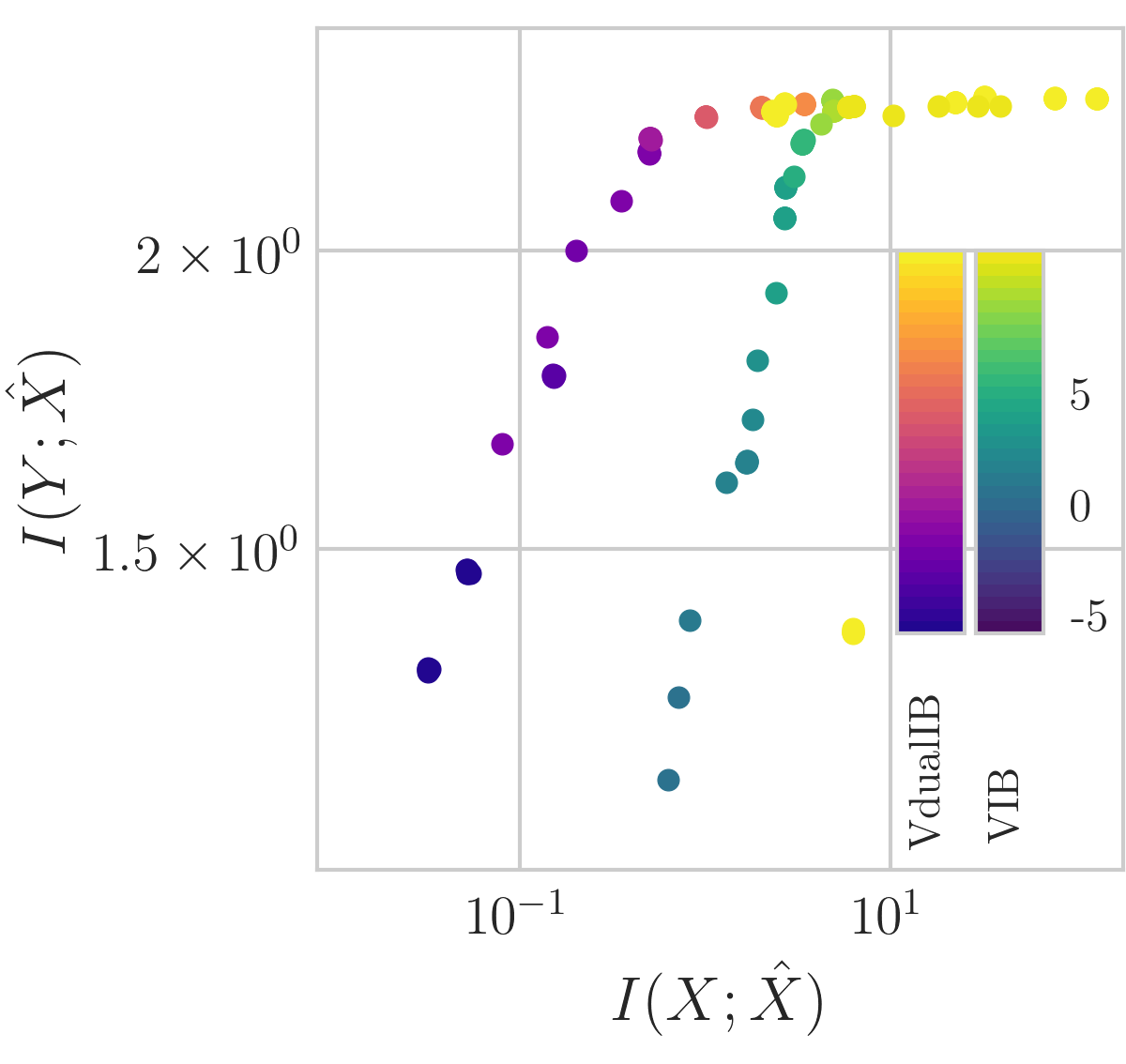}
    \caption{The information plane}
    \label{fig:inf_plane_mnist_beta}
\end{subfigure}
\begin{subfigure}{.3\textwidth}
    \centering
    \includegraphics[width=\textwidth]{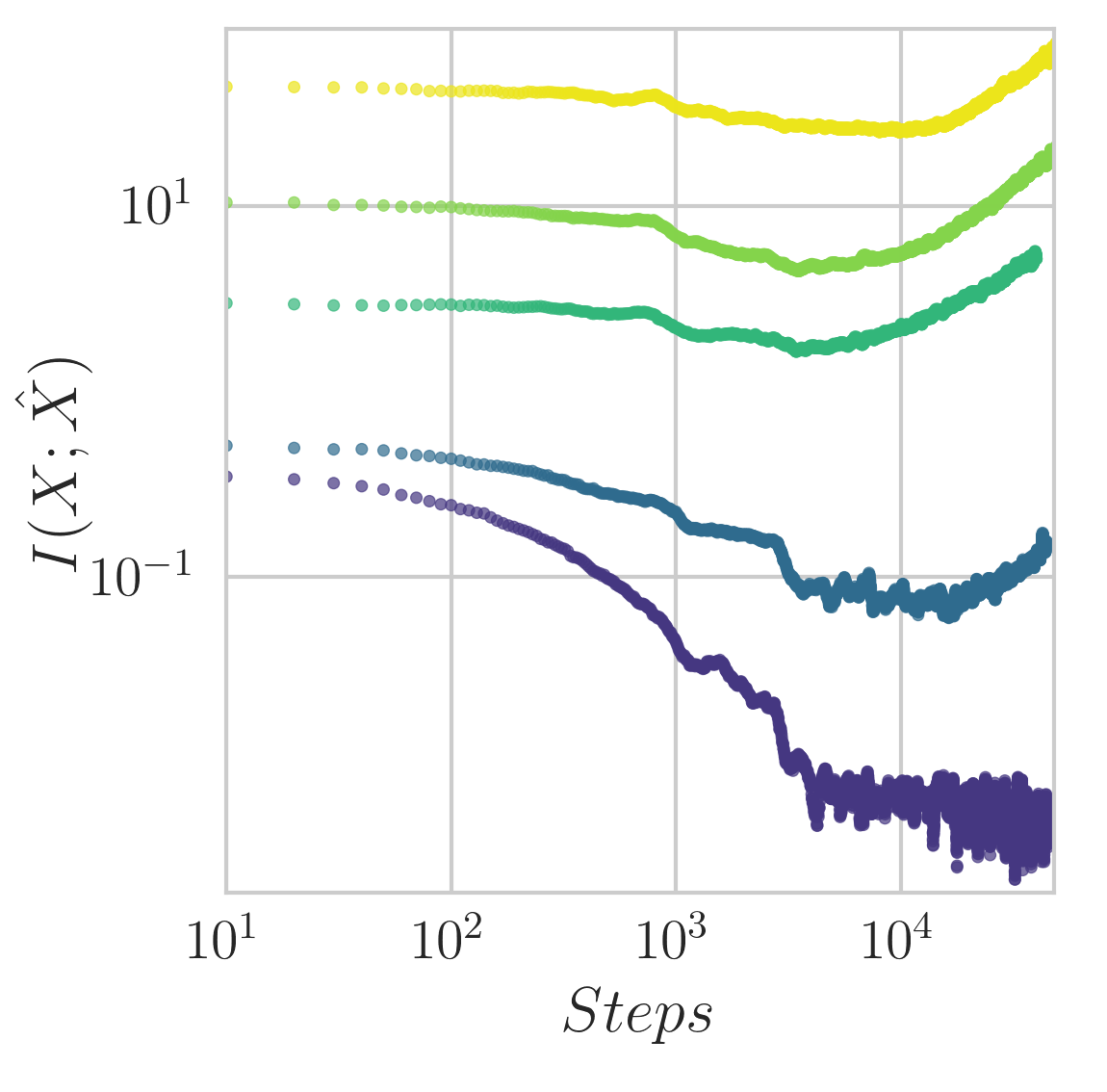}
    \caption{$I(X;\hX)$ vs. update steps}
    \label{fig:inf_plane_mnist_compare}
\end{subfigure}
\begin{subfigure}{.3\textwidth}
    \centering
    \includegraphics[width=\textwidth]{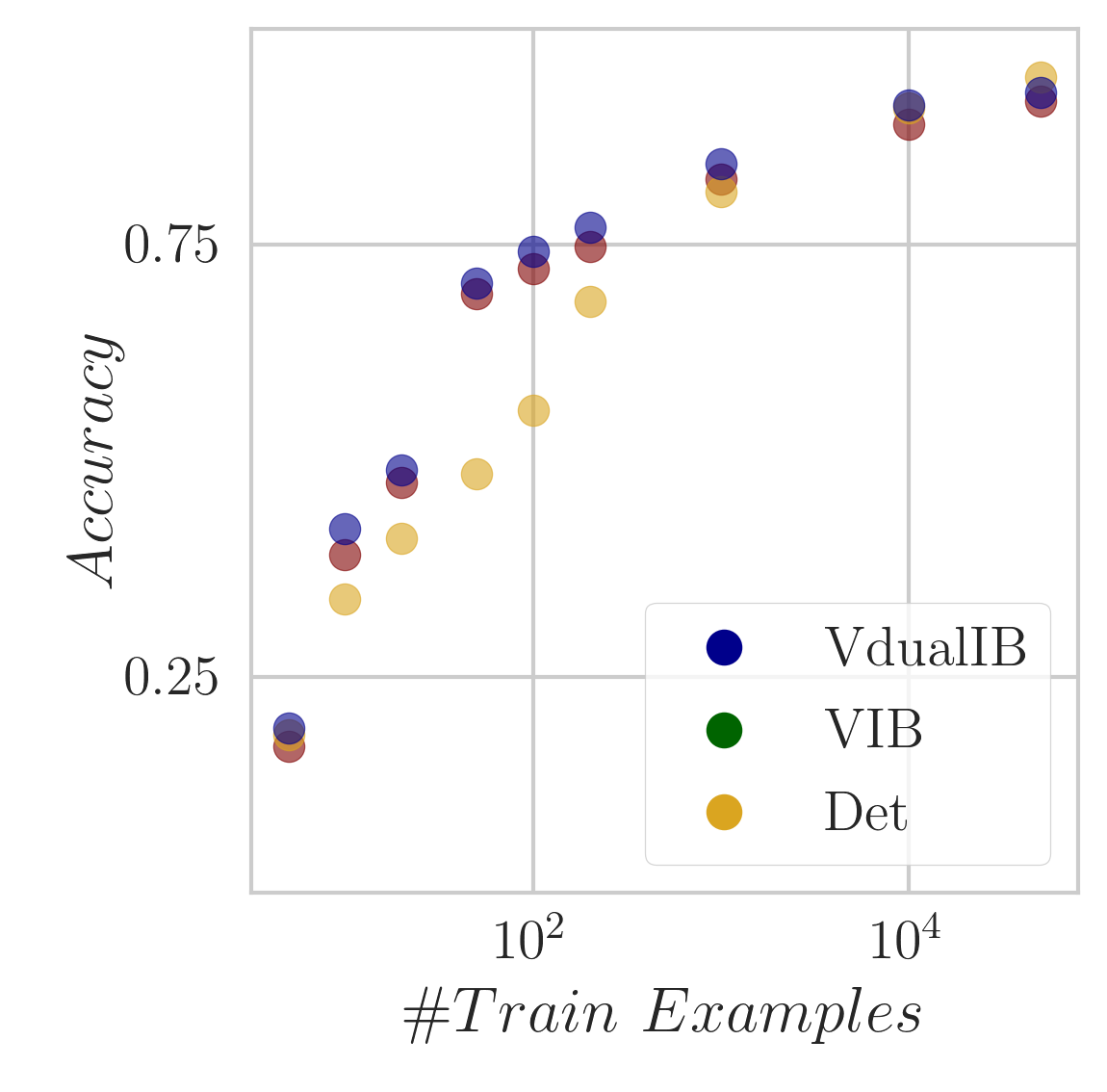}
    \caption{Accuracy vs. training size}
    \label{fig:train_examples_fasion}
\end{subfigure}
 \caption{Experiments over FashionMNIST. $(a)$ The information plane of the {\cvdib} and {\vib} for a range of $\beta$ values at the final training step. $(b)$ The evolution of the the {\cvdib}'s $I(X;\hX)$ along the optimization update steps. $(c)$ The models accuracy as a function of the training set size.}
 \label{fig:inf_plane_mnist}
\end{figure}

In \fref{fig:inf_plane_mnist} we present the information plane of the {\vdib} where the distribution $\tilde{p}\brk*{y \mid x}$ (the noise model) is a learnt confusion matrix, {\cvdib} (similarly to \citep{wu2020phase}). We compare it to the {\vib} over a range of $\beta$ values ($-5\leq\log{\beta} \leq 5$).
\fref{fig:inf_plane_mnist_beta} validates that, as expected, the information growth is approximately monotonic with $\beta$. Comparing the {\vdib} to the {\vib} model, we can see significant differences between their representations. The {\vdib} successfully obtains better compressed representations in comparison to the {\vib} performance, where only for large values of $I(X;\hX)$ their performances 
match.
As predicted by the theory, in the limit $\beta \rightarrow \infty$ the models behaviour match.
Furthermore, the {\vdib} values are smoother and they are spread over the information plane, making it easier to optimize for a specific value in it. In \fref{fig:inf_plane_mnist_compare} we consider the dynamics of $I(X;\hX)$  for several values of $\beta$. Interestingly, at the initial training stage the representation information for all values of $\beta$ decreases. However, as the training continues, the information increases only for high $\beta$s.

\subsubsection{The {\vdib} noise model}
\label{sec:cifar10_inf}
As mentioned above, learning with the {\vdib} objective requires 
 a choice of a noise model for the distribution $\tilde{p}\brk*{y\mid \hx}$. To explore the influence of different models on the learning we evaluate four types, with different assumptions on the access to the data. (i) Adding Gaussian noise to the one-hot vector of the true label ({\gvdib}); (ii) An analytic Gaussian integration of the log-loss around the one-hot labels; (iii) 
A pre-computed confusion matrix for the labels ({\cvdib}) as in  \citep{wu2020phase}; (iv) 
Using predictions of another trained model as the induced distribution. 
Where for (i) and (ii) the variance acts as a free parameter determining the noise level. The complexity of the noise models can be characterized by the additional prior knowledge on our data-set they require. While adding Gaussian noise does not require prior knowledge, using a trained model requires access to the prediction 
for every data sample. 
The use of a confusion matrix is an intermediate level 
of prior knowledge requiring access only to the $\abs{\mathcal{Y}}\times\abs{\mathcal{Y}}$ pre-computed matrix.
Here we present 
cases (i) and (iii) 
(see \S \ref{app:noise_model} for  (ii) and (iv)). Note that although using the {\vib} does not require the use of a noise model we incorporate it by replacing the labels with $\tilde{p}\brk*{y \mid x}$.
In the analysis below, the results are presented with the {\vib} trained with the same noise model as the {\vdib} (see \S \ref{app:noise_model} for  
a comparison between training {\vdib} with noise and {\vib} without it).

\begin{figure}
\begin{subfigure}{.3\textwidth}
    \centering
    \includegraphics[width=\textwidth]{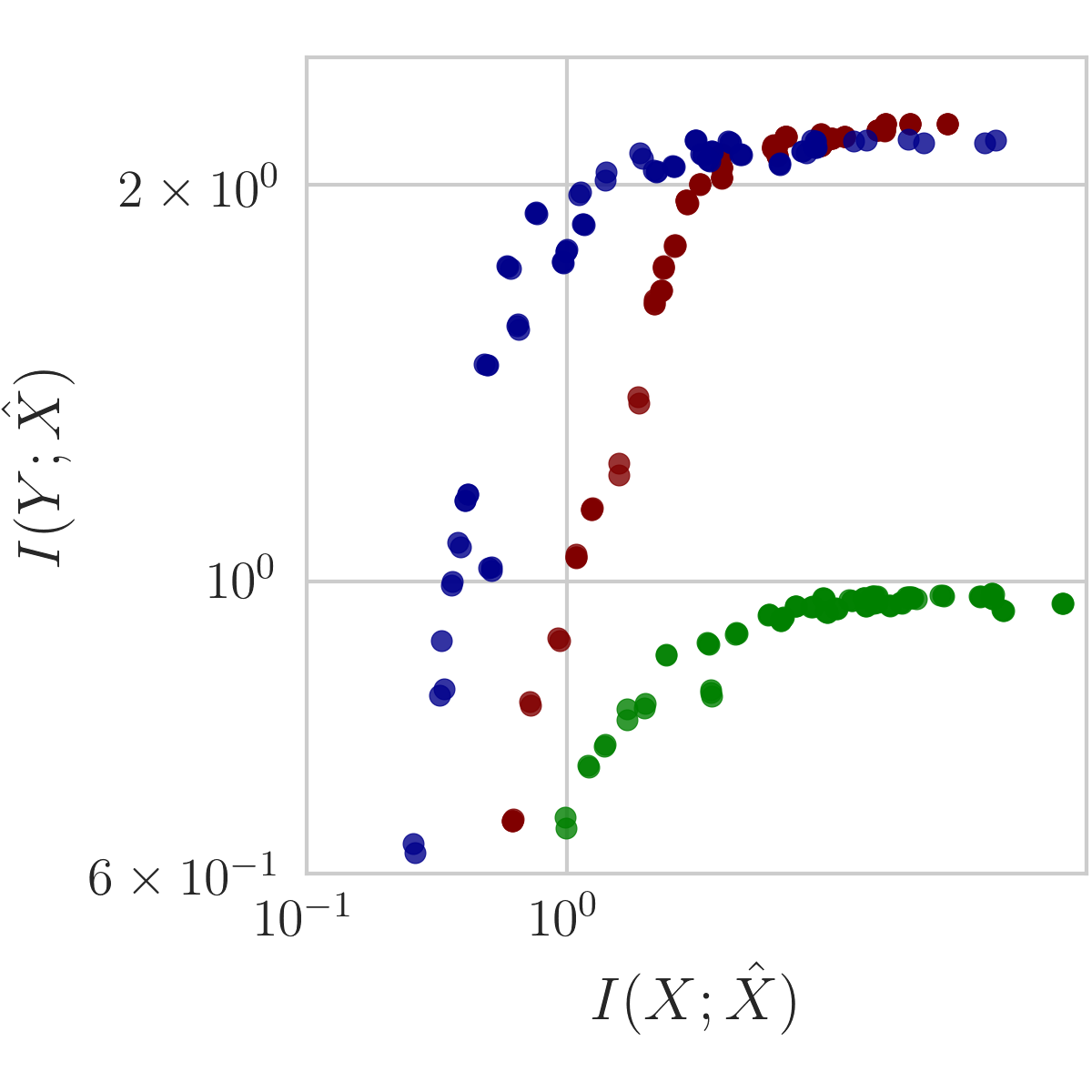}
    \caption{The information plane}
    \label{fig:inf_plane_all}
\end{subfigure}
\begin{subfigure}{.3\textwidth}
    \centering
    \includegraphics[width=\textwidth]{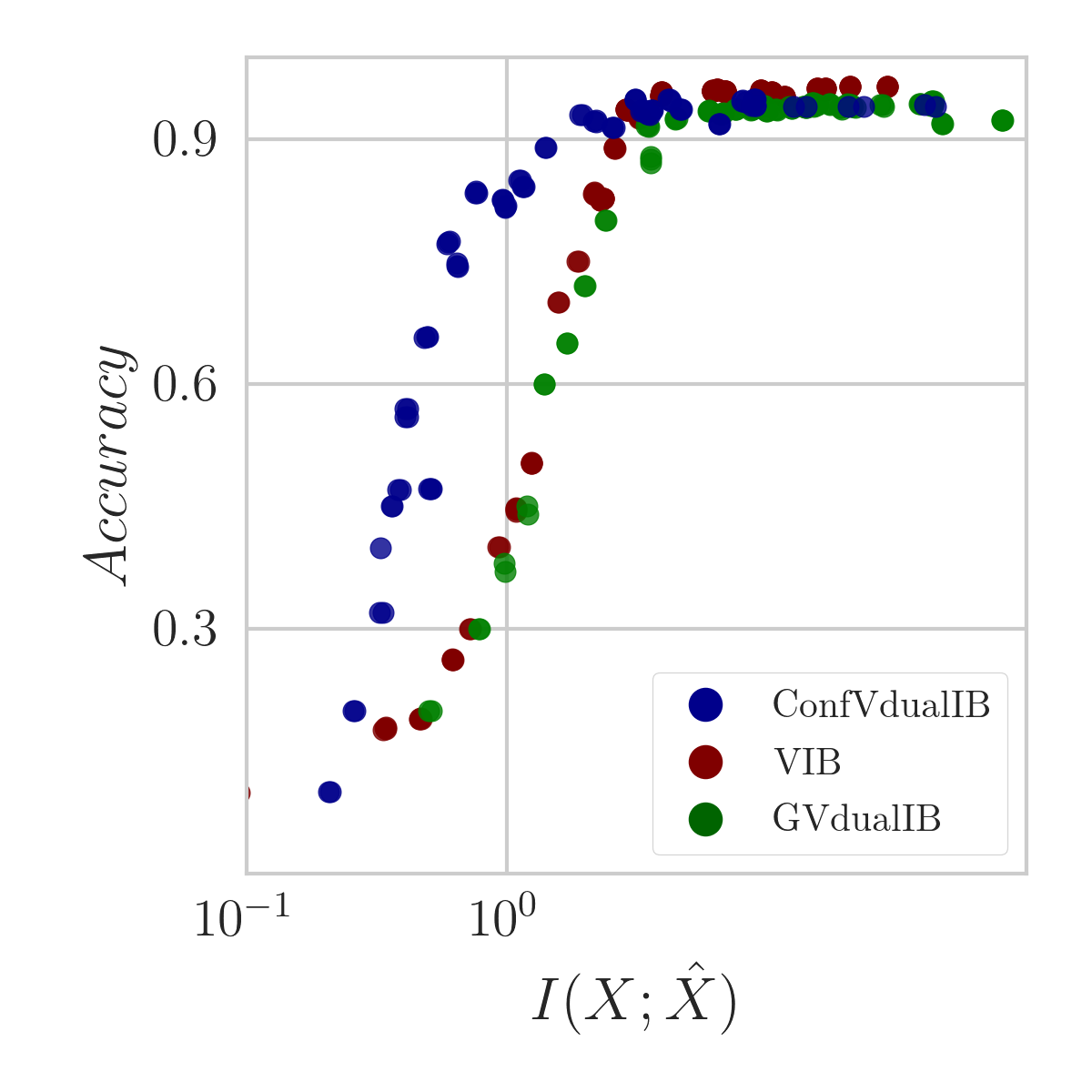}
    \caption{Accuracy vs. $I(X;\hX)$}
    \label{fig:inf_plane_acc_1}
\end{subfigure}
\begin{subfigure}{.3\textwidth}
    \centering
    \includegraphics[width=\textwidth]{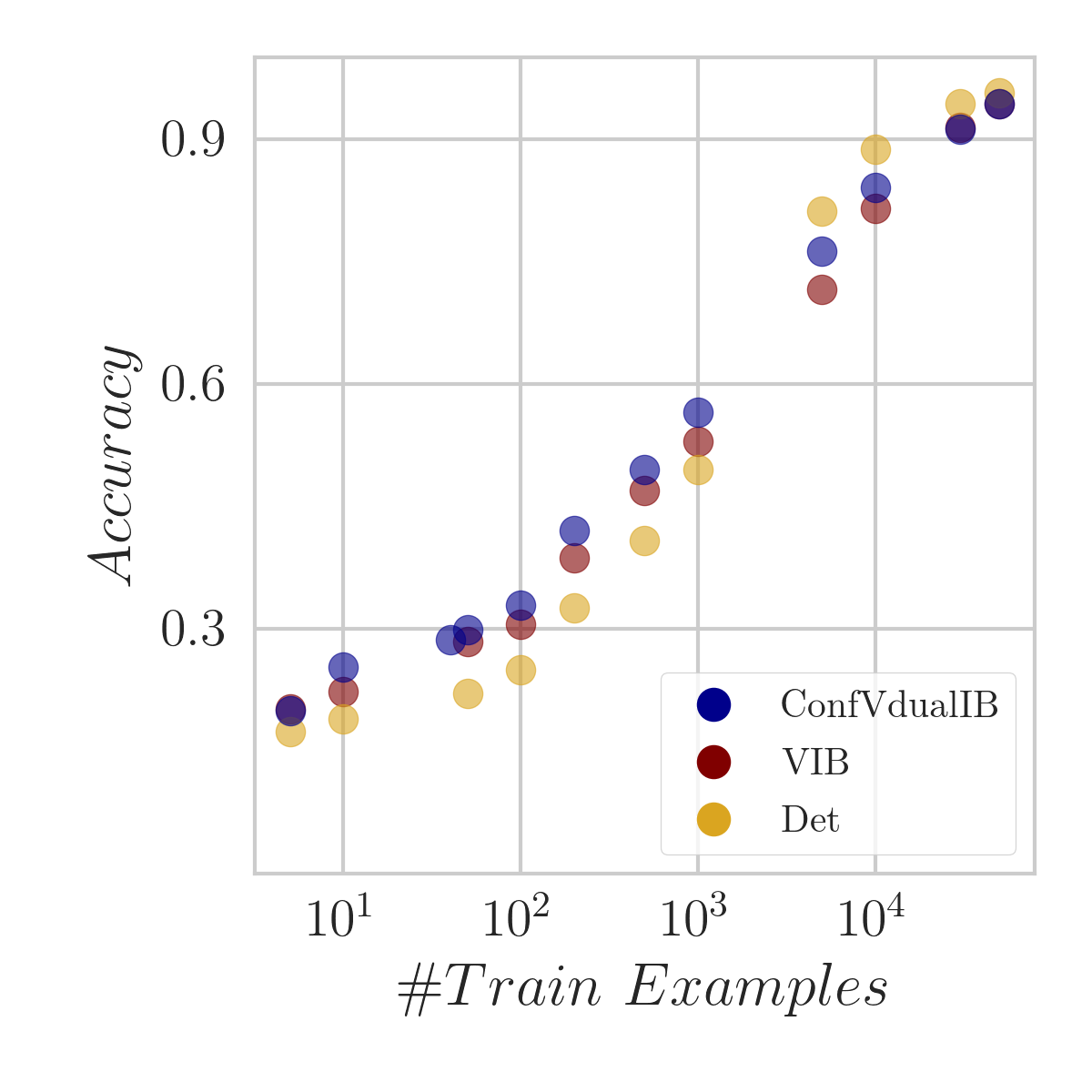}
    \caption{Accuracy vs. training size}
    \label{fig:train_examples_cifar}
\end{subfigure}
\caption{Experiments over CIFAR10. $(a)$ The information plane of the {\vib}, {\cvdib}, {\gvdib}  and {\vib} for a range of $\beta$ values. $(b)$ The accuracy of the models as a function of the mutual information, $I(X;\hX)$. $(c)$ The accuracy of the models as a function of the training set size.}
\label{fig:inf_plane}
\end{figure}

\fref{fig:inf_plane} depicts 
the information plane of the CIFAR10 data-set. \fref{fig:inf_plane_all} shows the information obtained from a range of $\beta$. The colors depict the different models {\cvdib}, {\gvdib} and the {\vib}. As we can see, training a {\vdib} with Gaussian noise achieves much less information with the labels at any given $I(X;\hX)$. We note that we verified that this behaviour is consistent over a wide range of variances.
The {\cvdib} model performance is similar to the {\vib}'s with the former showcasing more compressed representations. When we present the prediction accuracy (\fref{fig:inf_plane_acc_1}), here all  models attain roughly the same accuracy values. The discrepancy between the accuracy and information, $I(Y;\hX)$,
is similar to the one discussed in \citep{dusenberry2020efficient}. 

\subsubsection{Performance with different training set sizes}
\label{sec:experiment_setsize}
Our theoretical analysis (\S \ref{sec:min_err_exp}) shows that under given assumptions the {\dualib} bounds the optimal achievable error exponent on expectation hence it optimizes the error for a given data size $n$. We turn to test this in the {\vdib} setting. 
We train the models on a subset of the training set and evaluate them on the test set. We compare the {\vib} and the {\vdib} to a deterministic network (Det; Wide Res Net 28-10). Both the {\vib} and {\vdib} are trained over a wide range of $\beta$ values ($-5 \leq \log{\beta}\leq 6$). Presented is the best accuracy value for each model at a given $n$. 
\fref{fig:train_examples_fasion} and \fref{fig:train_examples_cifar} 
 show the accuracy of the models as a function of the training set size over FashionMNIST and CIFAR10 respectively. The {\vdib} performance is slightly better in comparison to the {\vib}, while the accuracy of the deterministic network is lower for small training sets. The superiority of the variational models over the deterministic network is not surprising as minimizing $I(X;\hX)$ acts as regularization.

\section*{Conclusions}
\addcontentsline{toc}{section}{Conclusions}

\label{sec:conclusions}
We present here the Dual Information Bottleneck ({\dualib}), a framework resolving some of the known drawbacks of the {\ib} obtained by a mere switch between the terms in the distortion function. 
We provide the {\dualib} self-consistent equations allowing 
us to obtain analytical solutions. 
A local stability analysis revealed the underlying structure of the critical points of the solutions, resulting in a full bifurcation diagram of the optimal pattern representations.
The study of the {\dualib} objective reveals several interesting properties.  First, when the data can be modeled in a parametric form the {\dualib} preserves this structure and it obtains the representation in terms of the original parameters, as given by the {\expib} equations. Second, it optimizes the mean prediction error exponent thus improving the accuracy of the predictions as a function of the data size.
In addition to the {\dualib} analytic solutions, we provide a variational {\dualib} ({\vdib}) framework, which optimizes the functional using DNNs. This framework enables practical implementation of the {\dualib} to real world data-sets. While a broader analysis is required, the {\vdib} experiments shown validate the theoretical predictions. Our results demonstrate the potential advantages and unique properties of the framework.

\bibliographystyle{dcu}
\bibliography{main}
 \clearpage

\section*{Appendix}
\addcontentsline{toc}{section}{Appendix}

\section*{Appendix A - The Information Bottleneck method}
\label{app:ib}
The Information Bottleneck ({\ib}) trade off between the encoder and decoder mutual information values is defined by the minimization of the Lagrangian:
\begin{align}
\label{eq:IB_L1}
    \mathcal{F}\brk[s]*{p_{\beta}\brk*{\hat{x} \mid x}; p_{\beta}\brk*{y \mid \hat{x} }}  =I(X;\hat{X}) - \beta I(Y; \hat{X})~,
\end{align}

independently over the convex sets of the normalized distributions, $\brk[c]*{p_{\beta}\brk*{\hat{x} \mid x}}$, $\brk[c]*{p_{\beta}\brk*{\hat{x}}}$ and $\brk[c]*{p_{\beta}\brk*{y \mid \hat{x}}}$, given a positive Lagrange multiplier $\beta$. As shown in \cite{tishby99information, DBLP:conf/alt/ShamirST08}, this is a natural generalization of the classical concept of \emph{Minimal Sufficient Statistics} \cite{Cover:2006:EIT:1146355}, where the estimated parameter is replaced by the output variable $Y$ and \emph{exact} statistical sufficiency is characterized by the mutual information equality: $I(\hX;Y)=I(X;Y)$. The minimality of the statistics is captured by the minimization of $I(X;\hX) $, due to the Data Processing Inequality (DPI). However, non-trivial minimal sufficient statistics only exist for very special parametric distributions known as exponential families \cite{Exp_forms}. Thus in general, the {\ib} relaxes the minimal sufficiency problem to a continuous family of representations $\hX$ which are characterized by the trade off between compression, $I(X;\hX)\equiv I_{X}$, and accuracy, $I(Y;\hX)\equiv I_{Y}$, along a convex line in the \emph{Information-Plane} ($I_{Y}$ vs. $I_{X}$). When the rule $p(x,y)$ is strictly stochastic, the convex optimal line is smooth and each point along the line is uniquely characterized by the value of $\beta$. We can then consider the optimal representations $\hx=\hx(\beta)$ as encoder-decoder pairs: $(p_{\beta}(x\mid \hx),p_{\beta}(y\mid \hx))$\footnote{Here we use the \emph{inverse encoder}, which is in the fixed dimension simplex of distributions over $X$.} - a point in the continuous manifold defined by the Cartesian product of these distribution simplexes. We also consider a small variation of these representations, $\delta\hx$, as an infinitesimal change in this (encoder-decoder) continuous manifold (not necessarily on the optimal line(s)).         

\subsection{IB and Rate-Distortion Theory}

The {\ib} optimization trade off can be considered as a generalized rate-distortion problem \cite{Cover:2006:EIT:1146355} with the distortion function between a data point, $x$ and a representation point $\hx$ taken as the KL-divergence between their predictions of the desired label $y$:
\begin{align}
d_{\ib}\brk*{x,\hx}&=D\brk[s]*{p\brk*{y \mid x}||p_{\beta}\brk*{y|\hx}} \nonumber \\
&=\sum_{y}p\brk*{y \mid x}\log \frac{p\brk*{y \mid x}}{p_{\beta}\brk*{y \mid \hx}}.  
\end{align}
The expected distortion $\mathbb{E}_{p_{\beta}(x,\hx)}\brk[s]*{d_{\ib}\brk*{x,\hx}}$ for the optimal decoder is simply the label-information loss: $I(X;Y)-I(\hX;Y)$, using the Markov chain condition. Thus minimizing the expected $\ib$ distortion is equivalent to maximizing $I(\hX;Y)$, or minimizing \eqref{eq:IB_L}.  
Minimizing this distortion is equivalent to minimizing the cross-entropy loss, and it provides an upper-bound to other loss functions such as the $\mathcal{L}_{1}$-loss (due to the Pinsker inequality, see also \cite{Painsky2019}). Pinsker implies that both orders of the cross-entropy act as an upper bound to the $\mathcal{L}_{1}$-loss,
 $\min\{D\brk[s]*{q||p},D\brk[s]*{p||q}\} \geq \frac{1}{2 \log 2} \|p - q \|_{1}^{2}~.$  
 
\subsection{The IB Equations}

For discrete $X$ and $Y$, a necessary condition for the $\ib$ (local) minimization is given by the three self-consistent equations for the optimal encoder-decoder pairs, known as the \emph{{\ib} equations}:
\begin{align} \label{eq:IB4}
	\begin{cases}
	\brk*{i}\ &p_{\beta}\brk*{\hat{x} \mid x} = \frac{p_{\beta}\brk*{\hat{x}}}{Z\brk*{x;\beta}} e^{-\beta D\brk[s]*{p\brk*{y\mid x} \| p_{\beta}\brk*{y \mid \hat{x}}}} \\
	\brk*{ii}\ &p_{\beta}\brk*{\hat{x}} = \sum_{x} p_{\beta}\brk*{\hat{x} \mid x} p\brk*{x} \\
	\brk*{iii}\ &p_{\beta}\brk*{y \mid \hat{x} } = \sum_{x} p\brk*{y\mid x} p_{\beta}\brk*{x \mid \hat{x} } 
	\end{cases}
,\end{align}
where $Z\brk*{x;\beta}$ is the normalization function. Iterating these equations is a generalized, Blahut-Arimoto, alternating projection algorithm \cite{CIS-58533,Cover:2006:EIT:1146355} and it converges to a stationary point of the Lagrangian, \eqref{eq:IB_L} \cite{tishby99information}. Notice that the minimizing decoder, (\eqref{eq:IB4}-$(iii)$), is precisely the \emph{Bayes optimal decoder} for the representation $\hx(\beta)$, given the Markov chain conditions.

\subsection{Critical points and critical slowing down}
\label{sec:bifurcation_points}

One of the most interesting aspects of the {\ib} equations is the existence of critical points along the optimal line of solutions in the information plane (i.e. the information curve). At these points the representations change topology and cardinality (number of clusters) \cite{ZaslavskyTishby:2019, parker} and they form the skeleton of the information curve and representation space.  
To identify such points we perform a perturbation analysis of the {\ib} equations:\footnote{We ignore here the possible interaction between the different representations, for simplicity.}:
 \begin{align}
    \delta \log p_{\beta}\brk*{x \mid \hx}  =&  \beta\sum_{y} p\brk*{y\mid x} \delta \log p_{\beta}\brk*{y \mid \hx}
    \label{eq:stab_anl_x_ib} 
    ,\end{align}
    \begin{align}
    \delta \log p_{\beta}\brk*{y \mid \hx} = \frac{1}{p_{\beta}\brk*{y\mid \hx}} \sum_{x} p\brk*{y\mid x}p_{\beta}\brk*{x\mid \hx} \delta \log p_{\beta}\brk*{x \mid \hx}  \label{eq:stab_anl_y_ib}
.\end{align}
Substituting \eqref{eq:stab_anl_y_ib} into \eqref{eq:stab_anl_x_ib} and vice versa one obtains:
\begin{align*}
    \delta \log p_{\beta}\brk*{x \mid \hx}  &= \beta\sum_{y,x'}  p\brk*{y\mid x'} \frac{p\brk*{y\mid x} }{p_{\beta}\brk*{y\mid \hx}} p_{\beta}\brk*{x'\mid \hx} \delta \log p_{\beta}\brk*{x' \mid \hx} \\
    \delta \log p_{\beta}\brk*{y \mid \hx} &= \beta \sum_{x,y'} p\brk*{y\mid x} \frac{p_{\beta}\brk*{x\mid \hx}}{p_{\beta}\brk*{y\mid \hx}}   p\brk*{y'\mid x} \delta \log p_{\beta}\brk*{y' \mid \hx}
\end{align*}
Thus by defining the matrices: 
\begin{align} \label{eq:c_mat}
	C^{\ib}_{xx'}(\hx,\beta) &= \sum_{y}p\brk*{y \mid x} \frac{p_{\beta}\brk*{x' \mid \hat{x}} }{p_{\beta}\brk*{y \mid \hat{x} }} p\brk*{y \mid x' } ~,~ 
	C^{\ib}_{yy'}(\hx,\beta) = \sum_{x}p\brk*{y \mid x} \frac{p_{\beta}\brk*{x \mid \hat{x} }}{p_{\beta}\brk*{y \mid \hat{x} }} p\brk*{ y' \mid x }
.\end{align}
We obtain the following nonlinear eigenvalues problem:
\begin{align}
    \label{eq:stab_anl}
	\brk[s]*{I - \beta 	C^{\ib}_{xx'} \brk*{\hat{x}, \beta }} \delta \log p_{\beta}\brk*{x' \mid \hat{x}}&=0 ~,~~~~ 
	\brk[s]*{I - \beta C^{\ib}_{y y'}\brk*{\hat{x}, \beta }} \delta \log p_{\beta}\brk*{y' \mid \hat{x}}=0
,\end{align}

These two matrices have the same eigenvalues and have non-trivial eigenvectors  (i.e., different co-existing optimal representations) at the critical values of $\beta$, the bifurcation points of the {\ib} solution. At these points the cardinality of the representation $\hX$ (the number of ``{\ib}-clusters") changes due to splits of clusters, resulting in topological phase transitions in the encoder. These critical points form the ``skeleton" of the topology of the optimal representations. Between critical points the optimal representations change continuously (with $\beta$). The important computational consequence of critical points is known as \emph{critical slowing down} \cite{CriticalSlowingDown:2004}. 
For binary $y$, near a critical point the convergence time, $\tau_{\beta}$, of the iterations of \eqref{eq:IB4} scales like:
$ \tau_{\beta} \sim {1}/{(1-\beta \lambda_2)}$,
where $\lambda_2$ is the second eigenvalue of either $C^{\ib}_{yy'}$ or $C^{\ib}_{xx'}$. At criticality, $\lambda_2(\hx)=\beta^{-1}$ and the number of iterations diverges. This phenomenon dominates any local minimization of \eqref{eq:IB4} which is based on alternate encoder-decoder optimization. 

The appearance of the critical points and the critical slowing-down is visualized in Figure $1$ in the main text.
\section*{Appendix B - The dualIB mathematical formulation}
\label{app:dual_ib}
The {\dualib} is solved with respect to the full Markov chain ($Y\rightarrow X \rightarrow \hX_{\beta} \rightarrow \hY$) in which we introduce the new variable, $\hy$, the \emph{predicted label}. Thus, in analogy to the $\ib$ we want to write the optimization problem in term of $\hY$.

Developing the expected distortion we find:
\begin{align*}
    \mathbb{E}_{p_{\beta}\brk*{x, \hat{x}}}\brk[s]*{d_{\dualib}\brk*{x, \hat{x}}} &= \sum_{x, \hat{x}}p_{\beta}\brk*{x, \hat{x}} \sum_{\hy} p_{\beta}\brk*{y=\hy \mid \hat{x}} \log \frac{p_{\beta}\brk*{y=\hy \mid \hat{x}} }{p\brk*{y=\hy \mid {x}}} \nonumber \\
     &=  \sum_{\hx, \hy} p_{\beta}\brk*{\hx} p_{\beta}\brk*{\hy \mid {\hx}} \log \frac{{p_{\beta}\brk*{\hy \mid \hx} }}{p_{\beta}\brk*{\hy}} - \sum_{x, \hy} p\brk*{x} p_{\beta}\brk*{\hy \mid {x}} \log \frac{{p_{\beta}\brk*{\hy \mid {x}} }}{p_{\beta}\brk*{\hy}} \nonumber \\
     &+ \sum_{x, \hy} p\brk*{x} p_{\beta}\brk*{\hy \mid {x}} \log \frac{p_{\beta}\brk*{\hy \mid {x}} }{p\brk*{y=\hy \mid {x}}} \nonumber \\
     &= I(\hX ; \hY) - I(X; \hY) + \mathbb{E}_{p\brk*{x }}\brk[s]*{D\brk[s]*{p_{\beta}\brk*{\hy \mid x} \| p\brk*{y=\hy \mid x} }}
.\end{align*}

Allowing the dual optimization problem to be written as:
  \begin{align*} 
      \mathcal{F}^{*}\brk[s]*{p\brk*{\hx \mid x} ;p\brk*{y \mid \hx }} &=  I(X;\hX) -\beta \brk[c]*{ I(X; \hY) - I(\hX ; \hY)- \mathbb{E}_{p\brk*{x }}\brk[s]*{D\brk[s]*{p_{\beta}\brk*{\hy \mid x} \| p\brk*{y=\hy \mid x} }}}
  .\end{align*}
\section*{Appendix C - The DualIB solutions}
\label{app:dual_ib_eq_proof}
To prove \emph{theorem} $2$ we want to obtain the normalized distributions minimizing the {\dualib} rate-distortion problem.
\begin{proof} 
$(i)$ Given that the problem is formulated as a rate-distortion problem the encoder's update rule must be the known minimizer of the distortion function. \cite{Cover:2006:EIT:1146355}. 
Thus the $\ib$ encoder with the dual distortion is plugged in.
$(ii)$ For the decoder, by considering a small perturbation in the distortion $d_{\dualib}\brk*{x, \hat{x}}$, with $\alpha\brk*{\hat{x}}$ the normalization Lagrange multiplier, we obtain:
\begin{align*}
    \delta d_{\dualib}\brk*{x, \hat{x}} &= \delta \brk*{\sum_{y} p_{\beta}\brk*{y \mid \hat{x}} \log \frac{p_{\beta}\brk*{y \mid \hat{x}} }{p\brk*{y \mid {x}}} + \alpha\brk*{\hat{x}} \brk*{\sum_{y}p_{\beta}\brk*{y \mid \hat{x}} - 1 } }\nonumber \\
    \frac{\delta d_{\dualib}\brk*{x, \hat{x}} }{\delta p_{\beta}\brk*{y \mid \hat{x}}} &= \log \frac{p_{\beta}\brk*{y \mid \hat{x}} }{p\brk*{y \mid {x}}} + 1  + \alpha\brk*{\hat{x}} 
.\end{align*}
Hence, minimizing the expected distortion becomes:
\begin{align*}
    0 &= \sum_{x}  p_{\beta}\brk*{x \mid \hat{x}}\brk[s]*{\log \frac{p_{\beta}\brk*{y \mid \hat{x}} }{p\brk*{y \mid {x}}} + 1}  + \alpha\brk*{\hat{x}} \nonumber \\
    &= \log p_{\beta}\brk*{y \mid \hat{x}} - \sum_{x}  p_{\beta}\brk*{x \mid \hat{x}} \log p\brk*{y \mid x} + 1 + \alpha\brk*{\hat{x}} 
,\end{align*}
which yields Algorithm $1$, row $6$.
\end{proof}

Considering the $\dualib$ encoder-decoder, Algorithm $1$, we find that $\mathbb{E}_{p_{\beta}\brk*{x, \hat{x}}}\brk[s]*{d_{\dualib}\brk*{x, \hat{x}}}$ reduces to the expectation of the decoder's $\log$ partition function:
   \begin{align*} 
      \mathbb{E}_{p_{\beta}\brk*{x, \hat{x}}}\brk[s]*{d_{\dualib}\brk*{x, \hat{x}}} &= \sum_{x, \hat{x}}p_{\beta}\brk*{x, \hat{x}} \sum_{y} p_{\beta}\brk*{{y} \mid \hat{x}} \log \frac{p_{\beta}\brk*{{y} \mid \hat{x}} }{p\brk*{{y} \mid {x}}} \nonumber \\
      &= - \mathbb{E}_{p_{\beta}\brk*{\hx}}\brk[s]*{\log Z_{\rvy\mid \hat{\rvx}}\brk*{\hat{x}; \beta}} + \sum_{\hat{x}, y} p_{\beta}\brk*{\hx } \brk[s]*{ \sum_{x'} p_{\beta}\brk*{x'\mid \hat{x}} \log p\brk*{y \mid x'}  - \sum_{x} p_{\beta}\brk*{x\mid \hat{x}} \log p\brk*{y \mid x}} \nonumber \\
      &= - \mathbb{E}_{p_{\beta}\brk*{\hx}}\brk[s]*{\log Z_{\rvy\mid \hat{\rvx}}\brk*{\hat{x}; \beta}}
  .\end{align*}
\section*{Appendix D - Stability analysis}
\label{app:stability_anl}
Here we provide the detailed stability analysis allowing the definition of the matrices $C^{\dualib}_{xx'}, C^{\dualib}_{yy'}$ (\emph{theorem $4$}) which allows us to claim that they obey the same rules as the $C$ matrices of the {\ib}. Similarly to the  {\ib} in this calculation we ignore second order contributions which arise form the normalization terms.
Considering a variation in $\hat{x}$ we get:
 \begin{align}
    \delta \log p_{\beta}\brk*{x \mid \hx} =& \beta\sum_{y}p_{\beta}\brk*{y \mid \hat{x}}   \brk*{\log \frac{ p\brk*{y \mid x} }{p_{\beta} \brk*{y \mid \hat{x}}} - 1}\delta \log p_{\beta}\brk*{y \mid \hx}\nonumber \\
    =&  \beta\sum_{y}p_{\beta}\brk*{y \mid \hat{x}} \brk[s]*{\log  p\brk*{y \mid x} - \sum_{\tilde{x}} p_{\beta} \brk*{\tilde{x} \mid \hx}\log  p\brk*{y \mid \tilde{x}} }
    \delta \log p_{\beta}\brk*{y \mid \hx} \nonumber \\
    +& \beta\sum_{y} \log  Z_{\rvy\mid \hat{\rvx}}\brk*{\hat{x}; \beta} \frac{\partial  p_{\beta}\brk*{y \mid \hat{x}}}{\partial \hat{x}} \nonumber \\
    =&  \beta\sum_{y,\tilde{x}}p_{\beta}\brk*{y \mid \hat{x}} p_{\beta} \brk*{\tilde{x} \mid \hx} \log \frac{ p\brk*{y \mid x} }{p \brk*{y \mid \tilde{x}}}\delta \log p_{\beta}\brk*{y \mid \hx}
    \label{eq:stab_anl_x} 
    ,\end{align}
    \begin{align}
    \delta \log p_{\beta}\brk*{y \mid \hx} =&- \frac{1}{ Z_{\rvy\mid \hat{\rvx}}\brk*{\hat{x}; \beta}}\frac{\partial Z_{\rvy\mid \hat{\rvx}}\brk*{\hat{x}; \beta}}{\partial \hat{x}}+  \sum_{x}p_{\beta}\brk*{x \mid \hat{x}}\log p\brk*{y \mid x}\delta \log p_{\beta}\brk*{x \mid \hx} \nonumber \\
    =&  -\sum_{\tilde{y}} p_{\beta}\brk*{\tilde{y} \mid \hat{x}} \sum_{x}p_{\beta}\brk*{x \mid \hat{x}} \log p\brk*{\tilde{y} \mid x} \delta \log p_{\beta}\brk*{x \mid \hx} \nonumber \\
    +&\sum_{x} p_{\beta}\brk*{x \mid \hat{x}} \log p\brk*{y \mid x}\delta \log p_{\beta}\brk*{x \mid \hx} 
    \nonumber \\
    =& \sum_{x, \tilde{y}} p_{\beta}\brk*{x \mid \hat{x}}p_{\beta}\brk*{\tilde{y} \mid \hat{x}}\log \frac{{p\brk*{y \mid x}}}{p\brk*{\tilde{y} \mid x}} \delta \log p_{\beta}\brk*{x \mid \hx} \label{eq:stab_anl_y}
.\end{align}

Substituting \eqref{eq:stab_anl_y} into \eqref{eq:stab_anl_x} and vice versa one obtains:
 \begin{align*} 
    \delta \log p_{\beta}\brk*{x \mid \hx} &= \beta\sum_{x',y,\tilde{y}, \tilde{x}}p_{\beta}\brk*{y \mid \hat{x}} p_{\beta} \brk*{\tilde{x} \mid \hat{x}}  \log \frac{ p\brk*{y \mid x} }{p \brk*{y \mid \tilde{x}}}   \nonumber \\
    &\cdot p_{\beta}\brk*{x' \mid \hat{x}}p_{\beta}\brk*{\tilde{y} \mid \hat{x}}  \log \frac{p\brk*{y \mid x'}}{p\brk*{\tilde{y} \mid x'} } {\delta \log p_{\beta}\brk*{x' \mid \hat{x}}} \nonumber \\
    \delta \log p_{\beta}\brk*{y \mid \hx} &= \beta \sum_{x ,y', \tilde{x}, \tilde{y}} p_{\beta}\brk*{x \mid \hat{x}}p_{\beta}\brk*{\tilde{y} \mid \hat{x}}  \log \frac{p\brk*{y \mid x}}{p\brk*{\tilde{y} \mid x} }  \nonumber \\
    &\cdot p_{\beta}\brk*{y' \mid \hat{x}} p_{\beta}\brk*{\tilde{x} \mid \hat{x}} \log \frac{ p\brk*{y' \mid x} }{p \brk*{y' \mid \tilde{x}}} {\delta \log p_{\beta}\brk*{y' \mid \hat{x}}}
.\end{align*}

We now define the $C^{\dualib}$ matrices as follows:
\begin{align*} 
	C^{\dualib}_{xx'}\brk*{\hat{x}; \beta} = &\sum_{y, \tilde{y}, \tilde{x}} p_{\beta}\brk*{y \mid \hat{x}}  p_{\beta}\brk*{\tilde{x} \mid \hat{x}}   \log\frac{ p\brk*{y \mid x} }{p \brk*{y \mid \tilde{x}}}  \cdot  p_{\beta}\brk*{x' \mid \hat{x}}p_{\beta}\brk*{\tilde{y} \mid \hat{x}} \log \frac{p\brk*{y \mid x'}}{p\brk*{\tilde{y} \mid x'} }
	 \nonumber \\
	C^{\dualib}_{yy'}\brk*{\hat{x}; \beta} = &\sum_{x, \tilde{x}, \tilde{y}} p_{\beta}\brk*{x \mid \hat{x}}  p_{\beta}\brk*{\tilde{y} \mid \hat{x}} \log \frac{p\brk*{y \mid x}}{p\brk*{\tilde{y} \mid x}}  
	\cdot p_{\beta}\brk*{y' \mid \hat{x}} p_{\beta}\brk*{\tilde{x} \mid \hat{x}} \log \frac{ p\brk*{y' \mid x} }{p \brk*{y' \mid \tilde{x}}} 
.\end{align*}
Using the above definition we have an equivalence to the {\ib} stability analysis in the form of: 
\begin{align*} 
	\brk[s]*{I - \beta 	C^{\dualib}_{xx'} \brk*{\hat{x}, \beta }} {\delta \log p_{\beta}\brk*{x' \mid \hat{x}}}&=0 ~,~~~~ 
	\brk[s]*{I - \beta C^{\dualib}_{y y'}\brk*{\hat{x}, \beta }} {\delta \log p_{\beta}\brk*{y' \mid \hat{x}}}=0
.\end{align*}
Note that for the binary case, the matrices may be simplified to:
\begin{align*} 
	C^{\dualib}_{xx'}\brk*{\hat{x}; \beta} = &\sum_{y, \tilde{x}} p_{\beta}\brk*{y \mid \hat{x}}  p_{\beta}\brk*{\tilde{x} \mid \hat{x}}   \log\frac{ p\brk*{y \mid x} }{p \brk*{y \mid \tilde{x}}}  \cdot  p_{\beta}\brk*{x' \mid \hat{x}}\brk*{1-p_{\beta}\brk*{y \mid \hat{x}}} \log \frac{p\brk*{y \mid x'}}{1-p\brk*{y \mid x'} }
	 \nonumber \\
	C^{\dualib}_{yy'}\brk*{\hat{x}; \beta} = &\sum_{x, \tilde{x}} p_{\beta}\brk*{x \mid \hat{x}}  \brk*{1-p_{\beta}\brk*{y \mid \hat{x}}} \log \frac{p\brk*{y \mid x}}{1-p\brk*{y \mid x}}  
	\cdot p_{\beta}\brk*{y' \mid \hat{x}} p_{\beta}\brk*{\tilde{x} \mid \hat{x}} \log \frac{ p\brk*{y' \mid x} }{p \brk*{y' \mid \tilde{x}}} 
.\end{align*}

We turn to show that the $C^{\dualib}$ matrices share the same eigenvalues with $\lambda_{1}\brk*{\hx} = 0$. 
\begin{proof}
The matrices, $C^{\dualib}_{xx'}\brk*{\hat{x}; \beta}$, $C^{\dualib}_{yy'}\brk*{\hat{x}; \beta}$, are given by: 
\begin{equation*}
    C^{\dualib}_{xx'}\brk*{\hat{x}; \beta} = A_{xy}\brk*{\hx;\beta} B_{yx'}\brk*{\hx;\beta} ~,~  
	C^{\dualib}_{yy'}\brk*{\hat{x}; \beta} = B_{yx}\brk*{\hx;\beta} A_{xy'}\brk*{\hx;\beta}
, \end{equation*}
with: \\
\begin{align*} 
	A_{xy}\brk*{\hx;\beta} = p_{\beta}\brk*{y \mid \hat{x}}\sum_{\tilde{x}}p_{\beta}\brk*{\tilde{x} \mid \hat{x}}   \log\frac{ p\brk*{y \mid x} }{p \brk*{y \mid \tilde{x}}} ~, ~
	B_{yx}\brk*{\hx;\beta} = p_{\beta} \brk*{x \mid \hat{x}} \sum_{\tilde{y}}p_{\beta}\brk*{\tilde{y} \mid \hat{x}}\log \frac{{p\brk*{y \mid x}}}{ p\brk*{\tilde{y} \mid x}}
.\end{align*}
Given that the matrices are obtained by the multiplication of the same matrices, it follows that they have the same eigenvalues $\brk[c]*{\lambda_i\brk*{\hx;\beta}}$. 

To prove that $\lambda_1\brk*{\hx;\beta} = 0$ we show that $\det(C^{\dualib}_{y y'})= 0$. We present the exact calculation for a binary label $y \in \brk[c]*{y_{0},y_{1}}$ (the argument for general $y$ follows by encoding the label as a sequence of bits and discussing the first bit only, as a binary case):
\begin{align*}
    	\det(C^{\dualib}_{yy'}\brk*{\hat{x}; \beta}) =& \sum_{x, \tilde{x}} p_{\beta}\brk*{x \mid \hat{x}}  p_{\beta}\brk*{y_{1} \mid \hat{x}} \log \frac{p\brk*{y_{0} \mid x}}{p\brk*{y_{1} \mid x}}  
	\cdot p_{\beta}\brk*{y_{0} \mid \hat{x}} p_{\beta}\brk*{\tilde{x} \mid \hat{x}} \log \frac{ p\brk*{y_{0} \mid x} }{p \brk*{y_{0} \mid \tilde{x}}} \\
	\cdot& \sum_{x', \tilde{x}',} p_{\beta}\brk*{{x'} \mid \hat{x}}  p_{\beta}\brk*{y_{0} \mid \hat{x}} \log \frac{p\brk*{y_{1} \mid {x'}}}{p\brk*{y_{0} \mid {x'}}}  
	\cdot p_{\beta}\brk*{y_{1} \mid \hat{x}} p_{\beta}\brk*{\tilde{x}' \mid \hat{x}} \log \frac{ p\brk*{y_{1} \mid {x'}} }{p \brk*{y_{1} \mid \tilde{x}'}} \\
	-& \sum_{x, \tilde{x}} p_{\beta}\brk*{x \mid \hat{x}}  p_{\beta}\brk*{y_{0} \mid \hat{x}} \log \frac{p\brk*{y_{1} \mid x}}{p\brk*{y_{0} \mid x}}  
	\cdot p_{\beta}\brk*{y_{0} \mid \hat{x}} p_{\beta}\brk*{\tilde{x} \mid \hat{x}} \log \frac{ p\brk*{y_{0} \mid x} }{p \brk*{y_{0} \mid \tilde{x}}} \\
	\cdot& \sum_{{x'}, \tilde{x}'} p_{\beta}\brk*{{x'} \mid \hat{x}}  p_{\beta}\brk*{y_{1} \mid \hat{x}} \log \frac{p\brk*{y_{0} \mid {x'}}}{p\brk*{y_{1} \mid {x}'}}  
	\cdot p_{\beta}\brk*{y_{1} \mid \hat{x}} p_{\beta}\brk*{\tilde{x}' \mid \hat{x}} \log \frac{ p\brk*{y_{1} \mid {x'}} }{p \brk*{y_{1} \mid \tilde{x}'}} \\
	=& \sum_{x, x', \tilde{x}, \tilde{x}'}
	p_{\beta}\brk*{x \mid \hat{x}}  p_{\beta}\brk*{{x'} \mid \hat{x}}  p^2_{\beta}\brk*{y_{0} \mid \hat{x}}  p^2_{\beta}\brk*{y_{1} \mid \hat{x}}  p_{\beta}\brk*{\tilde{x} \mid \hat{x}}\log \frac{ p\brk*{y_{0} \mid x} }{p \brk*{y_{0} \mid \tilde{x}}}  p_{\beta}\brk*{\tilde{x}' \mid \hat{x}} \log \frac{ p\brk*{y_{1} \mid {x'}} }{p \brk*{y_{1} \mid \tilde{x}'}}\\
	\cdot&  \brk[s]*{\log \frac{p\brk*{y_{0} \mid x}}{p\brk*{y_{1} \mid x}} \log \frac{p\brk*{y_{1} \mid {x'}}}{p\brk*{y_{0} \mid {x'}}}   -   \log \frac{ p\brk*{y_{0} \mid x} }{p \brk*{y_{1} \mid x}} \log \frac{p\brk*{y_{1} \mid {x'}}}{p\brk*{y_{0} \mid {x'}}}   } = 0
.\end{align*}
Given that the determinant is $0$ implies that $\lambda_{1}\brk*{\hx} = 0$. 
\end{proof}
For a binary problem we can describe the non-zero eigenvalue using $\lambda_{2}\brk*{\hx} =  \textrm{Tr}(C^{\dualib}_{yy'}\brk*{\hat{x}; \beta})$. That is:
\begin{align*}
    \lambda_{2}\brk*{\hx} =& \sum_{x, \tilde{x}} p_{\beta}\brk*{x \mid \hat{x}}  p_{\beta}\brk*{y_{1} \mid \hat{x}} \log \frac{p\brk*{y_{0} \mid x}}{p\brk*{y_{1} \mid x}}  
	\cdot p_{\beta}\brk*{y_{0} \mid \hat{x}} p_{\beta}\brk*{\tilde{x} \mid \hat{x}} \log \frac{ p\brk*{y_{0} \mid x} }{p \brk*{y_{0} \mid \tilde{x}}} \\
	+& \sum_{x, \tilde{x}} p_{\beta}\brk*{{x} \mid \hat{x}}  p_{\beta}\brk*{y_{0} \mid \hat{x}} \log \frac{p\brk*{y_{1} \mid {x}}}{p\brk*{y_{0} \mid {x}}}  
	\cdot p_{\beta}\brk*{y_{1} \mid \hat{x}} p_{\beta}\brk*{\tilde{x} \mid \hat{x}} \log \frac{ p\brk*{y_{1} \mid {x}} }{p \brk*{y_{1} \mid \tilde{x}}} \\
	=& p_{\beta}\brk*{y_{1} \mid \hat{x}}  
	p_{\beta}\brk*{y_{0} \mid \hat{x}}  \sum_{x, \tilde{x}} p_{\beta}\brk*{x \mid \hat{x}}  p_{\beta}\brk*{\tilde{x} \mid \hat{x}} \log \frac{p\brk*{y_{0} \mid x}}{p\brk*{y_{1} \mid x}} \brk[s]*{\log \frac{ p\brk*{y_{0} \mid x} }{p \brk*{y_{0} \mid \tilde{x}}} - \log \frac{ p\brk*{y_{1} \mid x} }{p \brk*{y_{1} \mid \tilde{x}}} }
.\end{align*}

\subsection{Definition of the sample problem}
\label{app:def_prob}
We consider a problem for a binary label $Y$ and $5$ possible inputs $X$ uniformly distributed, i.e.  $\forall x \in \mathcal{X}, p\brk*{x} = 1/5$ and the conditional distribution, $p\brk*{y\mid x}$, given by:
\begin{center}
            \begin{tabular}{c| c c c c c}
        & $x= 0$ & $x=1$ & $x=2 $ & $x=3$ & $x=4$ \\ \hline 
            $y=0$ &  0.12 & 0.23 & 0.4 & 0.6 & 0.76 \\ \hline 
            $y=1$ &  0.88 & 0.77 & 0.6 & 0.4 & 0.24 
        \end{tabular}
\end{center}
\section*{Appendix E - Information plane analysis}
\label{app:performance_anl}
We rely on known results for the rate-distortion problem and the information plane:
\begin{lemma} \label{lm:dist_convex}
 $I(x;\hX)$ is a non-increasing convex function of the distortion $\mathbb{E}_{p_{\beta}\brk*{x, \hx}}\brk[s]*{d\brk*{x, \hx}}$ with a slope of $-\beta$. 
\end{lemma} 
We emphasis that this is a general result of rate-distortion thus holds for the $\dualib$ as well.
\begin{lemma} \label{lm:ib_concave}
For a fixed encoder $p_{\beta}\brk*{\hx \mid x}$ and the Bayes optimal decoder $p_{\beta}\brk*{y \mid \hx}$:
\begin{align*}
    \mathbb{E}_{p_{\beta}\brk*{x, \hx}}\brk[s]*{d_{\ib}\brk*{x, \hx}} = I(X ; Y) - I(\hX ; Y)
.\end{align*}
Thus, the information curve, $I_{y}$ vs. $I_{x}$, is a non-decreasing concave function with a positive slope, $\beta^{-1}$. The concavity implies that $\beta$ increases along the curve.
\end{lemma} 
\cite{Cover:2006:EIT:1146355, Gilad-bachrach}.

\subsection{Proof of Lemma 3} 
\label{sec:lemma_info_proof}
In the following section we provide a proof to \emph{lemma} $3$,  for the $\ib$ and {\dualib} problems.
\begin{proof}
We want to analyze the behavior of $I_{x}\brk*{\beta}$, $I_{y}\brk*{\beta}$, that is the change in each term as a function of the corresponding $\beta$. From \lmref{lm:ib_concave}, the concavity of the information curve, we can deduce that both are non-decreasing functions of $\beta$. As the two $\beta$ derivatives are proportional it's enough to discuss the first one. 

Next, we focus on their behavior between two critical points. That is, where the cardinality of $\hX$ is fixed (clusters are "static"). 
For ''static" clusters, the $\beta$ derivative of $I_{x}$, along the optimal line is given by:
\begin{align*} 
 \frac{\partial  I(X ; \hat{X})  }{\partial \beta} &=-  \frac{\partial  }{\partial\beta} \brk[s]*{\sum_{x, \hat{x}} p_{\beta}\brk*{ x, \hat{x}} \brk*{\log {Z_{\hat{\rvx} \mid \rvx }\brk*{x;\beta} } + \beta d\brk*{x,\hx} } } \\
 &= - \beta \brk[a]*{ d\brk*{{x} ,\hx }\frac{\partial \log p_{\beta} \brk*{\hx \mid x}  }{\partial \beta}  }_{p_{\beta}\brk*{{x} ,\hx }} \\ 
 &\approx  \beta \brk[a]*{ d\brk*{{x} ,\hx } \brk[s]*{ \frac{\partial \log Z_{\hat{\rvx} \mid \rvx }\brk*{x;\beta}  }{\partial \beta} + d\brk*{{x} ,\hx } }  }_{p_{\beta}\brk*{{x} ,\hx }} \\
&\approx  \beta\brk[a]*{\underbrace{\brk[a]*{ d^2\brk*{{x} ,\hx }}_{p_{\beta}\brk*{\hx \mid x} } -\brk[a]*{ d\brk*{{x} ,\hx } }^2_{p_{\beta}\brk*{\hx \mid x } } }_{\textrm{Var}\brk*{d\brk*{x}} } }_{p\brk*{x}}
.\end{align*}
This first of all reassures that the function is non-decreasing as $\textrm{Var}\brk*{d(x)} \geq 0$.

The piece-wise concavity follows from the fact that when the number of clusters is fixed (between the critical points) - increasing $\beta$ decreases the clusters conditional entropy $H(\hX \mid x)$, as the encoder becomes more deterministic. The mutual information is bounded by $H(\hX)$ and it's $\beta$ derivative decreases. Further, between the critical points there are no sign changes in the second $\beta$ derivative.
\end{proof}

\subsection{Proof of Theorem 4} 
\begin{proof}
The proof follows from \textit{lemma} $3$ together with the critical points analysis above, and is only sketched here. 
As the encoder and decoder at the critical points, $\beta_{c}^{\ib}$ and $\beta_{c}^{\dualib}$, have different left and right derivatives, they form cusps in the curves of the mutual information ($I_{x}$ and $I_{y}$) as functions of $\beta$. These cusps can only be consistent with the optimality of the {\ib} curves ( implying that sub-optimal curves lie below it; i.e, the {\ib} slope is steeper) if $\beta_{c}^{\dualib} < \beta_{c}^{\ib}$ (this is true for any sub-optimal distortion), otherwise the curves intersect. 

Moreover, at the $\dualib$ critical points, the distance between the curves is minimized due to the strict concavity of the functions segments between the critical points. As the critical points imply discontinuity in the derivative, this results in a ''jump" in the information values. Therefore, at any $\beta_{c}^{\dualib}$ the distance between the curves has a (local) minimum.
This is depicted in Figure $4$ (in the main text), comparing $I_{x}\brk*{\beta}$ and $I_{y}\brk*{\beta}$ and their differences for the two algorithms.

The two curves approach each other for large $\beta$ since the two distortion functions become close in the low distortion limit (as long as $p\brk*{y \mid x}$ is bounded away from $0$).
\end{proof}
\section*{Appendix F - Derivation of the dualExpIB}
\label{app:dualexpIB}
We  provide elaborate derivations to \textit{theorem} $9$; that is, we obtain the {\dualib} optimal encoder-decoder under the exponential assumption over the data.
We use the notations defined in \S \emph{The Exponential Family dualIB}.
\begin{itemize}
    \item The \textit{decoder}. 
    Substituting the exponential assumption into the {\dualib} $\log$-decoder yields:
    \begin{align*} 
	\log p_{\beta}\brk*{y \mid \hat{x} } &=  \sum_{x} p_{\beta}\brk*{x \mid \hat{x}} \log p\brk*{y \mid x}  - \log Z_{\rvy\mid \hat{\rvx}} \brk*{\hat{x}; \beta}  \nonumber \\
	&= - \sum_{x} \sum_{r=0}^{d} p_{\beta}\brk*{x \mid \hat{x}}   {\lambda}^r\brk*{y} A_r \brk*{x} - \log Z_{\rvy\mid \hat{\rvx}} \brk*{\hat{x};\beta} \nonumber \\
	&= -  \sum_{r=1}^{d}  {\lambda}^r\brk*{y} A_{r,\beta} \brk*{\hx} - \mathbb{E}_{p_{\beta}\brk*{x\mid \hat{x}}} \brk[s]*{ {\lambda}^0_{\rvx}}- \log Z_{\rvy\mid \hat{\rvx}} \brk*{\hat{x};\beta}
.\end{align*}
Taking a closer look at the normalization term:
\begin{align*} 
	  Z_{\rvy\mid \hat{\rvx}} \brk*{\hat{x}; \beta} &= \sum_{y} e^{\sum_{x} p_{\beta}\brk*{x \mid \hat{x}} \log p\brk*{y \mid x}   } = e^{-\mathbb{E}_{p_{\beta}\brk*{x\mid \hat{x}}} \brk[s]*{ {\lambda}^0_{\rvx}}}\sum_{y} e^{-\sum_{r=1}^{d}  {\lambda}^r\brk*{y} A_{r, \beta} \brk*{\hx}} \nonumber \\
	  \log Z_{\rvy\mid \hat{\rvx}} \brk*{\hat{x};\beta} &= -\mathbb{E}_{p_{\beta}\brk*{x\mid \hat{x}}} \brk[s]*{ {\lambda}^0_{\rvx}} + \log \brk*{\sum_{y}e^{-\sum_{r=1}^{d}  {\lambda}^r\brk*{y} A_{r, \beta} \brk*{\hx}}} \nonumber
.\end{align*}
From which it follows that ${\lambda}^{0}_{\beta}\brk{ \hx}$ is given by:
\begin{align*} 
 {\lambda}^{0}_{\beta}\brk{\hx} &= \log \brk*{\sum_{y} e^{-\sum_{r=1}^{d}  {\lambda}^r\brk*{y} A_{r,\beta} \brk*{\hat{x}}}}
,\end{align*}
and we can conclude that the $\expib$ decoder takes the form:
\begin{align*} 
	\log p_{\beta}\brk*{y \mid \hat{x} } &=  -\sum_{r=1}^{d}  {\lambda}^r\brk*{y} A_{r,\beta}\brk*{\hat{x}} - {\lambda}^{0}_{\beta}\brk{\hx}
.\end{align*}
\item The \textit{encoder}. \\
The core of the encoder is the dual distortion function which may now be written as:
 \begin{align*} 
	 d_{\dualib}\brk*{x, \hat{x}} &= \sum_{y} p_{\beta}\brk*{y\mid \hat{x}} \log \frac{p_{\beta}\brk*{y\mid \hat{x}}}{{p}\brk*{y\mid {x}} } \nonumber \\
	 &=  \sum_{y} p_{\beta}\brk*{y\mid \hat{x}}\brk[s]*{ \brk*{{\lambda}^{0}_{\rvx}-{\lambda}^{0}_{\beta}\brk*{\hat{x}}}+\sum_{r=1}^{d} {\lambda}^r\brk*{y} \brk*{A_r \brk*{x} - A_{r,\beta}\brk*{\hat{x}} }}\nonumber\\
	 &= {\lambda}^{0}_{\rvx}-{\lambda}^{0}_{\beta}\brk*{\hx}  +  \sum_{r=1}^{d} {\lambda}_{\beta}^r\brk*{\hat{x}}   \brk*{A_r \brk*{x} - A_{r, \beta}\brk*{\hat{x}}} 
,\end{align*}
substituting this into the encoder's definition we obtain:
 \begin{align*} 
	 p_{\beta}\brk*{\hx\mid x} &= \frac{p_{\beta}\brk*{\hx}}{Z_{\hat{\rvx}\mid \rvx}\brk*{x;\beta} }  e^{ -\beta\brk[s]*{ {\lambda}^{0}_{\rvx}-{\lambda}^{0}_{\beta}\brk*{\hat{x}}+  \sum_{r=1}^{d} {\lambda}_{\beta}^r\brk*{\hat{x}}   \brk[s]*{A_r \brk*{x} - A_{r, \beta}\brk*{\hat{x}}} }} \\
	 &= \frac{p_{\beta}\brk*{\hx}e^{\beta {\lambda}^{0}_{\beta}\brk*{\hx}}}{Z_{\hat{\rvx}\mid \rvx}\brk*{x;\beta} }  e^{-\beta  \sum_{r=1}^{d} {\lambda}_{\beta}^r\brk*{\hat{x}}   \brk[s]*{A_r \brk*{x} - A_{r, \beta}\brk*{\hat{x}}} }
.\end{align*}
\end{itemize}
We can further write down the information quantities under these assumptions:
\begin{align*}
 I(X;\hX) &= \sum_{x,\hx} p_{\beta}\brk*{x, \hx } \log \frac{p_{\beta}\brk*{x \mid \hx}}{p\brk*{x } }  \\
        &= H\brk*{X} - \beta \sum_{r=1}^{d}  \sum_{\hx} p_{\beta}\brk*{\hx } {\lambda}_{\beta}^r\brk*{\hat{x}}   \brk[s]*{  \sum_{x} p_{\beta}\brk*{x\mid \hx } A_r \brk*{x} - A_{r, \beta}\brk*{\hat{x}}} + \\\beta  \mathbb{E}_{p_{\beta}\brk*{\hx}}\brk[s]*{{\lambda}^{0}_{\beta}\brk*{\hx} }  - \mathbb{E}_{p\brk*{x}}\brk[s]*{ \log Z_{\hat{\rvx}\mid \rvx}\brk*{x;\beta}} \\
        &= H\brk*{X} + \beta  \mathbb{E}_{p_{\beta}\brk*{\hx}}\brk[s]*{{\lambda}^{0}_{\beta}\brk*{\hx} }  - \mathbb{E}_{p\brk*{x}}\brk[s]*{ \log Z_{\hat{\rvx}\mid \rvx}\brk*{x;\beta}} \\
 I(Y;\hX) &= \sum_{y,\hx} p_{\beta}\brk*{y, \hx } \log \frac{p_{\beta}\brk*{y \mid \hx}}{p\brk*{y } }  \\
        &= H\brk*{Y} - \sum_{r=1}^{d} \sum_{\hx} p_{\beta}\brk*{ \hx }   \sum_{y}p_{\beta}\brk*{y \mid \hx } {\lambda}^r\brk*{y} A_{r,\beta}\brk*{\hat{x}} - \mathbb{E}_{p_{\beta}\brk*{\hx}}\brk[s]*{{\lambda}^{0}_{\beta}\brk*{\hx} } \\
        &=  H\brk*{Y} - \mathbb{E}_{p_{\beta}\brk*{\hx}}\brk[s]*{ \sum_{r=1}^{d}{\lambda}_{\beta}^r\brk*{\hat{x}}A_{r,\beta}\brk*{\hat{x}} + {\lambda}^{0}_{\beta}\brk*{\hx} }
\end{align*}
\section*{Appendix G - Optimizing the error exponent}
\label{app:err_exp}
We start by to expressing the Chernoff information for the binary hypothesis testing problem using $p\brk*{y
\mid x}$: 
\begin{align*}
C\brk*{p_{0}, p_1} &= \min_{\lambda \in \brk[s]*{0,1}} \log \brk*{\sum_{x} p\brk*{x \mid y_0}^{q_{\lambda}\brk*{y_0}} p\brk*{x \mid y_1}^{q_{\lambda}\brk*{y_1}} } \\
&= \min_{\lambda \in \brk[s]*{0,1}} \log \brk*{\sum_{x} p\brk*{y=0 \mid x}^{\lambda}p\brk*{x}^{\lambda} p\brk*{y=0}^{-\lambda}p\brk*{y=1 \mid x}^{1-\lambda}p\brk*{x}^{1-\lambda} p\brk*{y=1}^{\lambda-1}} \nonumber \\
&= \min_{\lambda \in \brk[s]*{0,1}} \log \brk*{\sum_{x}p\brk*{x} p\brk*{y=0 \mid x}^{\lambda} p\brk*{y=1 \mid x}^{1-\lambda} } -\log\brk*{p\brk*{y=0}^{\lambda}p\brk*{y=1}^{1-\lambda}} \nonumber \\
&= \min_{q_{\lambda}\brk*{y}} \log \brk*{\sum_{x} e^{q_{\lambda}\brk*{y_0} \log p\brk*{x \mid y_0} + q_{\lambda}\brk*{y_1}\log p\brk*{x \mid y_1}} } \\
&=\min_{q_{\lambda}\brk*{y}} \log \brk*{\sum_{x}  e^{-D\brk[s]*{q_{\lambda}\brk*{y} \| p\brk*{y \mid x}} + D\brk[s]*{q_{\lambda}\brk*{y} \| p{y}} + \log p\brk*{x}} }\\
&= \min_{q_{\lambda}\brk*{y}} \log \brk*{e^{D\brk[s]*{q_{\lambda}\brk*{y} \| p\brk*{y}}} \sum_{x}  e^{-D\brk[s]*{q_{\lambda}\brk*{y} \| p\brk*{y \mid x}}  + \log p\brk*{x}} } \\
&= \min_{q_{\lambda}\brk*{y}} \brk[c]*{ \log \brk*{ \sum_{x} p\brk*{x}e^{-D\brk[s]*{q_{\lambda}\brk*{y} \| p\brk*{y \mid x}}} } + D\brk[s]*{q_{\lambda}\brk*{y} \| p\brk*{y}}}
,\end{align*}
where $q_{\lambda}\brk*{y_0} = \lambda, q_{\lambda}\brk*{y_1} = 1-\lambda$.
Now, if we consider the mapping, $q_{\lambda}\brk*{y} =  p_{\beta}\brk*{y \mid \hx}$ we can write the above as:
\begin{align*}
C\brk*{p_{0}, p_1} &= \min_{p_{\beta}\brk*{y \mid \hx}}\brk[c]*{ \log \brk*{\sum_{x} p\brk*{x} e^{-D\brk[s]*{p_{\beta}\brk*{y \mid \hx}\| p\brk*{y \mid x}}} } + D\brk[s]*{ p_{\beta}\brk*{y \mid \hx} \| p\brk*{y}} }
.\end{align*}
The above term in minimization is proportional to $\log$-partition function of $p_{\beta}\brk*{x \mid \hx}$, namely we get the mapping $p_{\beta}\brk*{x \mid \hx} = p_{\lambda}$. 
Next we shall generalize the setting to the $M$-hypothesis testing problem. Having that solving for the Chernoff information is notoriously difficult we consider an upper bound to it, taking the expectation over the classes. Instead of  choosing $p_{\lambda^*}$ as the maximal value of the minimimum  $\brk[c]*{D\brk[s]*{p_{\lambda^*} \| p_0}, D\brk[s]*{p_{\lambda^*} \| p_1}}$ we consider it w.r.t the full set $\brk[c]*{D\brk[s]*{p_{\lambda^*} \| p_i}}_{i=1}^{M}$. Using the above mapping we must take the expectation also over the representation variable $\hx$. Thus we get the expression:
\begin{align*} \label{eq:chernoff_dual}
    D^{*}\brk*{\beta} &= \min_{p_{\beta}\brk*{y\mid \hx}, p_{\beta}\brk*{ \hx \mid x}}\mathbb{E}_{p_{\beta}\brk*{y, \hx}}\brk[s]*{D\brk[s]*{ p_{\beta}\brk*{ x \mid \hx} \mid p\brk*{x\mid y}}}
.\end{align*}
  From the definition of $D^{*}\brk*{\beta}$ we obtain the desired bound of the {\dualib}:
  \begin{align*} 
    D^{*}\brk*{\beta} &= \min_{p_{\beta}\brk*{y\mid \hx}, p_{\beta}\brk*{ \hx \mid x}}\mathbb{E}_{p_{\beta}\brk*{y, \hx}}\brk[s]*{D\brk[s]*{ p_{\beta}\brk*{ x \mid \hx} \| p\brk*{x\mid y}}} \\
    &= \min_{p_{\beta}\brk*{y\mid \hx}, p_{\beta}\brk*{ \hx \mid x}}\sum_{x, y,\hx} p_{\beta}\brk*{y \mid\hx}p_{\beta}\brk*{\hx} \brk[s]*{D\brk[s]*{ p_{\beta}\brk*{ x \mid \hx} \mid p\brk*{x\mid y}}} \\
    &= \min_{p_{\beta}\brk*{y\mid \hx}, p_{\beta}\brk*{ \hx \mid x}}\sum_{x, y,\hx} p_{\beta}\brk*{y \mid\hx}p_{\beta}\brk*{\hx}p_{\beta}\brk*{x \mid\hx} \brk[c]*{ \log \frac {p_{\beta}\brk*{y \mid\hx}}{p\brk*{y\mid x}} + \log \frac {p_{\beta}\brk*{x \mid\hx}}{p_{\beta}\brk*{y \mid\hx}} + \log \frac{p\brk*{y}}{p\brk*{x}} } \\
    &=  \min_{p_{\beta}\brk*{y\mid \hx}, p_{\beta}\brk*{ \hx \mid x}}\brk[c]*{I(X;\hX) + \mathbb{E}_{p_{\beta}\brk*{x, \hx}}\brk[s]*{D\brk[s]*{ p_{\beta}\brk*{ y \mid \hx} \| p\brk*{y \mid x}}} + H(Y \mid \hX) + \mathbb{E}_{p_{\beta}\brk*{y}}\brk[s]*{\log p\brk*{y}} }\\
    &\leq \min_{p_{\beta}\brk*{y\mid \hx}, p_{\beta}\brk*{ \hx \mid x}}\brk[c]*{I(X;\hX) + \mathbb{E}_{p_{\beta}\brk*{x, \hx}}\brk[s]*{D\brk[s]*{ p_{\beta}\brk*{ y \mid \hx} \| p\brk*{y \mid x}}}} \\
    &\leq \mathcal{F}^{*}\brk[s]*{p\brk*{\hat{x} \mid x}; p\brk*{y \mid \hat{x} }}
.\end{align*}
  
  \subsection{Error exponent optimization example}
\begin{figure}[htb!]
    \begin{center}
        \includegraphics[scale = 0.42]{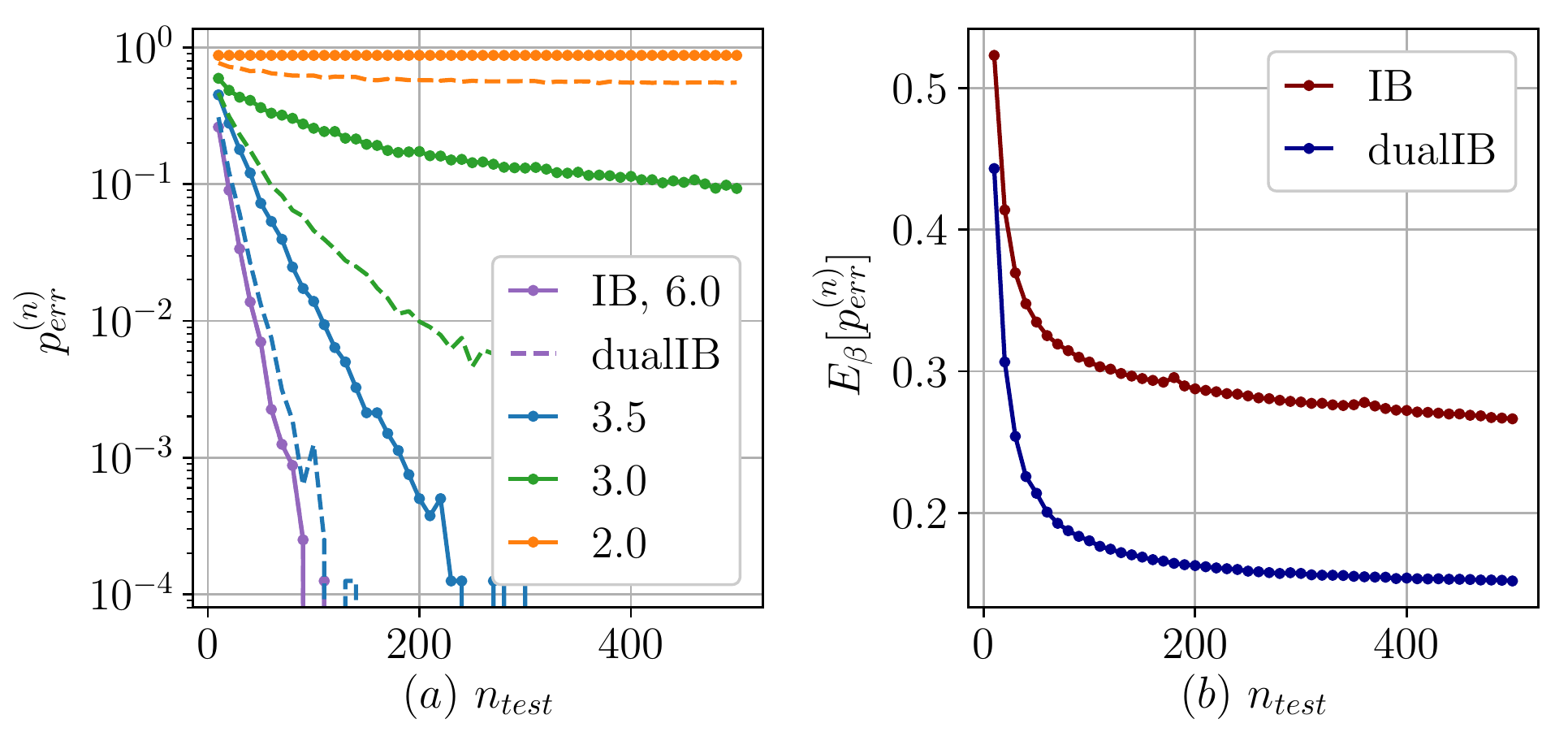}
    \end{center}
        
        \caption{The probability of error, $p_{err}^{\brk*{n}}$, as a function of test sample size, $n_{test}$. $\brk*{a}$ The exponential decay of error for representative $\beta$ values ($\log_2(\beta)$ reported in the legend). For a given $\beta$ the {\ib} performance is plotted in solid line and the {\dualib} in dashed (for $\log_{2}\brk*{\beta} = 6$ the lines overlap). $\brk*{b}$ The expectation of the error over all $\beta$'s ($\log_2\beta \in \brk[s]*{1, 6}$).}
        
        \label{fig:p_err}
  \end{figure}

To demonstrate the above properties we consider a classification problem with $M=8$ classes, each class characterized by $p_{i} = p\brk*{x \mid y_i}$. The \emph{training} is performed according to the above algorithms to obtain the {\ib} ({\dualib}) encoder and decoder. For the prediction, given a new sample $x^{\brk*{n}} \overset{i.i.d}{\sim} p\brk*{x\mid y}$ defining an empirical distribution $\hat{p}\brk*{x}$ the prediction is done by first evaluating $\hat{p}_{\beta}\brk*{\hx} = \sum_{x} p_{\beta}\brk*{\hx \mid x}\hat{p}\brk*{x}$. Next, using the (representation) optimal decision rule, we obtain the prediction:
\begin{align*} 
    \hat{H}_{\beta} = \arg \min_{i} D\brk[s]*{\hat{p}_{\beta}\brk*{\hx} \| p_{\beta}\brk*{\hx \mid y_i}  }  
    ,\end{align*}
and we report $p_{err}^{\brk*{n}}$, the probability of miss-classification. 
This represents the most general classification task; the distributions $p_i$ represent the empirical distributions over a training data-set and then testing is performed relative to a test set.
Looking at the results, \fref{fig:p_err}, it is evident that indeed the {\dualib} improves the prediction error (at $log_2\brk*{\beta} = 6$ the algorithms performance is identical due to the similarity of the algorithms behavior as $\beta$ increases).

\section*{Appendix H - The variational {\dualib} } 
\subsection*{Derivation of the {\vdib} objective}
\label{app:vdib_obj}
Just as \citep{fischer2020ceb} did, we can variationally upper bound the information of the input with the representation variable using:
\begin{align*}
    I(\hX;X \mid Y)=\mathbb{E}_{p(x,y)p(\hx\mid x)}\left[\log\frac{p(\hx \mid x,y)}{p(\hat{x}\mid y)}\right]\leq \mathbb{E}_{\tilde{p}(y\mid x) p\brk*{x}p(\hat{x}\mid x)}\left[\log\frac{p(\hat{x}\mid x)}{q(\hat{x}\mid y)}\right]
\end{align*}
where $q(\hx \mid y)$ is a variational class conditional marginal.
In contradiction to the CEB, in order to bound the {\dualib} distortion, we replace the bound on $I(\hX;Y)$ with a bound over the expected {\dualib} distortion. Here, given the assumption of a noise model $\tilde{p}\brk*{y \mid x}$ which we evaluate the expected distortion with respect to it:
\begin{align*}
    \mathbb{E}_{p\brk*{x, \hx}}\brk[s]*{d_{\dualib}\brk*{x, \hx}} &=  \mathbb{E}_{{p}\brk*{y\mid\hx}p\brk*{\hx\mid x} p\brk*{x}}\brk[s]*{\log \frac{p\brk*{y\mid \hx}}{\tilde{p}\brk*{y\mid x}}} 
\end{align*}
Combining the above together gives the variational upper bound to the {\dualib} as the following objective:
\begin{align*}
    I(X;\hat{X}) + \beta \mathbb{E}_{p\brk*{x, \hx}}\brk[s]*{d_{\dualib}\brk*{x, \hx}}  \leq   
   {\mathbb{E}_{\tilde{p}(y\mid x)p\brk*{\hx\mid x} p\brk*{x}}\brk[s]*{
   \log\frac{p\brk*{\hx\mid x}}{q\brk*{\hx\mid y}} } + \beta \mathbb{E}_{{p}\brk*{y\mid\hx}p\brk*{\hx\mid x}p\brk*{x}}\brk[s]*{\log \frac{p\brk*{y\mid \hx}}{\tilde{p}\brk*{y\mid x}}}}
\end{align*}
\subsection*{Experimental setup}
\label{app:exp_setup}
For  both CIFAR10 and FasionMNIST We trained a set of 30 $28-10$ Wide ResNet
models in a range of values of $\beta$ ($-5\leq \log\beta \leq 5$). The training was doneusing Adam \citep{kingma2014adam} at a base learning rate of $10^{-4}$. We lowered the learning rate two times by a factor of $0.3$ each time.
Additionally,  following \cite{fischer2020ceb}, we use a jump-start method for $\beta<100$. We start the training with $\beta=100$, anneal down  to the target $\beta$ over 1000 steps. The training includes data augmentation with horizontal flip and width height shifts. Note, that we exclude from the analysis runs that  didn't succeed to learn at all (for which the results look as random points).
\subsection*{The variational information plane}
Note that, for the information plane analysis, there were several runs that failed to achieved more than random accuracy. In such cases, we remove them. 
The confusion matrix used for the FashionMNIST data-set is:
\begin{align*}
\begin{pmatrix}
0.828 & 0.013 & 0.012 & 0.011 & 0.018 & 0.& 0.002 & 0.004 & 0.085 & 0.027  \\
0.01&0.91&0.&0.005&0.001&0.001&0.&0.001&0.011&0.061  \\
0.047&0.001&0.708&0.064&0.088&0.014&0.063&0.004&0.008&0.003 \\
0.003&0.004&0.016&0.768&0.033&0.093&0.05&0.019&0.004&0.01 \\
0.01&0.&0.039&0.043&0.788&0.012&0.057&0.043&0.006&0.002 \\
0.002&0.&0.01&0.137&0.029&0.777&0.008&0.033&0.&0.004 \\
0.007&0.002&0.01&0.054&0.029&0.007&0.888&0.001&0.001&0.001 \\
0.024&0.002&0.014&0.039&0.076&0.017&0.004&0.818&0.002&0.004 \\
0.027&0.013&0.&0.007&0.003&0.&0.003&0.&0.933&0.014. \\
0.019&0.064&0.001&0.007&0.002&0.001&0.001&0.&0.018&0.887
\end{pmatrix}    
\end{align*}
\subsection*{The {\vdib} noise models}
\label{app:noise_model}
\begin{figure}[htb!]
    \centering
    \includegraphics[width=\textwidth]{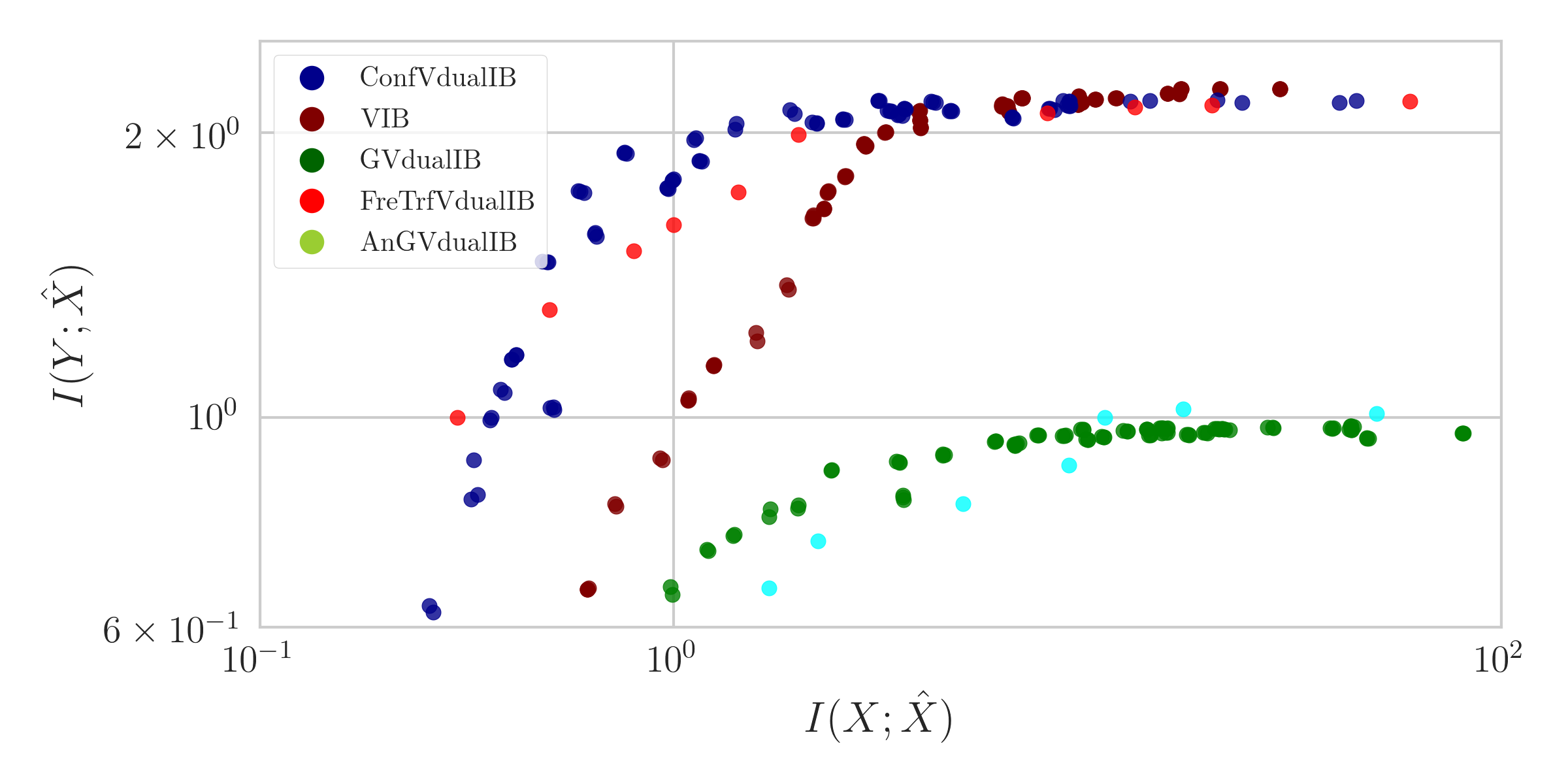}
 \caption{The information plane for the different noise models.}
 \label{fig:noise_models}
\end{figure}

As described in the main text we consider two additional noise models; (i) An analytic Gaussian integration of the log-loss around the one-hot labels ({\anvdib}) (ii) 
Using predictions of another trained model as the induced distribution ({\prdvib}). In this case, we use a deterministic wide ResNet $28-10$ network that achieved $95.8\%$ accuracy on CIFAR10. In \fref{fig:noise_models} we can see all the different models, $4$ noise models for the {\vdib} and the {\vib}). As expected, we can see that analytic Gaussian integration  noise model obtains similar results to adding Gaussian noise to the one-hot vector of the true label, while the performance of the noise models that are based on a trained network are similar to the {\cvdib}. 

\subsubsection{Training {\vib} model with noise}
\begin{figure}[htb!]
    \centering
    \includegraphics[width=\textwidth]{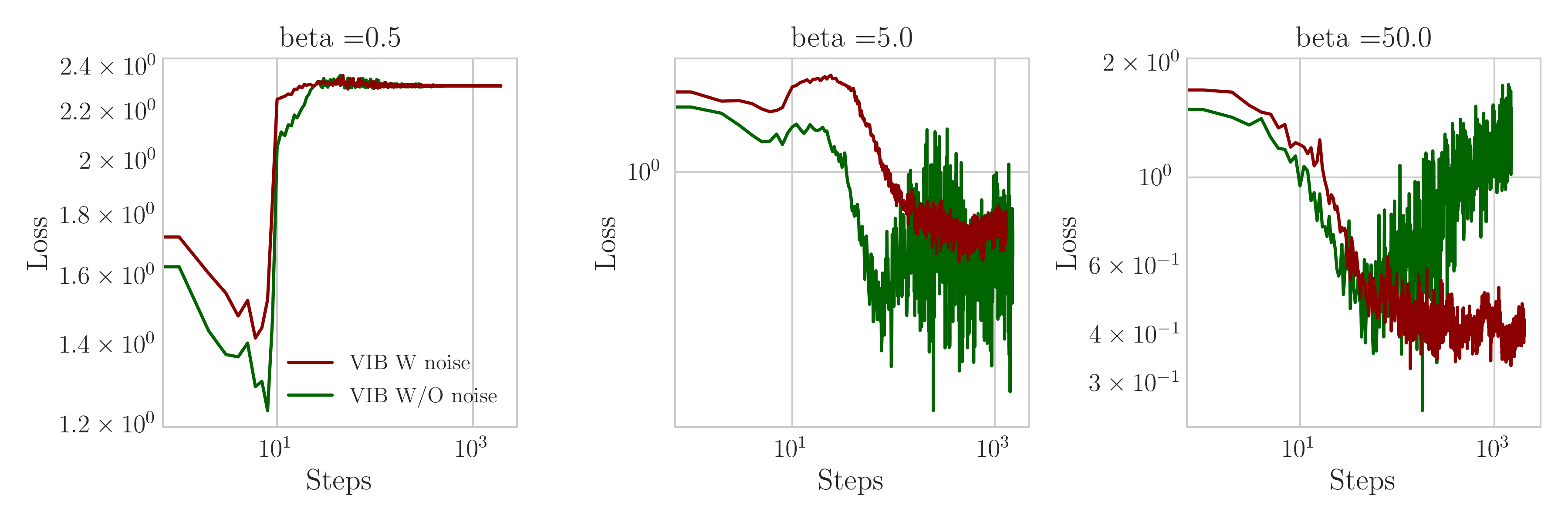}
 \caption{The influence of a noise model on the {\vib} performance. Loss as function of the update steps for different values of $\beta$, $\beta = 0.5, 5.0. 50.0$ from left to right.}
 \label{fig:label_smooth}
\end{figure}

In our analysis, we train a {\vib} model with the same noise model as the {\vdib}. Namely, instead of training with a deterministic label (one-hot vector of zeros and ones), we use our noise model also for the {\vib}. As mentioned in the text, this training procedure is closely related to label smoothing. In \fref{fig:label_smooth}, we present the loss function of the {\vib} on CIFAR10 with and without the noise models along the training process for $3$  different values of $\beta$. For a small $\beta$ (left) both regimes under-fit the data as expected. However, when we enlarge $\beta$, we can see that the labels' noise makes the training more stable and for a high value of $\beta$ (right) training without noise over-fits the data and the loss increases.

The confusion matrix for the CIFAR10 data set is:
\begin{align*}
  \begin{pmatrix}
0.878 & 0.  & 0.017& 0.013&0.002& 0.001& 0.082& 0.   & 0.007&
        0.   \\
       0.  & 0.984& 0.002&0.009& 0.001& 0.   & 0.003& 0.   & 0.   &
        0.   \\
       0.013& 0.001& 0.896& 0.009& 0.038& 0.   & 0.043& 0.   & 0.   &
        0.   \\
       [0.022& 0.004& 0.011& 0.913& 0.023& 0.   & 0.027& 0.   & 0.001&
        0.   \\
       0.   & 0.   & 0.072& 0.022& 0.85 & 0.   & 0.058& 0.   & 0.   &
        0.\\   
       0.   & 0.   & 0.  & 0.   & 0.   & 0.982& 0.   & 0.011& 0.   &
        0.007\\
       0.099& 0.001& 0.049& 0.021& 0.055& 0.   & 0.768& 0.   & 0.005&
        0.   \\
       0.   & 0.   & 0.   & 0.   & 0.   & 0.006& 0.   & 0.976& 0.   &
        0.019\\
       0.004& 0.001& 0.001& 0.001& 0.004&0.002&0.003& 0.001& 0.98 &
        0.001\\
       0.   & 0.   & 0.   & 0.   & 0.   & 0.004& 0.   & 0.02 & 0.001&
        0.974
\end{pmatrix}  
\end{align*}
\subsection*{CIFAR100 results}
\label{app:cifar100}
\begin{figure}[htb!]
\begin{subfigure}{.5\textwidth}
    \centering
    \includegraphics[width=\textwidth]{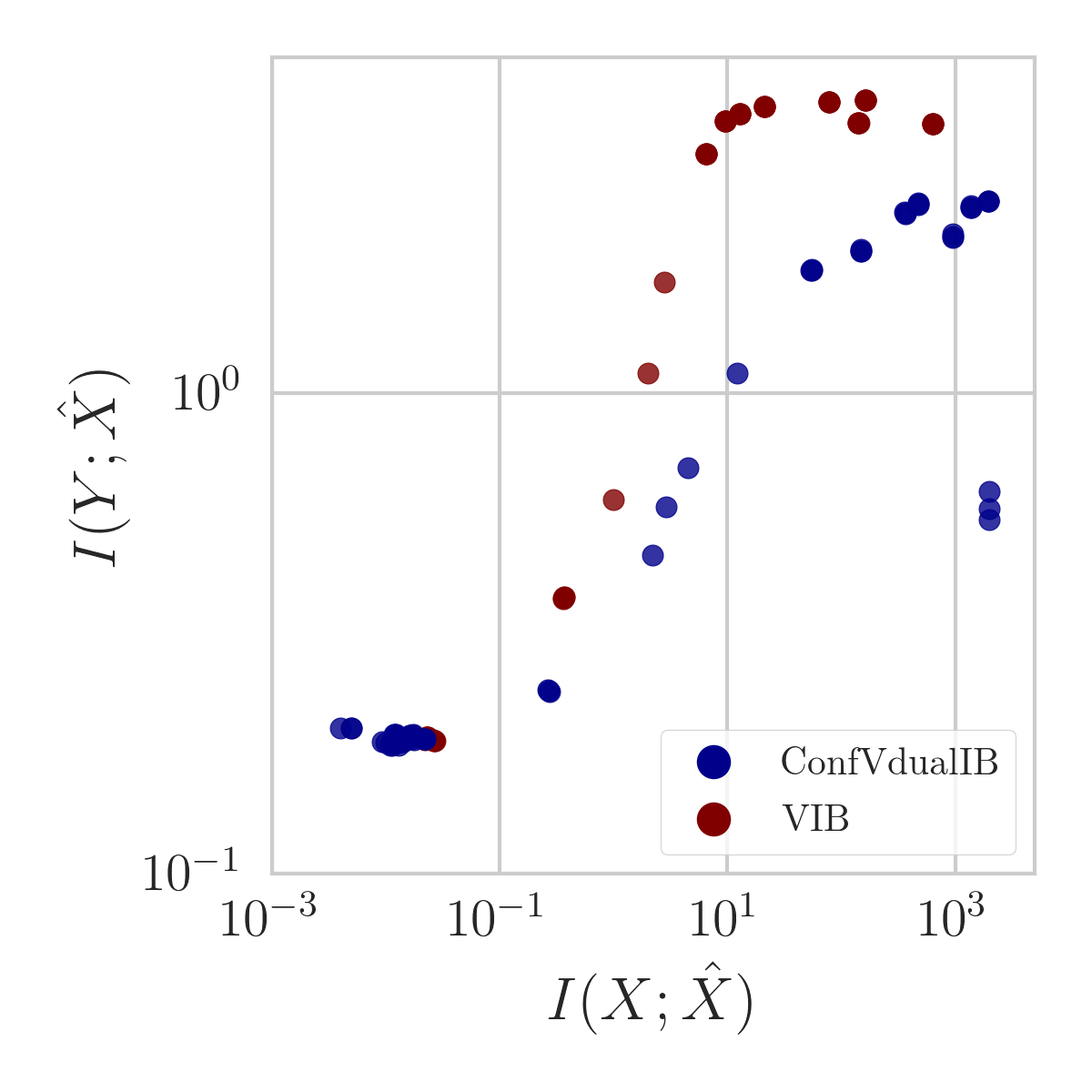}
    \caption{The information plane}
    \label{fig:inf_plane_cifar100_beta}
\end{subfigure}
\begin{subfigure}{.5\textwidth}
    \centering
    \includegraphics[width=\textwidth]{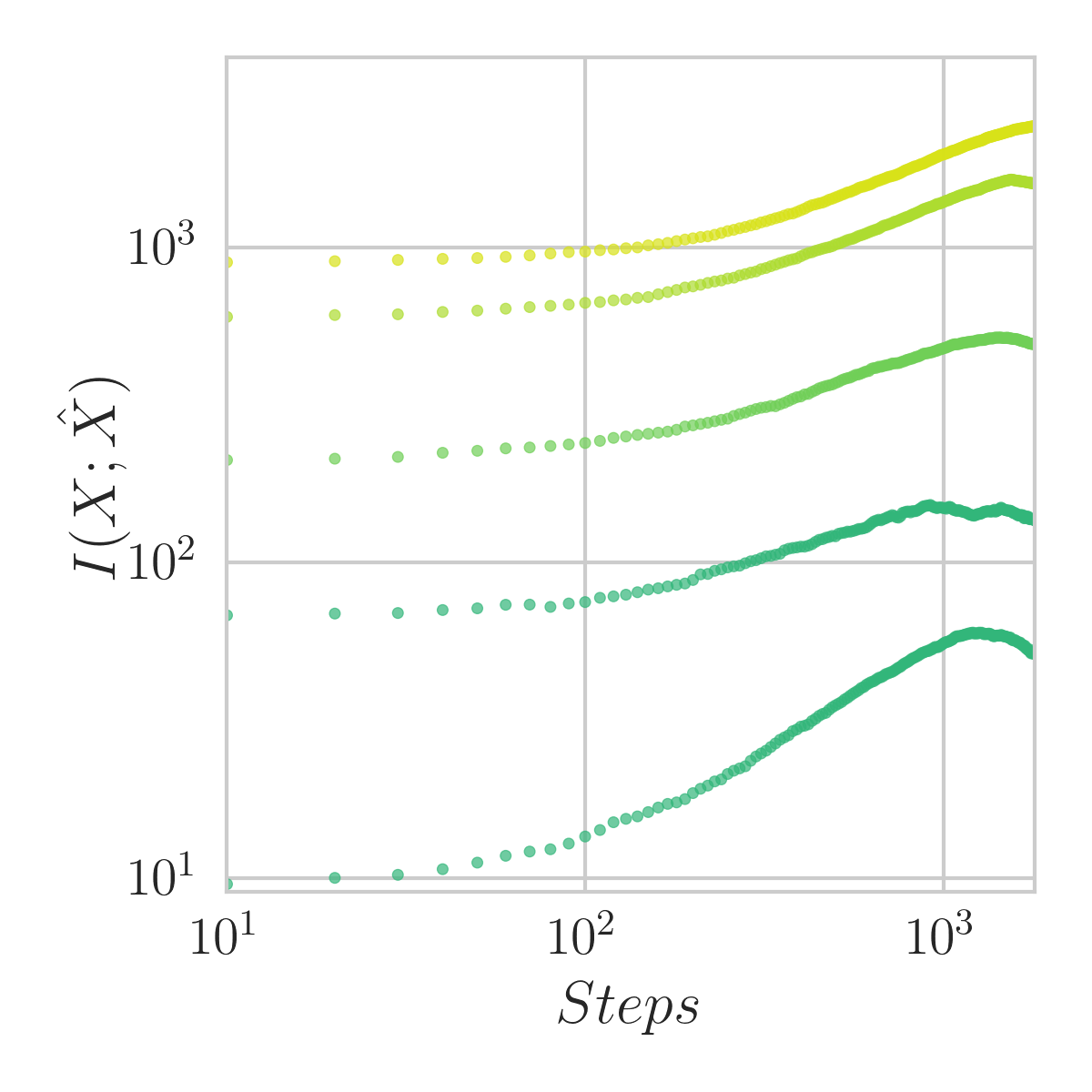}
    \caption{$I(X;\hX)$ vs. update steps for different $beta$'s}
    \label{fig:inf_plane_cifar100_steps}
\end{subfigure}
 \caption{Experiments over CIFAR100. $(a)$ The information plane of the {\cvdib} and {\vib} for a range of $\beta$ values at the final training step. $(b)$ The evolution of the the {\cvdib}'s $I(X;\hX)$ along the optimization update steps.}
 \label{fig:inf_plane_cifar100}
\end{figure}

As mentioned in the text, we trained {\vdib} networks also on CIFAR100. For this, we used the  same  $28-10$ Wide ResNet with a confusion matrix as our noise model.  The confusion matrix was calculated based on the predictions of a deterministic network. The deterministic network achieved $80.2\%$ accuracy on CIFAR100.
In \ref{fig:inf_plane_cifar100_beta}, we can see the information plane for both {\vdib} and the {\vib} models. As we can see, both models are monotonic with $I(X;\hX)$, however the {\vib}'s performance is better. The {\vib} achieves higher values of information with the labels along with more compressed representation at any given level of predication. Although a broader analysis is required, and possible further parameter tuning of the architecture, we hypothesize that the caveat is in the noise model used for the {\vdib}. Using a noise model which is based on a network that achieves almost $20\%$ error might be insufficient in this case. It might be that ``errors'' in the noise model becomes similar to random errors, similar to the Gaussian case, and hence depicting similar learning performance to the {\gvdib} case. 

When we look at the information with the input as a function of time ,\fref{fig:inf_plane_cifar100_steps}, we see that similar to the FasionMNIST and CIFAR10 results, the information saturates for small values of $\beta$, but over-fits for higher values of it. 

%% file: Chapters/general_discussion.tex
\chapter*{General Discussion} 
\addcontentsline{toc}{chapter}{  General Discussion}

This thesis explored DNNs via an information theory perspective. Using the Information Bottleneck (IB) principle, we explored the underlying behavior of DNNs. We formulated deep learning as an information-theoretic trade-off between compression and prediction, which gives an optimal representation for each layer. Our theory is rooted in the idea that DNN learning with SGD aims to learn optimal representations in the IB sense. At its core, our theory aims to summarize each hidden layer using  mutual information with the input and output.  In the first section of this thesis, we combined the novel perspective with an empirical case study to make claims about phase transitions in SGD optimization dynamics, the computational benefits of deep networks, and the relation between generalization and compressed representation. These observations were collected to a new information-theoretic paradigm to explain deep learning, which inspired multiple follow-up works. As several researchers have noticed, measuring information in deterministic DNNs is hard \citep{goldfeld2018estimating}. To overcome this problem, in the second section of this thesis, we utilized the NTK framework to derive many information-theoretic quantities in infinitely wide neural networks. These quantities allow us to explore where the information is in DNNs and the relationship between generalization, compression, and information. By ensemble over different initial conditions, we tried to find the hyperparameters' effect on the network's information.  Our analysis revealed several interesting connections between the different information-theoretic quantities in this framework, the optimally of DNNs, their generalization ability, and capacity.

In the last section of this thesis, we presented the dualIB framework, which enabled an optimal representation that resolves some of the original IB's drawbacks. We provided the dualIB self-consistent equations, which allowed us to obtain analytical solutions; we characterized the structure of the critical points of the solutions, resulting in a full bifurcation diagram of its representation; we derived several interesting properties of the dualIB: First, when the data can be modeled in a parametric form, it preserves the original distribution's statistics. Second, it optimizes the mean prediction error exponent, therefore improving the predictions as a function of the data size. Additionally, we provide a variational dualIB framework. By optimizing its functional using DNNs, we can apply the dualIB for real-world datasets. These results demonstrate the potential advantages of the framework in the context of information in DNNs.

Although many works have been done based on ideas from this thesis,  the framework, and the experiments laid out in it open several more important avenues for future research. A few examples are outlined below.
\begin{itemize}
    \item \textbf{Generalization and compression} -- One of the most important questions in DNNs is their generalization ability. The followed-up works that have presented empirically investigations of the connection between compression of the information and generalization have attained mixed results. While the authors of \cite{entropy2019} saw a clear connection, the authors of \cite{gabrie2018entropy} concluded that compression and generalization may not be linked. This contradicts many PAC-Bayes bounds on the information, which bound the generalization gap between train and test error by these information-theoretic quantities. The natural question that arises is the reason for this discrepancy. Is it due to poor information estimators in DNNs? Are these bounds not tight enough? This research line can shed light on the important information quantities and how useful they can explain DNNs. Our work involving infinitely-wide neural networks is the first step in this direction. However, a broader analysis is needed to understand the connections between the different factors. 

    \item \textbf{Multi-domain, semi-supervised representation learning } -- Semi-supervised learning takes advantage of a large amount of unlabeled data that are available for many uses in addition to typically smaller sets of labeled data \citep{van2020survey}. However, semi-supervised learning's advantages are unclear from an information and probabilistic perspective. Taking advantage of this approach to multi-domain data allows us to better control information in our network. Using information-theoretic principles and self-supervised approaches, we can create compressed representation learning from multi-domain and multi-task data. Several questions remain to be examined: How should data from different modalities be stored? For each domain, what are the main factors leading to better generalizations? By compressing irrelevant information within each domain, how can we make our models more robust by using semi-supervised learning? The generalization and robustness should be improved with the integration of multimodality models using semi-supervised learning.

    \item \textbf{The benefits of the hidden layers in DNNs} -- This thesis suggested that one of the benefits of hidden layers is computational; by adding layers, the amount of compression each layer needs to do is reduced, resulting in a dramatic reduction in training time. Nevertheless, training very neural networks is challenging, and the network's performance does not monotonically increase with layers \citep{zagoruyko2017diracnets}. Understanding this trade-off between learning from finite samples and computational benefit will allow us to develop better principle designs for DNNs. 
    \item \textbf{Finite-sample information plane} -- In sections $1$ and $2$, we illustrated the network's behavior in the information plane for different dataset sizes. It remains a challenge, however, to pinpoint the exact relationship between the two. We need to investigate further how the finite-sample problem affects the optimization dynamics and whether a corresponding IB problem exists for each dataset size.      
    \item \textbf{Variational dualIB} --  Our thesis presented a variational {\dualib} formulation. A study comparing the differences between the IB and the {\dualib} models would be an interesting future direction. According to our initial analysis, the variational {\dualib} model compresses better than the regular VIB model. It will be interesting to explore it on larger datasets, more noise models, and different architectures. In addition, the question whether the differences between the models affect network properties, such as robustness, remains open. 

\item \textbf{Biologically plausible models } -- 
 The question of how the brain processes sensory input and elevates it was at the heart of much of the early interest in neural networks. In spite of being inspired by brains, deep artificial neural networks do not exhibit brain-like characteristics. Adapting biologically plausible deep learning algorithms to information-theoretic learning principles would be an interesting future direction. Gradient back-propagation, for example, relies on mechanisms that seem biologically implausible. It is not possible to alternate between a bottom-up forward pass and a top-down backward pass, and the labels are not available for each example. These problems may be solveable by the IB principle. For example, by combining a bottleneck objective with self-supervised learning, neural networks can be trained layer-wise without labels presented and through alternative bottom-up forward passes and top-down backward passes \citep{pogodin2020kernelized}.

More generally, in this thesis, we outlined the foundation for a novel, comprehensive theory of large-scale learning via deep neural networks that builds upon the correspondence between deep learning and the information bottleneck framework. One of the most important and challenging directions in this field is to use theoretical tools from other fields to analyze deep networks. Even though most deep learning studies use practical applications, they can only be effective if they are backed up by good theoretical knowledge.  Our theory offers a number of benefits, such as providing a deeper understanding of the information that resides within DNNs, identifying different explanations for their behavior, and opening up theoretical and practical research opportunities in the field. Hence, many open questions remain for further research, and by combining these directions, we may arrive at a stronger theory of the field that can also bring practical benefits and specific design principles. 

\end{itemize}
\bibliographystyle{dcu}

\bibliography{main}